\documentclass[10pt,twocolumn,letterpaper]{article}

\usepackage[final]{cvpr}      
\definecolor{cvprblue}{rgb}{0.21,0.49,0.74}
\usepackage[pagebackref,breaklinks,colorlinks,allcolors=cvprblue]{hyperref}

\def\paperID{} 
\def\confName{CVPR}
\def\confYear{2026}

\title{Why MLLMs Struggle to Determine Object Orientations}

\author{
Anju Gopinath \quad Nikhil Krishnaswamy \quad Bruce Draper\\
Department of Computer Science\\
Colorado State University\\
Fort Collins, CO, USA\\
{\tt\small anju@colostate.edu}
}

\begin{document}
\maketitle
\begin{abstract}
Multimodal Large Language Models (MLLMs) struggle with tasks that require reasoning about 2D object orientation in images, as documented in prior work. \citeauthor{tong2024eyes} and \citeauthor{nichols2025right} hypothesize that these failures originate in the visual encoder, since commonly used encoders such as CLIP and SigLIP are trained for image–text semantic alignment rather than geometric reasoning. We design a controlled empirical protocol to test this claim by measuring 
whether rotations can be recovered from encoder representations.
In particular, we examine SigLIP and ViT features from LLaVA-OneVision and Qwen2.5-VL-7B-Instruct models, respectively, using full images, and examine CLIP representations in LLaVA-1.5 and 1.6 using rotated foreground patches against natural background images. Our null hypothesis is that orientation information is not preserved in the encoder embeddings and we test this by training linear regressors to predict object orientation from encoded features. Contrary to the hypothesis, we find that orientation information is recoverable from encoder representations: simple linear models accurately predict object orientations from embeddings. This contradicts the assumption that MLLM orientation failures originate in the visual encoder. 

Having rejected the accepted hypothesis that MLLMs struggle with 2D orientation tasks because of visual encoder limitations, we still don't know why they fail. Although a full explanation is beyond the scope of this paper, we show that although present, orientation information is spread diffusely across tens of thousands of features. This may or may not be while MLLMs fail to exploit the available orientation information.

\end{abstract}    
\section{Introduction}
\label{sec:intro}

 Multimodal Large Language Models (MLLMs) struggle with tasks that require identifying orientations of objects in images \cite{nichols2025right,tong2024eyes,yang2025thinking,niu2025rotbench,wu2025spatial,wang2025spatialviz,zhu2025internvl3,chenspatial,lian2025euclid, zheng2025multimodal, ma2026attention, yu2025far, daehyunaligning,jia2025omnispatial}. For example, \citeauthor{nichols2025right} tested 15 MLLMs on questions requiring granular knowledge of object orientations and found that the best systems could only answer $34\%$ of the questions correctly \cite{nichols2025right}.
 Similarly, \citeauthor{kamoi2024visonlyqa} evaluated 23 MLLMs on visual question answering (VQA) tasks that require knowledge of shapes and angles, and found that the best were acccurate only $60\%$ of the time \cite{kamoi2024visonlyqa}. On the Visual-Spatial Intelligence Benchmark (VSI-Bench), \citeauthor{yang2025thinking} discovered that MLLMs performed near or below chance-level performance and \cite{zhu2025internvl3} report the highest score of 46.3 for a proprietary model and 43.3 for an open source model (7B parameters) vs. 39.5 (78B parameters) for relative directional tasks among 11 MLLMs~\cite{yang2025thinking,zhu2025internvl3} . This observation is corroborated by \citeauthor{zhang2025mllms}'s finding that increasing the training data alone is not sufficient to improve performance across all spatial tasks \cite{zhang2025mllms}. Even worse, linguistic reasoning techniques such as \textit{Chain-of-Thought (CoT), Self-Consistency} and \textit{Tree-of-Thoughts} led to a significant performance degradation on the benchmark, highlighting the lack of a robust spatial reasoning module in the architecture of MLLMs \cite{yang2025thinking}. 
 
 The reason why MLLMs fail at 2D orientation tasks is unknown, but \citet{tong2024eyes} and \citet{nichols2025right} suggest that the initial visual encoders may be at fault. Most visual encoders are not trained on orientation tasks, and if object orientation information is not embedded in the visual encoding, there is no way for an MLLM to recover it.

This paper empirically tests the hypothesis that MLLMs struggle with orientation tasks because their visual encoders fail to embed orientation information. In particular, we measure how well a linear regressor can predict the orientation of an object or image from its visual embedding. Since this requires access to a network's internal representations, we test four open-source systems (LLaVA-OneVision \cite{li2024llava}, Qwen2.5-VL-7B \cite{bai2025qwen2}, LLaVA-v1.5-13B \cite{liu2024improved}, and LLaVA-v1.6-vicuna-13B \cite{liu2024llavanext}) which between them use three open-source visual encoders (ViT \cite{dosovitskiy2020image}, SigLIP \cite{zhai2023sigmoid}, and CLIP \cite{radford2021learning}). 
For systems that take two images as input, we give them an image and a rotated version of the same image and ask for the degree of rotation, while for single image input systems, we rotate a foreground object and ask for the rotation of the foreground object relative to the (unrotated) background. Not surprisingly, all four systems fail at their assignments -- we have already noted that MLLMs struggle with orientation tasks. What we measure is how well linear regressors can
predict the rotations from the internal visual embeddings.


Our key finding is that linear regressors {\em can} predict image and object rotations to within $\pm 3^\circ$ from the visual embeddings. This contradicts the original hypothesis and raises a new question: if the fault is not with the visual encoders, why do MLLMs struggle with orientation tasks? We do not conclusively answer this question in its entirety, but we explore some properties of rotation information in visual embeddings. We discover (1) that errors in rotation estimation are not only small but approximately Gaussian and (2) that orientation information is diffusely distributed across tens of thousands of values in the embedding vector.

To summarize, our contributions are:

(1) We show that the SigLIP, ViT and CLIP vision encoders capture the orientations of images and foreground objects with a high degree of accuracy (MAE $<3^{\circ}$). 

(2) We show that the errors in orientation predictions based on SigLIP, ViT and CLIP encoding are roughly Gaussian and distributed across tens of thousands of features.\\
 For anyone wishing to replicate this work, the images and code can be found in \href{https://github.com/anjugopinath/MLLM\_Orientation}{https://github.com/anjugopinath/MLLM\_Orientation}.


\section{Related Work}
\label{sec:relatedWork}

\subsection{Orientation Estimation Capabilities of Multimodal Large Language Models}
\label{sub:relWork_orienEstMLLM}
Existing works have identified the gap in visual-spatial intelligence of MLLMs and have attributed it to their limited
spatial reasoning capabilities \cite{yang2025thinking,qu2025spatialvla,zhang2025embodied,yuan2025depthvla,bigverdi2025perception,chenspatial, ghaffari2024large, zheng2025multimodal, ma2026attention, yu2025far, daehyunaligning,jia2025omnispatial}. Orientation (or relative direction) estimation is one of the metrics in these works that performs poorly.

Previous works addressing object orientation have constructed benchmarking datasets containing coarse-grained questions (e.g. left vs. right, above vs. below, and front vs. behind) \cite{chenspatial,shiri2024empirical,jung2025isright} that test the
understanding of relative orientation and fine-grained questions ($0^{\circ}$, $90^{\circ}$, $180^{\circ}$, $270^{\circ}$)  \cite{niu2025rotbench} that test the accuracy in estimating a few canonical orientations. \citeauthor{jung2025isright} argue that inconsistent data annotation leads to poor results on object orientation estimation tasks \cite{jung2025isright}. \citet{nichols2025right} and \citet{tong2024eyes} hypothesize that the reason why MLLMs perform well on categorical clustering of directions into ``left" and ``right" but fail at angular estimation is
because they are pretrained on CLIP-like models which focus on image-text semantic alignment rather than on geometric understanding. Similary, \citeauthor{yoon2025visual} conclude that MLLMs lose fine-grained visual information due to the visual instruction tuning paradigm \cite{yoon2025visual}, as does \citet{liu2026spatial}. \citeauthor{huynh2025vision} analyzes orientation estimation capabilities of LVLMs using odd-one-out experiments and concludes that the poor performance is due to the information provided by the vision encoder not being discriminative enough \cite{huynh2025vision}. We note that these works do not analyze the performance of the vision encoder in MLLMs in isolation on orientation estimation. In contrast, we perform a fine-grained analysis for every 1$^{\circ}$ orientation between $0-360^{\circ}$ and conclude
that even though the language models perform poorly the vision encoders embed orientations to a high degree of accuracy.
\subsection{Feature/Latent Space Manipulation}
For both LLMs and vision-only models, there exist methods for interpretability and feature analysis. In vision-only models, disentangling CNN representations have been studied to understand how a visual
pattern might describe object parts or textures \cite{zhang2018visual}. In LLMs, methods such as Representation Control \cite{bartoszcze2025representation} involves editing the latent space of a neural network to steer the  model towards a particular output by injecting a vector in between the layers of a model at inference. 
While these methods 
involve architectural changes to analyze the interpretability of Large Language Models, analyzing and controlling representations in vision-language models
remains challenging \cite{tian2025representation}.
\citeauthor{cuidual} performs interchange intervention to analyze ordering information contained in VLMs \cite{cuidual}. Faced with the unique challenge of identifying how vision-language models such as LLaVA-LLaMA and Qwen2.5-VL-7B-Instruct might be encoding orientation awareness in their vision encoders, we analyze the visual embeddings and perform feature substitution by iteratively substituting features from an anchor embedding into a target embedding with the goal of identifying the features that 
encode orientation.

\subsection{Orientation Estimation by Neural Networks}
There is an older line of research asking whether Convolutional Neural Nets (CNNs) can tell if an image has been rotated from its original orientation. \citet{sun2017orientation} perform orientation estimation using image features on the Outdoor Images dataset collected from the Flickr1M dataset. The orientations were manually annotated. The work titled OSKDet performs orientation-sensitive keypoint based rotated object detector \cite{lu2022oskdet}. \citeauthor{fischer2015image} perform exact orientation estimation on general natural images. Images from the Microsoft COCO dataset were artificially rotated and in the test set, slanted images, framed images and images which do not have a well-defined orientation were discarded \cite{fischer2015image}. Their method estimates orientation with an average accuracy of 3° in the setting with $\pm$30°. Their model was built on the AlexNet architecture pretrained on ImageNet. While existing works on orientation estimation that utilize features from image-based feature extractors are predominantly vision-only models, in this work, we study orientation estimation using features from LLaVA's vision encoder.


\section{Methodology}
\label{sec:methodology-orientation}
\begin{table}[h]
\centering
\footnotesize
\renewcommand{\arraystretch}{1.1}
\setlength{\tabcolsep}{6pt}
\begin{tabular}{|l|p{6cm}|}
\hline
\textbf{Image Set} & \textbf{Prompt} \\ 
\hline
Whole Images & How much is the 2nd image rotated clockwise when compared to the first image?\\
\hline
\makecell[l]{In-Place Rotated\\Images}& How much is the $<$object(s)$>$ rotated clockwise in degrees when compared to the first image?\\
\hline
\hline
dog on beach & By how much is the dog in the center inside the circle rotated if it is known that the rotation is zero degrees when the dog's legs are vertical? \\ 
\hline
lizard on fish & By how much is the lizard in the center inside the circle rotated if it is known that the rotation is zero degrees when the lizard is roughly horizontal with its tail pointing left and its head pointing right? \\ 
\hline
train on indoor & By how much is the train in the center inside the circle rotated if it is known that the rotation is zero degrees when the train is vertical with the train tracks at the bottom and the steam from the train going vertically upwards? \\ 
\hline
\end{tabular}
\caption{Image sets and corresponding prompts to LLaVA/Qwen2.5-VL-7B-Instruct.}
\label{tab:image_prompts}
\end{table}

\begin{figure*}[h!]
  \centering
   \includegraphics[width=16cm, height=8.5cm]{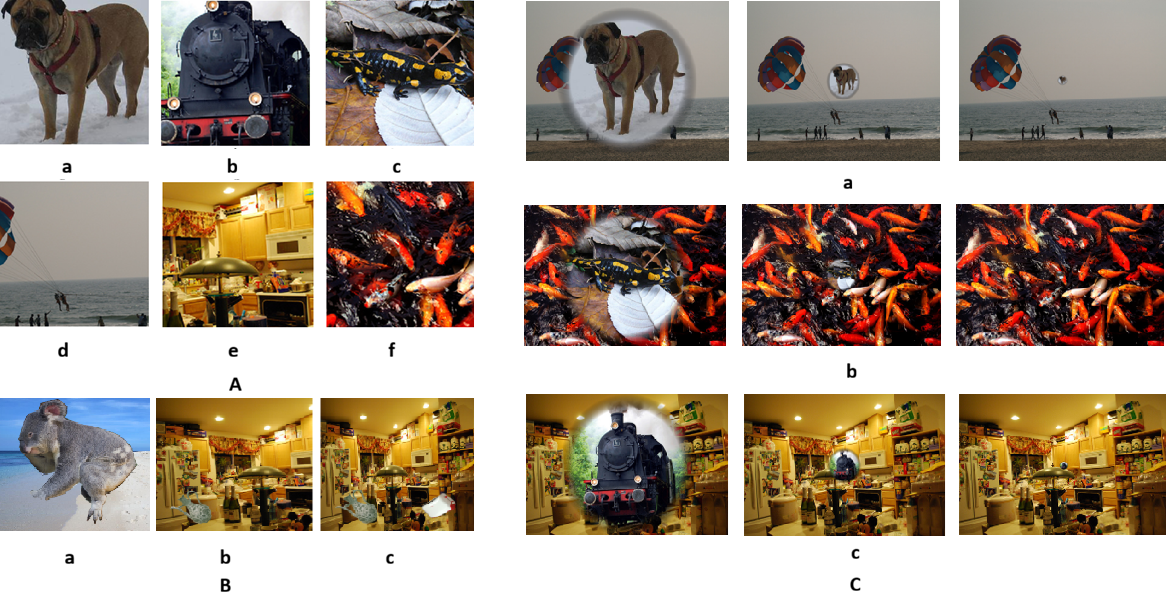}
   \caption{Set of images - Sets A and B are used for experiments with LLaVA-OV and Qwen2.5-VL-7B-Instruct, and set C is used for LLaVA1.5 and LLaVA1.6. Since LLaVA-OV and Qwen2.5-VL-7B-Instruct can be prompted with 2 images, real images were used. But since LLaVA1.5 and 1.6 can be prompted with only 1 image, synthetic images with reduced artifacts were used to improve the prediction accuracy.\\
\textbf{[A: Whole Images]}: (a) dog (b) train (c) lizard (d) beach (e) indoor (f) fish. \\
\textbf{[B: In-Place Rotated Objects]}: (a) koala on beach (b) vase on indoor (c) vase and toaster on indoor. \\
\textbf{[C: Blended Images with foreground on background]}: (a) dog on beach (b) lizard on fish (c) train on indoor environment, with foreground scales 1, 2 and 3 from left to right.}
   \vspace{-2mm}
   \label{fig:blended_dataset}
   \hfill
\end{figure*}
We test the hypothesis suggested by \citeauthor{nichols2025right} that MLLMs struggle with orientation queries because CLIP-style encoders do not preserve object orientation information \cite{nichols2025right}. It is an appealing hypothesis because CLIP was trained for semantic contrastive alignment, not orientation tasks, but we find the hypothesis is not true. To test it, we use natural images for LLaVA-OneVision and Qwen2.5-VL-7B-Instruct (2 image version) and superimpose circular patches of foreground images onto natural background images for LLaVA1.5 and 1.6 (single image version). The superimposed images look odd (see Figure~\ref{fig:blended_dataset}, section C), but they allow us to apply arbitrary rotations to the foreground without introducing rotation-induced artifacts at patch boundaries. We present these images with a text query to LLaVA-LLaMA models (OneVision, 1.5 and 1.6) and Qwen2.5-VL-7B-Instruct and extract the embedding vectors produced by the visual encoder (SigLIP for LLaVA-OneVision, ViT for Qwen and CLIP for LLaVA 1.5 and 1.6). We then train a linear ridge regressor to predict orientations of the image (2 image version - LLaVA-OneVision and Qwen2.5-VL-7B-Instruct) or foreground orientations given inputs composed of a single background image with a foreground patch superimposed at different orientations (single image version - LLaVA 1.5 and 1.6), and test it using novel whole image or foreground patch orientations respectively. The objective is to test whether the orientation of the whole image or of the foreground patch is encoded in the visual embedding vector when all other factors are held constant. 

Probing further, we estimate the accuracy of the orientation predictions for 6 image sets (2 image version - LLaVA-OneVision and Qwen2.5-VL-7B-Instruct) and at different foreground scales and with three different foreground/background pairs (single image version - LLaVA 1.5 and 1.6). We find that the mean average prediction error is always less than $3^{\circ}$ and that the errors are roughly Gaussian. This implies that LLaVA's and Qwen2.5-VL-7B-Instruct's visual encoder does preserve orientation information, and that the orientation signal is accurate and well-behaved. The rest of this section describes the experiment and its analysis in more detail.

\subsection{Constructing Image Samples}
We construct 3 distinct types of image sets for experiments as detailed in the subsections below. Models that accept multiple image inputs are prompted to determine the orientation of a rotated image compared to an unrotated image, while models that take only a single image input are asked for the orientation of a rotated foreground patch relative to an unrotated background image. 

\subsubsection{Cropped Whole Images and Images with In-Place Rotated Objects for LLaVA-OneVision and Qwen2.5-VL-7B}
We perform experiments with two categories of images - whole images where the whole image is rotated and images with in-place rotated objects where selected objects are rotated keeping the background static.
For the first category, we selected six images from ImageNet \cite{ILSVRC15}. Each image was rotated in 1$^{\circ}$ increments, and the central region was cropped to remove blank canvas artifacts introduced by rotation, which could otherwise affect regression accuracy. After preprocessing, each image set contained 180 samples. For the second category, we utilize the SI-Score dataset \cite{yung2021si} and code, and generate 3 image sets, each with 180 samples. Two of them - \textit{koala} and \textit{vase} are instances of a single object rotated in-place while \textit{vase and toaster} is an instance of two objects rotated in-place.
The resultant images from both categories are either 200 by 200 or 250 by 250 in size. The first image from each set from the two categories are shown in Figure \ref{fig:blended_dataset}, sections A and B.

\subsubsection{Blended Image Samples for LLaVa1.5 and 1.6}

\begin{figure}[t]
\centering
\includegraphics[width=6cm,height=2cm]{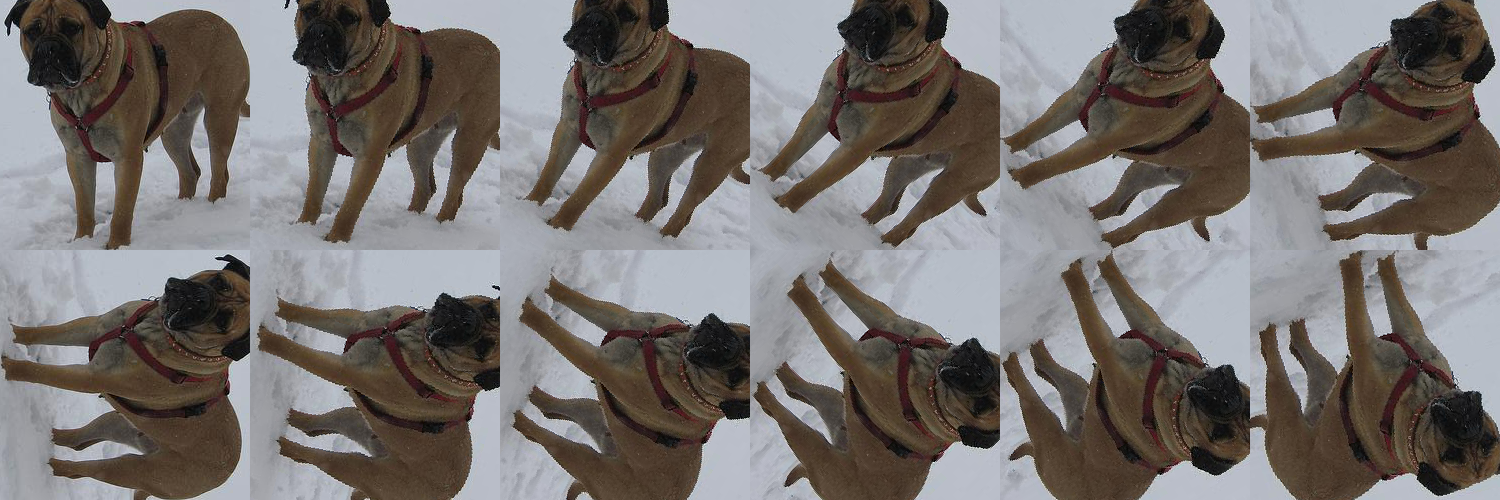}

\vspace{0.3cm}
 \includegraphics[width=6cm,height=2cm]{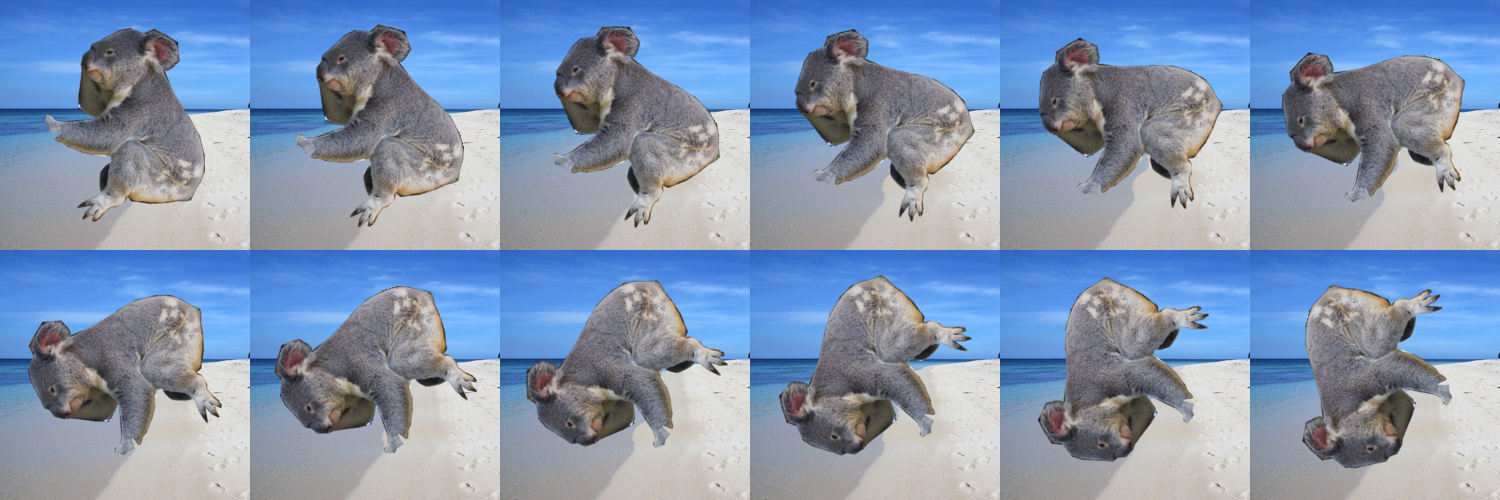}

\caption{Collage of every 15th image from Sections A (a) dog (whole image) and B (a) koala (in-place rotated) of Figure \ref{fig:blended_dataset}.}
\label{fig:collage_whole_in_place}
\end{figure}

Newer MLLMs that take multiple image inputs do not use the older CLIP encoder. To test whether the CLIP encoder preserves foreground orientation information, we created test images where the 2D orientation of the foreground could be carefully controlled. We started with three natural images from ImageNet \cite{ILSVRC15} for the background: a beach scene, a top-view of a koi pond, and an indoor picture of a kitchen. Next, the rectangular source foreground image (also from ImageNet) is padded using black pixels to obtain a square. A black square array having the same dimensions as the padded source is created on which a circle is drawn using alpha blending with a mask value of 1.0 (opaque) inside the circle which gradually fades to 0.0 (transparent) outside it ($Blended = Foreground \times Mask + Background \times (1 - Mask)$). This ensures a smooth transition, avoiding jagged boundaries. This process is repeated every time as the foreground patch is rotated in $1^\circ$ increments. The final output is a rectangular background (bg) image with a circular patch (fg) in the center which serves as the foreground object. We used three sizes (in pixels) of fg patches: large (l), medium (m), small (s) - measured by pixel diameter, with the resulting combinations being :
dog images [$500 \times 375$ (bg) $|$ fg patches : 272 (l), 68 (m), 18(s)], lizard images [$500 \times 333$ (bg) $|$ fg patches : 264 (l), 66 (m), 16(s)],
train images [$500 \times 334$ (bg) $|$ fg patches : 264 (l), 66 (m), 16(s)]. The resulting $3,240$ images (3 foreground/background pairs $\times$ 3 scales $\times$ 360 orientations), of which the first image from each set is shown in Figure \ref{fig:blended_dataset} section C, do not look natural but that is not the point. By controlling the size of the foreground patch, we can rotate it without introducing orientation-specific artifacts. Figure \ref{fig:collage_whole_in_place} shows every 15th image from two categories of the image sets - whole images and in-place rotated objects. Figure \ref{fig:collage} in the supplementary shows 18 rotations of the dog superimposed on the beach (blended image category). Note that we did not sample a large number of images because we are not interested in the contents of the images. We are interested in how the image encoding changes as the whole image/object/foreground is rotated, and whether these changes can be used to predict the orientation.  Hence lots of orientations but few base images. The images include both indoor and outdoor scenes and both structured and less organized content.

\section{Orientation Prediction Results}
\begin{figure}[t]
  \centering
  \includegraphics[width=7.5cm, height=5cm]{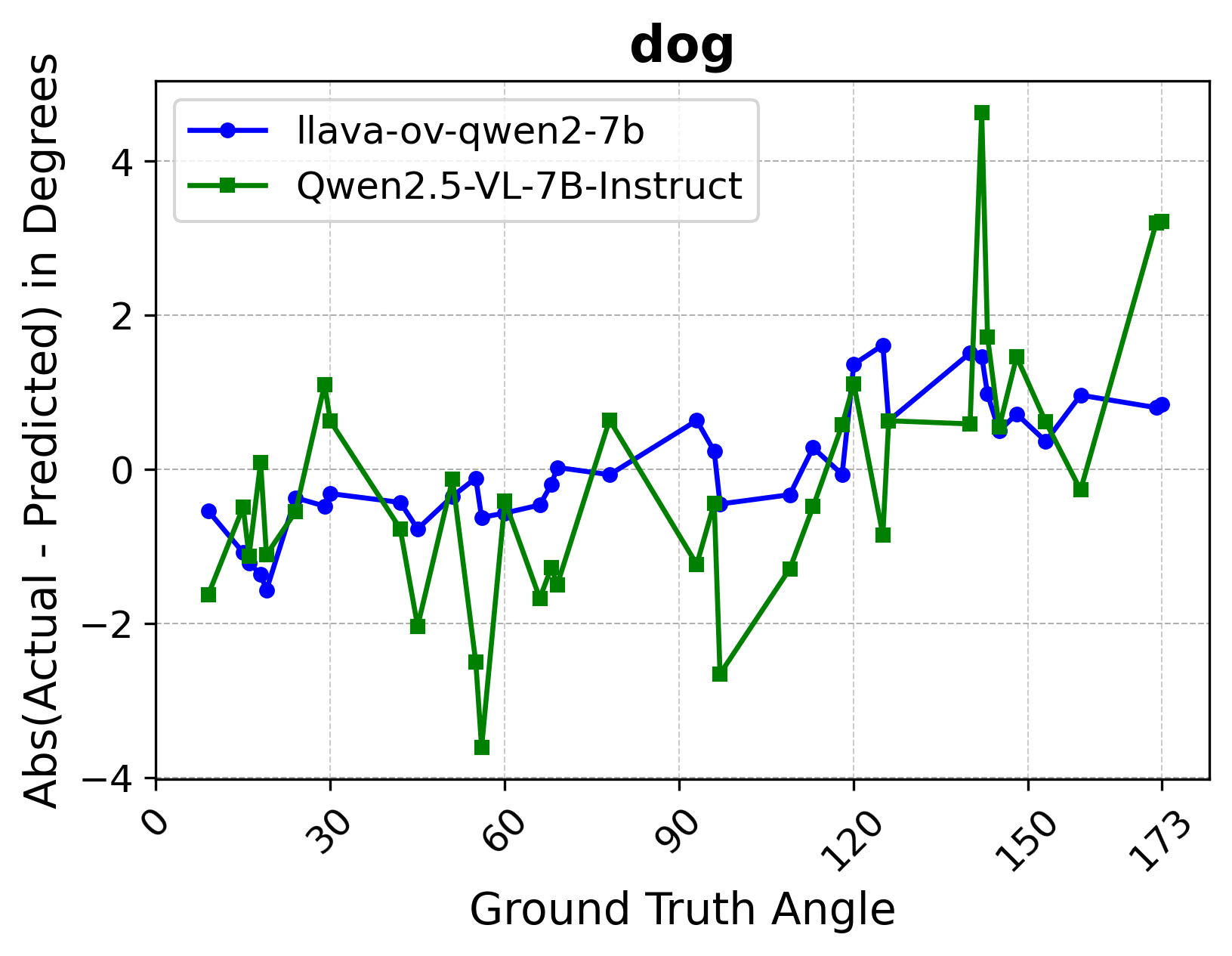}
  \vspace{-2mm}
  \caption{2D orientation estimation performance comparison of LLaVA-OneVision and Qwen2.5-Vl-7B-Instruct on the images with the dog for 36 randomly selected images.}
  \label{fig:regression_comparison_llava1.5_1.6_dog-on-beach}
\end{figure}

The goal of this experiment is to determine if the orientation of the whole image and in-place rotated object is preserved by LLaVA-One Vision and Qwen2.5-VL-7B-Instruct and similarly, if the orientation of the foreground patch is preserved by LLaVA 1.5 and 1.6. LLaVA, Qwen2.5-VL-7B-Instruct and its CLIP/SigLIP/ViT encoders are not retrained or altered in any way for this experiment. Instead, the original and rotated images, along with the textual prompt, are presented to LLaVA-OneVision and Qwen2.5-VL-7B-Instruct or in the case of LLaVA 1.5 and 1.6, for every background image, the variations with different foreground rotations are presented along with textual prompts (see Table~\ref{tab:image_prompts} for the prompts). For each input sample, the feature vector computed by CLIP/SigLIP/ViT is extracted from the network, flattened, and analyzed. Before flattening, the embeddings for LLaVA-OneVision are $4\times729\times1152$ values, $392\times 1280$ or $648 \times 1280$ values for Qwen2.5-VL-7B-Instruct depending on the image size, $576 \times 1024$ values for LLAVA 1.5; for LLaVA 1.6 the embedding size depends on the image size and is $5 \times 576 \times 1024$ for the beach images, and $3 \times 576 \times 1024$ for the fish and kitchen images. The embedding vectors for each image set are divided 80:20 into training and test sets and normalized. Two independent Linear Ridge Regressors are used: one trained to predict the sine of the angle and the other to predict the cosine. Both models are provided with a predefined set of regularization parameters ($\alpha$) and are trained using K-fold cross-validation to determine their respective optimal L2 regularization strengths ($\alpha$). The resulting optimal $\alpha$ values (0.005 for both models) are then used to refit each separate Ridge Regression model on the entire training dataset to yield the final predictions for the sine and cosine of the rotation angle of the rotated image.
We compare the performance of LLaVA-OneVision and Qwen2.5-VL-7B-Instruct vision encoders for the image set with the dog in Figure \ref{fig:regression_comparison_llava1.5_1.6_dog-on-beach}. Additional plots are presented in the supplementary material in sections \ref{app:reg-comp-llavaOV-qwen} and \ref{app:reg-comp-llava1.5-1.6}. The Mean Absolute Error (MAE) of the predictions along with the maximum and minimum values for all image sets are presented in Table \ref{tab:llava_qwen_comparison}. The mean errors are all less then $2^\circ$ except for the whole images with the fish scene and the images with the lizard foreground with the biggest foreground patch for the blended images.

\begin{table*}[ht]
\centering
\footnotesize
\setlength{\tabcolsep}{3pt}

\begin{tabular}{cc}

\begin{tabular}{|c| >{\columncolor{maecolor}}c >{\columncolor{maxcolor}}c c| >{\columncolor{maecolor}}c >{\columncolor{maxcolor}}c c|}
\hline
\multicolumn{7}{|c|}{\cellcolor{titlecolor}\textbf{Two Image Version}} \\
\hline
\multicolumn{7}{|c|}{\textbf{Whole Images}} \\
\hline
 & \multicolumn{3}{c|}{\textbf{LLaVA-OV}} & \multicolumn{3}{c|}{\textbf{Qwen Instruct}} \\

\hline
 & MAE & Max & Min & MAE & Max & Min \\
\hline
\textbf{Dog}     & 0.67 & 1.61 & 0.03 & 1.30 & 4.63 & 0.09 \\
\hline
\textbf{Lizard}  & 0.44 & 1.63 & 0.01 & 2.20 & 6.31 &0.06  \\
\hline

\textbf{Train}   & 0.56 & 2.18 & .01 &2.43  & 10.01 & 0.29 \\
\hline

\textbf{Beach}   & 0.74 & 2.67 & 0.03 &1.63  &4.98  & 0.03 \\
\hline

\textbf{Indoor}  & 0.81 & 2.92 & 0.003 & 1.32 & 4.96 & 0.01 \\
\hline
\textbf{Fish}    & 2.31 & 5.31 &  0.12& 2.85 &8.62  & 0.17 \\
\hline
\multicolumn{7}{|c|}{\textbf{In-Place Rotated Object(s)}} \\
\hline
\textbf{Koala}   & 0.36 & 1.47 & 0.01 & 1.01 &3.66  & 0.06 \\

\textbf{Vase}    & 0.48 & 1.76 & 0.003 & 0.90 & 4.18 &0.01  \\
\hline

\textbf{Vase \& Toaster} & 0.65 & 1.88 & 0.02 & 1.58 &5.81  &0.02  \\
\hline
\end{tabular}

&
\hspace{1cm}

\begin{tabular}{|c|c| >{\columncolor{maecolor}}c >{\columncolor{maxcolor}}c c| >{\columncolor{maecolor}}c >{\columncolor{maxcolor}}c c|}
\hline
\multicolumn{8}{|c|}{\cellcolor{titlecolor}\textbf{Single Image Version}} \\
\hline
\multicolumn{8}{|c|}{\textbf{Blended Images}} \\
\hline
 & & \multicolumn{3}{c|}{\textbf{LLaVA 1.5}} & \multicolumn{3}{c|}{\textbf{LLaVA 1.6}} \\
\hline

 & & MAE & Max & Min & MAE & Max & Min \\
\hline

\multirow{3}{*}{\textbf{Dog}}
 & Scale 1 & 1.41 &4.19 &0.06 & 0.62 &2.40 &0.001 \\
\cline{2-8}
 & Scale 2 & 0.89 &3.20 &0.0001 & 0.6&2.01 &0.01 \\
\cline{2-8}
 & Scale 3 & 0.67 &2.52 &0.02 & 0.52 &2.2 & 0.03\\
\hline
\multirow{3}{*}{\textbf{Train}}

 & Scale 1 & 1.3  &3.79 &0.02 & 0.78 &2.12 &0.02 \\
\cline{2-8}

 & Scale 2 & 0.9  &2.69 &0.003 & 0.53 & 1.72& 0.0002\\
\cline{2-8}

 & Scale 3 & 0.72 & 2.12& 0.008& 0.42 & 1.2&0.005 \\
\hline

\multirow{3}{*}{\textbf{Lizard}}

 & Scale 1 & 2.08 &7.24 & 0.0003& 0.87 & 3.15&0.04 \\
\cline{2-8}

 & Scale 2 & 1.25 &4.67 & 0.005& 0.64 & 2.42& 0.02\\
\cline{2-8}
 & Scale 3 & 1.13 &3.53 &0.04 & 0.71 &1.69 &0.05 \\
\hline
\end{tabular}

\end{tabular}
\caption{\textbf{LLaVA-OneVision}, \textbf{Qwen2.5-VL-7B-Instruct}, \textbf{LLaVA 1.5} and \textbf{LLaVA 1.6} vision encoders estimate 2D orientation to a high degree of accuracy across diverse image sets.}
\label{tab:llava_qwen_comparison}
\end{table*}
\subsection{Error Distributions}
Surprisingly, the linear regressor was able to predict the orientation of foreground patches from the vision encoder embeddings, at least on average. This implies that LLaVA and  Qwen2.5-VL-7B-312
Instruct should be able determine the 2D orientation of familiar objects, at least within familiar settings. But how easy is this information to use? Predictions with normal error distributions are easily exploitable, so how are the errors of our trained ridge regressor distributed?

\begin{figure}[t]
  \centering
\includegraphics[width=7cm, height=6cm]{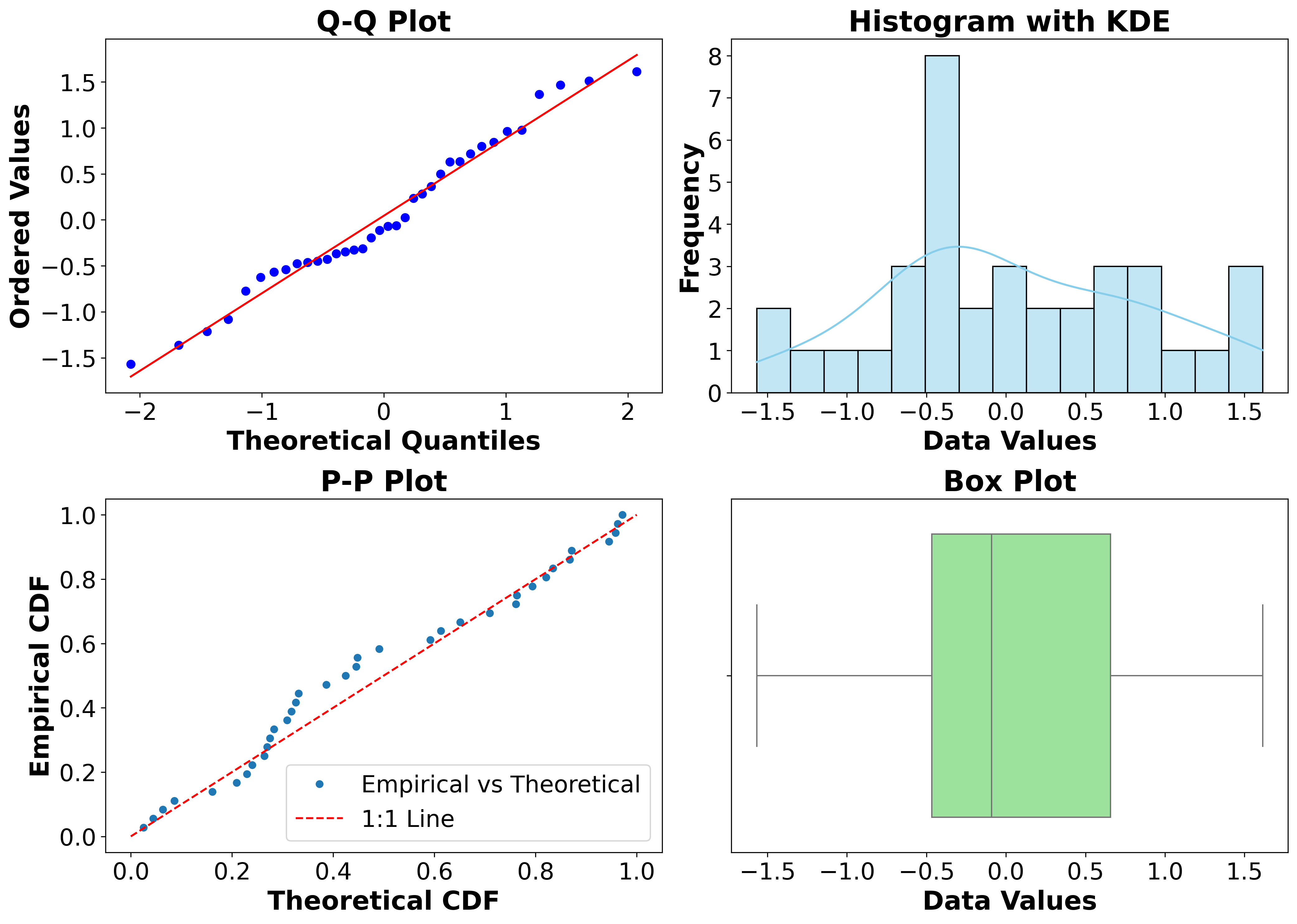}
   \caption{Statistical Analysis using visual plots for LLaVA-OneVision - results for images with dog. Together with the numerical results in Table \ref{tab:ks_test} and the visual plots in this Figure, we can conclude that the residual error distribution (Table \ref{tab:llava_qwen_comparison}) is random/Gaussian.}
   \label{fig:stat_analysisllavaOV_dog}
   \hfill
\end{figure}

To test the assumption of normality, we used a combination of statistical and 
visual assessment tests, as recommended by \cite{habibzadeh2024data}. In addition to histograms, Q-Q plots, P-P plots, and Box plots, we also used the Kolmogorov-Smirnov (K-S) test. For normally distributed data, in a Q-Q plot and P-P plot, observed data would be approximate to the expected data (an approximate straight line). For a box plot, the median line would be approximately at the centre of box with symmetric whiskers. In the histogram, the graph would be approximately bell-shaped and symmetric about the mean \cite{mishra2019descriptive}. The results of the K-S test are given in Table \ref{tab:ks_test}. With a low K-S statistic and a $p$-value greater than the significance level (alpha) of 0.05, there is no strong evidence to reject the null hypothesis of normality of the residuals \cite{massey1951kolmogorov,mishra2019descriptive}. The visual assessment plots for LlaVA-OneVision and Qwen2.5-VL-7B-Instruct for the dog images are present in Figures \ref{fig:stat_analysisllavaOV_dog} and \ref{fig:stat_analysisQwen2.5-VL-7B-Instruct_dog}. The results for other images and models are present in the supplementary material in Sections \ref{app:llava-qwen} and \ref{app:llava}. We therefore conclude that the prediction errors are random and normally distributed, at least approximately, i.e., it is the expected stochasticity associated with a neural network.
\begin{table*}[ht]
\centering
\footnotesize
\setlength{\tabcolsep}{3pt}

\begin{tabular}{cc}

\begin{tabular}{|c| cc| cc|}
\hline
\multicolumn{5}{|c|}{\cellcolor{titlecolor}\textbf{Two Image Version}} \\
\hline
\multicolumn{5}{|c|}{\textbf{Whole Images}} \\
\hline
 & \multicolumn{2}{c|}{\textbf{LLaVA-OV}} & \multicolumn{2}{c|}{\textbf{Qwen Instruct}} \\
\hline
 & K-S & p-value & K-S & p-value \\
\hline

\textbf{Dog}              & 0.11&0.69 &0.12 &0.6 \\
\hline

\textbf{Lizard}           & 0.14& 0.43&0.09 & 0.9\\
\hline

\textbf{Train}            & 0.11&0.74 &0.11 &0.7 \\
\hline

\textbf{Beach}            & 0.14&0.44 &0.11 &0.78 \\
\hline

\textbf{Indoor}           & 0.09&0.87 &0.09 &0.89 \\
\hline

\textbf{Fish}             & 0.06&0.997 &0.12 &0.62 \\
\hline
\multicolumn{5}{|c|}{\textbf{In-Place Rotated Object(s)}} \\
\hline
\textbf{Koala}            & 0.08&0.94 &0.12 &0.61 \\
\hline

\textbf{Vase}             & 0.08&0.97 &0.16&0.32 \\
\hline

\textbf{Vase \& Toaster}  & 0.09&0.93 &0.09 & 0.9\\
\hline
\end{tabular}

&
\hspace{1cm}

\begin{tabular}{|c|c| cc| cc|}
\hline
\multicolumn{6}{|c|}{\cellcolor{titlecolor}\textbf{Single Image Version}} \\
\hline
\multicolumn{6}{|c|}{\textbf{Blended Images}} \\
\hline
 & & \multicolumn{2}{c|}{\textbf{LLaVA 1.5}} & \multicolumn{2}{c|}{\textbf{LLaVA 1.6}} \\
\hline
 & & K-S & p-value & K-S & p-value \\
\hline
\multirow{3}{*}{\textbf{Dog}}
 & Scale 1 &0.09 &0.51 &0.05 &0.99 \\
\cline{2-6}
 & Scale 2 & 0.07&0.81 &0.08 &0.7 \\
\cline{2-6}
 & Scale 3 &0.1 &0.46 &0.12 &0.37 \\
\hline
\multirow{3}{*}{\textbf{Train}}
 & Scale 1 &0.06 &0.95 &0.06 &095 \\
\cline{2-6}
 & Scale 2 & 0.07&0.9 &0.09 &0.56 \\
\cline{2-6}
 & Scale 3 & 0.08&0.75 &0.1 &0.41 \\
\hline
\multirow{3}{*}{\textbf{Lizard}}
 & Scale 1 & 0.06&0.96 &0.08 &0.74 \\
\cline{2-6}
 & Scale 2 &0.08 &0.68 &0.04 &1 \\
\cline{2-6}
 & Scale 3 &0.06 &0.95 &0.08 &0.69 \\
\hline
\end{tabular}

\end{tabular}
\caption{Low Kolmogorov–Smirnov (K-S) statistic coupled with $p$-value $>$ alpha (0.05) means there's no reason to reject the null hypothesis that the error distribution for \textbf{LLaVA-OneVision}, \textbf{Qwen2.5-VL-7B-Instruct}, \textbf{LLaVA 1.5} and \textbf{1.6} are purely random (Gaussian).}
\label{tab:ks_test}
\end{table*}

\begin{figure}[t]
  \centering
\includegraphics[width=7cm, height=6cm]{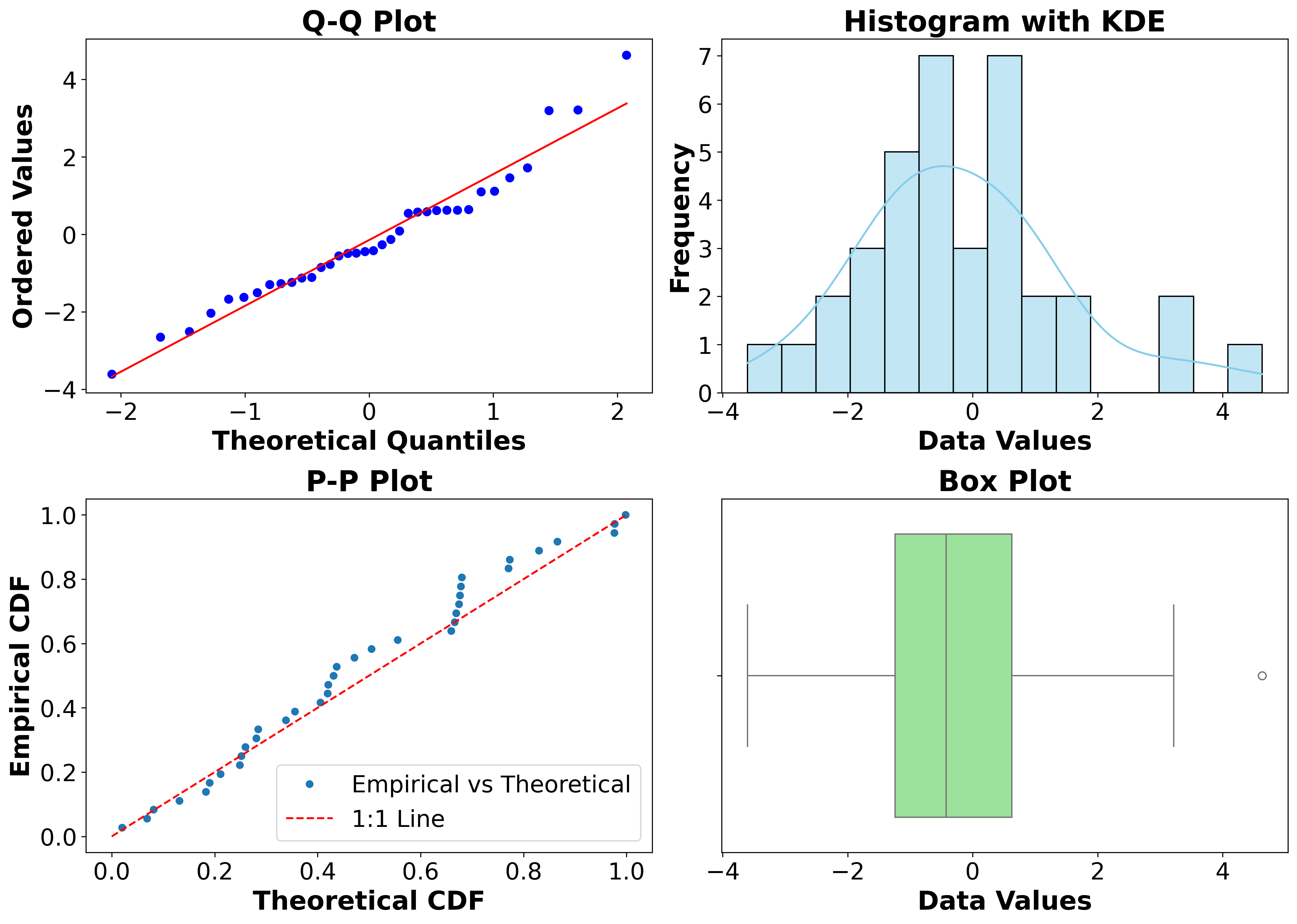}
   \caption{Statistical Analysis using visual plots for Qwen2.5-VL-7B-Instruct - results for images with dog. Together with the numerical results in Table \ref{tab:ks_test} and the visual plots in this Figure, we can conclude that the residual error distribution (Table \ref{tab:llava_qwen_comparison}) is random/Gaussian.}
   \label{fig:stat_analysisQwen2.5-VL-7B-Instruct_dog}
   \hfill
\end{figure}

\subsection{Models used for experiments}
\label{sub:result_orientation+models}
We performed experiments with LLaVA-OneVision, LLaVA 1.5, LLaVA 1.6 (LLaVA-NeXT), and Qwen2.5-VL-7B-Instruct. LLaVA-OneVision utilizes the SigLIP vision encoder, whereas LLaVA 1.5 and 1.6 use the CLIP vision encoder. In contrast, Qwen2.5-VL-7B-Instruct uses a native Vision Transformer (ViT) backbone. While CLIP uses a softmax-based contrastive loss that requires global pairwise similarities within a batch, SigLIP introduces a sigmoid loss that operates independently on each image-text pair. This makes SigLIP more memory-efficient and performant at smaller batch sizes \cite{zhai2023sigmoid}. Despite these algorithmic differences, both SigLIP and CLIP utilize contrastive learning to achieve image-text semantic alignment \cite{nichols2025right}.
\subsubsection{LLaVA-OneVision}
LLaVa-OneVision uses the SigLIP vision encoder and the Qwen2 LLM backbone \cite{li2024llava}. It uses the AhnyRes-9 technique. For multi-image settings, it takes the original low-resolution image and several high-resolution patches which is a function of the source image size. The final feature is obtained by concatenating the global and the local features. In our experiments, the shape of the vision encoder output is (4,729,1152).
\subsubsection{Qwen2.5-VL-7B-Instruct}
The architecture of the Qwen2.5-VL series is characterized by several key technical advancements aimed at enhancing multimodal efficiency and temporal precision \cite{bai2025qwen2}. First, the Vision Transformer (ViT) was modified to incorporate efficient attention mechanisms, SwiGLU activations, and RMSNorm, significantly optimizing inference throughput. To support native input resolutions, the vision encoder utilizes 2D-RoPE and processes images via a dynamic resolution strategy.
Second, the model extends its dynamic resolution capabilities and the spatial-temporal representation for better video understanding. Finally, these architectural improvements are supported by a massive scaling of the pre-training corpus, with the resulting visual features integrated into a Qwen2.5 language model decoder that has been specifically adapted for multimodal alignment.

\subsubsection{LLaVA 1.5 vs. LLaVA 1.6}

We tested 2D orientation estimation on 2 different LLaVA architectures with the CLIP vision encoder - 1.5 and 1.6.
Compared with LLaVA 1.5 (llava-v1.5-13B), LLaVA 1.6 (llava-v1.6-vicuna-13B) adopts the AnyRes technique where an image is split into a grid configuration of \{2×2,1×\{2,3,4\},\{2,3,4\}×1\} before being fed to the language model resulting in significantly more tokens (4-5x depending on the image size for our problem) in the feature space \cite{liu2024llavanext}. 
\subsection{Summary}
Contrary to our initial hypothesis, we show that CLIP-style encoders  (including SigLIP and ViT are trained similarly to CLIP, as detailed in Section \ref{sub:result_orientation+models}) do preserve the orientation of familiar foreground objects with a high degree of accuracy. In fact, foreground orientations can be recovered  with an MAE $< 3^{\circ}$ (Table \ref{tab:llava_qwen_comparison}) from LLaVA-OneVision, Qwen2.5-VL-7B-Instruct, LLaVA1.5 and LLaVA1.6 encodings. But this does not negate the findings in the literarure that MLLMs, including LLaVA, perform poorly on object orientation tasks. So if orientation information is preserved in visual embeddings, why can't LLaVA do better with orientation queries?

\section{Discussion}
\label{sec:methodology-patch_analysis}
\vspace{-2mm}
\begin{table}[h]
    \centering
    \footnotesize
    \setlength{\tabcolsep}{2pt} 
    \begin{tabular*}{\columnwidth}{@{\extracolsep{\fill}}cccc}
        \toprule
        \multicolumn{2}{c}{\textbf{LLaVA OneVision}} & \multicolumn{2}{c}{\textbf{Qwen2.5-VL-7B-Instruct}} \\
        \cmidrule(lr){1-2} \cmidrule(lr){3-4}
        \textbf{Angle ($^{\circ}$)} & \textbf{Count} & \textbf{Angle ($^{\circ}$)} & \textbf{Count} \\
        \midrule
        90 & 149 & 90 & 142 \\
        180 & 28 & 10-15  &  14\\
        0 & 2 & 10 & 11 \\
        45 & 1  & 180 &6  \\
        \bottomrule
    \end{tabular*}
    \vspace{-2mm}
    \caption{Top LLaVA-OneVision and Qwen2.5-VL-7B-Instruct query responses for the 180 samples with the dog scene. Both models frequently respond that the 2D orientation is 90$^{\circ}$.}
\label{tab:llavaOV_qwen_metrics}
\end{table}

\begin{figure*}[h!]
    \begin{subfigure}{0.33\textwidth}
        \includegraphics[width=\linewidth]{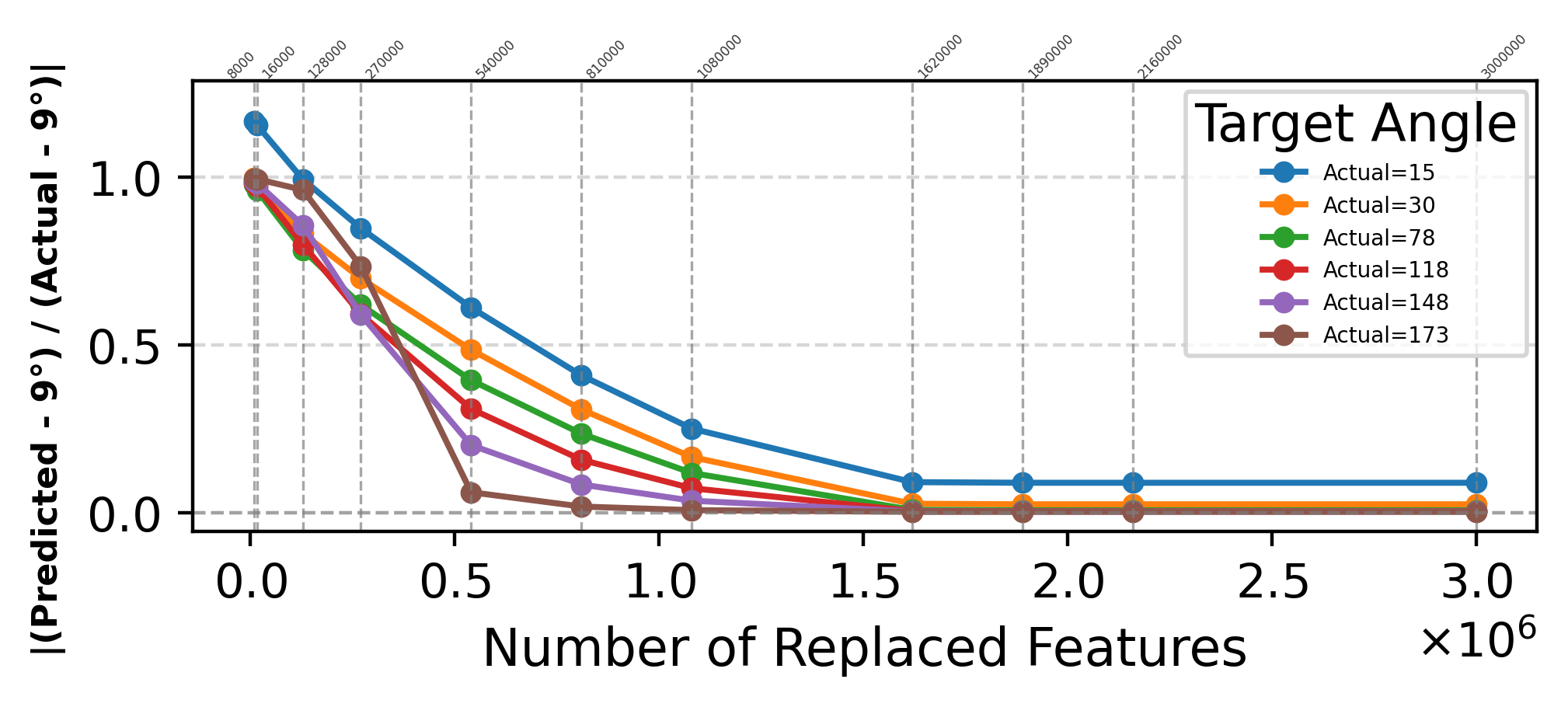}
        \caption{Ordered By Model Weight}
    \end{subfigure}%
    \hspace{-0.5em} 
    \begin{subfigure}{0.33\textwidth}
        \includegraphics[width=\linewidth]{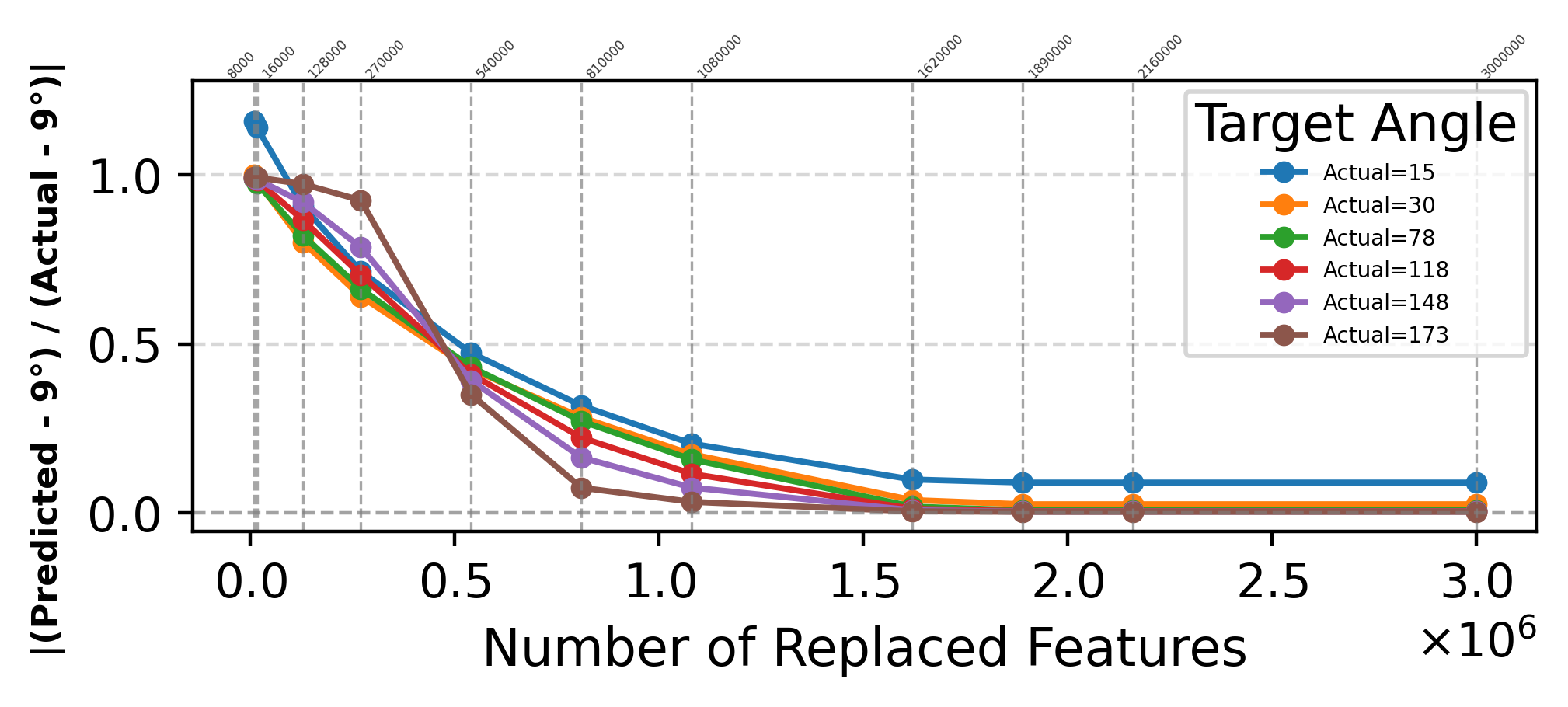}
        \caption{Ordered By Value Difference}
    \end{subfigure}%
    \hspace{-0.5em} 
    \begin{subfigure}{0.33\textwidth}
        \includegraphics[width=\linewidth]{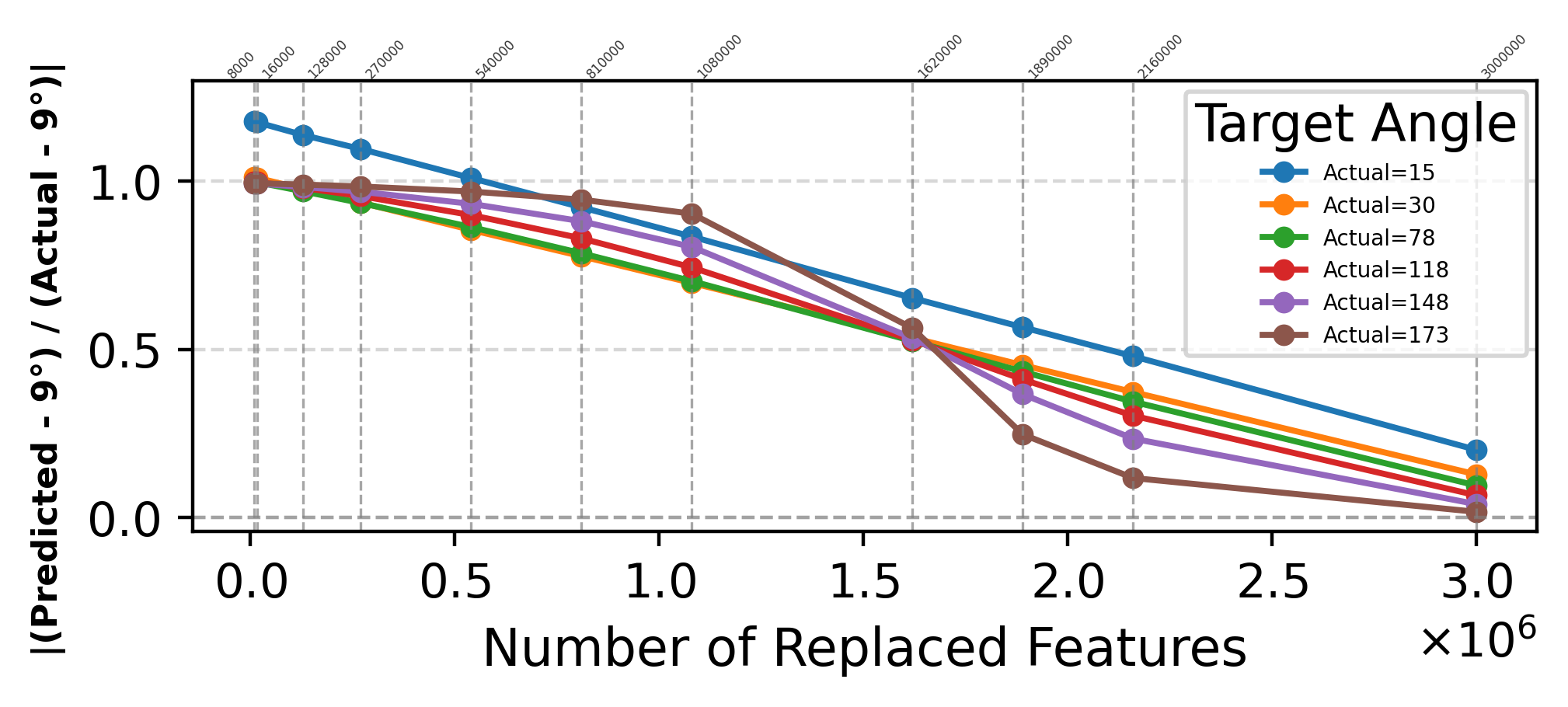}
        \caption{Picked Randomly}
    \end{subfigure}
   \vspace{-2mm}
    \caption{Incremental feature substitution for LLaVA-OneVision on images with the dog scene. On the y axis, when y = 1, predicted value matches the target orientation and when y = 0, predicted value matches the anchor orientation. No matter how the features are selected (according to the magnitude of the weights in the regressor or the absolute difference between anchor and target feature values, or randomly). 540,000 features or more must be replaced to fool the predictor. (Note that the x-axis is the number of feature substitutions times $10^6$.) This implies the orientation information is highly diffuse.}
   \vspace{-2mm}
    \label{fig:patch_analysis_llava-ov_dog}
\end{figure*}

\begin{figure*}[h!]
    \begin{subfigure}{0.33\textwidth}
        \includegraphics[width=\linewidth]{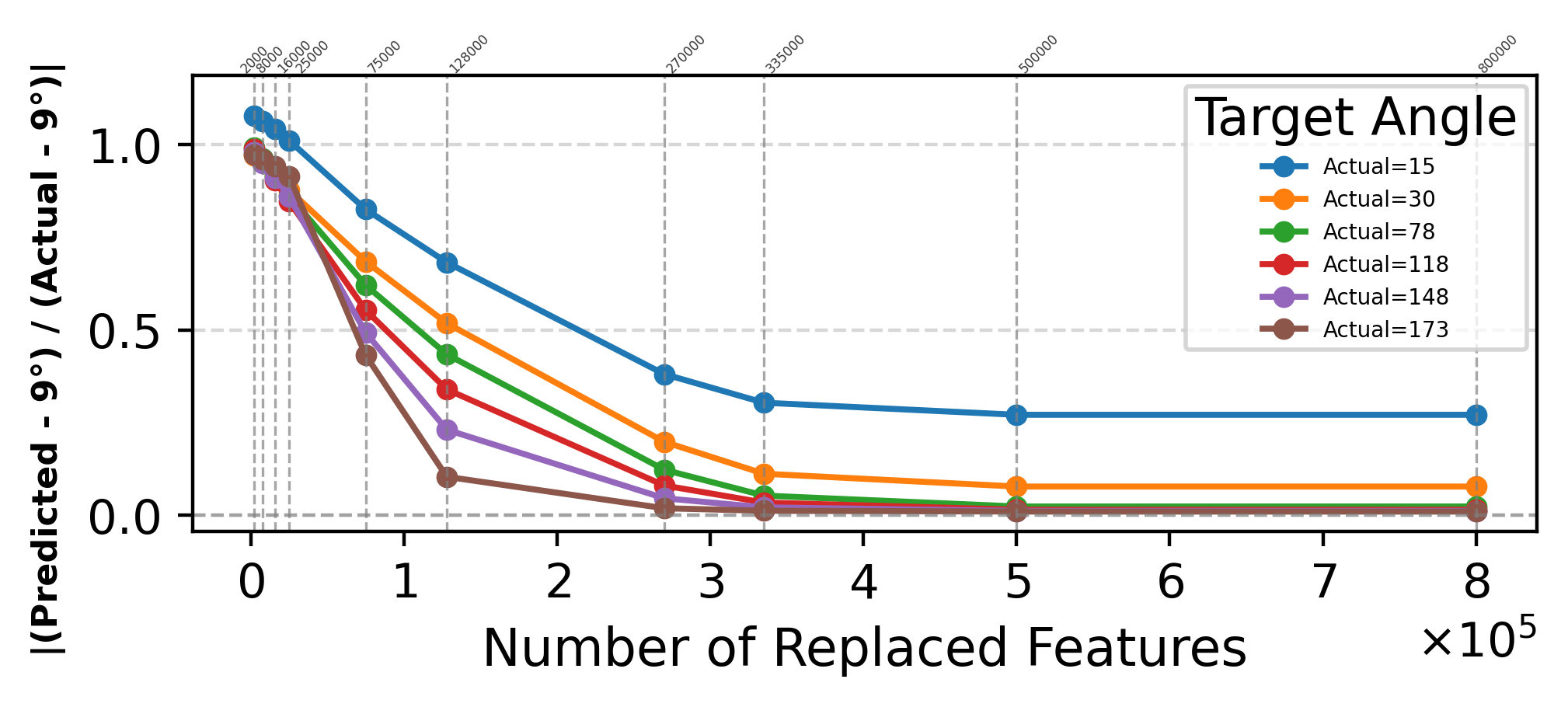}
        \caption{Ordered By Model Weight}
    \end{subfigure}%
    \hspace{-0.5em} 
    \begin{subfigure}{0.33\textwidth}
        \includegraphics[width=\linewidth]{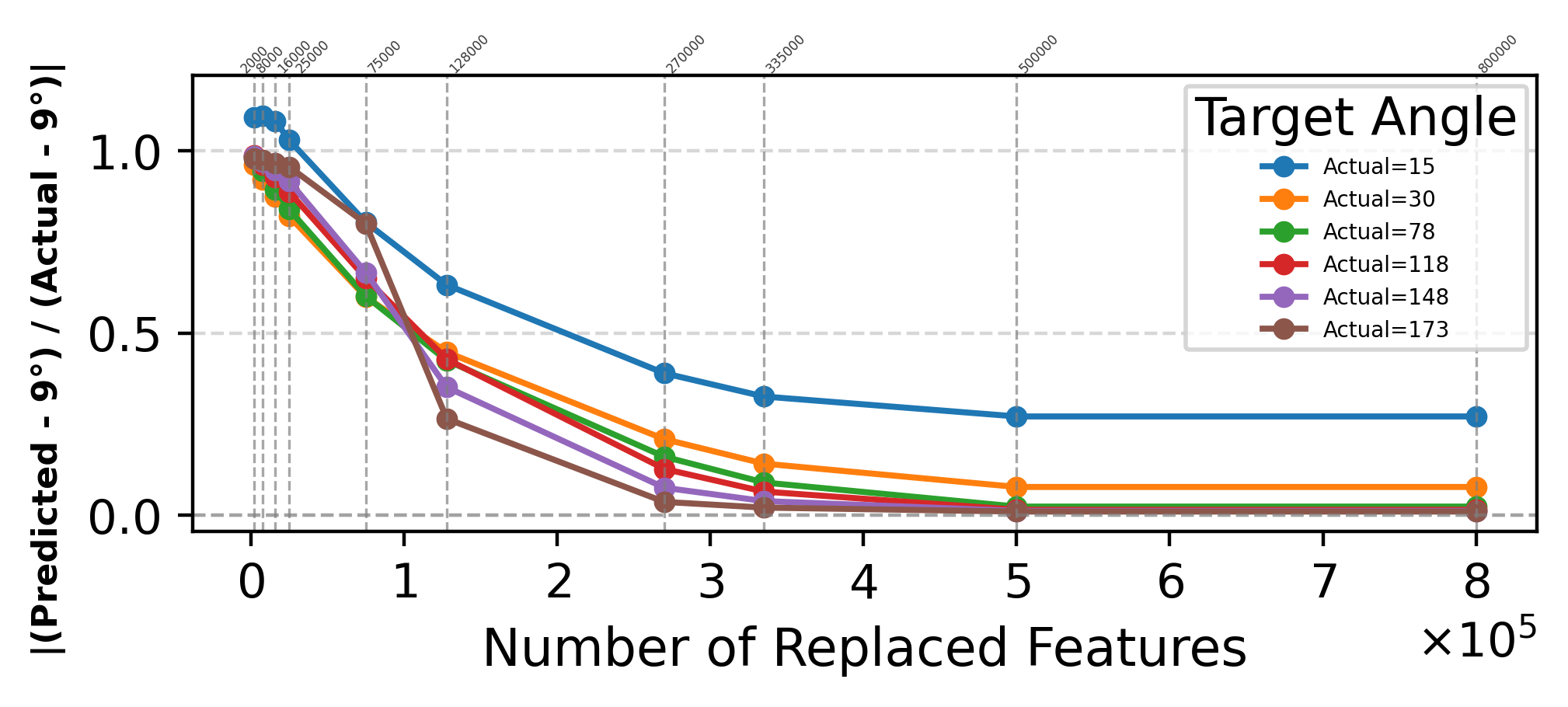}
        \caption{Ordered By Value Difference}
    \end{subfigure}%
    \hspace{-0.5em} 
    \begin{subfigure}{0.33\textwidth}
        \includegraphics[width=\linewidth]{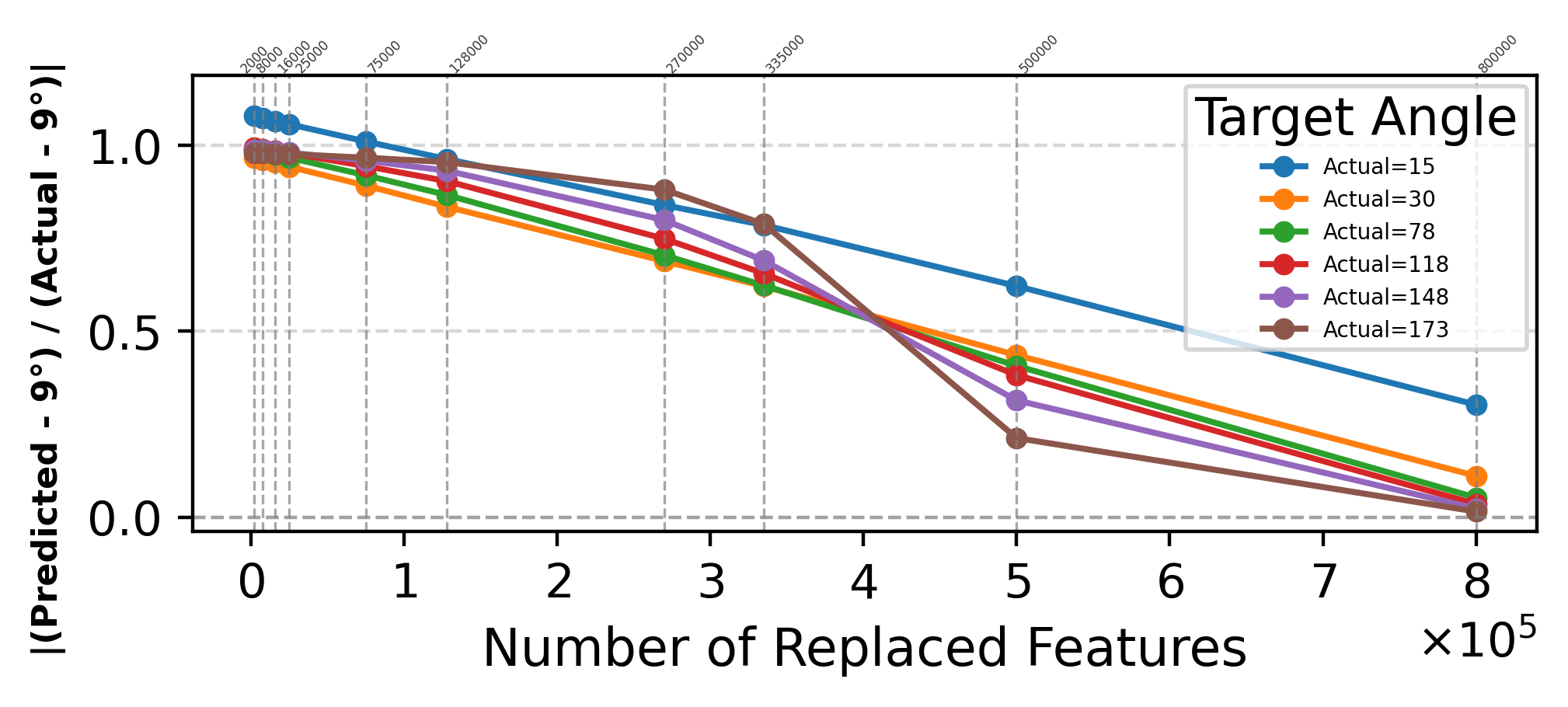}
        \caption{Picked Randomly}
    \end{subfigure}
   \vspace{-2mm}
    \caption{Incremental feature substitution for Qwen2.5-VL-7B-Instruct on images with the dog scene. On the y axis, when y = 1, predicted value matches the target orientation and when y = 0, predicted value matches the anchor orientation. No matter how the features are selected (according to the magnitude of the weights in the regressor or the absolute difference between anchor and target feature values, or randomly). 128,000 features or more must be replaced to fool the predictor. (Note that the x-axis is the number of feature substitutions times $10^5$.) This implies the orientation information is highly diffuse.}
   \vspace{-2mm}
    \label{fig:patch_analysis_qwen_dog}
\end{figure*}

Table~\ref{tab:llavaOV_qwen_metrics} shows
the query responses when estimating the 2D orientations of rotated images from the visual embeddings computed by LLaVA-OneVision and Qwen2.5-VL-7B-Instruct. Results for LLaVA 1.5 and 1.6 are given in Table \ref{tab:llama-llava-72} in the supplementary.

Since orientation information is encoded in the embedding vector and we are able to predict it accurately (Table \ref{tab:llava_qwen_comparison}), we next try to understand why the LLaVA-LLaMA models and Qwen2.5-VL-7B-Instruct perform so poorly on orientation tasks. We perform feature substitution to determine how many features are used when encoding orientation and show that the orientation information is spread diffusely across thousands of features. This may be one reason LLAVA-LLaMA and Qwen2.5-VL-7B-Instruct are unable to exploit it. We also discovered that estimates of foreground rotation depend on the background being in its standard orientation as detailed in Section \ref{app:bg-fg} in the supplementary. Any significant rotation of the background causes the foreground estimate to become nearly random.

\paragraph{Feature Substitution Experiments}
We selected an anchor (in this case, the vision encoder embedding for the image where the foreground patch is rotated $9^{\circ}$ clockwise) and replace values from the anchor's embedding vector into the embeddings for other images. The goal is to see how many embedding values have to be replaced in a non-anchor image in order for the linear predictor to believe the foreground is at $9^{\circ}$. We perform this experiment in 3 modes:
\begin{itemize}
    \item Select the {\em n} embedding features with the highest weights in the linear ridge regressor, i.e., the features the predictor is more reliant on. 
    \item Select the {\em n} embedding features with the highest absolute difference between the anchor and the target vectors, i.e. the features that change the most from anchor to target.
    \item Select {\em n} embedding features randomly (as a control).
    \vspace{0.001cm}
\end{itemize}
In all 3 modes, n is varied incrementally and patch values of the target embedding are replaced with the patch values from the corresponding locations of the anchor. The results for LLaVA-OneVision and Qwen2.5-VL-7B-Instruct for the images with the dog scene are shown in Figures \ref{fig:patch_analysis_llava-ov_dog} and \ref{fig:patch_analysis_qwen_dog} respectively. Plots for the other images are presented in the supplementary material in sections \ref{app:ftr-subs-llavaOV-qwen} and \ref{app:ftr-subs-llava1.5-1.6}. The x-axis shows the number of features from the anchor's embeddings (rotated 9$^\circ$) that were substituted into the target embedding. The y-axis shows the ratio of the prediction's deviation from the anchor to the actual value's deviation from the anchor. In other words, the y axis is 1 when the predicted value matches the target orientation and 0 when it matches the anchor orientation.  The more features we substitute from the anchor's embedding into the target embeddings, the closer the predicted orientation is to 9$^\circ$. As expected, if mode "random selection" is considered the baseline, we see that modes "ordered by model weight" and "ordered by value difference" converge faster, with the former converging the fastest.

\vspace{-2mm}
\section{Conclusion}
\label{sec:conclusion}
Many previous works have concluded that Multimodal Large Language Models (MLLMs) perform poorly on questions about orientations of objects~\cite{nichols2025right,tong2024eyes,yang2025thinking,niu2025rotbench,wu2025spatial,wang2025spatialviz,zhu2025internvl3,chenspatial,lian2025euclid}. Some works hypothesize that it is due to the vision encoders being pre-trained on CLIP-like models (Section \ref{sub:relWork_orienEstMLLM}). This work tested and rejected this hypothesis, at least for 3 versions of LLaVA-LLaMA (OneVision, 1.5 and 1.6) and Qwen2.5-VL-7B-Instruct models. It shows that the vision encoder (CLIP/SigLIP/ViT) embeddings of these models encode the orientation of foreground patches in familiar images to within $\pm 3^\circ$. This rejects the hypothesis, but doesn't explain why LLaVA-LLaMA and Qwen2.5-VL-7B-Instruct struggle with orientation tasks.  This work can not provide a definitive causal explanation, but it does note some interesting properties of CLIP/SigLIP/ViT embeddings that may make it difficult for LLaVA-LLaMA and Qwen2.5-VL-7B-Instruct to fully exploit them. Using feature substitution, we show that orientation information is distributed across tens of thousands of features, which may make it hard for LLaVA-LLaMA and Qwen2.5-VL-7B-Instruct to learn to recover rotations. We also show that foreground orientations are sensitive to canonical background orientations, which may make them unreliable in some circumstances.

It is possible that other MLLMs may behave differently. We note, however, the orientation issues have been documented in over two dozen MLLMs (including LLaVA-LLaMA), and that many MLLMs use some version of the CLIP/SigLIP/ViT encoders.
In future work, we will expand on the preliminary analysis, which showed that orientation information is diffusely encoded, which might be the reason for the poor performance of the language model.

\newpage

{
    \small
    \bibliographystyle{ieeenat_fullname}
    \bibliography{main}

@String(IJCV = {Int. J. Comput. Vis.})

@String(IJCV  = {IJCV})

@inproceedings{yang2025thinking,
  title={Thinking in space: How multimodal large language models see, remember, and recall spaces},
  author={Yang, Jihan and Yang, Shusheng and Gupta, Anjali W and Han, Rilyn and Fei-Fei, Li and Xie, Saining},
  booktitle={Proceedings of the Computer Vision and Pattern Recognition Conference},
  pages={10632--10643},
  year={2025}
}

@article{wu2025spatial,
  title={Spatial-mllm: Boosting mllm capabilities in visual-based spatial intelligence},
  author={Wu, Diankun and Liu, Fangfu and Hung, Yi-Hsin and Duan, Yueqi},
  journal={arXiv preprint arXiv:2505.23747},
  year={2025}
}

@inproceedings{jung2025isright,
  title={IsRight'Right? Enhancing Object Orientation Understanding in Multimodal Large Language Models through Egocentric Instruction Tuning},
  author={Jung, Ji Hyeok and Kim, Eun Tae and Kim, Seoyeon and Lee, Joo Ho and Kim, Bumsoo and Chang, Buru},
  booktitle={Proceedings of the Computer Vision and Pattern Recognition Conference},
  pages={14257--14267},
  year={2025}
}

@article{nichols2025right,
  title={Right side up? disentangling orientation understanding in mllms with fine-grained multi-axis perception tasks},
  author={Nichols, Keanu and Tasnim, Nazia and Yan, Yuting and Ikechukwu, Nicholas and Zou, Elva and Ghadiyaram, Deepti and Plummer, Bryan A},
  journal={arXiv preprint arXiv:2505.21649},
  year={2025}
}

@article{qu2025spatialvla,
  title={Spatialvla: Exploring spatial representations for visual-language-action model},
  author={Qu, Delin and Song, Haoming and Chen, Qizhi and Yao, Yuanqi and Ye, Xinyi and Ding, Yan and Wang, Zhigang and Gu, JiaYuan and Zhao, Bin and Wang, Dong and others},
  journal={arXiv preprint arXiv:2501.15830},
  year={2025}
}

@article{zhang2025embodied,
  title={Embodied-reasoner: Synergizing visual search, reasoning, and action for embodied interactive tasks},
  author={Zhang, Wenqi and Wang, Mengna and Liu, Gangao and Huixin, Xu and Jiang, Yiwei and Shen, Yongliang and Hou, Guiyang and Zheng, Zhe and Zhang, Hang and Li, Xin and others},
  journal={arXiv preprint arXiv:2503.21696},
  year={2025}
}

@article{yuan2025depthvla,
  title={DepthVLA: Enhancing Vision-Language-Action Models with Depth-Aware Spatial Reasoning},
  author={Yuan, Tianyuan and Liu, Yicheng and Lu, Chenhao and Chen, Zhuoguang and Jiang, Tao and Zhao, Hang},
  journal={arXiv preprint arXiv:2510.13375},
  year={2025}
}

@inproceedings{liu2024improved,
  title={Improved baselines with visual instruction tuning},
  author={Liu, Haotian and Li, Chunyuan and Li, Yuheng and Lee, Yong Jae},
  booktitle={Proceedings of the IEEE/CVF conference on computer vision and pattern recognition},
  pages={26296--26306},
  year={2024}
}

@inproceedings{bigverdi2025perception,
  title={Perception tokens enhance visual reasoning in multimodal language models},
  author={Bigverdi, Mahtab and Luo, Zelun and Hsieh, Cheng-Yu and Shen, Ethan and Chen, Dongping and Shapiro, Linda G and Krishna, Ranjay},
  booktitle={Proceedings of the Computer Vision and Pattern Recognition Conference},
  pages={3836--3845},
  year={2025}
}

@inproceedings{sun2017orientation,
  title={Orientation estimation network},
  author={Sun, Jie and Zhou, Wengang and Li, Houqiang},
  booktitle={International Conference on Image and Graphics},
  pages={151--162},
  year={2017},
  organization={Springer}
}

@inproceedings{lu2022oskdet,
  title={OSKDet: Orientation-sensitive keypoint localization for rotated object detection},
  author={Lu, Dongchen and Li, Dongmei and Li, Yali and Wang, Shengjin},
  booktitle={Proceedings of the IEEE/CVF Conference on Computer Vision and Pattern Recognition},
  pages={1182--1192},
  year={2022}
}

@inproceedings{fischer2015image,
  title={Image orientation estimation with convolutional networks},
  author={Fischer, Philipp and Dosovitskiy, Alexey and Brox, Thomas},
  booktitle={German conference on pattern recognition},
  pages={368--378},
  year={2015},
  organization={Springer}
}

@article{mishra2019descriptive,
  title={Descriptive statistics and normality tests for statistical data},
  author={Mishra, Prabhaker and Pandey, Chandra M and Singh, Uttam and Gupta, Anshul and Sahu, Chinmoy and Keshri, Amit},
  journal={Annals of cardiac anaesthesia},
  volume={22},
  number={1},
  pages={67--72},
  year={2019},
  publisher={Medknow}
}

@article{habibzadeh2024data,
  title={Data distribution: normal or abnormal?},
  author={Habibzadeh, Farrokh},
  journal={Journal of Korean medical science},
  volume={39},
  number={3},
  year={2024},
  publisher={The Korean Academy of Medical Sciences}
}

@article{wang2025spatialviz,
  title={SpatialViz-Bench: An MLLM Benchmark for Spatial Visualization},
  author={Wang, Siting and Pei, Minnan and Sun, Luoyang and Deng, Cheng and Shao, Kun and Tian, Zheng and Zhang, Haifeng and Wang, Jun},
  journal={arXiv preprint arXiv:2507.07610},
  year={2025}
}

@misc{liu2024llavanext,
    title={LLaVA-NeXT: Improved reasoning, OCR, and world knowledge},
    url={https://llava-vl.github.io/blog/2024-01-30-llava-next/},
    author={Liu, Haotian and Li, Chunyuan and Li, Yuheng and Li, Bo and Zhang, Yuanhan and Shen, Sheng and Lee, Yong Jae},
    month={January},
    year={2024}
}

@article{zhu2025internvl3,
  title={Internvl3: Exploring advanced training and test-time recipes for open-source multimodal models},
  author={Zhu, Jinguo and Wang, Weiyun and Chen, Zhe and Liu, Zhaoyang and Ye, Shenglong and Gu, Lixin and Tian, Hao and Duan, Yuchen and Su, Weijie and Shao, Jie and others},
  journal={arXiv preprint arXiv:2504.10479},
  year={2025}
}

@article{kamoi2024visonlyqa,
  title={Visonlyqa: Large vision language models still struggle with visual perception of geometric information},
  author={Kamoi, Ryo and Zhang, Yusen and Das, Sarkar Snigdha Sarathi and Zhang, Ranran Haoran and Zhang, Rui},
  journal={arXiv preprint arXiv:2412.00947},
  year={2024}
}

@article{zhang2025mllms,
  title={Why do mllms struggle with spatial understanding? a systematic analysis from data to architecture},
  author={Zhang, Wanyue and Huang, Yibin and Xu, Yangbin and Huang, JingJing and Zhi, Helu and Ren, Shuo and Xu, Wang and Zhang, Jiajun},
  journal={arXiv preprint arXiv:2509.02359},
  year={2025}
}

@inproceedings{chenspatial,
  title={Why Is Spatial Reasoning Hard for VLMs? An Attention Mechanism Perspective on Focus Areas},
  author={Chen, Shiqi and Zhu, Tongyao and Zhou, Ruochen and Zhang, Jinghan and Gao, Siyang and Niebles, Juan Carlos and Geva, Mor and He, Junxian and Wu, Jiajun and Li, Manling},
  booktitle={Forty-second International Conference on Machine Learning},
year={2025}
}

@article{niu2025rotbench,
  title={RotBench: Evaluating Multimodal Large Language Models on Identifying Image Rotation},
  author={Niu, Tianyi and Cho, Jaemin and Stengel-Eskin, Elias and Bansal, Mohit},
  journal={arXiv preprint arXiv:2508.13968},
  year={2025}
}

@inproceedings{shiri2024empirical,
  title={An empirical analysis on spatial reasoning capabilities of large multimodal models},
  author={Shiri, Fatemeh and Guo, Xiao-Yu and Far, Mona and Yu, Xin and Haf, Reza and Li, Yuan-Fang},
  booktitle={Proceedings of the 2024 Conference on Empirical Methods in Natural Language Processing},
  pages={21440--21455},
  year={2024}
}

@inproceedings{tong2024eyes,
  title={Eyes wide shut? exploring the visual shortcomings of multimodal llms},
  author={Tong, Shengbang and Liu, Zhuang and Zhai, Yuexiang and Ma, Yi and LeCun, Yann and Xie, Saining},
  booktitle={Proceedings of the IEEE/CVF Conference on Computer Vision and Pattern Recognition},
  pages={9568--9578},
  year={2024}
}

@inproceedings{huynh2025vision,
  title={Vision-Language Models Can't See the Obvious},
  author={Huynh, Ngoc Dung and Yasser Dahou and Le-Khac, Phuc H and Para, Wamiq Reyaz and Singh, Ankit and Narayan, Sanath},
  booktitle={Proceedings of the IEEE/CVF International Conference on Computer Vision},
  pages={24159--24169},
  year={2025}
}

@article{lian2025euclid,
  title={Euclid's Gift: Enhancing Spatial Perception and Reasoning in Vision-Language Models via Geometric Surrogate Tasks},
  author={Lian, Shijie and Wu, Changti and Yang, Laurence Tianruo and Yuan, Hang and Yu, Bin and Zhang, Lei and Chen, Kai},
  journal={arXiv preprint arXiv:2509.24473},
  year={2025}
}

@article{ILSVRC15,
Author = {Olga Russakovsky and Jia Deng and Hao Su and Jonathan Krause and Sanjeev Satheesh and Sean Ma and Zhiheng Huang and Andrej Karpathy and Aditya Khosla and Michael Bernstein and Alexander C. Berg and Li Fei-Fei},
Title = {{ImageNet Large Scale Visual Recognition Challenge}},
Year = {2015},
journal   = {International Journal of Computer Vision (IJCV)},
doi = {10.1007/s11263-015-0816-y},
volume={115},
number={3},
pages={211-252}
}

@article{massey1951kolmogorov,
  title={The Kolmogorov-Smirnov test for goodness of fit},
  author={Massey Jr, Frank J},
  journal={Journal of the American statistical Association},
  volume={46},
  number={253},
  pages={68--78},
  year={1951},
  publisher={Taylor \& Francis}
}

@article{bartoszcze2025representation,
  title={Representation Engineering for Large-Language Models: Survey and Research Challenges},
  author={Bartoszcze, Lukasz and Munshi, Sarthak and Sukidi, Bryan and Yen, Jennifer and Yang, Zejia and Williams-King, David and Le, Linh and Asuzu, Kosi and Maple, Carsten},
  journal={arXiv preprint arXiv:2502.17601},
  year={2025}
}

@article{tian2025representation,
  title={Why Representation Engineering Works: A Theoretical and Empirical Study in Vision-Language Models},
  author={Tian, Bowei and Lyu, Xuntao and Liu, Meng and Wang, Hongyi and Li, Ang},
  journal={arXiv preprint arXiv:2503.22720},
  year={2025}
}

@article{zhang2018visual,
  title={Visual interpretability for deep learning: a survey},
  author={Zhang, Quan-shi and Zhu, Song-Chun},
  journal={Frontiers of Information Technology \& Electronic Engineering},
  volume={19},
  number={1},
  pages={27--39},
  year={2018},
  publisher={Springer}
}

@article{yung2021si,
  title={Si-score: An image dataset for fine-grained analysis of robustness to object location, rotation and size},
  author={Yung, Jessica and Romijnders, Rob and Kolesnikov, Alexander and Beyer, Lucas and Djolonga, Josip and Houlsby, Neil and Gelly, Sylvain and Lucic, Mario and Zhai, Xiaohua},
  journal={arXiv preprint arXiv:2104.04191},
  year={2021}
}

@article{li2024llava,
  title={Llava-onevision: Easy visual task transfer},
  author={Li, Bo and Zhang, Yuanhan and Guo, Dong and Zhang, Renrui and Li, Feng and Zhang, Hao and Zhang, Kaichen and Zhang, Peiyuan and Li, Yanwei and Liu, Ziwei and others},
  journal={arXiv preprint arXiv:2408.03326},
  year={2024}
}

@inproceedings{zhai2023sigmoid,
  title={Sigmoid loss for language image pre-training},
  author={Zhai, Xiaohua and Mustafa, Basil and Kolesnikov, Alexander and Beyer, Lucas},
  booktitle={Proceedings of the IEEE/CVF international conference on computer vision},
  pages={11975--11986},
  year={2023}
}

@article{bai2025qwen2,
  title={Qwen2. 5-VL Technical Report},
  author={Bai, Shuai and Chen, Keqin and Liu, Xuejing and Wang, Jialin and Ge, Wenbin and Song, Sibo and Dang, Kai and Wang, Peng and Wang, Shijie and Tang, Jun and others},
  journal={arXiv preprint arXiv:2502.13923},
  year={2025}
}

@inproceedings{ghaffari2024large,
  title={Large language models are challenged by habitat-centered reasoning},
  author={Ghaffari, Sadaf and Krishnaswamy, Nikhil},
  booktitle={Findings of the Association for Computational Linguistics: EMNLP 2024},
  pages={13047--13059},
  year={2024}
}

@article{zheng2025multimodal,
  title={Multimodal spatial reasoning in the large model era: A survey and benchmarks},
  author={Zheng, Xu and Dongfang, Zihao and Jiang, Lutao and Zheng, Boyuan and Guo, Yulong and Zhang, Zhenquan and Albanese, Giuliano and Yang, Runyi and Ma, Mengjiao and Zhang, Zixin and others},
  journal={arXiv preprint arXiv:2510.25760},
  year={2025}
}

@inproceedings{cuidual,
  title={The Dual Mechanisms of Spatial Reasoning in Vision--Language Models},
  author={Cui, Kelly and Prakash, Nikhil and Raina, Ayush and Bau, David and Torralba, Antonio and Shaham, Tamar Rott},
  booktitle={The First Workshop on Efficient Spatial Reasoning}
}

@article{ma2026attention,
  title={Attention in Space: Functional Roles of VLM Heads for Spatial Reasoning},
  author={Ma, Xueqi and Yang, Shuo and Jiang, Yanbei and Liu, Shu and Liu, Zhenzhen and Ao, Jiayang and Ma, Xingjun and Erfani, Sarah Monazam and Bailey, James},
  journal={arXiv preprint arXiv:2603.20662},
  year={2026}
}

@article{yu2025far,
  title={How far are vlms from visual spatial intelligence? a benchmark-driven perspective},
  author={Yu, Songsong and Chen, Yuxin and Ju, Hao and Jia, Lianjie and Zhang, Fuxi and Huang, Shaofei and Wu, Yuhan and Cui, Rundi and Ran, Binghao and Zhang, Zaibin and others},
  journal={arXiv preprint arXiv:2509.18905},
  year={2025}
}

@article{yoon2025visual,
  title={Visual representation alignment for multimodal large language models},
  author={Yoon, Heeji and Jung, Jaewoo and Kim, Junwan and Choi, Hyungyu and Shin, Heeseong and Lim, Sangbeom and An, Honggyu and Kim, Chaehyun and Han, Jisang and Kim, Donghyun and others},
  journal={arXiv preprint arXiv:2509.07979},
  year={2025}
}

@article{liu2026spatial,
  title={Spatial Intelligence in Vision-Language Models: A Comprehensive Survey},
  author={Liu, Disheng and Liang, Tuo and Hu, Zhe and Peng, Jierui and Lu, Yiren and Xu, Yi and Fu, Yun and Yin, Yu},
  year={2026}
}

@article{daehyunaligning,
  title={Aligning Vision-Language Models With Human Directional Reference},
  author={Daehyun, KIM and Kim, Hyounghun}
}

@article{jia2025omnispatial,
  title={Omnispatial: Towards comprehensive spatial reasoning benchmark for vision language models},
  author={Jia, Mengdi and Qi, Zekun and Zhang, Shaochen and Zhang, Wenyao and Yu, Xinqiang and He, Jiawei and Wang, He and Yi, Li},
  journal={arXiv preprint arXiv:2506.03135},
  year={2025}
}

@article{dosovitskiy2020image,
  title={An image is worth 16x16 words: Transformers for image recognition at scale},
  author={Dosovitskiy, Alexey and Beyer, Lucas and Kolesnikov, Alexander and Weissenborn, Dirk and Zhai, Xiaohua and Unterthiner, Thomas and Dehghani, Mostafa and Minderer, Matthias and Heigold, Georg and Gelly, Sylvain and others},
  journal={arXiv preprint arXiv:2010.11929},
  year={2020}
}

@inproceedings{radford2021learning,
  title={Learning transferable visual models from natural language supervision},
  author={Radford, Alec and Kim, Jong Wook and Hallacy, Chris and Ramesh, Aditya and Goh, Gabriel and Agarwal, Sandhini and Sastry, Girish and Askell, Amanda and Mishkin, Pamela and Clark, Jack and others},
  booktitle={International conference on machine learning},
  pages={8748--8763},
  year={2021},
  organization={PmLR}
}
}

 \clearpage
\setcounter{page}{1}
\maketitlesupplementary


\section{Hypothesis: CLIP Fails to Encode Object Orientation Information}

\subsection{Collage of Images for LLaVA 1.5 and 1.6}

\begin{figure}[h!]
  \centering
   \includegraphics[width=7cm, height=2.5cm]{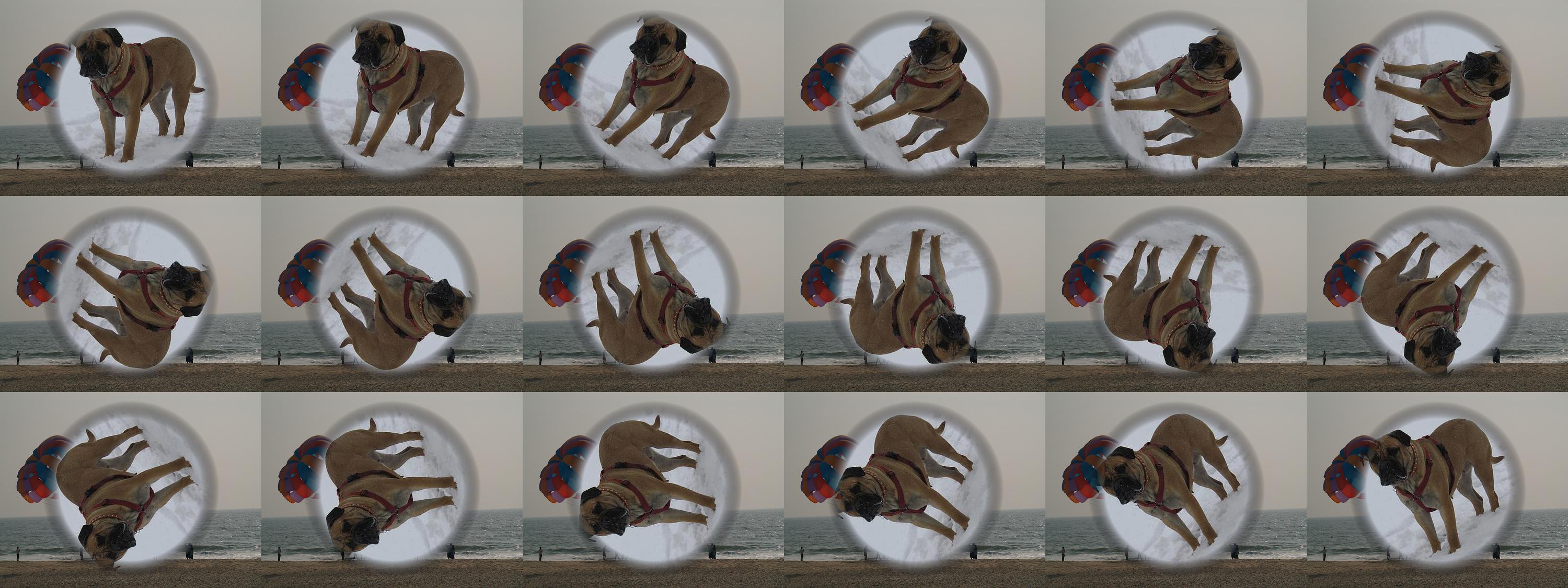}
   \caption{Collage of every 20th image from the images with the dog foreground (biggest foreground)}
   \label{fig:collage}
   \hfill
\end{figure}

\subsection{Plots showing Regression comparison between LLaVA OneVision and Qwen2.5-VL-7B-Instruct}
\label{app:reg-comp-llavaOV-qwen}
\begin{figure}[h]
  \centering
  \includegraphics[width=7.5cm, height=5cm]{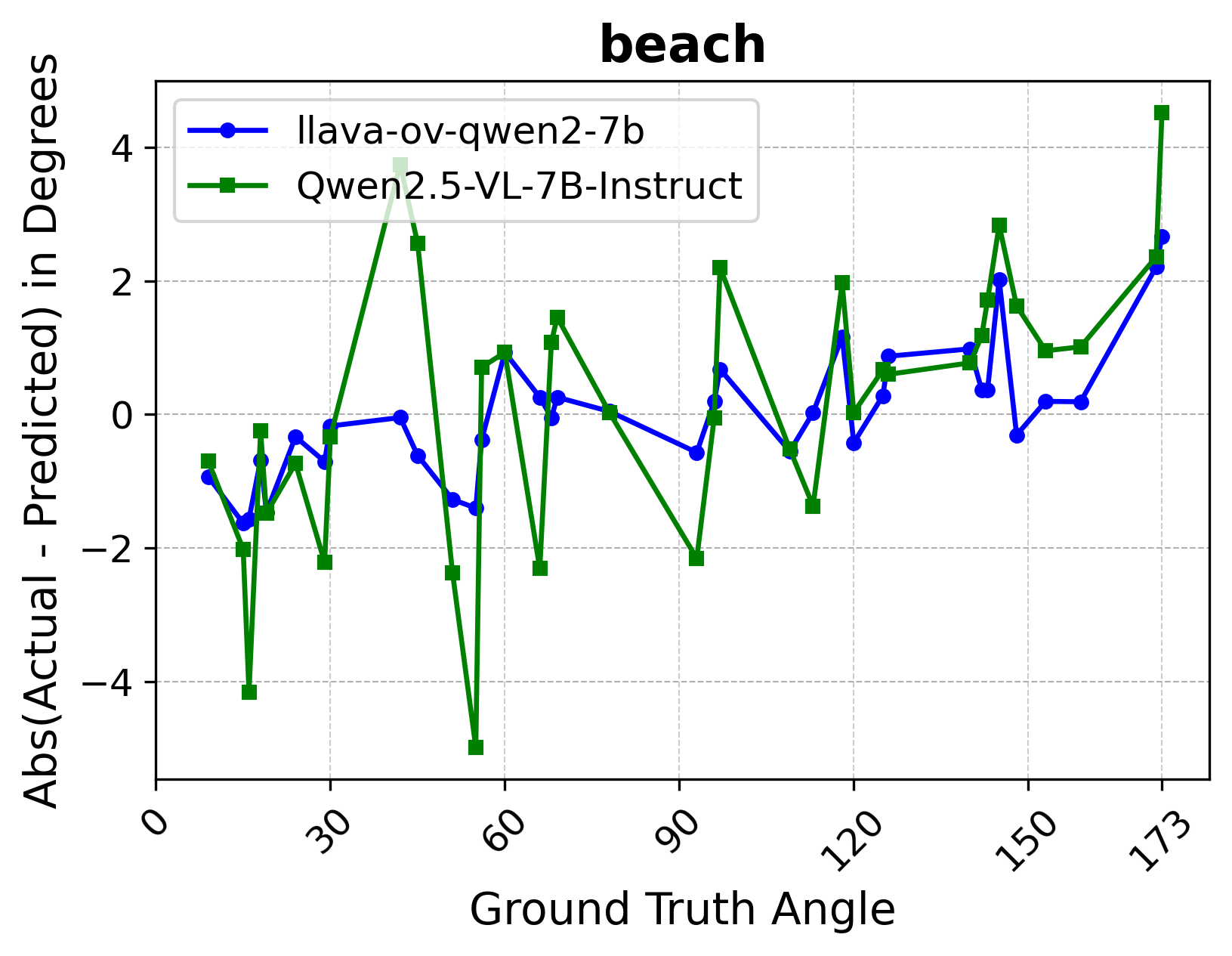}
  \vspace{-2mm}
  \caption{2D orientation estimation performance comparison between LLaVA OneVision and Qwen2.5-VL-7B-Instruct on the images with dog for 36 randomly selected images}
  \label{fig:regression_comparison_llavaOV_Qwen_dog}
\end{figure}

\begin{figure}[h]
  \centering
  \includegraphics[width=7.5cm, height=5cm]{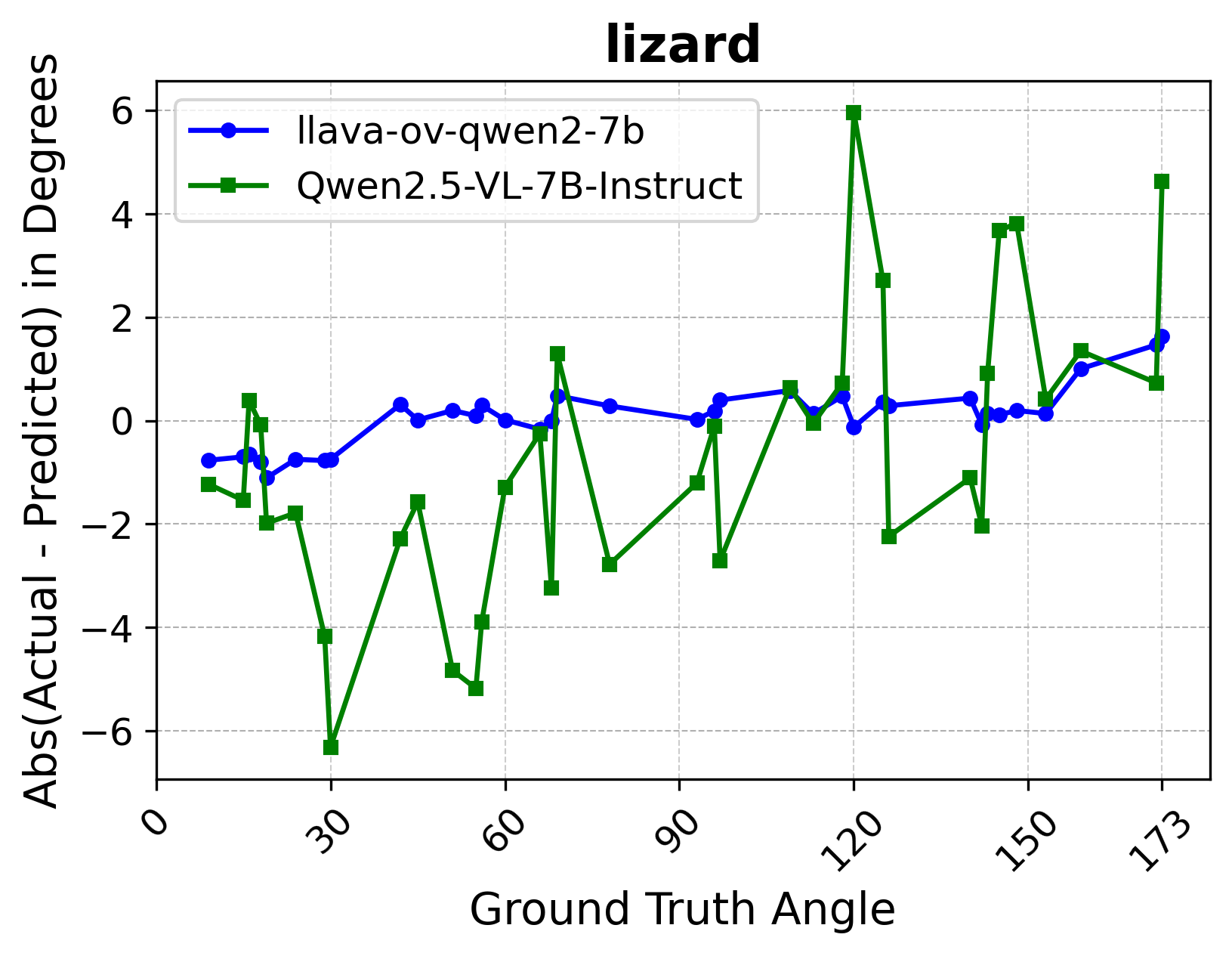}
  \vspace{-2mm}
  \caption{2D orientation estimation performance comparison between LLaVA OneVision and Qwen2.5-VL-7B-Instruct on the images with lizard for 36 randomly selected images}
  \label{fig:regression_comparison_llavaOV_Qwen_lizard}
\end{figure}

\begin{figure}[h]
  \centering
  \includegraphics[width=7.5cm, height=5cm]{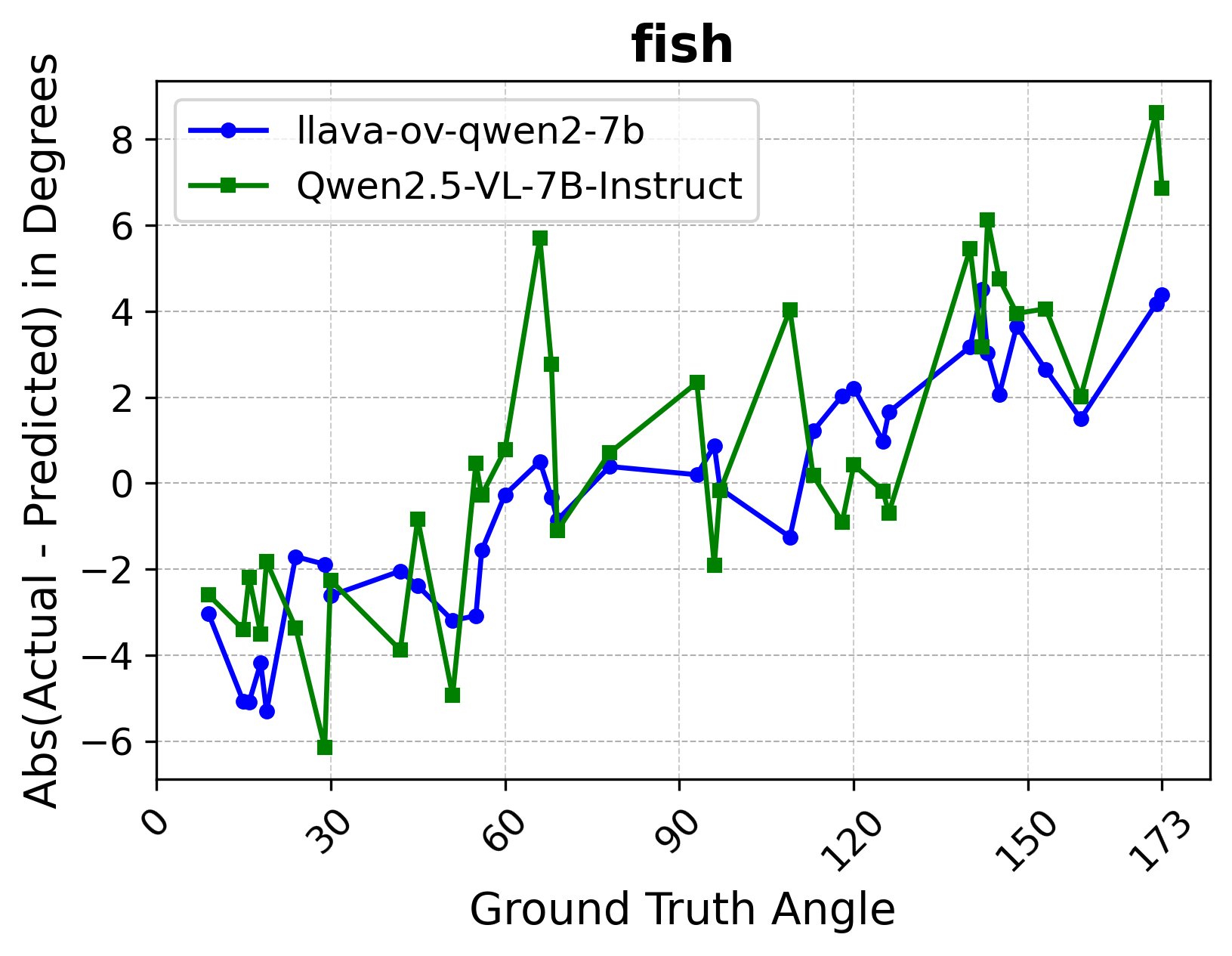}
  \vspace{-2mm}
  \caption{2D orientation estimation performance comparison between LLaVA OneVision and Qwen2.5-VL-7B-Instruct on the images with fish for 36 randomly selected images}
  \label{fig:regression_comparison_llavaOV_Qwen_fish}
\end{figure}

\begin{figure}[h]
  \centering
  \includegraphics[width=7.5cm, height=5cm]{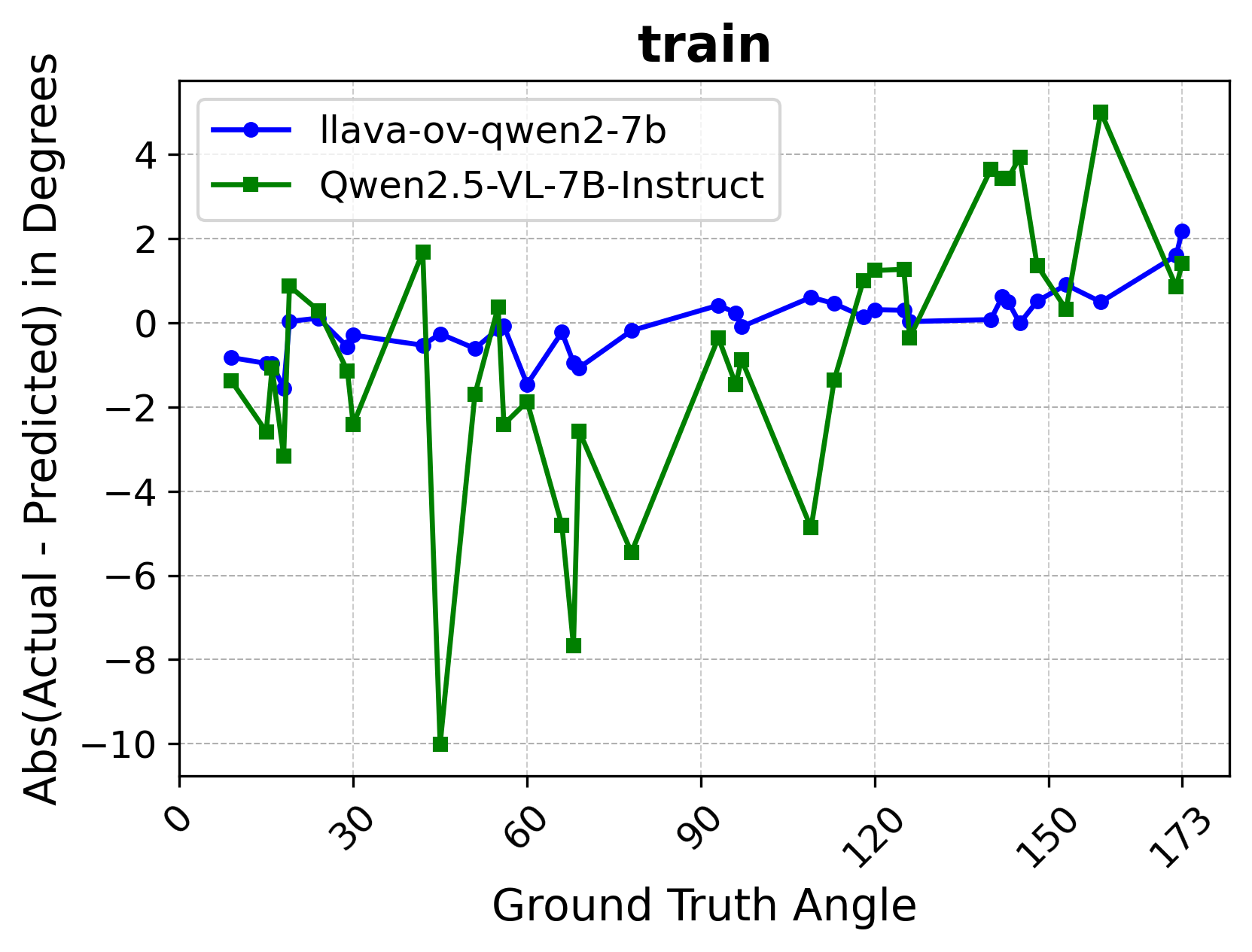}
  \vspace{-2mm}
  \caption{2D orientation estimation performance comparison between LLaVA OneVision and Qwen2.5-VL-7B-Instruct on the images with train for 36 randomly selected images}
  \label{fig:regression_comparison_llavaOV_Qwen_train}
\end{figure}

\begin{figure}[h]
  \centering
  \includegraphics[width=7.5cm, height=5cm]{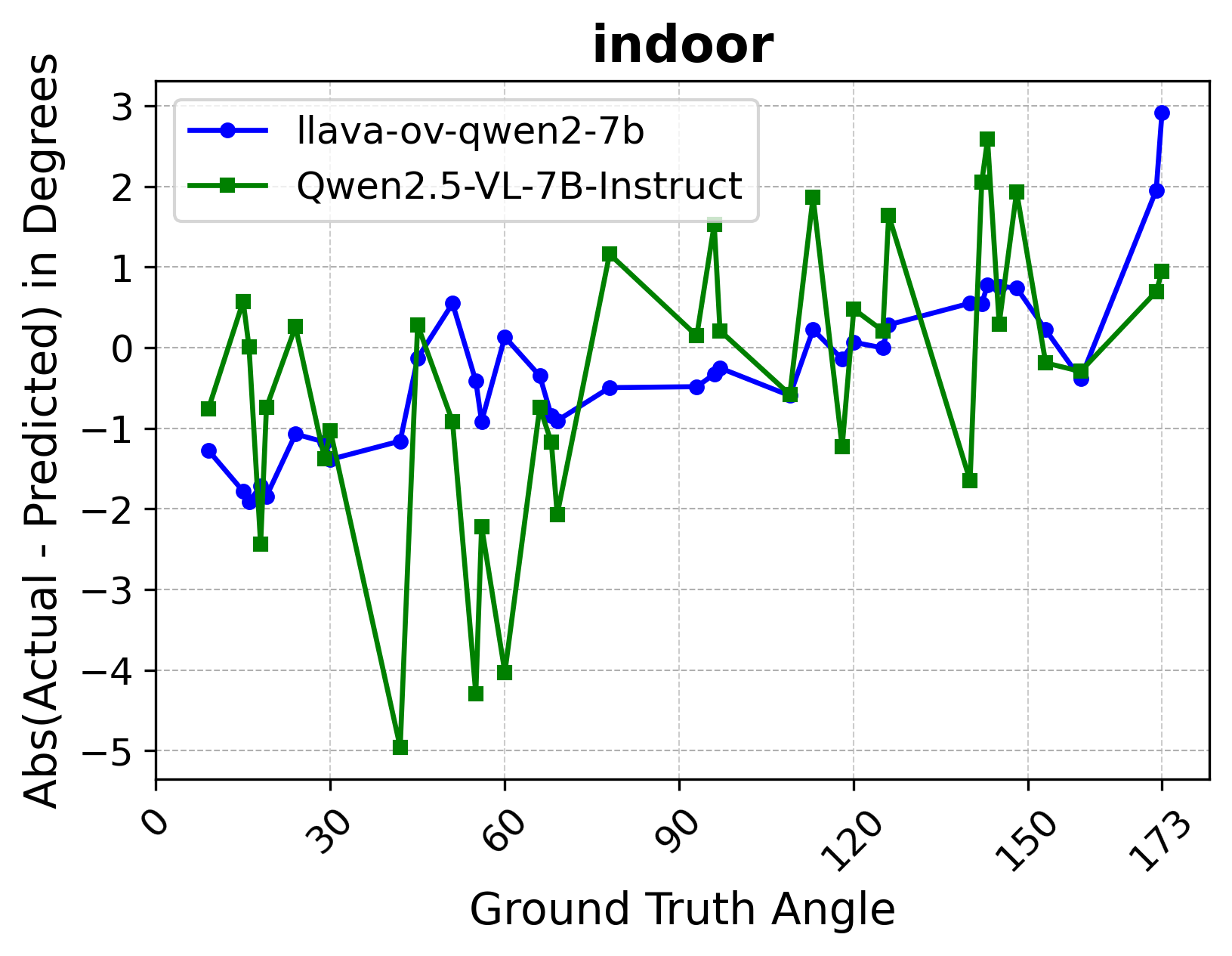}
  \vspace{-2mm}
  \caption{2D orientation estimation performance comparison between LLaVA OneVision and Qwen2.5-VL-7B-Instruct on the images with indoor for 36 randomly selected images}
  \label{fig:regression_comparison_llavaOV_Qwen_indoor}
\end{figure}

\begin{figure}[h]
  \centering
  \includegraphics[width=7.5cm, height=5cm]{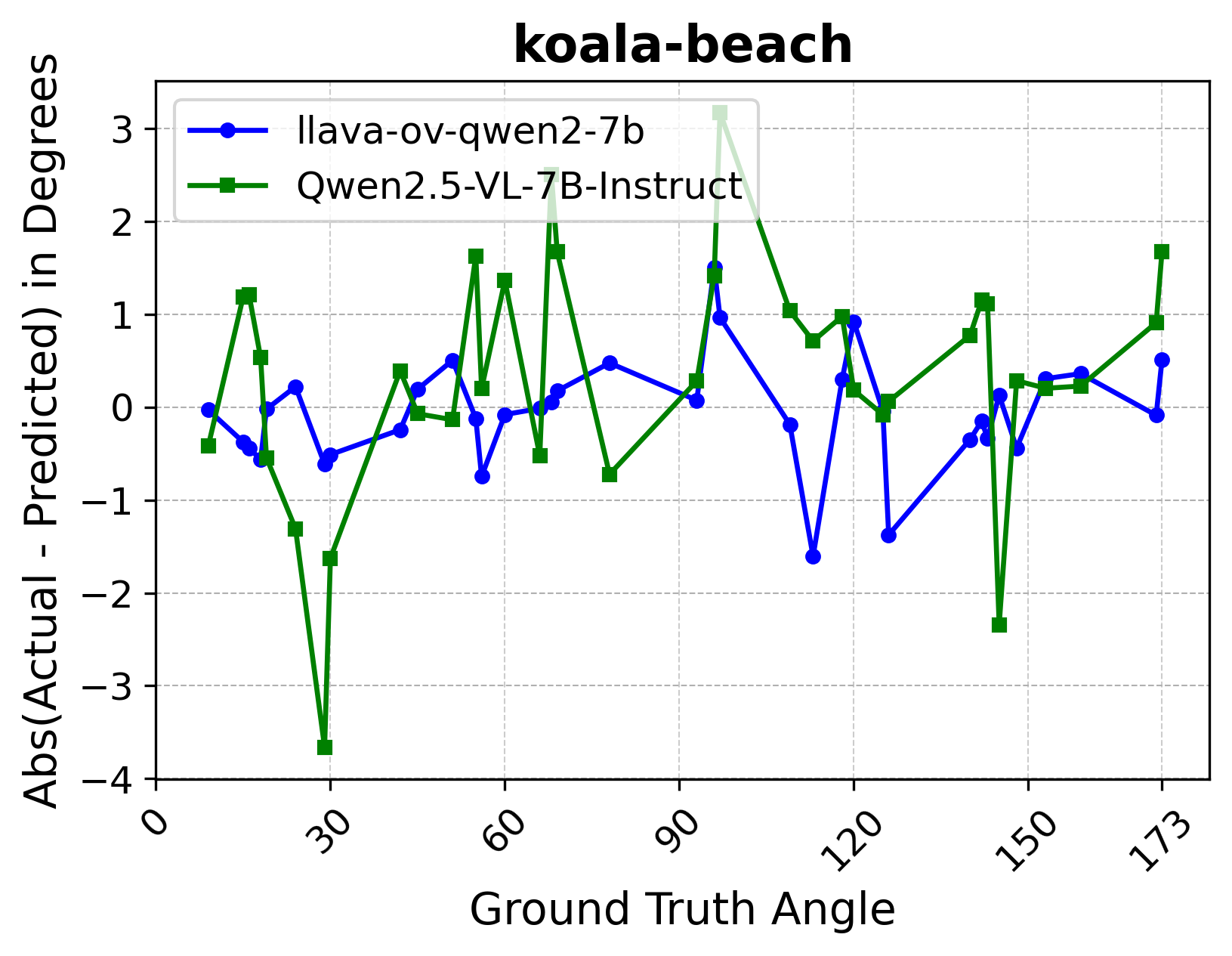}
  \vspace{-2mm}
  \caption{2D orientation estimation performance comparison between LLaVA OneVision and Qwen2.5-VL-7B-Instruct on the images with koala for 36 randomly selected images}
  \label{fig:regression_comparison_llavaOV_Qwen_koala_beach}
\end{figure}

\begin{figure}[h]
  \centering
  \includegraphics[width=7.5cm, height=5cm]{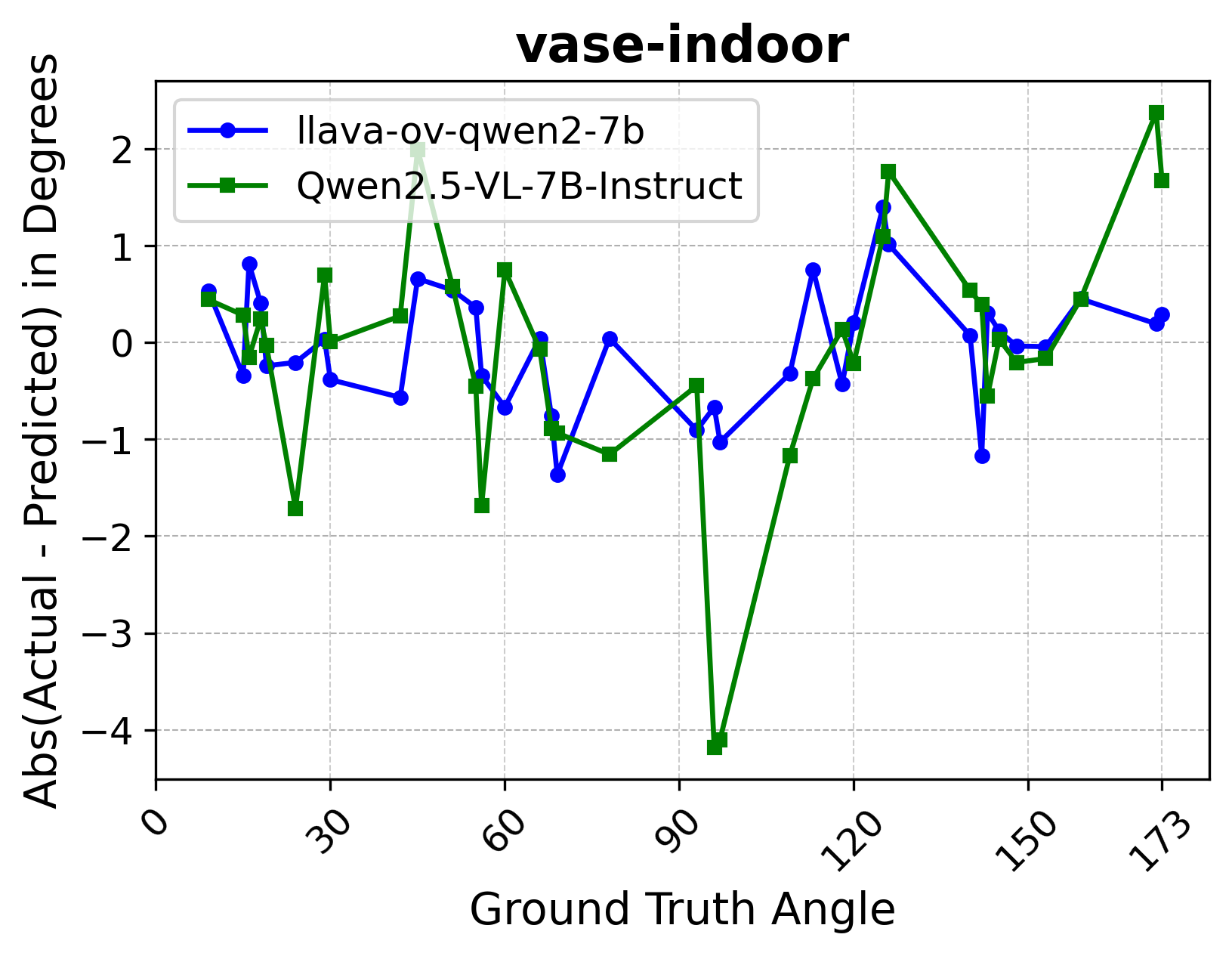}
  \vspace{-2mm}
  \caption{2D orientation estimation performance comparison between LLaVA OneVision and Qwen2.5-VL-7B-Instruct on the images with vase for 36 randomly selected images}
  \label{fig:regression_comparison_llavaOV_Qwen_vase}
\end{figure}

\begin{figure}[h]
  \centering
  \includegraphics[width=7.5cm, height=5cm]{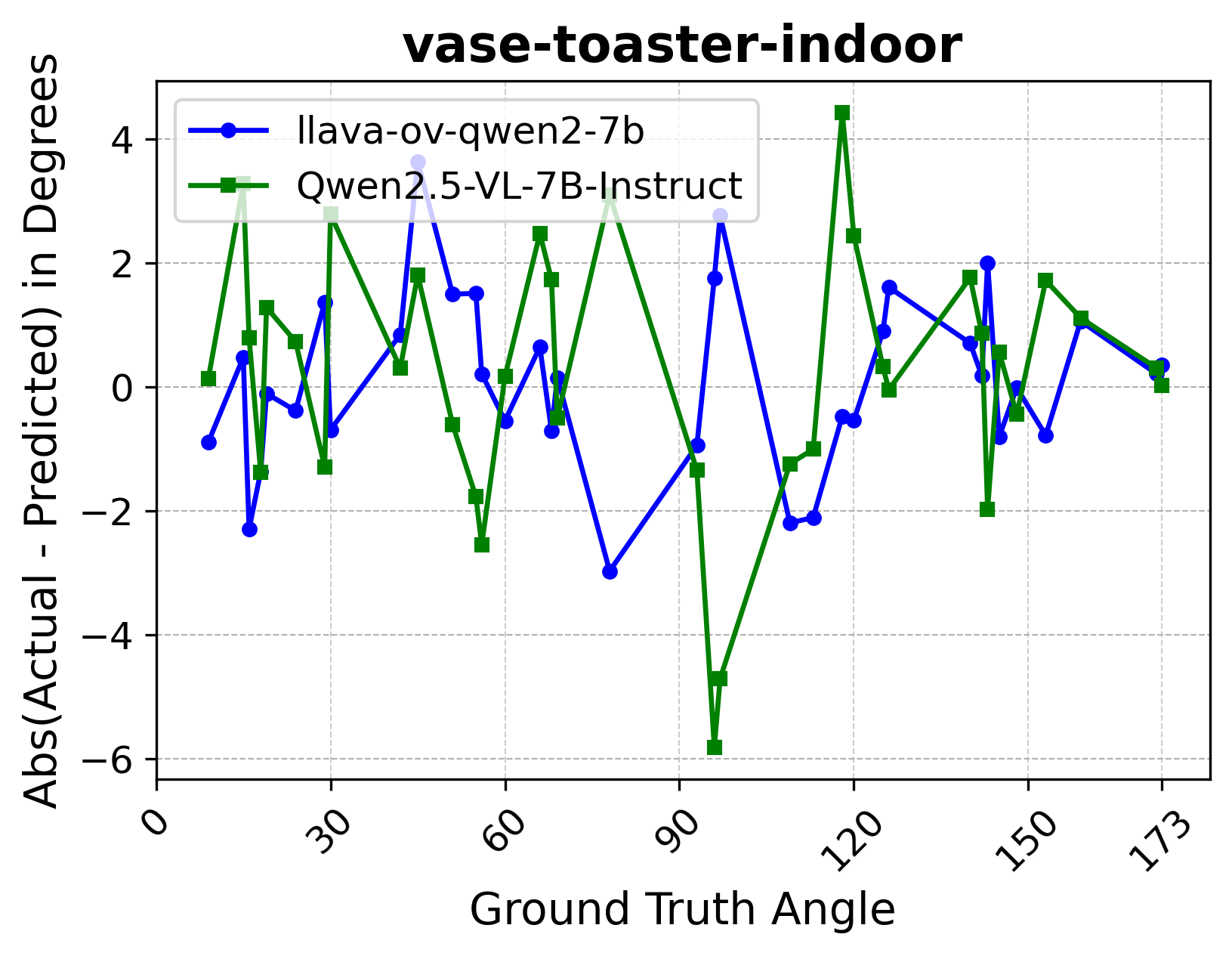}
  \vspace{-2mm}
  \caption{2D orientation estimation performance comparison between LLaVA OneVision and Qwen2.5-VL-7B-Instruct on the images with vase and toaster for 36 randomly selected images}
  \label{fig:regression_comparison_llavaOV_Qwen_vase_toaster}
\end{figure}
\clearpage

\subsection{Plots showing Regression comparison between LLaVA 1.5 and 1.6}
\label{app:reg-comp-llava1.5-1.6}
\begin{figure}[h!]
  \centering
  \includegraphics[width=7.5cm, height=5cm]{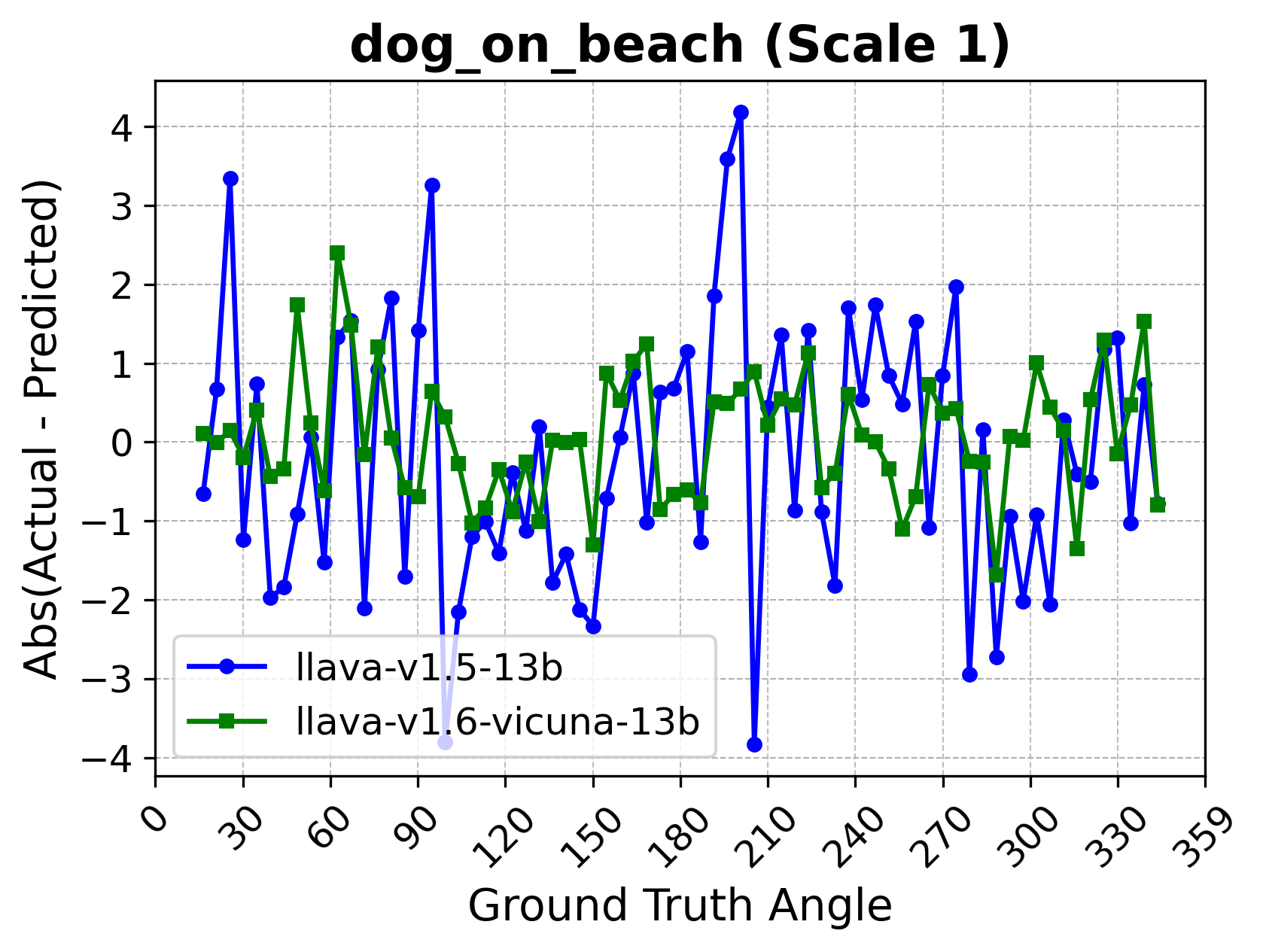}
  \vspace{-2mm}
  \caption{2D orientation estimation performance comparison between LLaVA 1.5 and 1.6 on the images with dog foregrounds (scale 1) for 72 randomly selected images}
  \label{fig:regression_comparison_llava1.5_1.6_dog-on-beach_scale1}
\end{figure}

\begin{figure}[h]
  \centering
  \includegraphics[width=7.5cm, height=5cm]{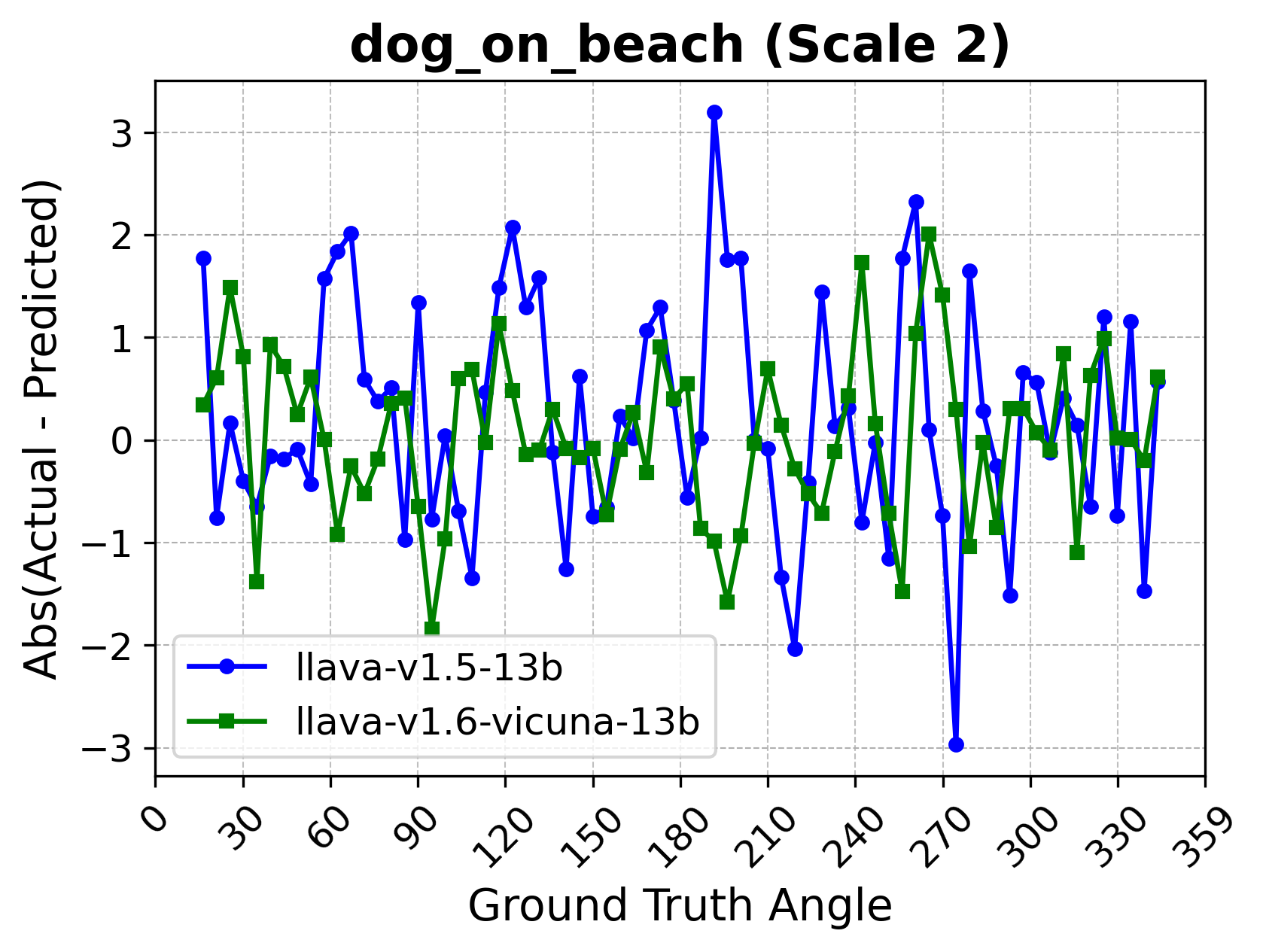}
  \vspace{-2mm}
  \caption{2D orientation estimation performance comparison between LLaVA 1.5 and 1.6 on the images with dog foregrounds (scale 2) for 72 randomly selected images}
  \label{fig:regression_comparison_llava1.5_1.6_dog-on-beach_scale2}
\end{figure}

\begin{figure}[h]
  \centering
  \includegraphics[width=7.5cm, height=5cm]{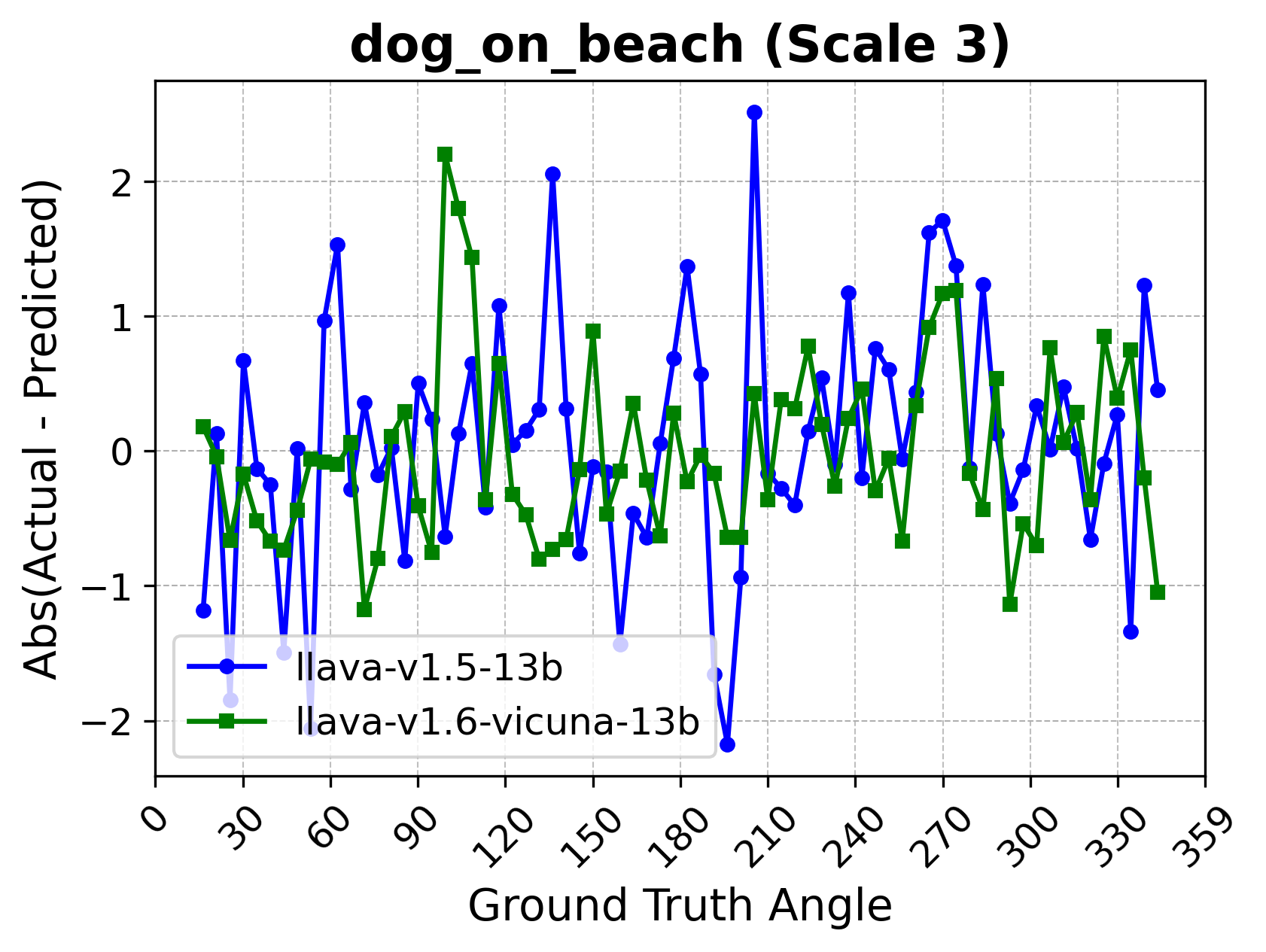}
  \vspace{-2mm}
  \caption{2D orientation estimation performance comparison between LLaVA 1.5 and 1.6 on the images with dog foregrounds (scale 3) for 72 randomly selected images}
  \label{fig:regression_comparison_llava1.5_1.6_dog-on-beach_scale3}
\end{figure}
 
\begin{figure}[htbp]
  \centering
  \includegraphics[width=7.5cm, height=5cm]{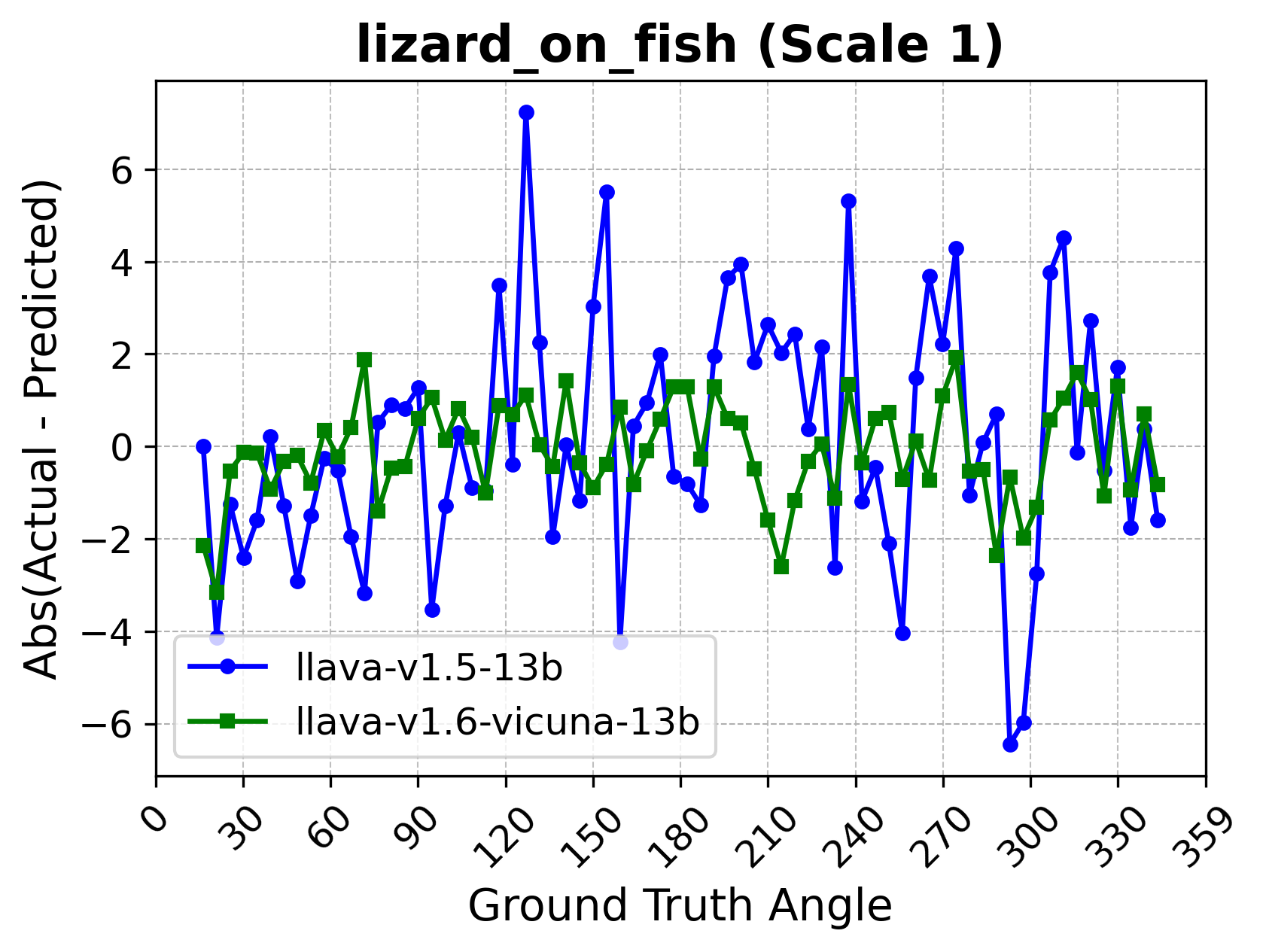}
  \vspace{-2mm}
  \caption{2D orientation estimation performance comparison between LLaVA 1.5 and 1.6 on the images with lizard foregrounds (scale 1) for 72 randomly selected images}
  \label{fig:regression_comparison_llava1.5_1.6_lizard-on-fish_scale1}
\end{figure}

\begin{figure}[h!]
  \centering
  \includegraphics[width=7.5cm, height=5cm]{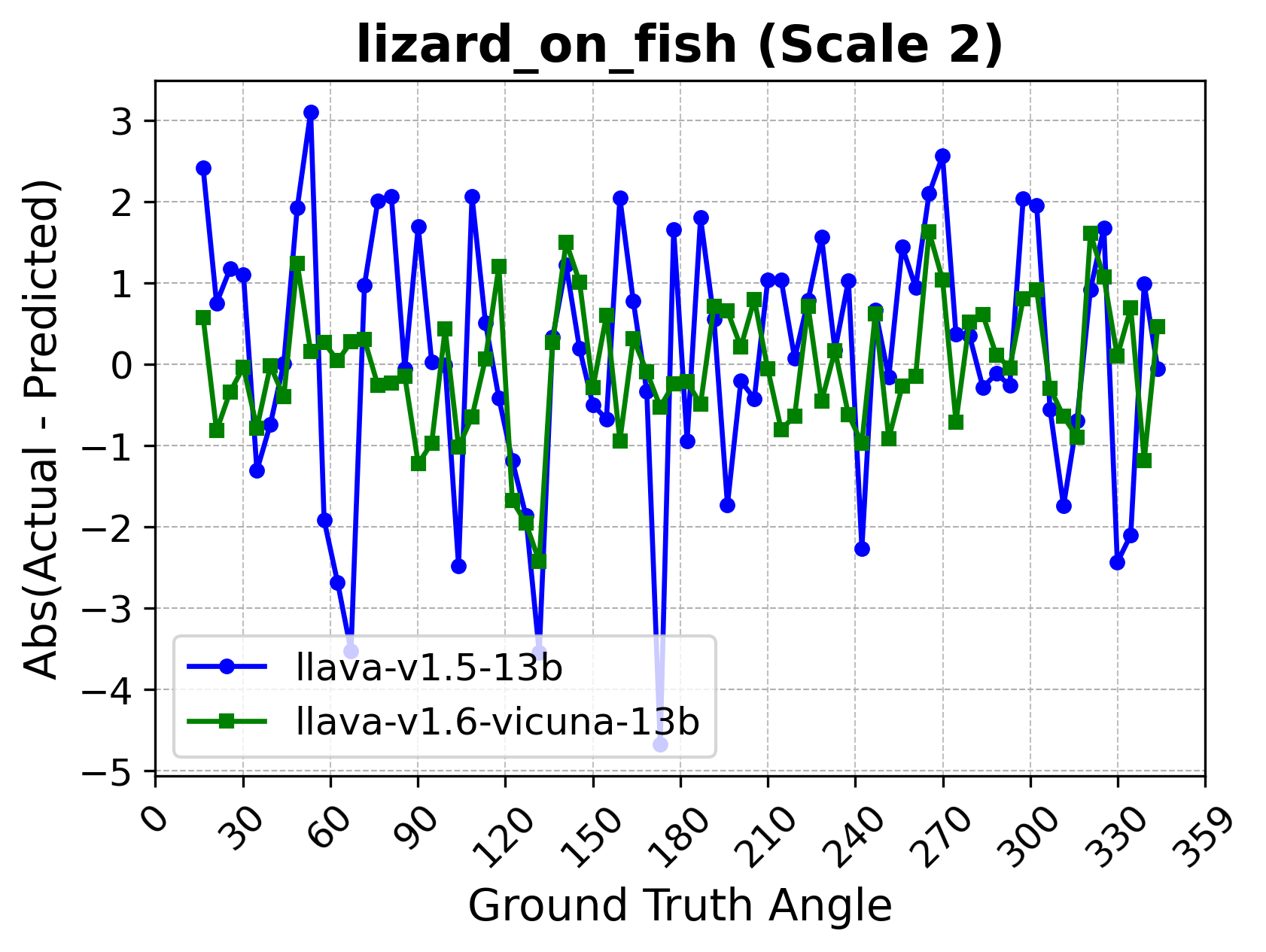}
  \vspace{-2mm}
  \caption{2D orientation estimation performance comparison between LLaVA 1.5 and 1.6 on the images with lizard foregrounds (scale 2) for 72 randomly selected images}
  \label{fig:regression_comparison_llava1.5_1.6_lizard-on-fish_scale2}
\end{figure}

\begin{figure}[h!]
  \centering
  \includegraphics[width=7.5cm, height=5cm]{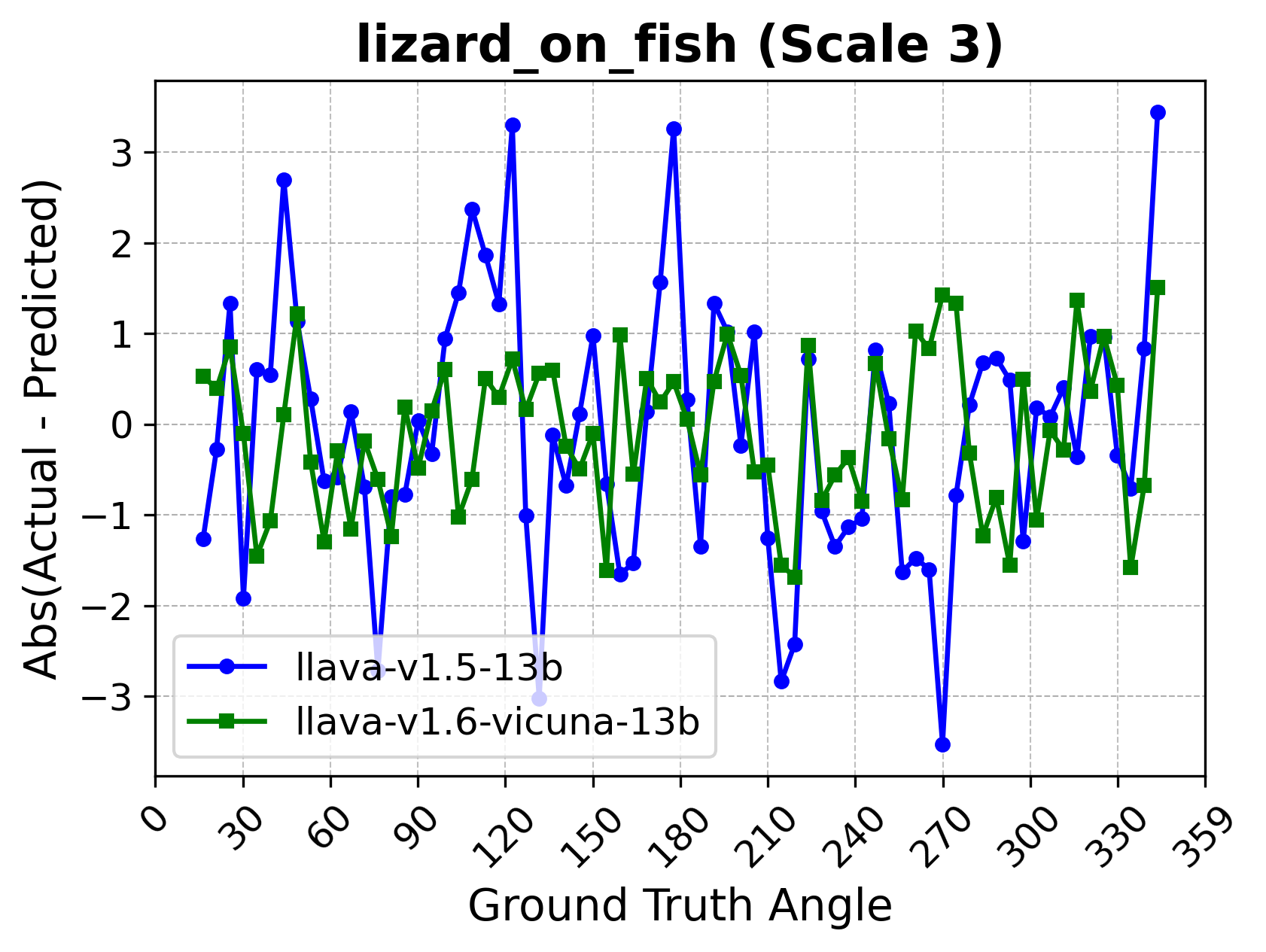}
  \vspace{-2mm}
  \caption{2D orientation estimation performance comparison between LLaVA 1.5 and 1.6 on the images with lizard foregrounds (scale 3) for 72 randomly selected images}
  \label{fig:regression_comparison_llava1.5_1.6_lizard-on-fish_scale3}
\end{figure}

\begin{figure}[h!]
  \centering
  \includegraphics[width=7.5cm, height=5cm]{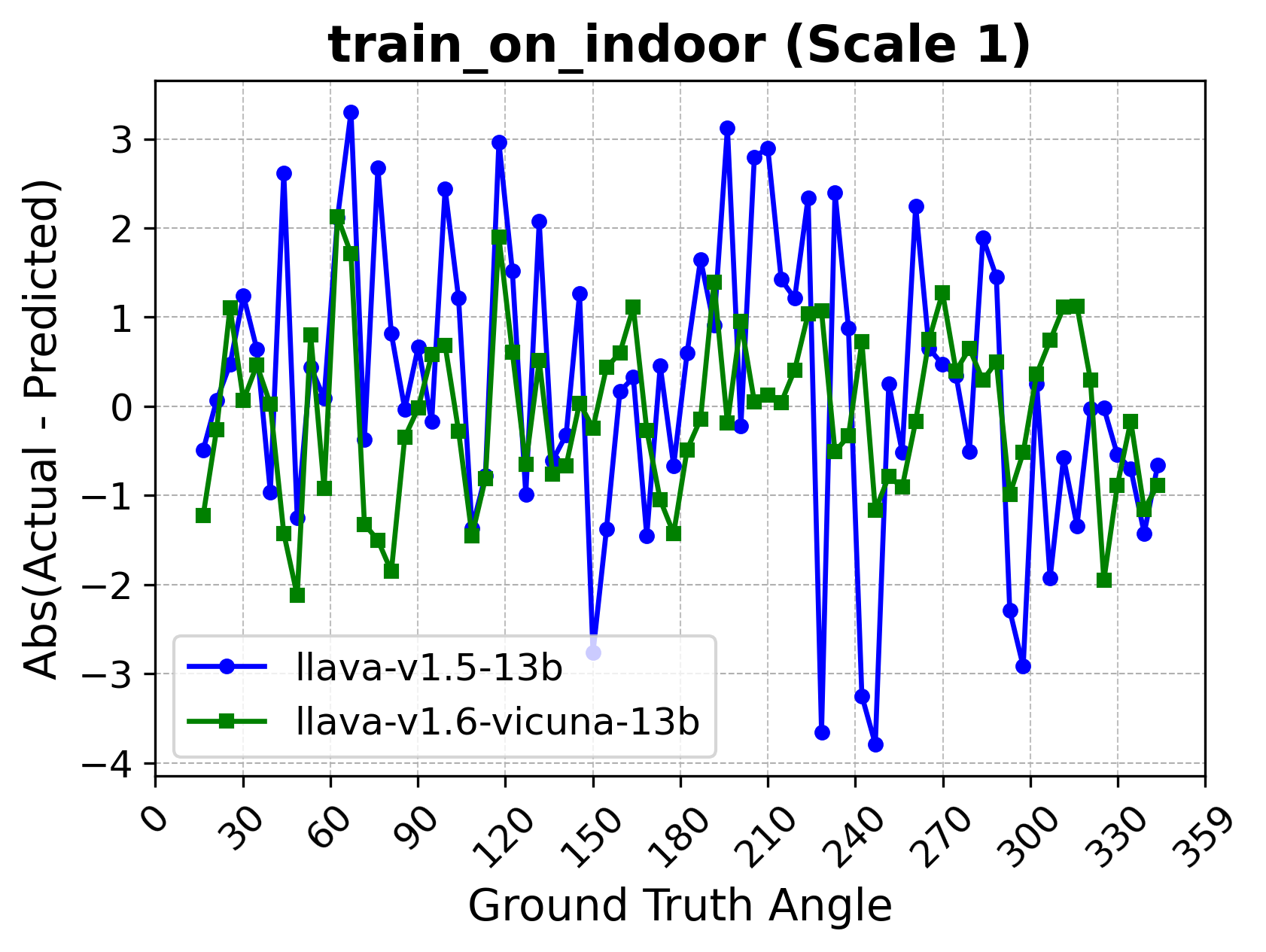}
  \vspace{-2mm}
  \caption{2D orientation estimation performance comparison between LLaVA 1.5 and 1.6 on the images with train foregrounds (scale 1) for 72 randomly selected images}
  \label{fig:regression_comparison_llava1.5_1.6_train_on_indoor_scale1}
\end{figure}

\begin{figure}[h!]
  \centering
  \includegraphics[width=7.5cm, height=5cm]{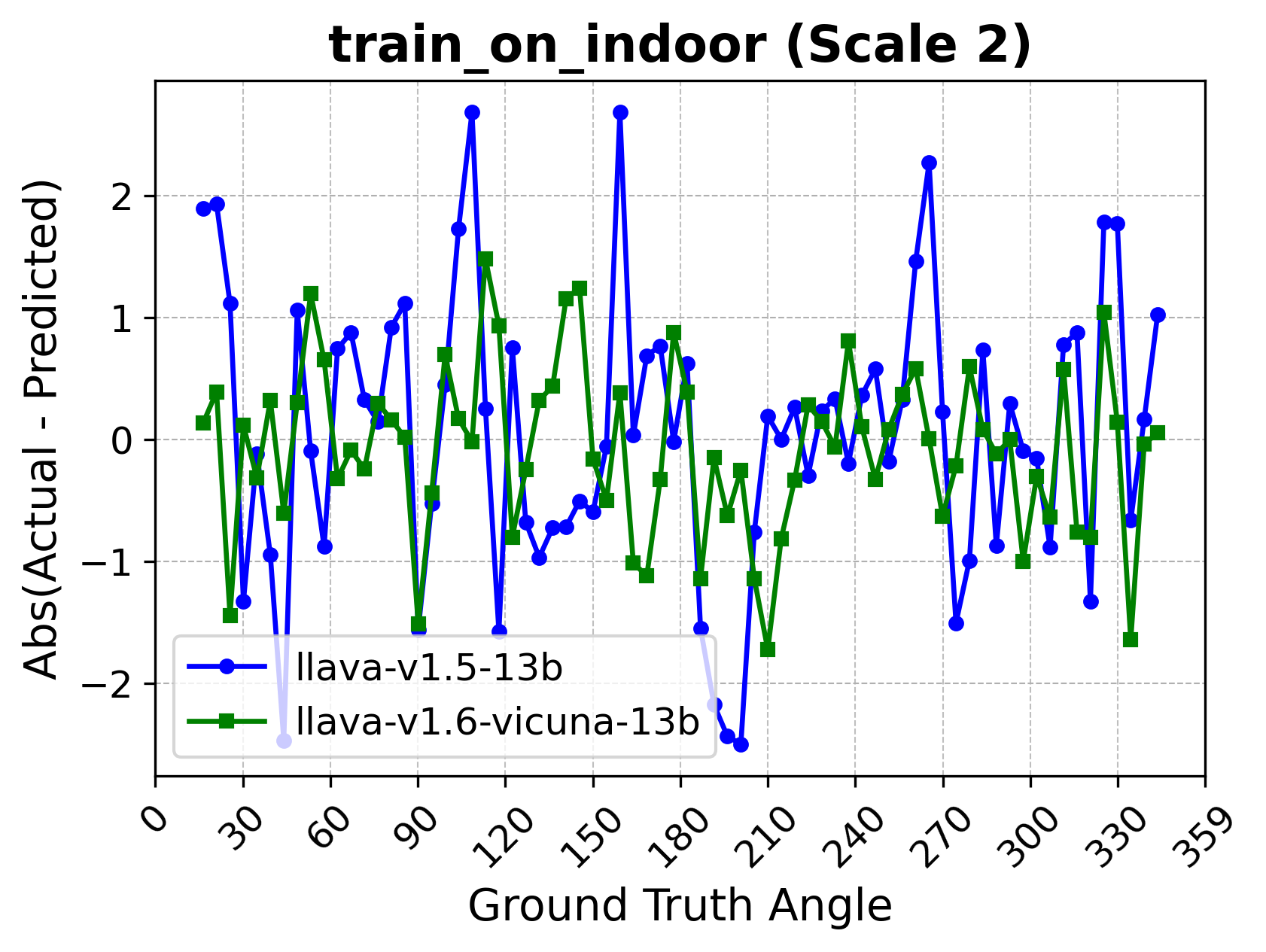}
  \vspace{-2mm}
  \caption{2D orientation estimation performance comparison between LLaVA 1.5 and 1.6 on the images with train foregrounds (scale 2) for 72 randomly selected images}
  \label{fig:regression_comparison_llava1.5_1.6_train_on_indoor_scale2}
\end{figure}

\begin{figure}[h!]
  \centering
  \includegraphics[width=7.5cm, height=5cm]{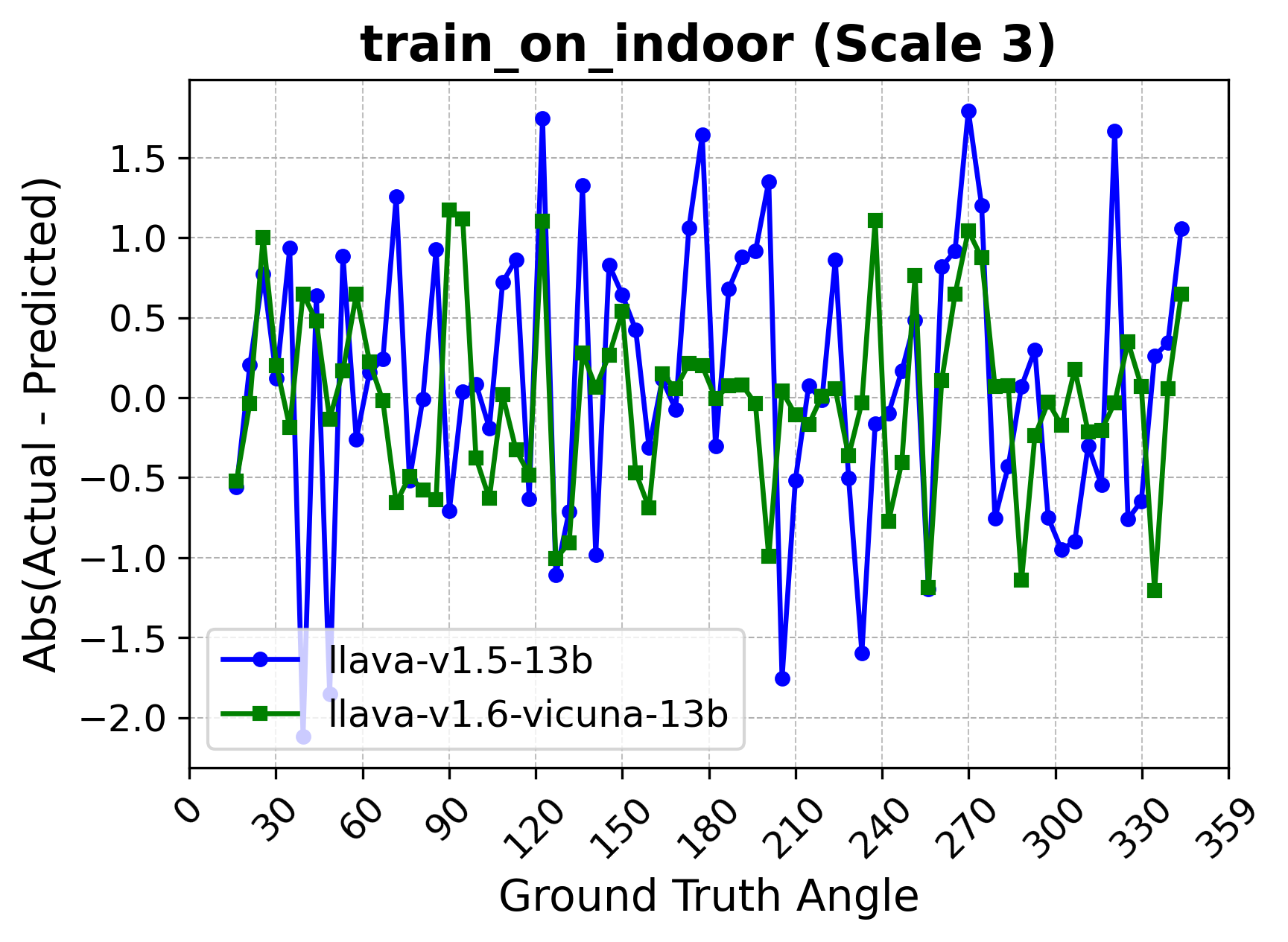}
  \vspace{-2mm}
  \caption{2D orientation estimation performance comparison between LLaVA 1.5 and 1.6 on the images with train foregrounds (scale 3) for 72 randomly selected images}
  \label{fig:regression_comparison_llava1.5_1.6_train_on_indoor_scale3}
\end{figure}

\clearpage

\subsection{Plots Showing Statistical Analysis for LLaVA-OneVision and Qwen2.5-VL-7B-Instruct}
\label{app:llava-qwen}

\begin{figure}[H]
  \centering
\includegraphics[width=7cm, height=5.5cm]{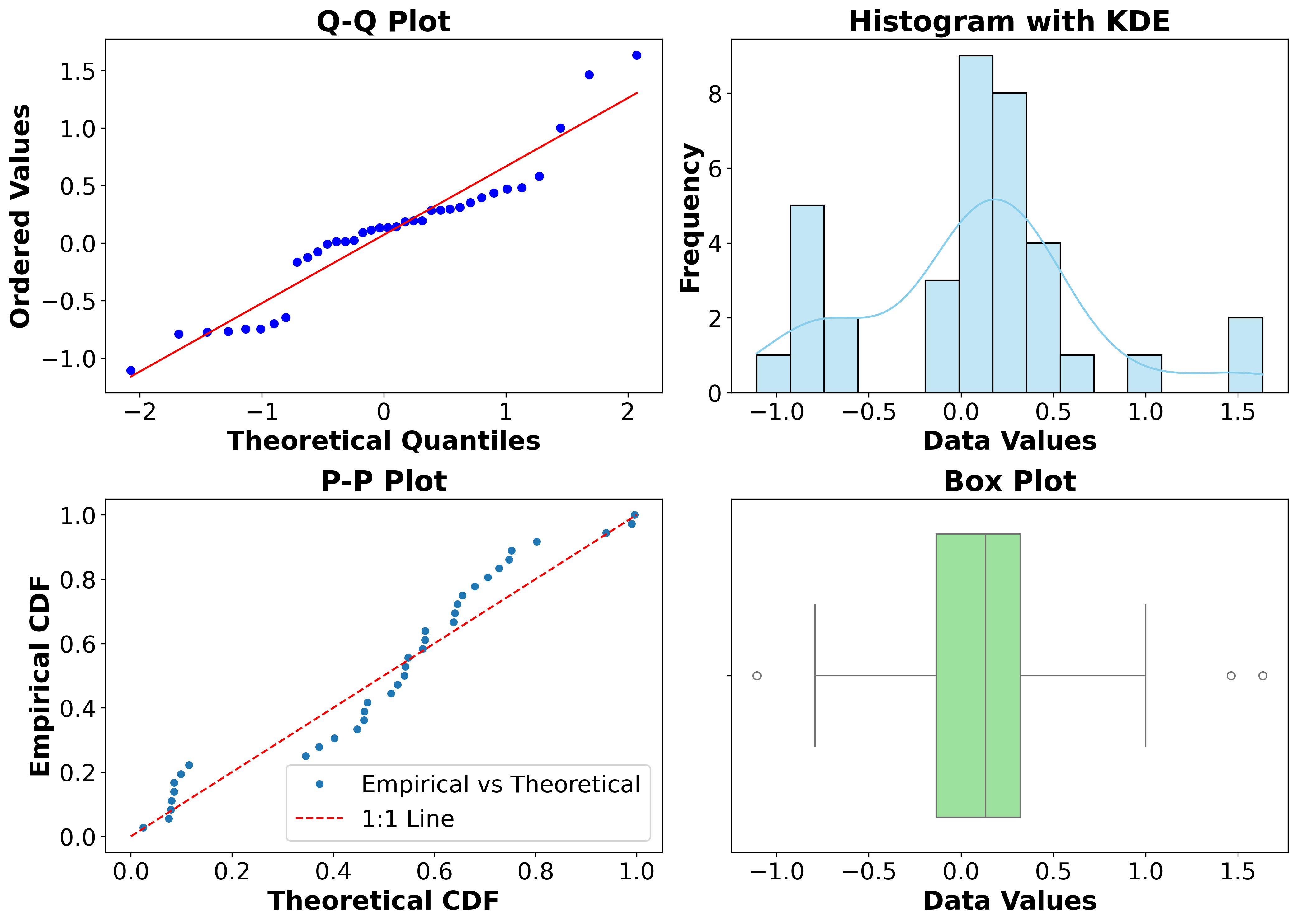}
   \caption{Statistical Analysis using visual plots for LLaVA-OneVision - results for images with lizard. Together with the numerical results in Table \ref{tab:ks_test} and the visual plots in this Figure, we can conclude that the residual error distribution (Table \ref{tab:llava_qwen_comparison}) is random/Gaussian}
   \label{fig:stat_analysisLLaVA-OneVision_lizard}
   \hfill
\end{figure}
\begin{figure}[H]
  \centering
\includegraphics[width=7cm, height=5.5cm]{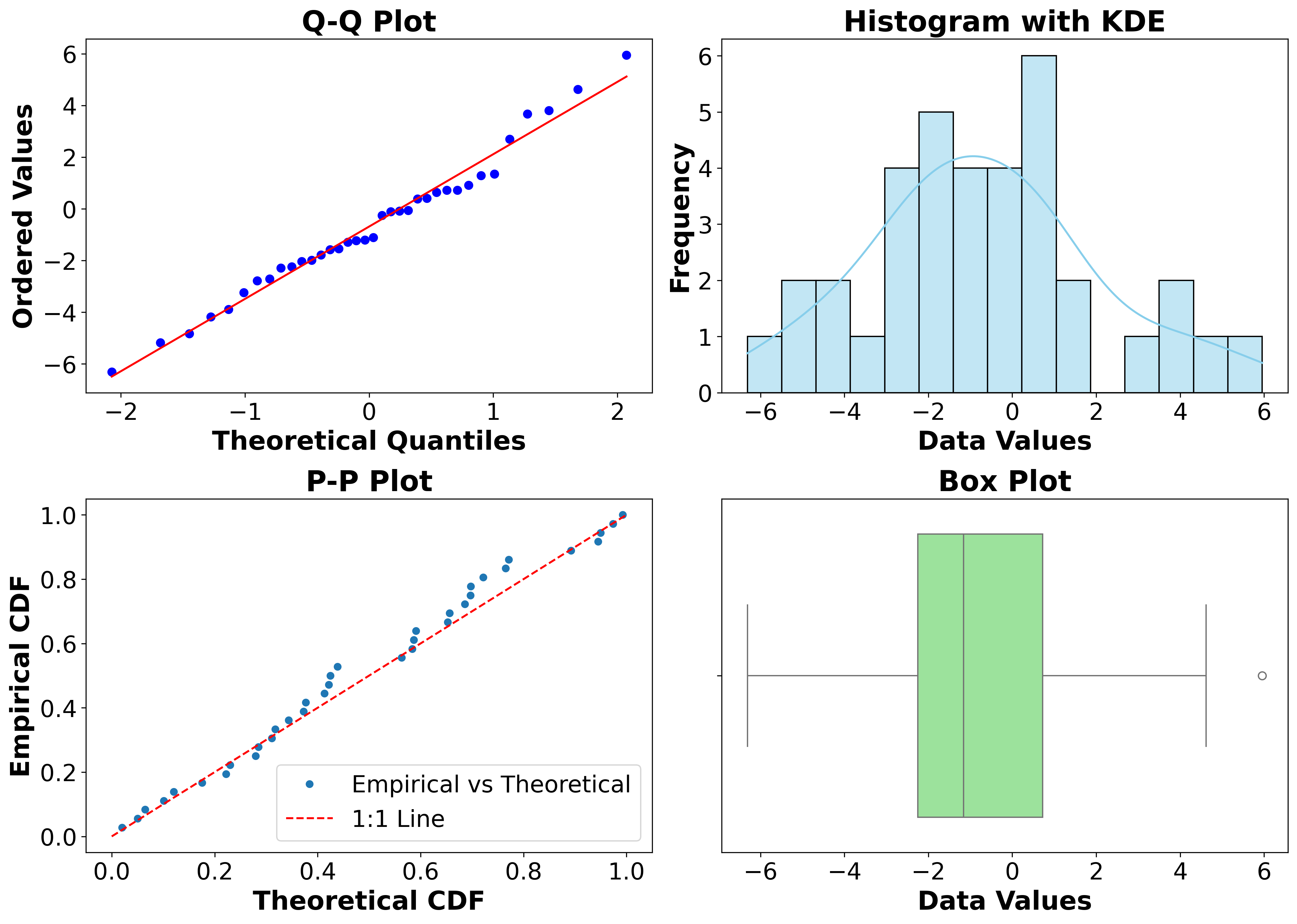}
   \caption{Statistical Analysis using visual plots for Qwen2.5-VL-7B-Instruct - results for images with lizard. Together with the numerical results in Table \ref{tab:ks_test} and the visual plots in this Figure, we can conclude that the residual error distribution (Table \ref{tab:llava_qwen_comparison}) is random/Gaussian}
   \label{fig:stat_analysisQwen2.5-VL-7B-Instruct_lizard}
   \hfill
\end{figure}
\clearpage
\begin{figure}[t]
  \centering
\includegraphics[width=7cm, height=5.5cm]{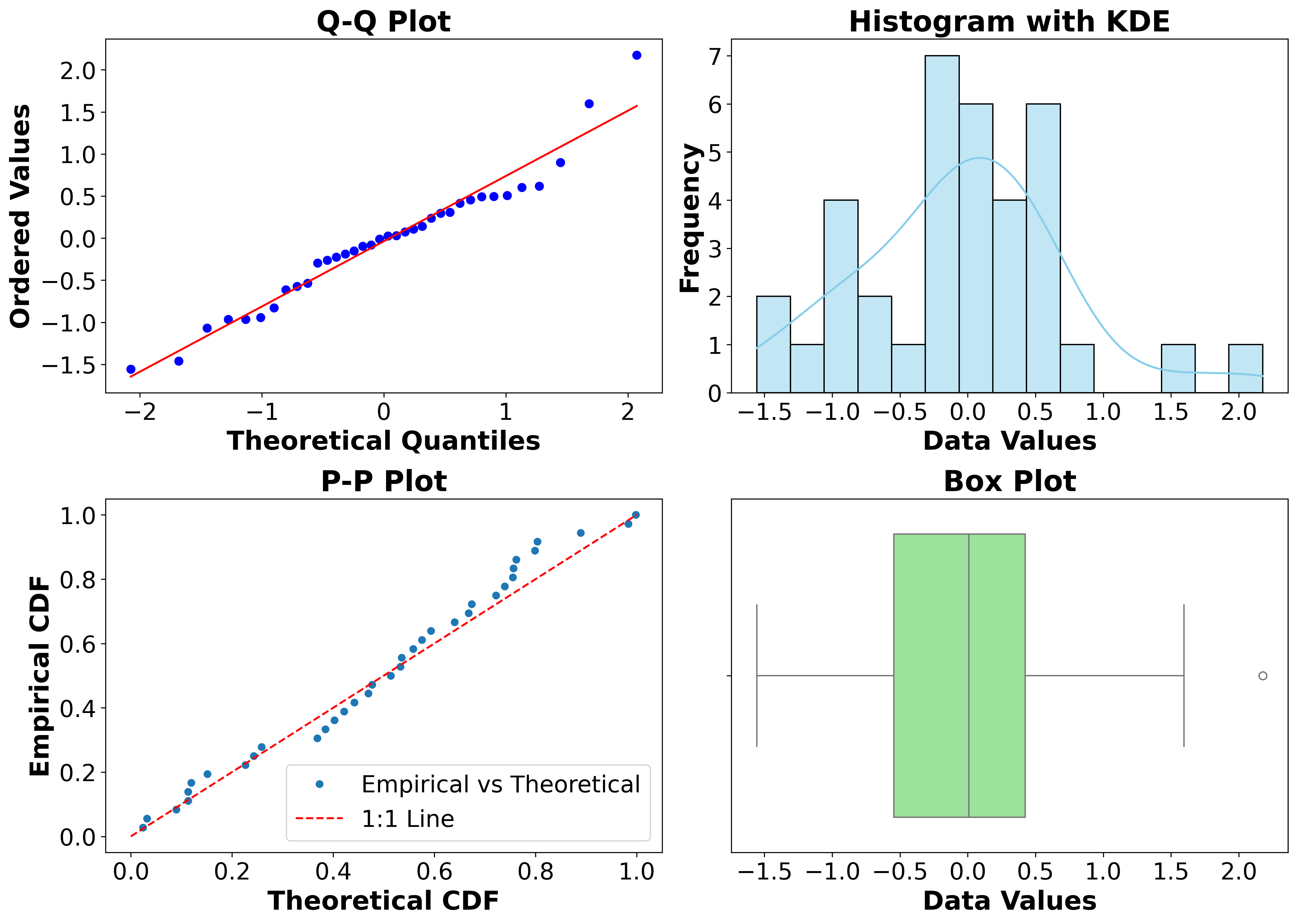}
   \caption{Statistical Analysis using visual plots for LLaVA-OneVision - results for images with train. Together with the numerical results in Table \ref{tab:ks_test} and the visual plots in this Figure, we can conclude that the residual error distribution (Table \ref{tab:llava_qwen_comparison}) is random/Gaussian}
   \label{fig:stat_analysisLLaVA-OneVision_train}
   \hfill
\end{figure}

\begin{figure}[t]
  \centering
\includegraphics[width=7cm, height=5.5cm]{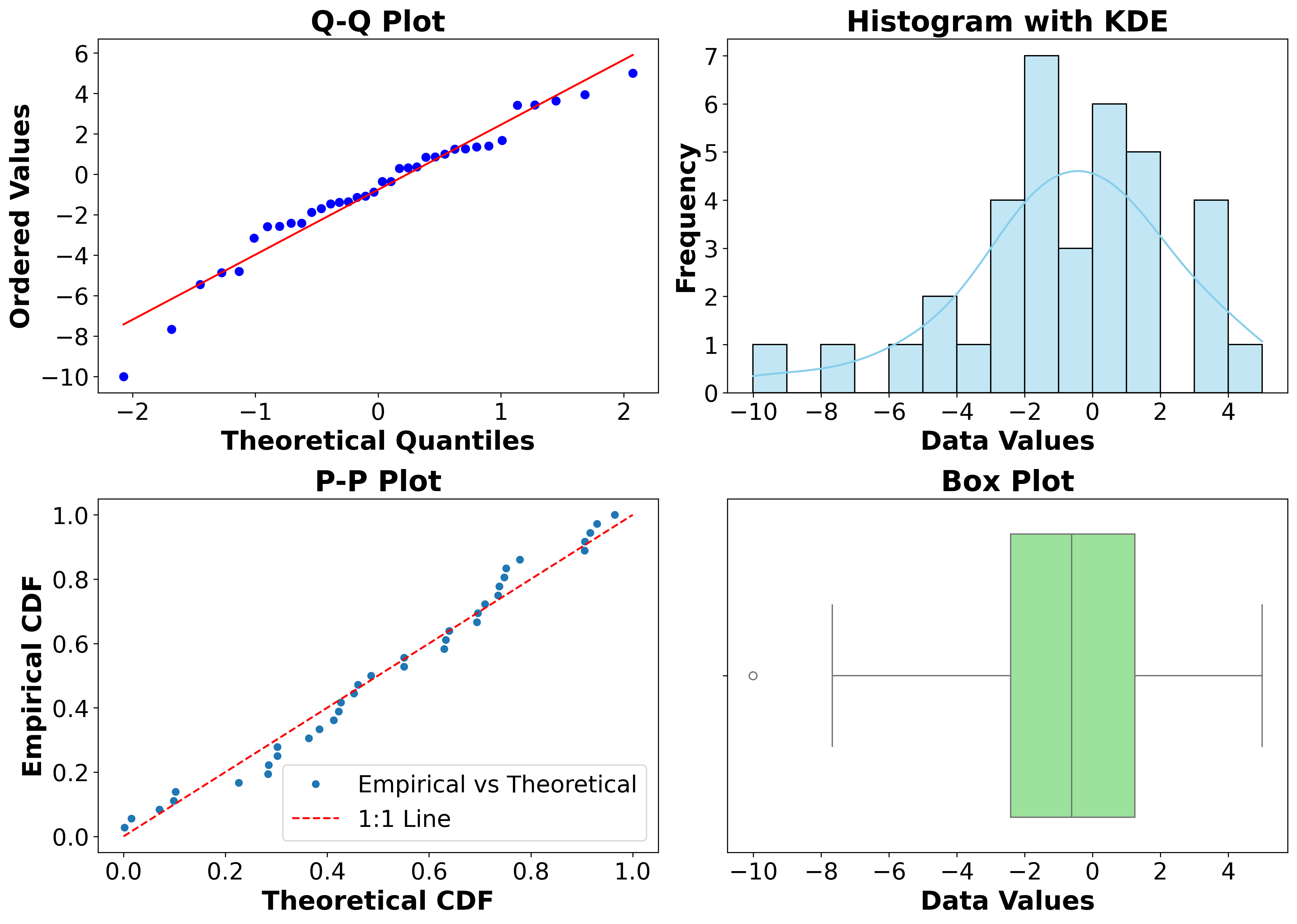}
   \caption{Statistical Analysis using visual plots for Qwen2.5-VL-7B-Instruct - results for images with train. Together with the numerical results in Table \ref{tab:ks_test} and the visual plots in this Figure, we can conclude that the residual error distribution (Table \ref{tab:llava_qwen_comparison}) is random/Gaussian}
   \label{fig:stat_analysisQwen2.5-VL-7B-Instruct_train}
   \hfill
\end{figure}
\clearpage
\begin{figure}[t]
  \centering
\includegraphics[width=7cm, height=5.5cm]{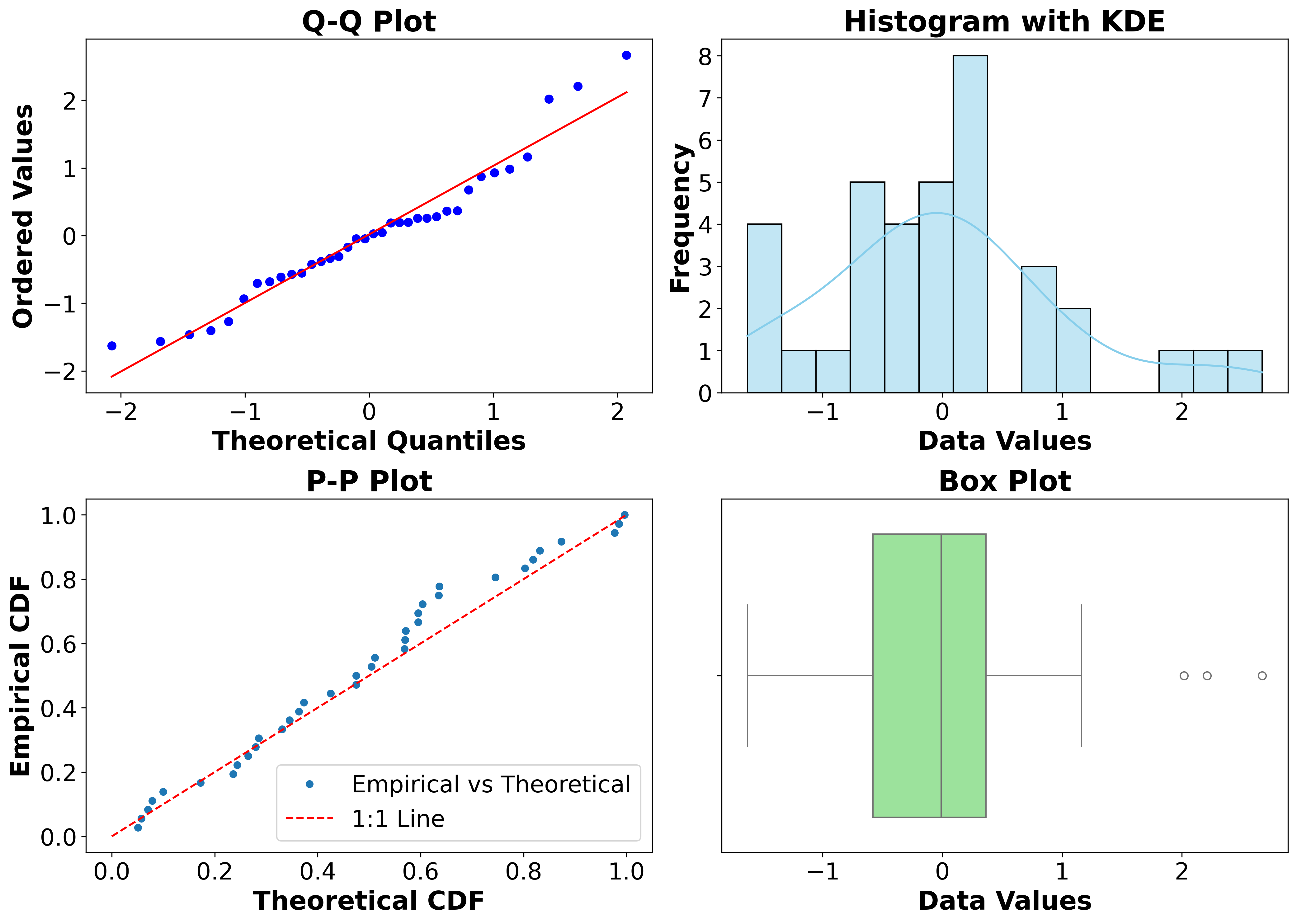}
   \caption{Statistical Analysis using visual plots for LLaVA-OneVision - results for images with beach. Together with the numerical results in Table \ref{tab:ks_test} and the visual plots in this Figure, we can conclude that the residual error distribution (Table \ref{tab:llava_qwen_comparison}) is random/Gaussian}
   \label{fig:stat_analysisLLaVA-OneVision_beach}
   \hfill
\end{figure}
\begin{figure}[t]
  \centering
\includegraphics[width=7cm, height=5.5cm]{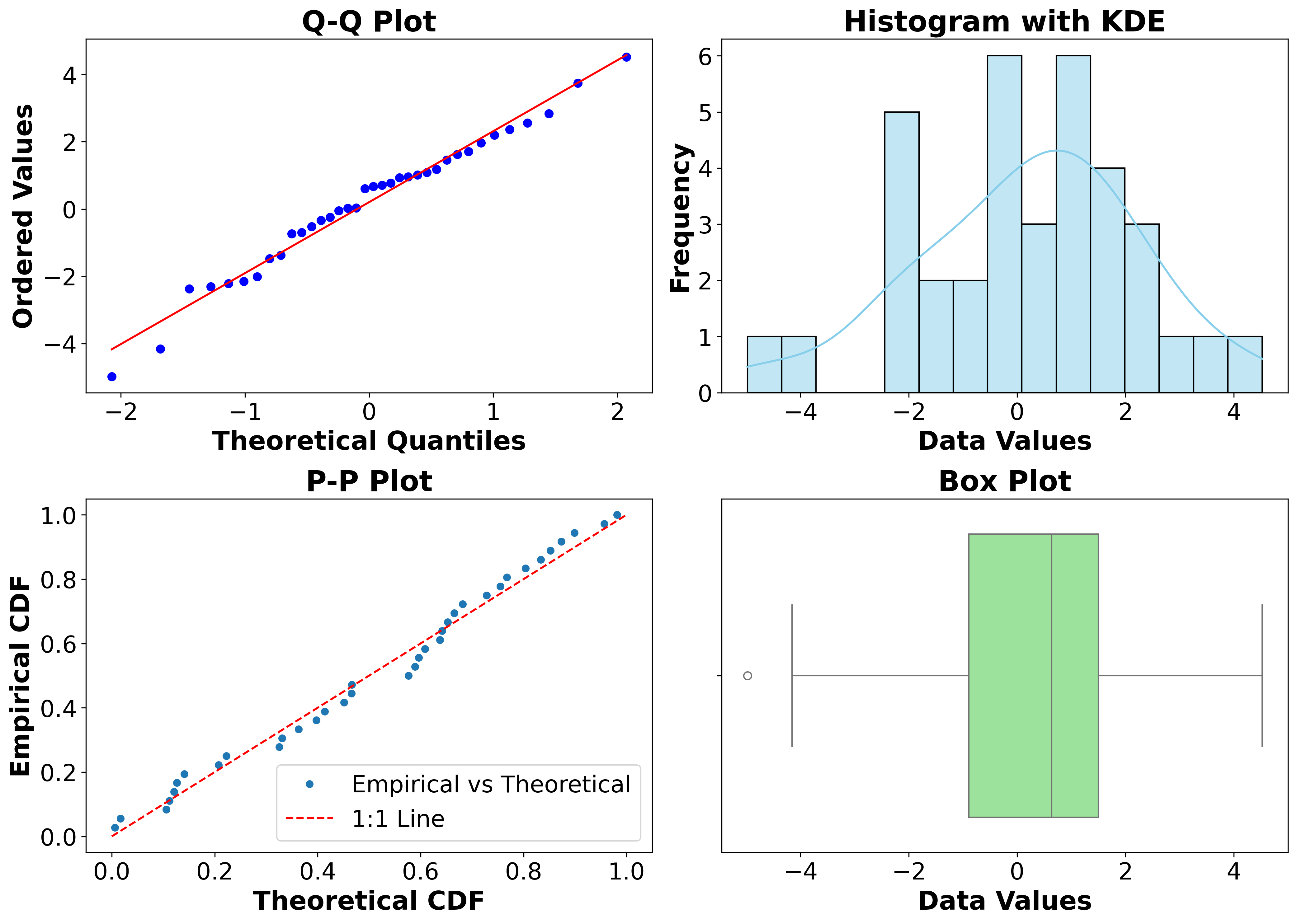}
   \caption{Statistical Analysis using visual plots for Qwen2.5-VL-7B-Instruct - results for images with beach. Together with the numerical results in Table \ref{tab:ks_test} and the visual plots in this Figure, we can conclude that the residual error distribution (Table \ref{tab:llava_qwen_comparison}) is random/Gaussian}
   \label{fig:stat_analysisQwen2.5-VL-7B-Instruct_beach}
   \hfill
\end{figure}

\begin{figure}[t]
  \centering
\includegraphics[width=7cm, height=5.5cm]{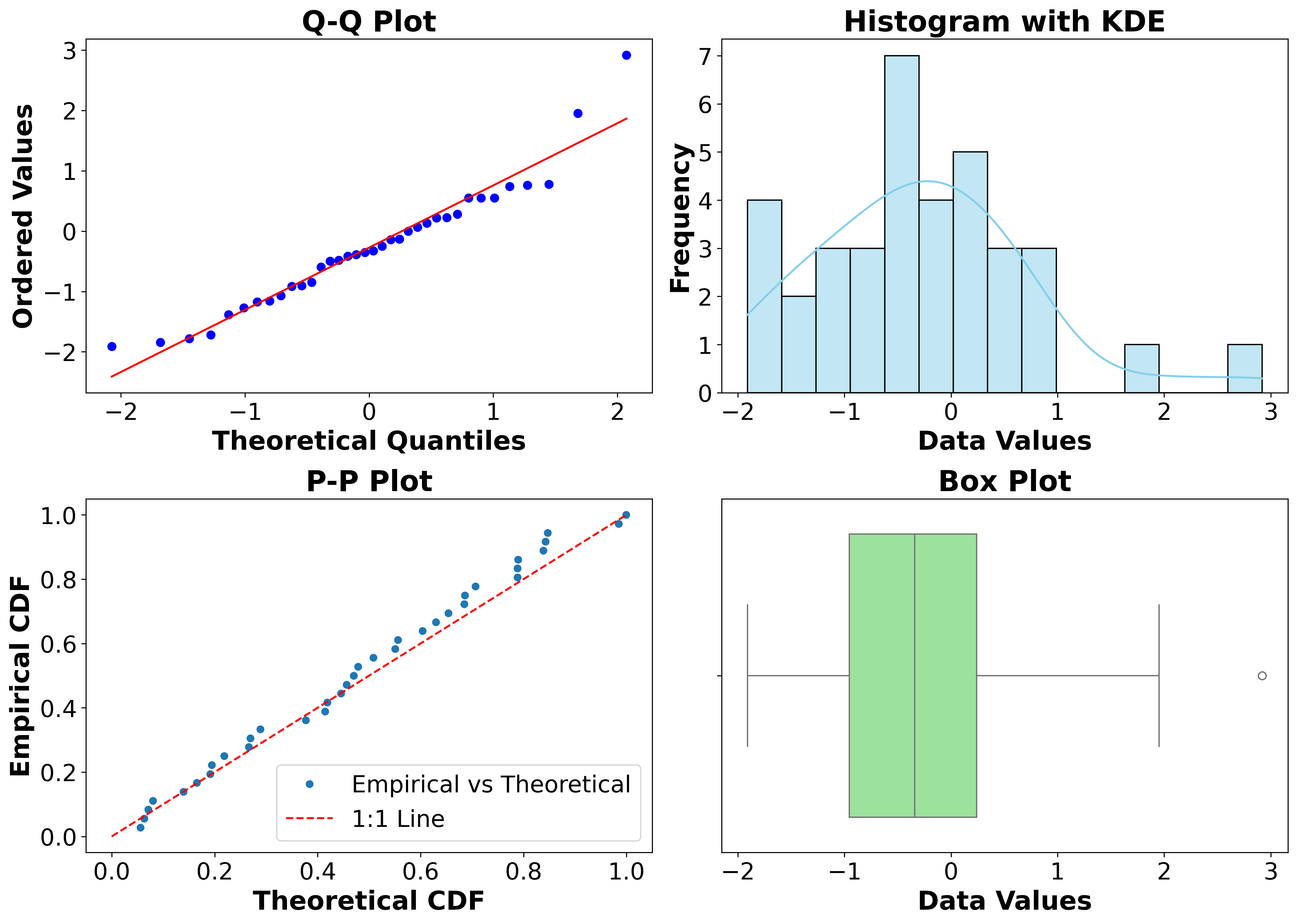}
   \caption{Statistical Analysis using visual plots for LLaVA-OneVision - results for images with indoor. Together with the numerical results in Table \ref{tab:ks_test} and the visual plots in this Figure, we can conclude that the residual error distribution (Table \ref{tab:llava_qwen_comparison}) is random/Gaussian}
   \label{fig:stat_analysisLLaVA-OneVision_indoor}
   \hfill
\end{figure}
\begin{figure}[t]
  \centering
\includegraphics[width=7cm, height=5.5cm]{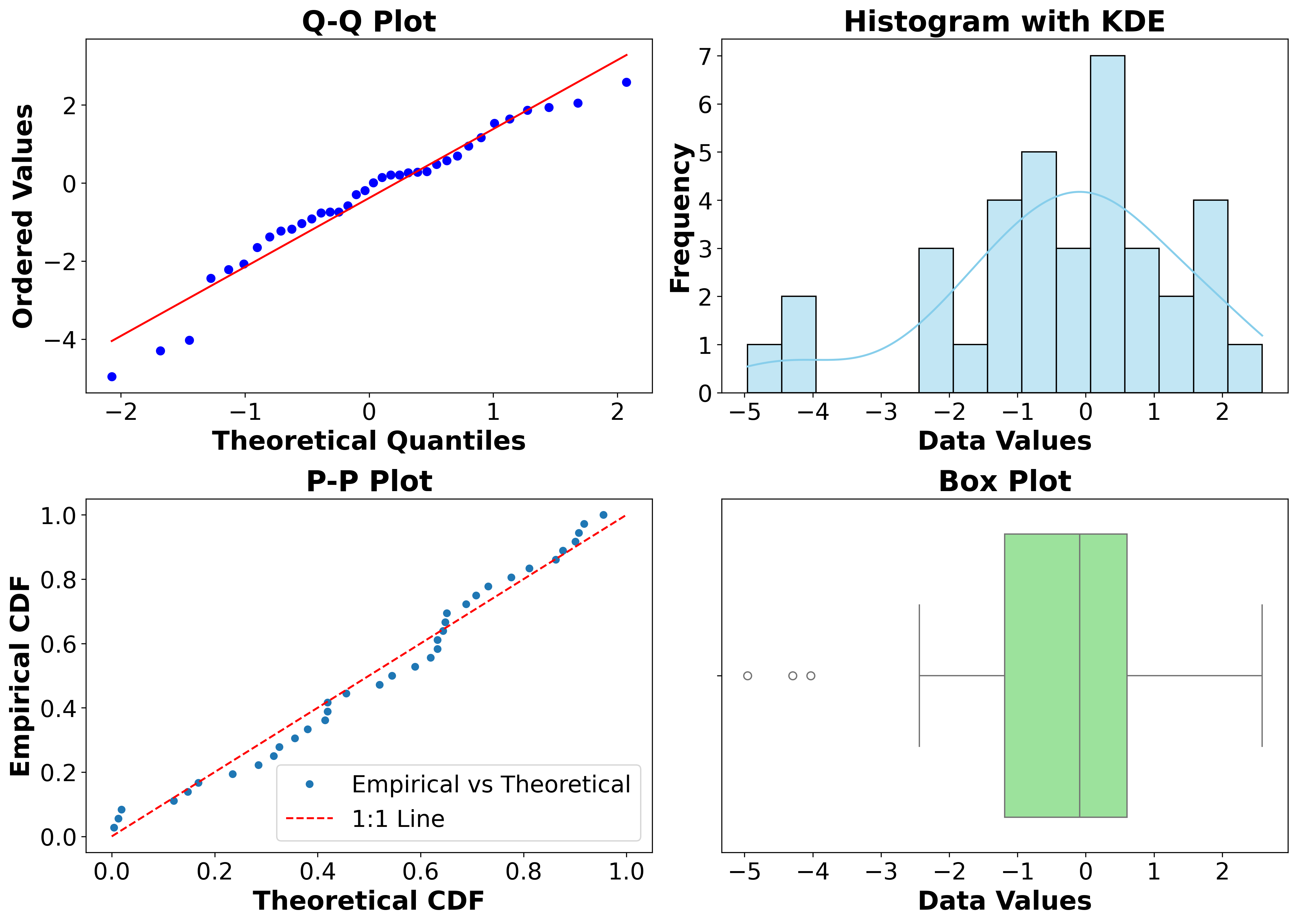}
   \caption{Statistical Analysis using visual plots for Qwen2.5-VL-7B-Instruct - results for images with indoor. Together with the numerical results in Table \ref{tab:ks_test} and the visual plots in this Figure, we can conclude that the residual error distribution (Table \ref{tab:llava_qwen_comparison}) is random/Gaussian}
   \label{fig:stat_analysisQwen2.5-VL-7B-Instruct_indoor}
   \hfill
\end{figure}

\begin{figure}[t]
  \centering
\includegraphics[width=7cm, height=5.5cm]{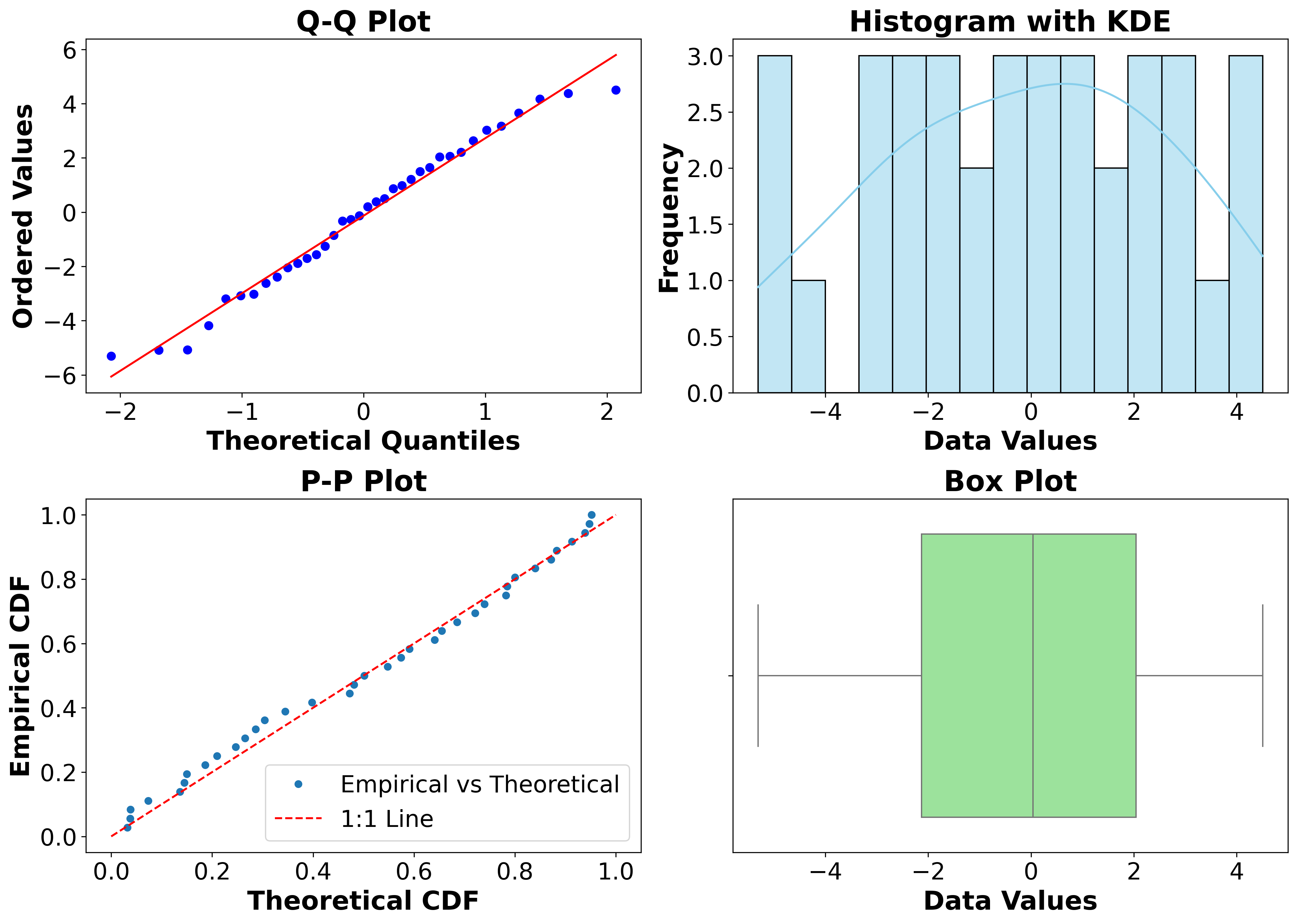}
   \caption{Statistical Analysis using visual plots for LLaVA-OneVision - results for images with fish. Together with the numerical results in Table \ref{tab:ks_test} and the visual plots in this Figure, we can conclude that the residual error distribution (Table \ref{tab:llava_qwen_comparison}) is random/Gaussian}
   \label{fig:stat_analysisLLaVA-OneVision_fish}
   \hfill
\end{figure}
\begin{figure}[t]
  \centering
\includegraphics[width=7cm, height=5.5cm]{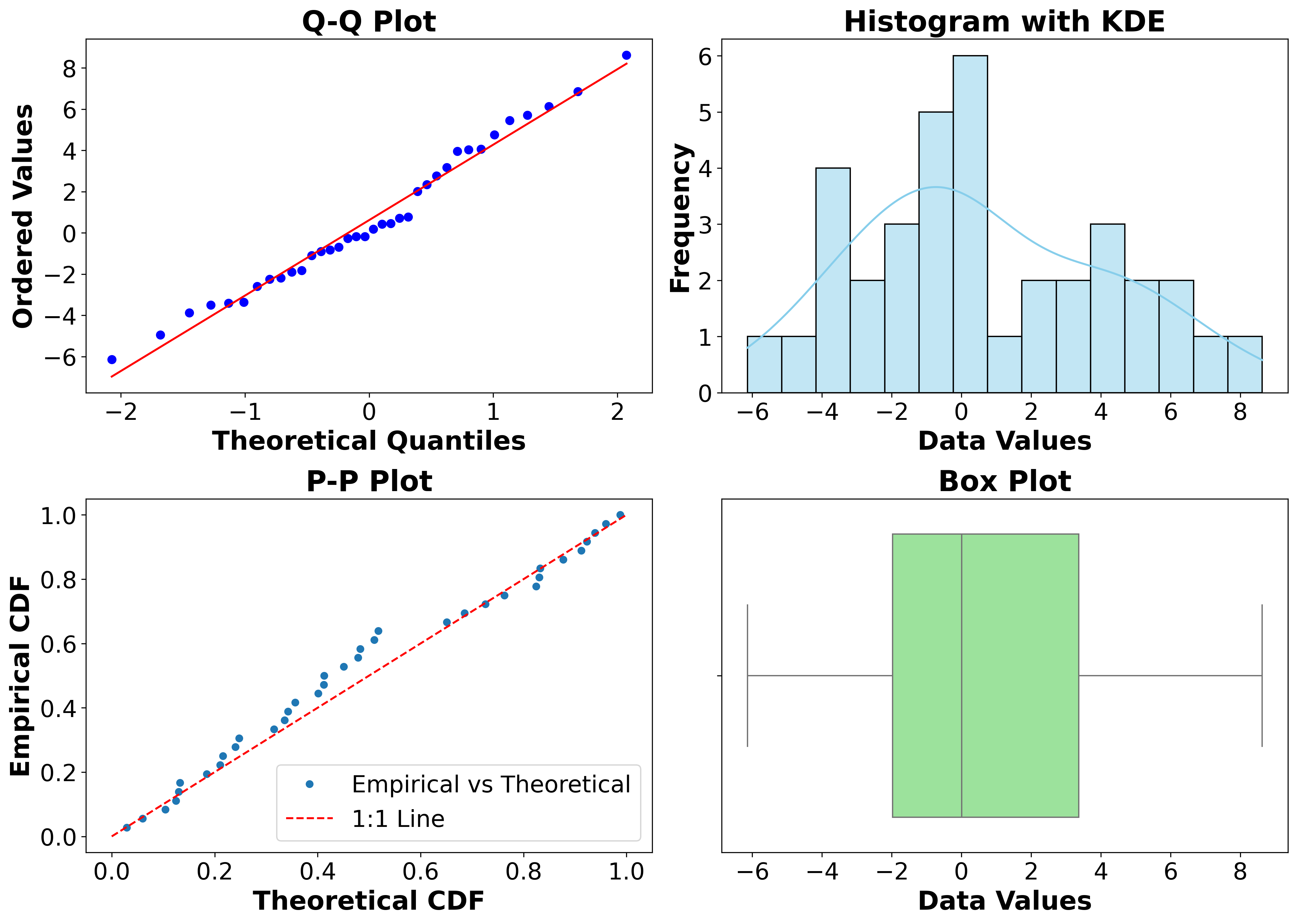}
   \caption{Statistical Analysis using visual plots for Qwen2.5-VL-7B-Instruct - results for images with fish. Together with the numerical results in Table \ref{tab:ks_test} and the visual plots in this Figure, we can conclude that the residual error distribution (Table \ref{tab:llava_qwen_comparison}) is random/Gaussian}
   \label{fig:stat_analysisQwen2.5-VL-7B-Instruct_fish}
   \hfill
\end{figure}
\begin{figure}[t]

  \centering
\includegraphics[width=7cm, height=5.5cm]{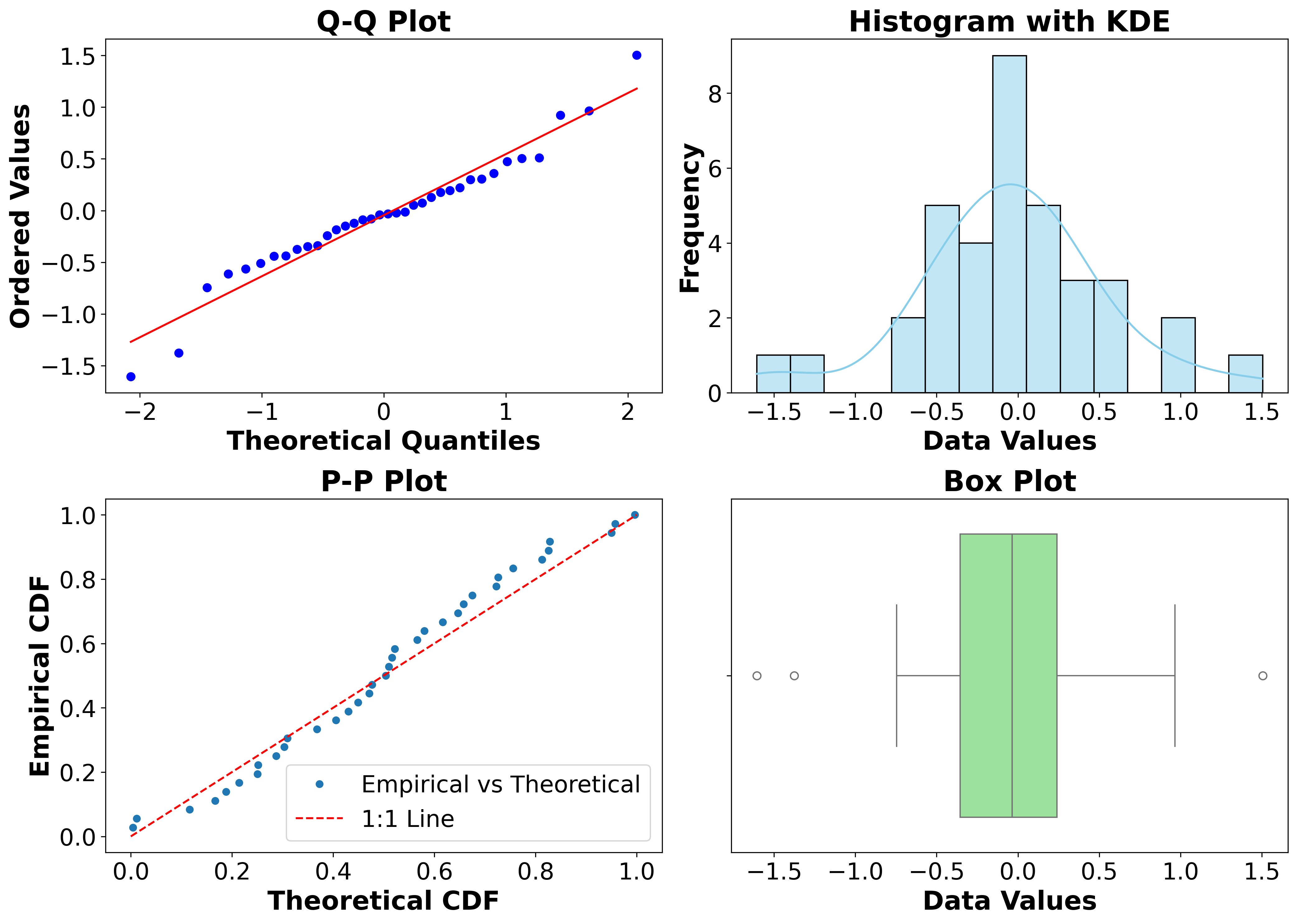}
   \caption{Statistical Analysis using visual plots for LLaVA-OneVision - results for images with koala-beach. Together with the numerical results in Table \ref{tab:ks_test} and the visual plots in this Figure, we can conclude that the residual error distribution (Table \ref{tab:llava_qwen_comparison}) is random/Gaussian}
   \label{fig:stat_analysisLLaVA-OneVision_koala-beach}
   \hfill
\end{figure}
\begin{figure}[t]
  \centering
\includegraphics[width=7cm, height=5.5cm]{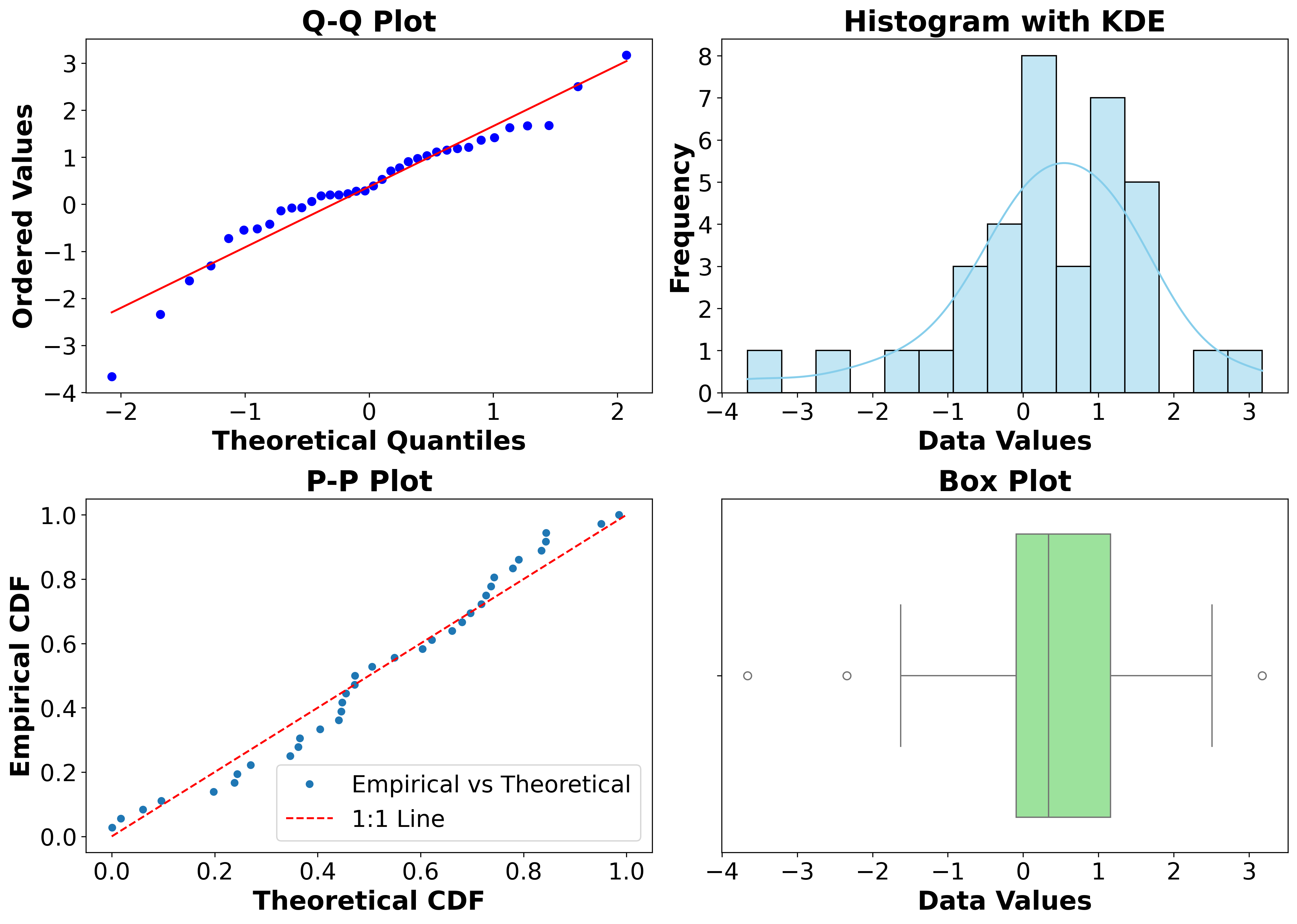}
   \caption{Statistical Analysis using visual plots for Qwen2.5-VL-7B-Instruct - results for images with koala-beach. Together with the numerical results in Table \ref{tab:ks_test} and the visual plots in this Figure, we can conclude that the residual error distribution (Table \ref{tab:llava_qwen_comparison}) is random/Gaussian}
   \label{fig:stat_analysisQwen2.5-VL-7B-Instruct_koala-beach}
   \hfill
\end{figure}

\begin{figure}[t]

  \centering
\includegraphics[width=7cm, height=5.5cm]{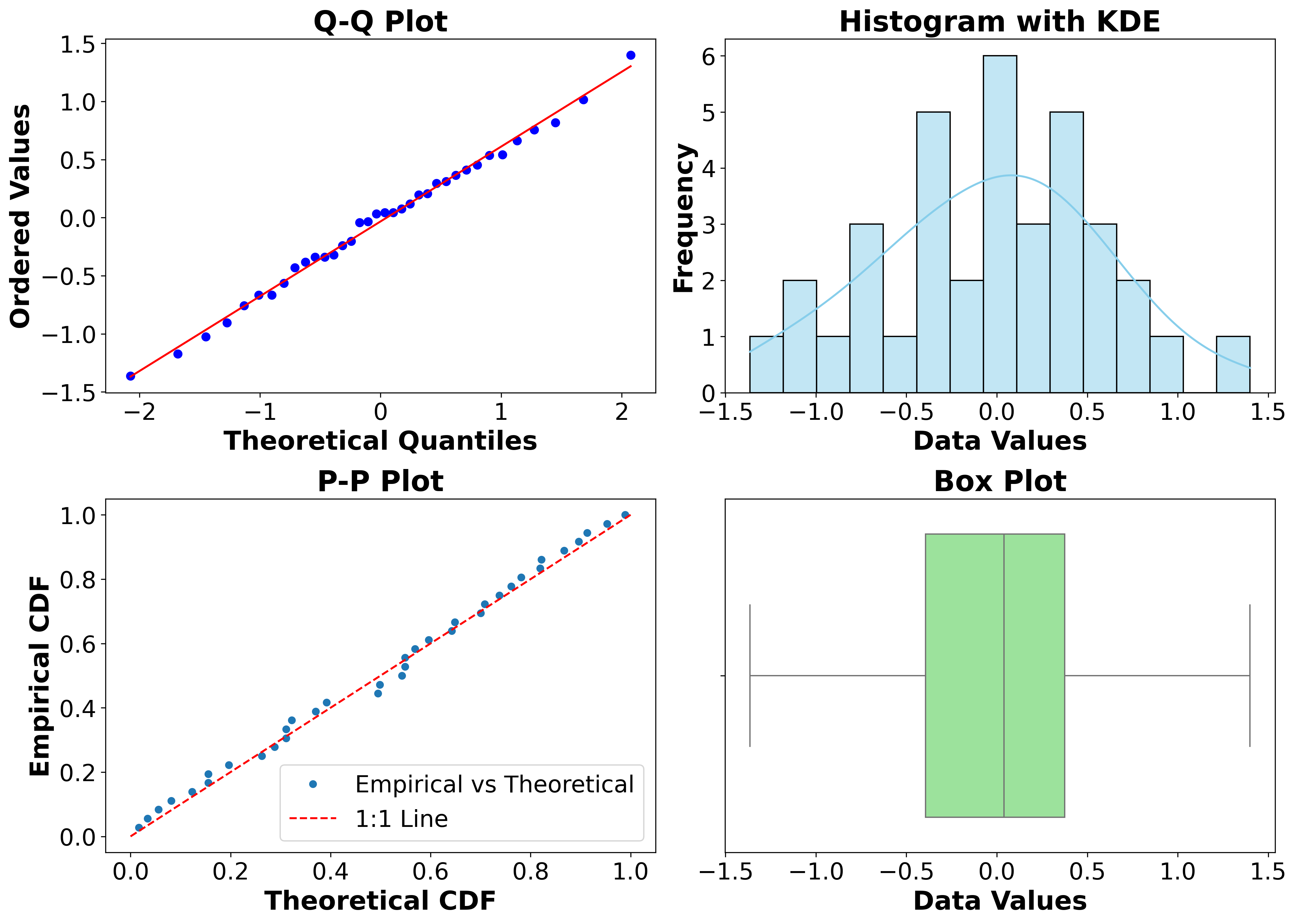}
   \caption{Statistical Analysis using visual plots for LLaVA-OneVision - results for images with vase-indoor. Together with the numerical results in Table \ref{tab:ks_test} and the visual plots in this Figure, we can conclude that the residual error distribution (Table \ref{tab:llava_qwen_comparison}) is random/Gaussian}
   \label{fig:stat_analysisLLaVA-OneVision_vase-indoor}
   \hfill
\end{figure}
\begin{figure}[t]
  \centering
\includegraphics[width=7cm, height=5.5cm]{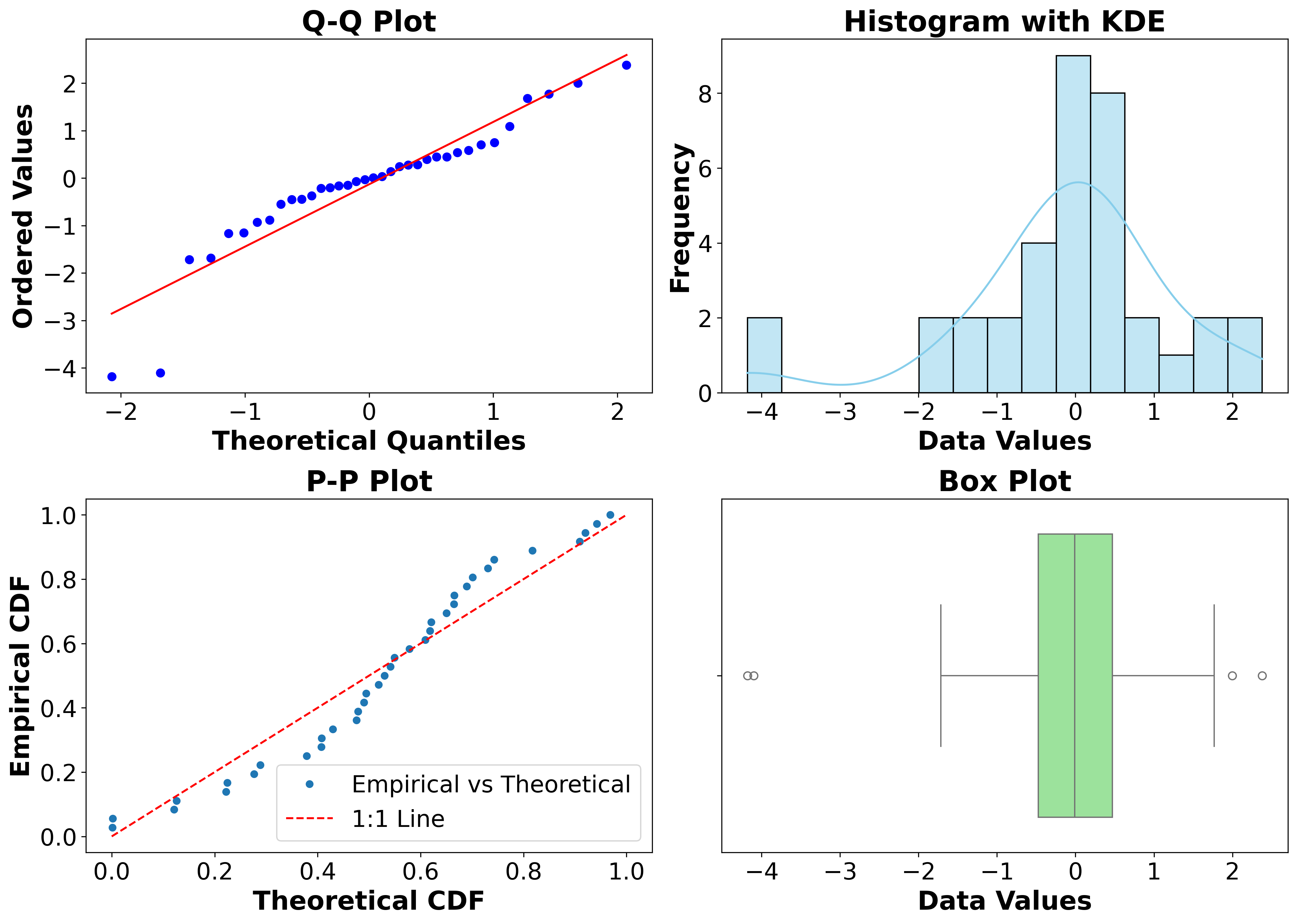}
   \caption{Statistical Analysis using visual plots for Qwen2.5-VL-7B-Instruct - results for images with vase-indoor. Together with the numerical results in Table \ref{tab:ks_test} and the visual plots in this Figure, we can conclude that the residual error distribution (Table \ref{tab:llava_qwen_comparison}) is random/Gaussian}
   \label{fig:stat_analysisQwen2.5-VL-7B-Instruct_vase-indoor}
   \hfill
\end{figure}

\begin{figure}[t]

  \centering
\includegraphics[width=7cm, height=5.5cm]{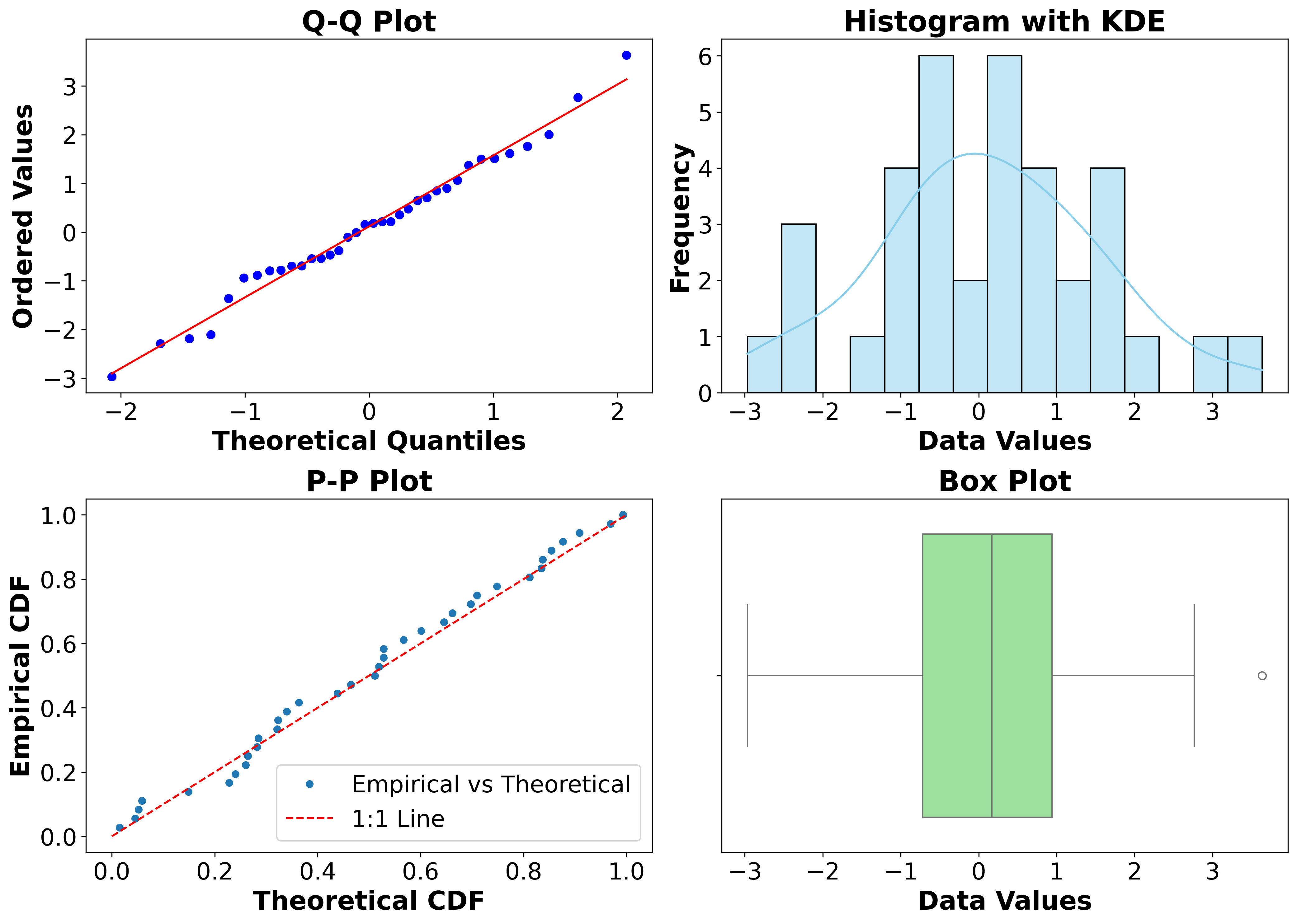}
   \caption{Statistical Analysis using visual plots for LLaVA-OneVision - results for images with vase-toaster-indoor. Together with the numerical results in Table \ref{tab:ks_test} and the visual plots in this Figure, we can conclude that the residual error distribution (Table \ref{tab:llava_qwen_comparison}) is random/Gaussian}
   \label{fig:stat_analysisLLaVA-OneVision_vase-toaster-indoor}
   \hfill
\end{figure}
\begin{figure}[t]
  \centering
\includegraphics[width=7cm, height=5.5cm]{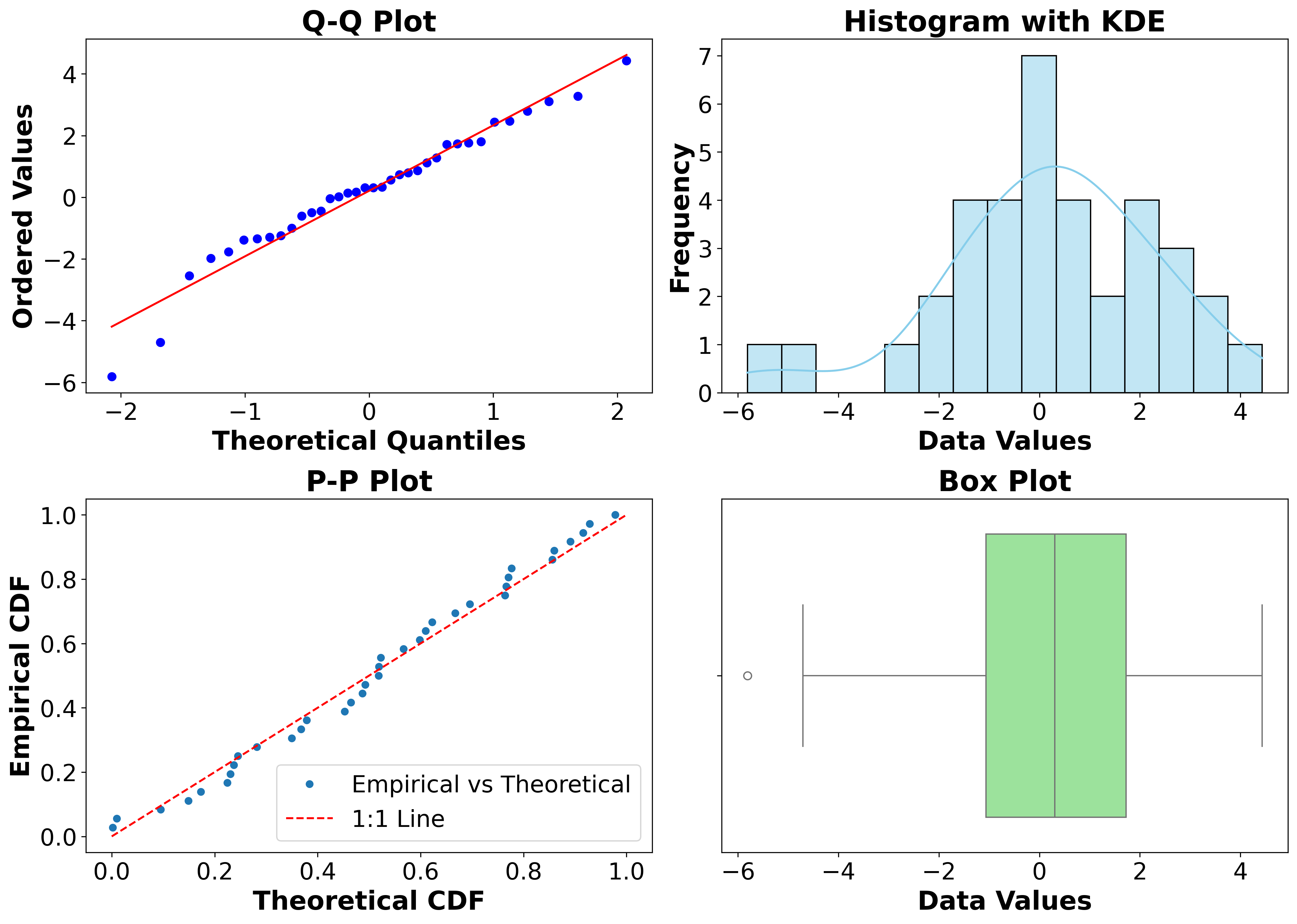}
   \caption{Statistical Analysis using visual plots for Qwen2.5-VL-7B-Instruct - results for images with vase-toaster-indoor. Together with the numerical results in Table \ref{tab:ks_test} and the visual plots in this Figure, we can conclude that the residual error distribution (Table \ref{tab:llava_qwen_comparison}) is random/Gaussian}
   \label{fig:stat_analysisQwen2.5-VL-7B-Instruct_vase-toaster-indoor}
   \hfill
\end{figure}

\clearpage

\subsection{Plots Showing Statistical Analysis for LLaVA 1.5 and 1.6}
\label{app:llava}

\begin{figure}[h]
  \centering
\includegraphics[width=8.5cm, height=7cm]{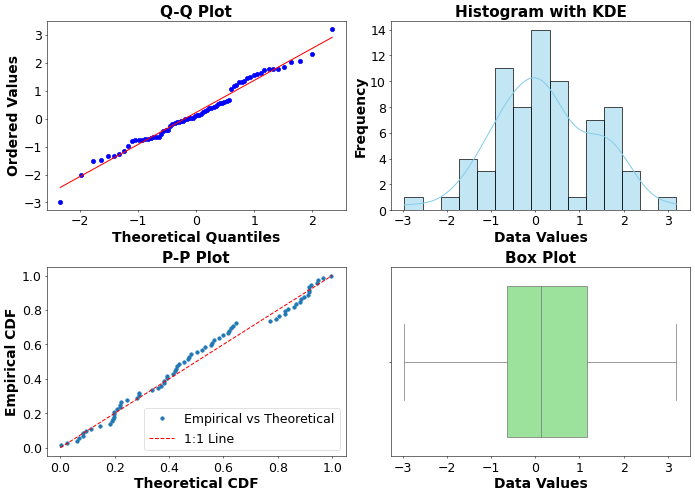}
   \caption{Statistical Analysis using visual plots for LLaVA 1.5 - results for images with dog foregrounds (Scale 2). Together with the numerical results in Table \ref{tab:ks_test} and the visual plots in this Figure, we can conclude that the residual error distribution (Table \ref{tab:llava_qwen_comparison}) is random/Gaussian}
   \label{fig:stat_analysisllava1.5_dog-on-beach_scale2}
   \hfill
\end{figure}


\begin{figure}[h]
  \centering
\includegraphics[width=8.5cm, height=7cm]{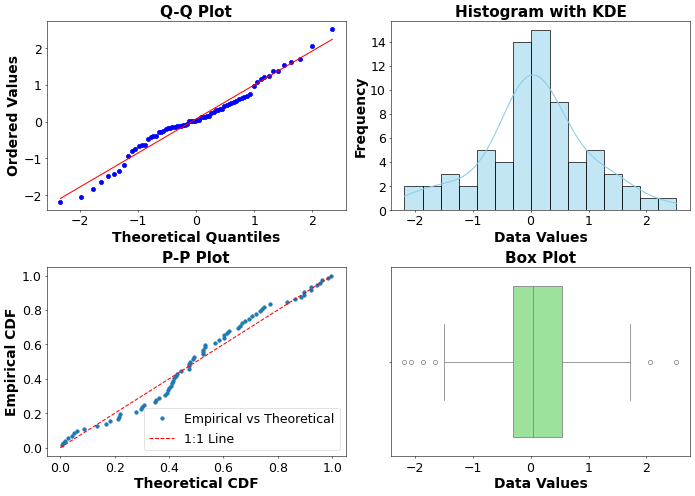}
   \caption{Statistical Analysis using visual plots for LLaVA 1.5 - results for images with dog foregrounds (Scale 3). Together with the numerical results in Table \ref{tab:ks_test} and the visual plots in this Figure, we can conclude that the residual error distribution (Table \ref{tab:llava_qwen_comparison}) is random/Gaussian}
   \label{fig:stat_analysisllava1.5_dog-on-beach_scale3}
   \hfill
\end{figure}

\begin{figure}[h]
  \centering
\includegraphics[width=8.5cm, height=7cm]{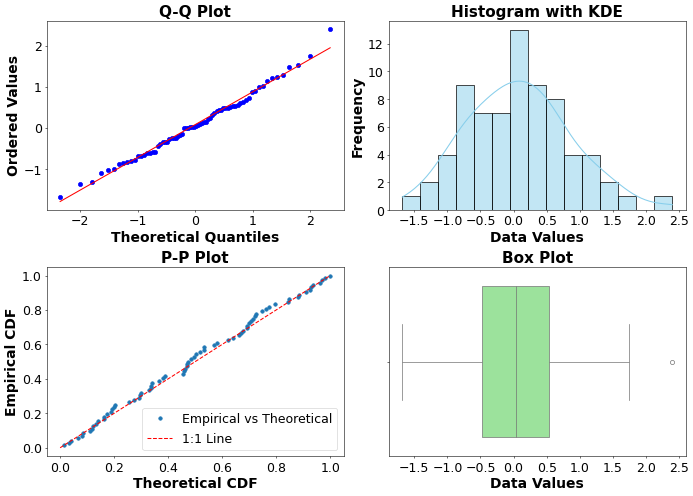}
   \caption{Statistical Analysis using visual plots for LLaVA 1.6 - results for images with dog foregrounds (Scale 1). Together with the numerical results in Table \ref{tab:ks_test} and the visual plots in this Figure, we can conclude that the residual error distribution (Table \ref{tab:llava_qwen_comparison}) is random/Gaussian}
   \label{fig:stat_analysisllava1.6_dog-on-beach_scale1}
   \hfill
\end{figure}

\begin{figure}[h]
  \centering
\includegraphics[width=8.5cm, height=7cm]{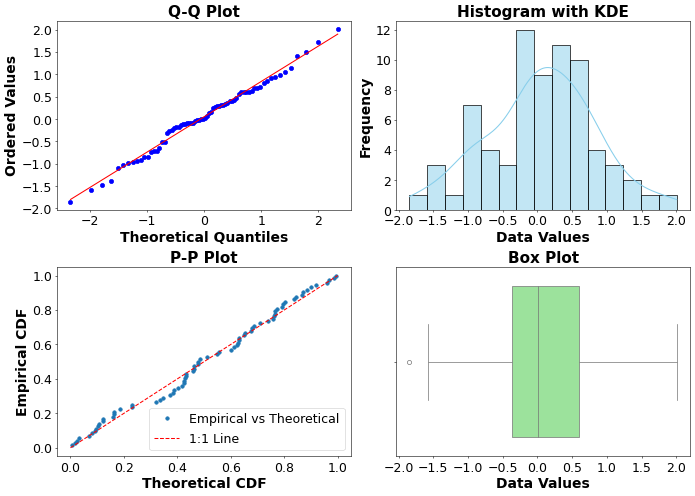}
   \caption{Statistical Analysis using visual plots for LLaVA 1.6 - results for images with dog foregrounds (Scale 2). Together with the numerical results in Table \ref{tab:ks_test} and the visual plots in this Figure, we can conclude that the residual error distribution (Table \ref{tab:llava_qwen_comparison}) is random/Gaussian}
   \label{fig:stat_analysisllava1.6_dog-on-beach_scale2}
   \hfill
\end{figure}

\begin{figure}[h]
  \centering
\includegraphics[width=8.5cm, height=7cm]{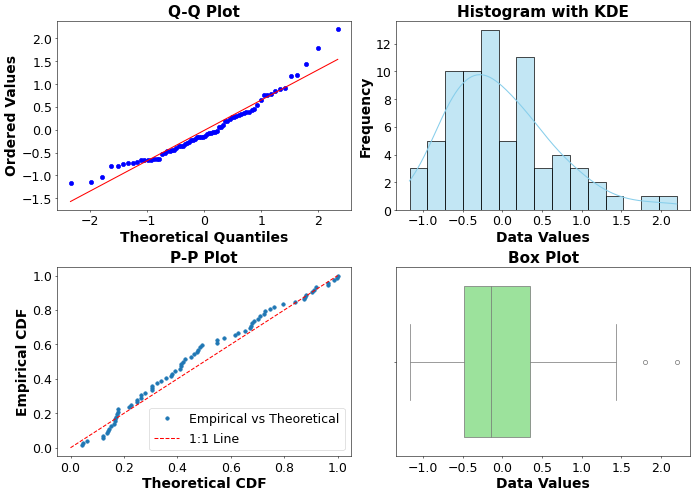}
   \caption{Statistical Analysis using visual plots for LLaVA 1.6 - results for images with dog foregrounds (Scale 3). Together with the numerical results in Table \ref{tab:ks_test} and the visual plots in this Figure, we can conclude that the residual error distribution (Table \ref{tab:llava_qwen_comparison}) is random/Gaussian}
   \label{fig:stat_analysisllava1.6_dog-on-beach_scale3}
   \hfill
\end{figure}

\begin{figure}[h]
  \centering
\includegraphics[width=8.5cm, height=7cm]{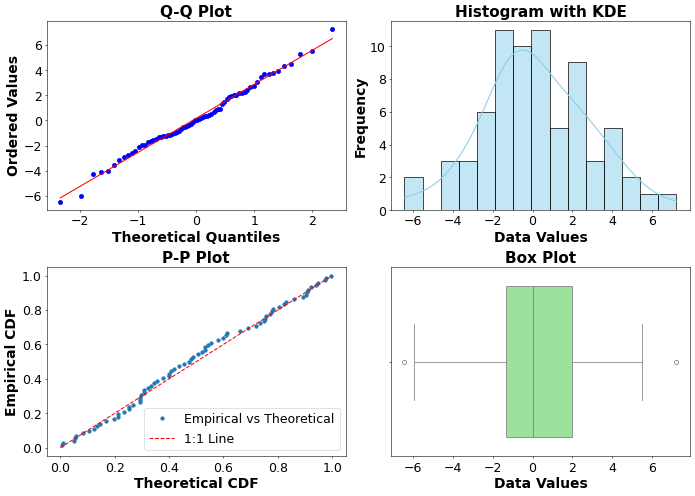}
   \caption{Statistical Analysis using visual plots for LLaVA 1.5 - results for images with lizard foregrounds (Scale 1). Together with the numerical results in Table \ref{tab:ks_test} and the visual plots in this Figure, we can conclude that the residual error distribution (Table \ref{tab:llava_qwen_comparison}) is random/Gaussian}
   \label{fig:stat_analysisllava1.5_lizard_on_fish_scale1}
   \hfill
\end{figure}

\begin{figure}[h]
  \centering
\includegraphics[width=8.5cm, height=7cm]{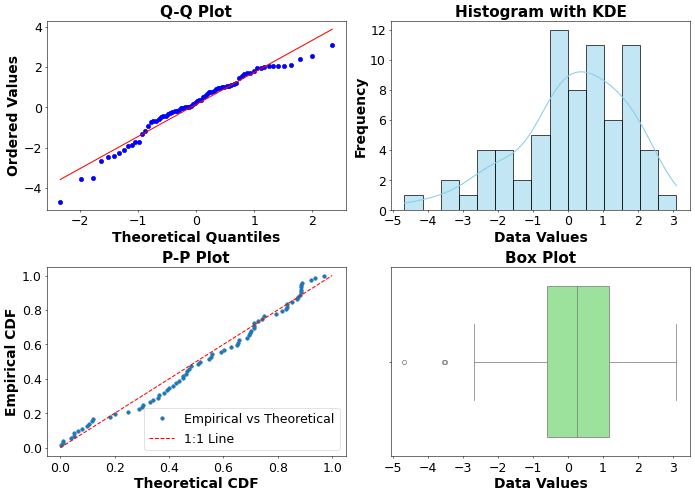}
   \caption{Statistical Analysis using visual plots for LLaVA 1.5 - results for images with lizard foregrounds (Scale 2). Together with the numerical results in Table \ref{tab:ks_test} and the visual plots in this Figure, we can conclude that the residual error distribution (Table \ref{tab:llava_qwen_comparison}) is random/Gaussian}
   \label{fig:stat_analysisllava1.5_lizard_on_fish_scale2}
   \hfill
\end{figure}


\begin{figure}[h]
  \centering
\includegraphics[width=8.5cm, height=7cm]{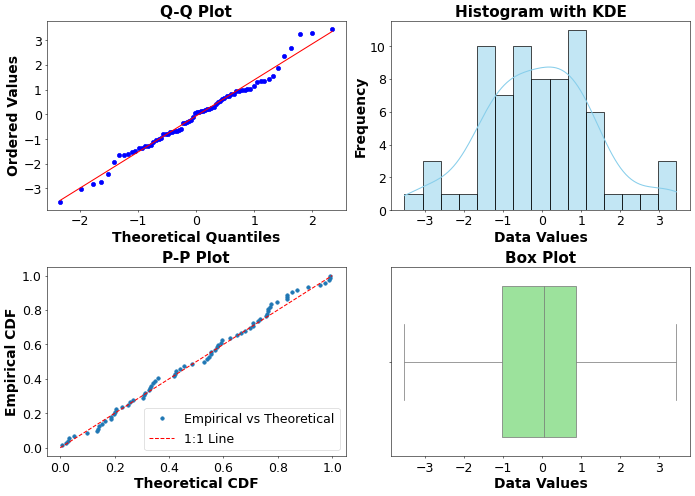}
   \caption{Statistical Analysis using visual plots for LLaVA 1.5 - results for images with lizard foregrounds (Scale 3). Together with the numerical results in Table \ref{tab:ks_test} and the visual plots in this Figure, we can conclude that the residual error distribution (Table \ref{tab:llava_qwen_comparison}) is random/Gaussian}
   \label{fig:stat_analysisllava1.5_lizard_on_fish_scale3}
   \hfill
\end{figure}

\begin{figure}[h]
  \centering
\includegraphics[width=8.5cm, height=7cm]{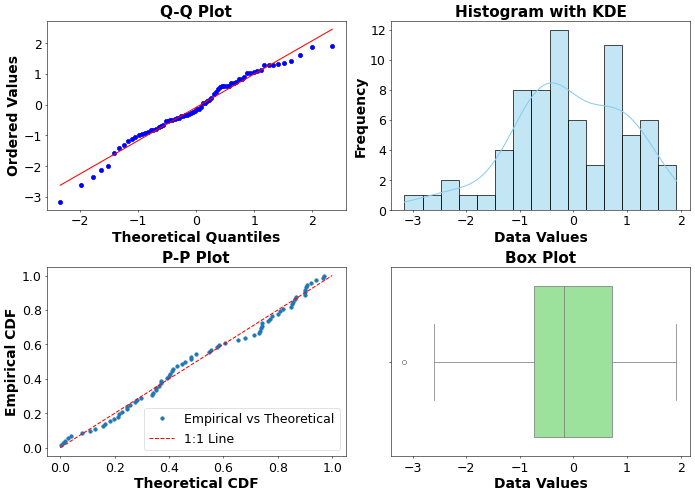}
   \caption{Statistical Analysis using visual plots for LLaVA 1.6 - results for images with lizard foregrounds (Scale 1). Together with the numerical results in Table \ref{tab:ks_test} and the visual plots in this Figure, we can conclude that the residual error distribution (Table \ref{tab:llava_qwen_comparison}) is random/Gaussian}
   \label{fig:stat_analysisllava1.6_lizard_on_fish_scale1}
   \hfill
\end{figure}

\begin{figure}[h]
  \centering
\includegraphics[width=8.5cm, height=7cm]{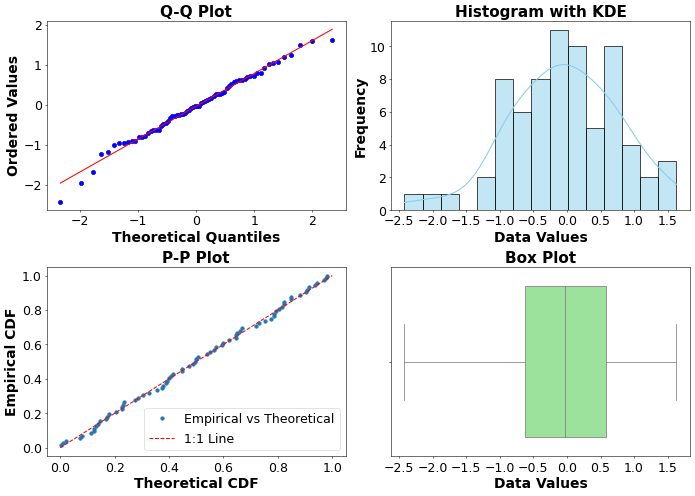}
   \caption{Statistical Analysis using visual plots for LLaVA 1.6 - results for images with lizard foregrounds (Scale 2). Together with the numerical results in Table \ref{tab:ks_test} and the visual plots in this Figure, we can conclude that the residual error distribution (Table \ref{tab:llava_qwen_comparison}) is random/Gaussian}
   \label{fig:stat_analysisllava1.6_lizard_on_fish_scale2}
   \hfill
\end{figure}

\begin{figure}[h]
  \centering
\includegraphics[width=8.5cm, height=7cm]{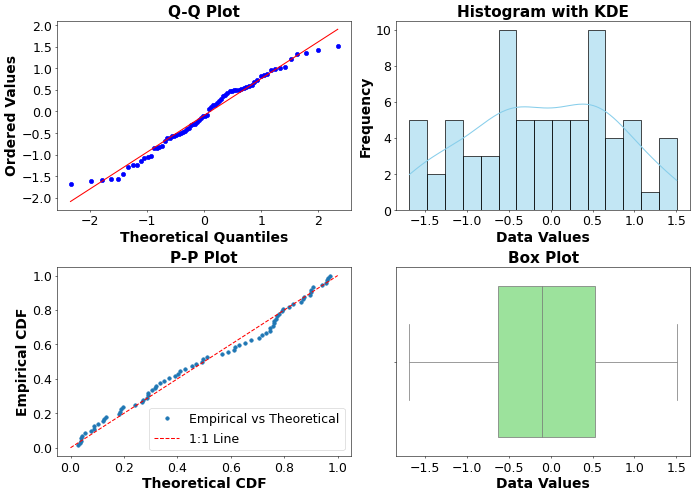}
   \caption{Statistical Analysis using visual plots for LLaVA 1.6 - results for images with lizard foregrounds (Scale 3). Together with the numerical results in Table \ref{tab:ks_test} and the visual plots in this Figure, we can conclude that the residual error distribution (Table \ref{tab:llava_qwen_comparison}) is random/Gaussian}
   \label{fig:stat_analysisllava1.6_lizard_on_fish_scale3}
   \hfill
\end{figure}

\begin{figure}[h]
  \centering
\includegraphics[width=8.5cm, height=7cm]{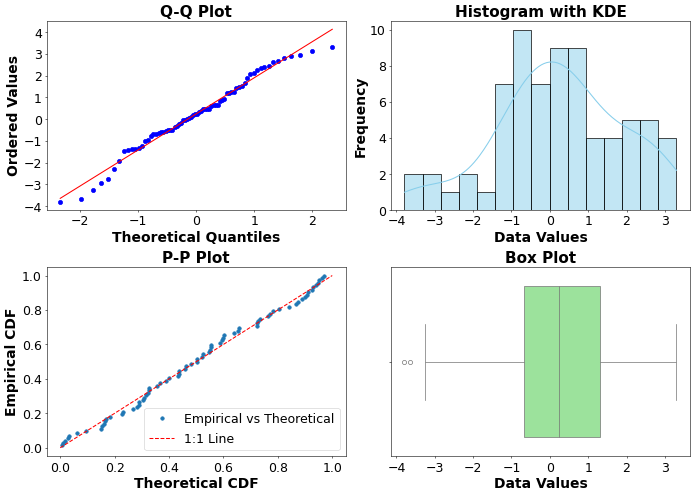}
   \caption{Statistical Analysis using visual plots for LLaVA 1.5 - results for images with train foregrounds (Scale 1). Together with the numerical results in Table \ref{tab:ks_test} and the visual plots in this Figure, we can conclude that the residual error distribution (Table \ref{tab:llava_qwen_comparison}) is random/Gaussian}
   \label{fig:stat_analysisllava1.5_train_on_indoor_scale1}
   \hfill
\end{figure}

\begin{figure}[h]
  \centering
\includegraphics[width=8.5cm, height=7cm]{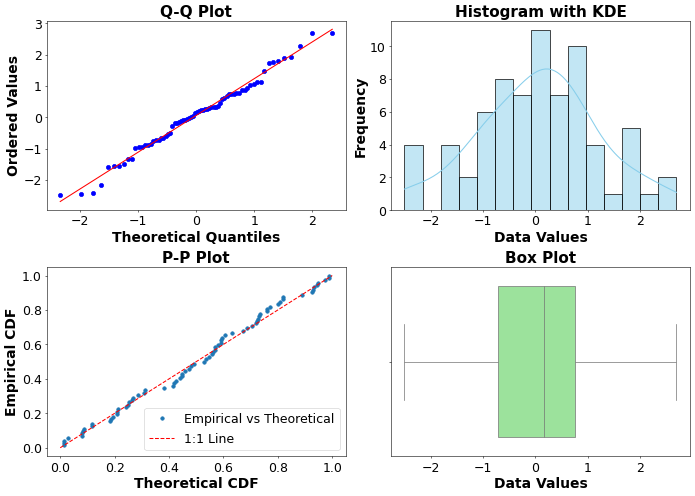}
   \caption{Statistical Analysis using visual plots for LLaVA 1.5 - results for images with train foregrounds (Scale 2). Together with the numerical results in Table \ref{tab:ks_test} and the visual plots in this Figure, we can conclude that the residual error distribution (Table \ref{tab:llava_qwen_comparison}) is random/Gaussian}
   \label{fig:stat_analysisllava1.5_train_on_indoor_scale2}
   \hfill
\end{figure}


\begin{figure}[h]
  \centering
\includegraphics[width=8.5cm, height=7cm]{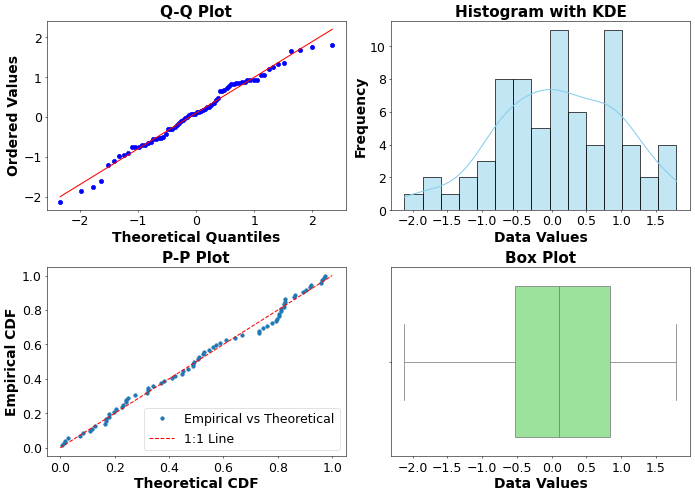}
   \caption{Statistical Analysis using visual plots for LLaVA 1.5 - results for images with train foregrounds (Scale 3). Together with the numerical results in Table \ref{tab:ks_test} and the visual plots in this Figure, we can conclude that the residual error distribution (Table \ref{tab:llava_qwen_comparison}) is random/Gaussian}
   \label{fig:stat_analysisllava1.5_train_on_indoor_scale3}
   \hfill
\end{figure}

\begin{figure}[h]
  \centering
\includegraphics[width=8.5cm, height=7cm]{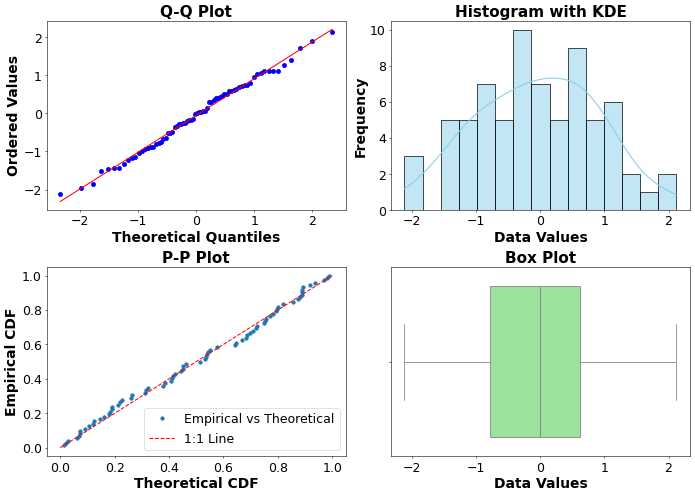}
   \caption{Statistical Analysis using visual plots for LLaVA 1.6 - results for images with train foregrounds (Scale 1). Together with the numerical results in Table \ref{tab:ks_test} and the visual plots in this Figure, we can conclude that the residual error distribution (Table \ref{tab:llava_qwen_comparison}) is random/Gaussian}
   \label{fig:stat_analysisllava1.6_train_on_indoor_scale1}
   \hfill
\end{figure}

\begin{figure}[h]
  \centering
\includegraphics[width=8.5cm, height=7cm]{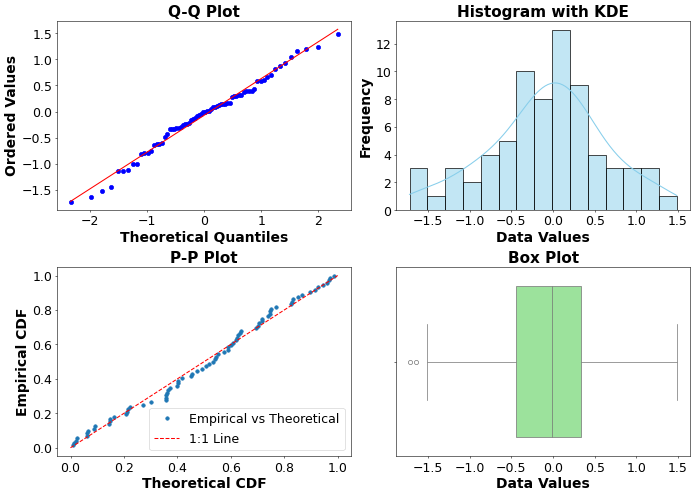}
   \caption{Statistical Analysis using visual plots for LLaVA 1.6 - results for images with train foregrounds (Scale 2). Together with the numerical results in Table \ref{tab:ks_test} and the visual plots in this Figure, we can conclude that the residual error distribution (Table \ref{tab:llava_qwen_comparison}) is random/Gaussian}
   \label{fig:stat_analysisllava1.6_train_on_indoor_scale2}
   \hfill
\end{figure}

\begin{figure}[h]
  \centering
\includegraphics[width=8.5cm, height=7cm]{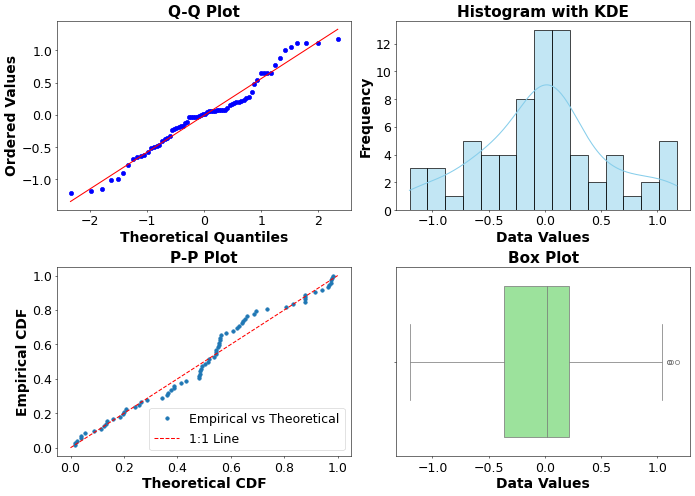}
   \caption{Statistical Analysis using visual plots for LLaVA 1.6 - results for images with train foregrounds (Scale 3). Together with the numerical results in Table \ref{tab:ks_test} and the visual plots in this Figure, we can conclude that the residual error distribution (Table \ref{tab:llava_qwen_comparison}) is random/Gaussian}
   \label{fig:stat_analysisllava1.6_train_on_indoor_scale3}
   \hfill
\end{figure}

\clearpage
\subsection{LLaVA 1.5 and 1.6 query responses}

 \begin{table}[h!]
    \centering
    \footnotesize
    \setlength{\tabcolsep}{2pt} 
    \begin{tabular*}{\columnwidth}{@{\extracolsep{\fill}}cccc}
        \toprule
        \multicolumn{2}{c}{\textbf{LLaVA 1.5}} & \multicolumn{2}{c}{\textbf{LLaVA 1.6}} \\
        \cmidrule(lr){1-2} \cmidrule(lr){3-4}
        \textbf{Angle ($^{\circ}$)} & \textbf{Count} & \textbf{Angle ($^{\circ}$)} & \textbf{Count} \\
        \midrule
        90 & 25 & 90 & 40 \\
        not possible & 10 & not possible & 17 \\
        45 & 9 & 0 & 6 \\
        0 & 8 & no answer & 4 \\
        no answer & 4 & 180 & 2 \\
        \midrule
        \multicolumn{2}{@{}l}{\textbf{Summary Performance}} & \textbf{LLaVA 1.5} & \textbf{LLaVA 1.6} \\
        \midrule
        \multicolumn{2}{l}{Correct Answers $(|diff| \le 5)$} & 3 & 2 \\
        \multicolumn{2}{l}{Incorrect Answers $(|diff| > 5)$} & 55 & 49 \\
        \midrule
        \multicolumn{2}{l}{Correct Answers $(|diff| \le 20)$} & 7 & 7 \\
        \multicolumn{2}{l}{Incorrect Answers $(|diff| > 20)$} & 51 & 44 \\
        \midrule
        \multicolumn{2}{l}{Correct Answers $(|diff| \le 45)$} & 16 & 17 \\
        \multicolumn{2}{l}{Incorrect Answers $(|diff| > 45)$} & 42 & 34 \\
        \bottomrule
    \end{tabular*}
    \vspace{-2mm}
    \caption{\label{tab:llama-llava-72}LLaVA-LLaMA results for the 72 test samples with the beach scene in the background. LLaMA frequently responds that the 2D orientation is 90$^{\circ}$ or ``not possible to determine". The number of correct answers (among valid responses) is very low, even with high thresholds}
\label{tab:llava_llm_metrics}
\end{table}

\newpage

\FloatBarrier
\section{Orientation Encoding Properties}
\subsection{Feature Substitution Plots for LLaVA-OneVision and Qwen2.5-VL-7B-Instruct}
\label{app:ftr-subs-llavaOV-qwen}

\begin{figure*}[t]
    \begin{subfigure}{0.33\textwidth}
        \includegraphics[width=\linewidth]{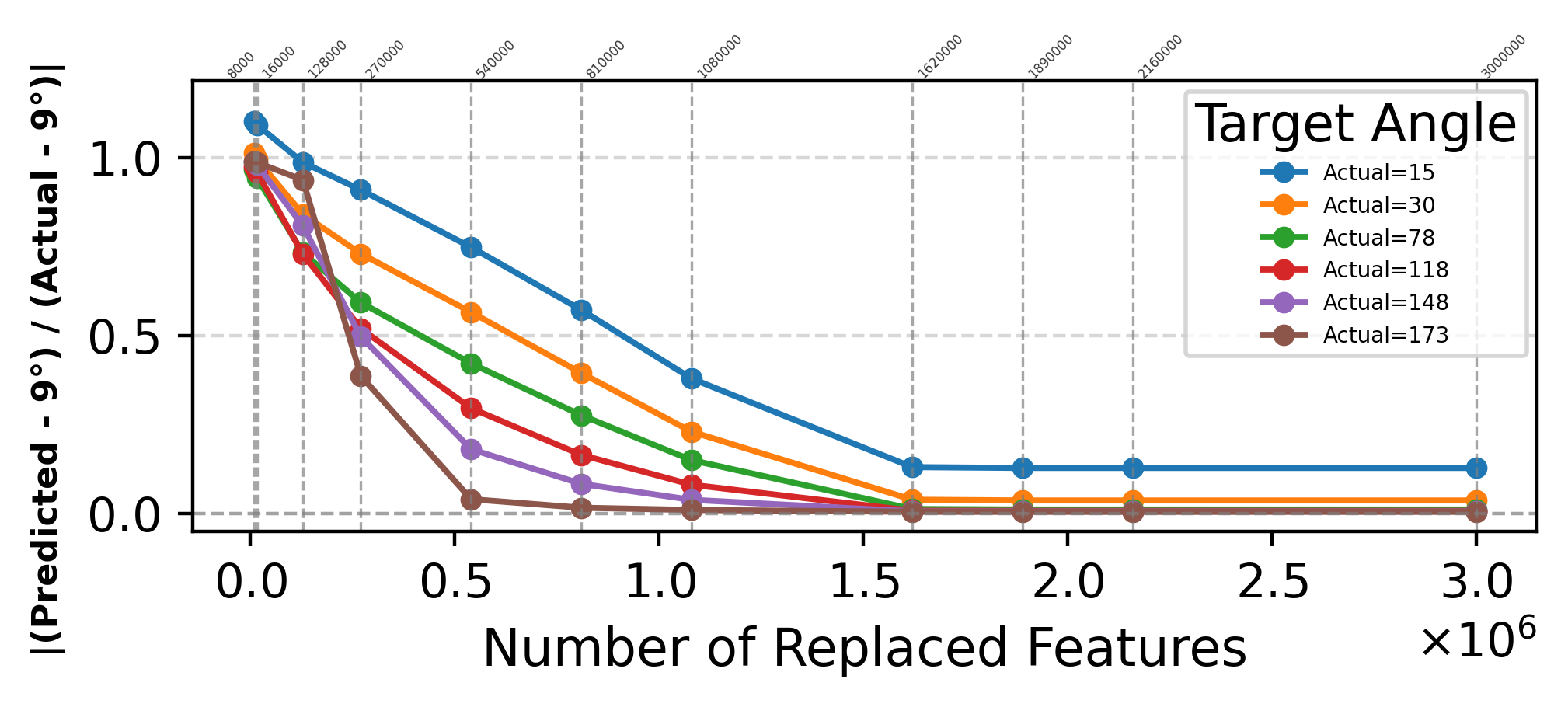}
        \caption{Ordered By Model Weight}
    \end{subfigure}%
    \hspace{-0.5em} 
    \begin{subfigure}{0.33\textwidth}
        \includegraphics[width=\linewidth]{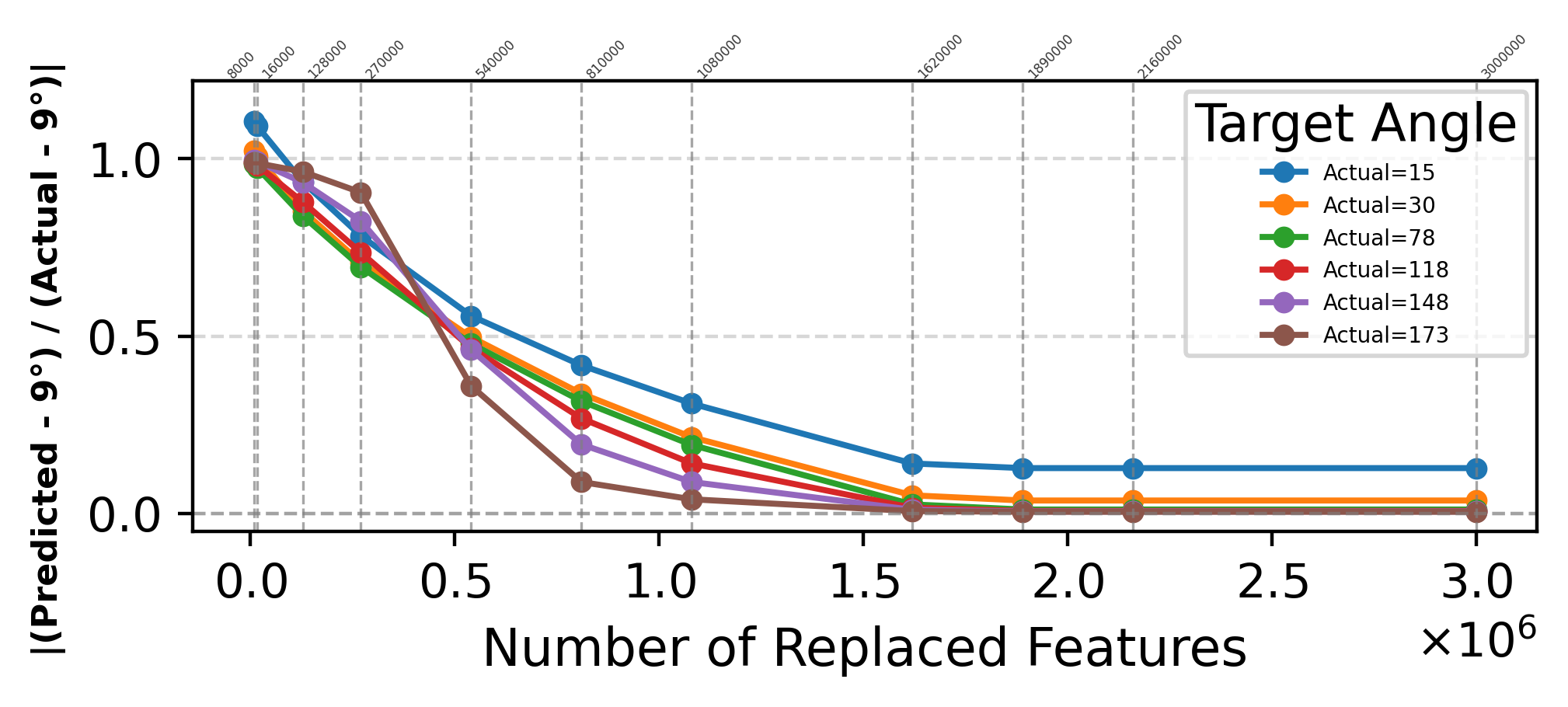}
        \caption{Ordered By Value Difference}
    \end{subfigure}%
    \hspace{-0.5em} 
    \begin{subfigure}{0.33\textwidth}
        \includegraphics[width=\linewidth]{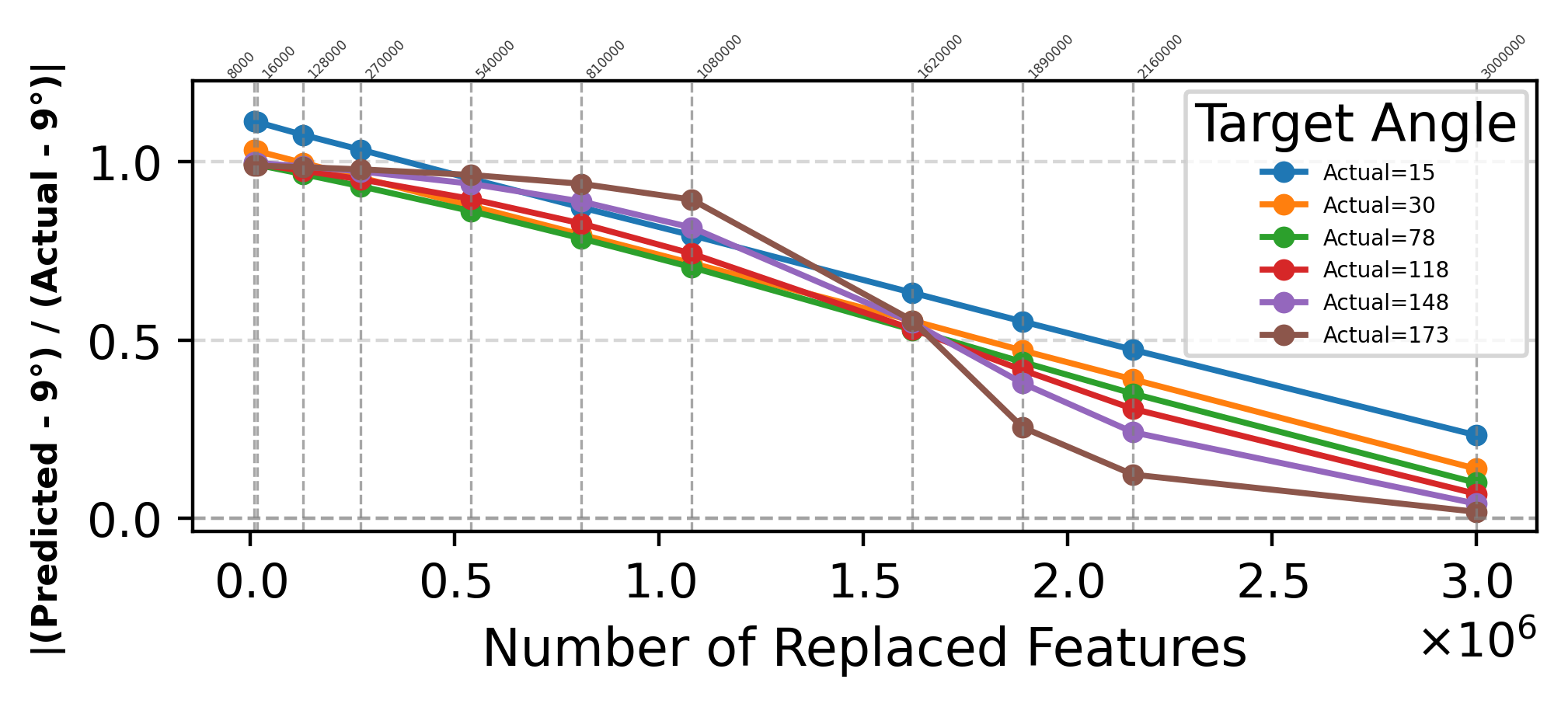}
        \caption{Picked Randomly}
    \end{subfigure}
   \vspace{-2mm}
    \caption{Incremental feature substitution for LLaVA-OneVision on images with the lizard scene. No matter how the features are selected (according to the magnitude of the weights in the regressor or the absolute difference between anchor and target feature values, or randomly). 540,000 features or more must be replaced to fool the predictor. (Note that the x-axis is the number of feature substitutions times $10^6$.) This implies the orientation information is highly diffuse.}
   \vspace{-2mm}
    \label{fig:patch_analysis_llava-ov_lizard}
\end{figure*}
\begin{figure*}[t]
    \begin{subfigure}{0.33\textwidth}
        \includegraphics[width=\linewidth]{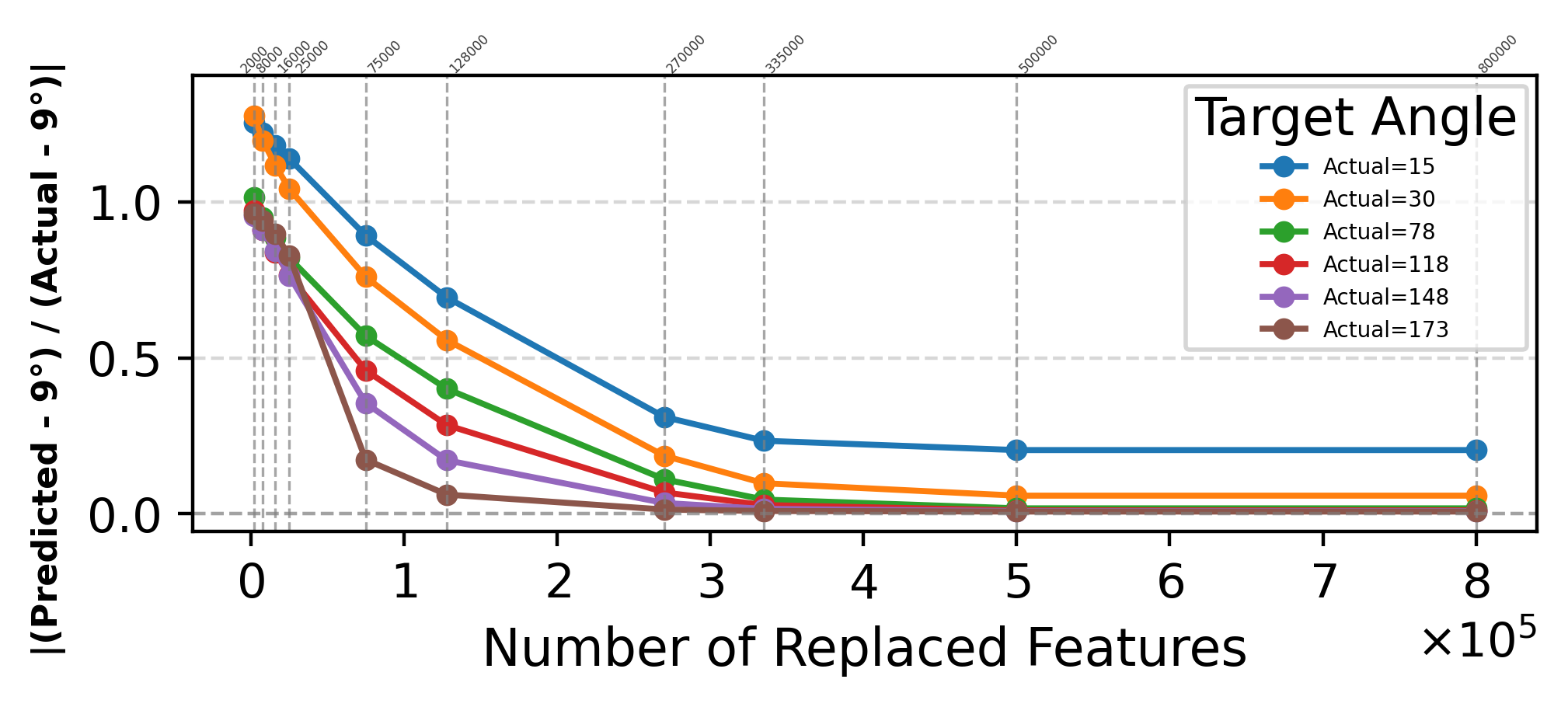}
        \caption{Ordered By Model Weight}
    \end{subfigure}%
    \hspace{-0.5em} 
    \begin{subfigure}{0.33\textwidth}
        \includegraphics[width=\linewidth]{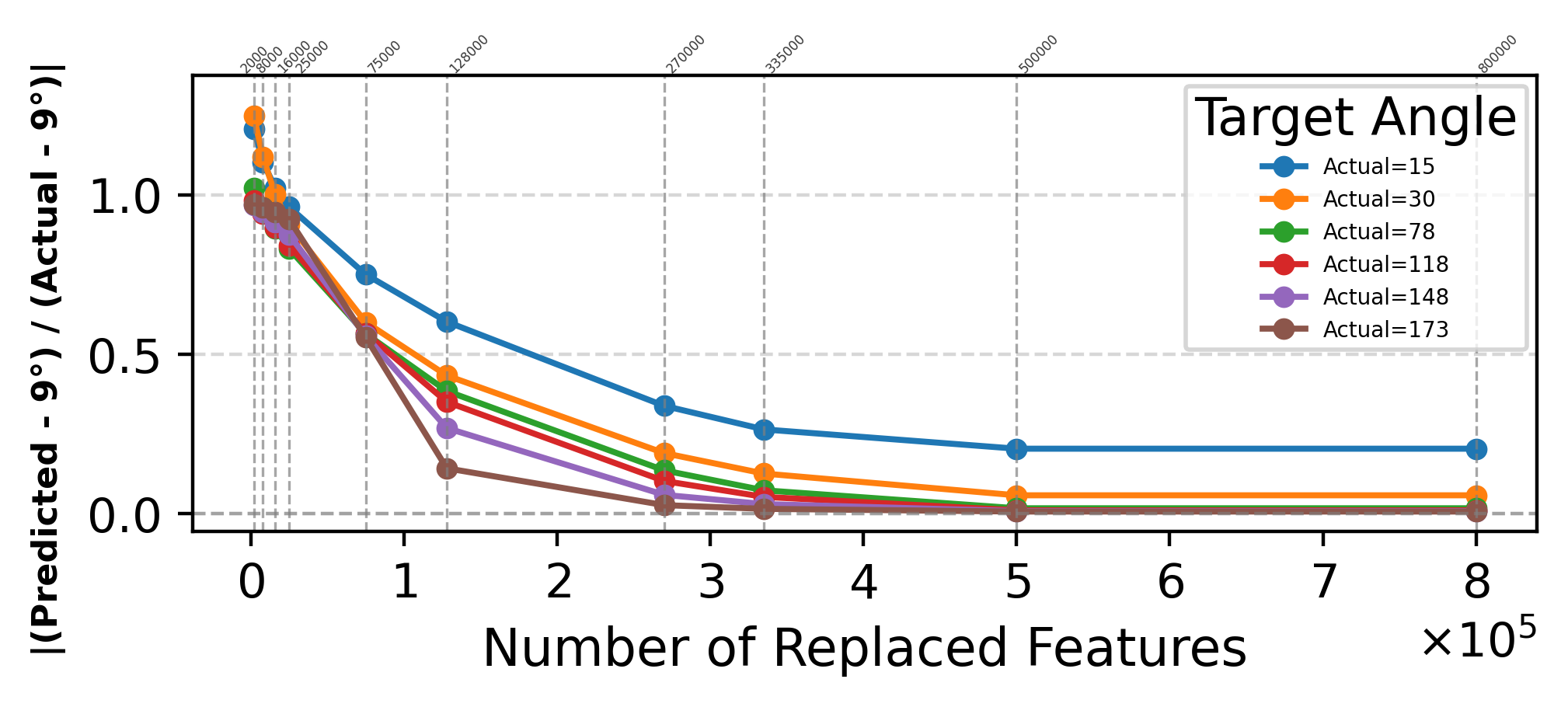}
        \caption{Ordered By Value Difference}
    \end{subfigure}%
    \hspace{-0.5em} 
    \begin{subfigure}{0.33\textwidth}
        \includegraphics[width=\linewidth]{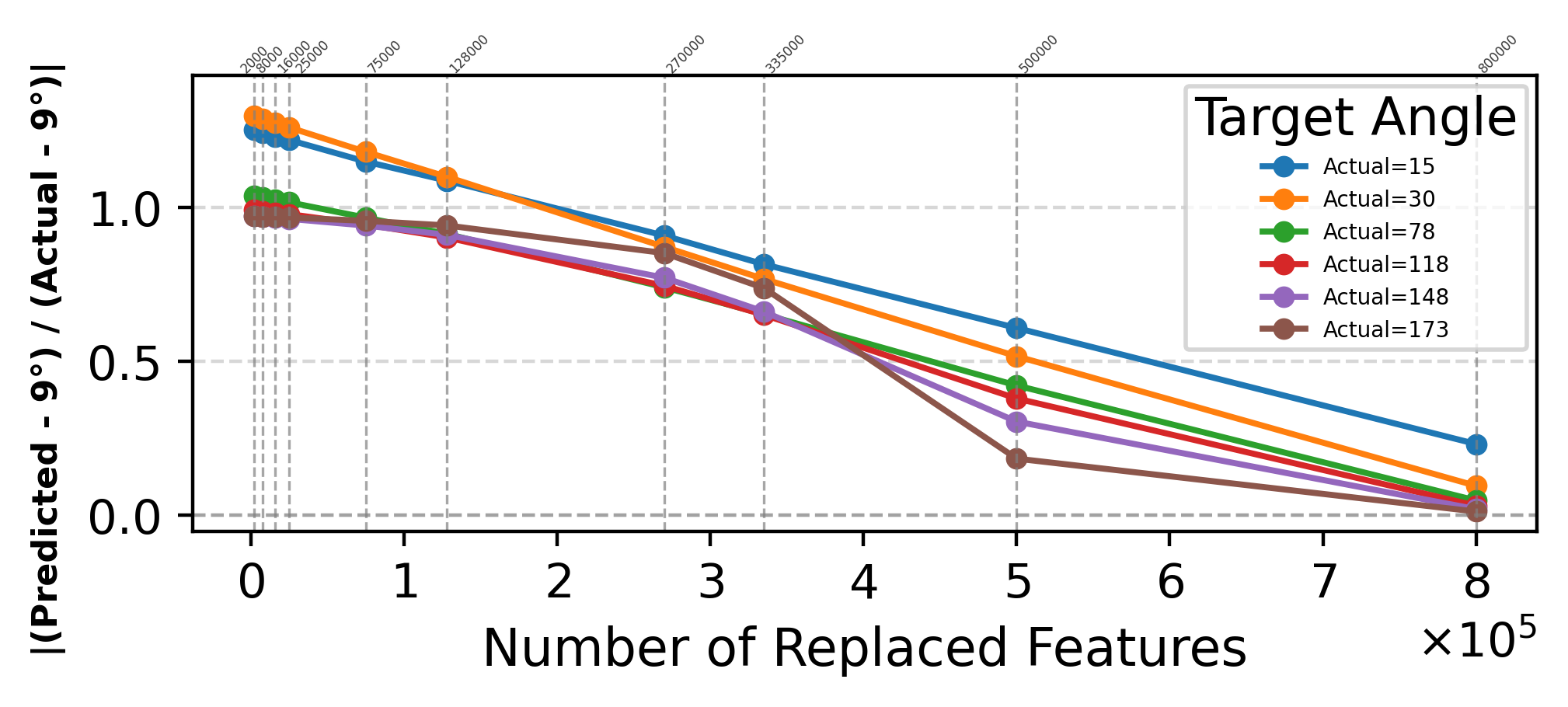}
        \caption{Picked Randomly}
    \end{subfigure}
   \vspace{-2mm}
    \caption{Incremental feature substitution for Qwen2.5-VL-7B-Instruct on images with the lizard scene. No matter how the features are selected (according to the magnitude of the weights in the regressor or the absolute difference between anchor and target feature values, or randomly). 128,000 features or more must be replaced to fool the predictor. (Note that the x-axis is the number of feature substitutions times $10^5$.) This implies the orientation information is highly diffuse.}

    \label{fig:patch_analysis_qwen_lizard}
\end{figure*}
\newpage

\begin{figure*}[H]
    \begin{subfigure}{0.33\textwidth}
        \includegraphics[width=\linewidth]{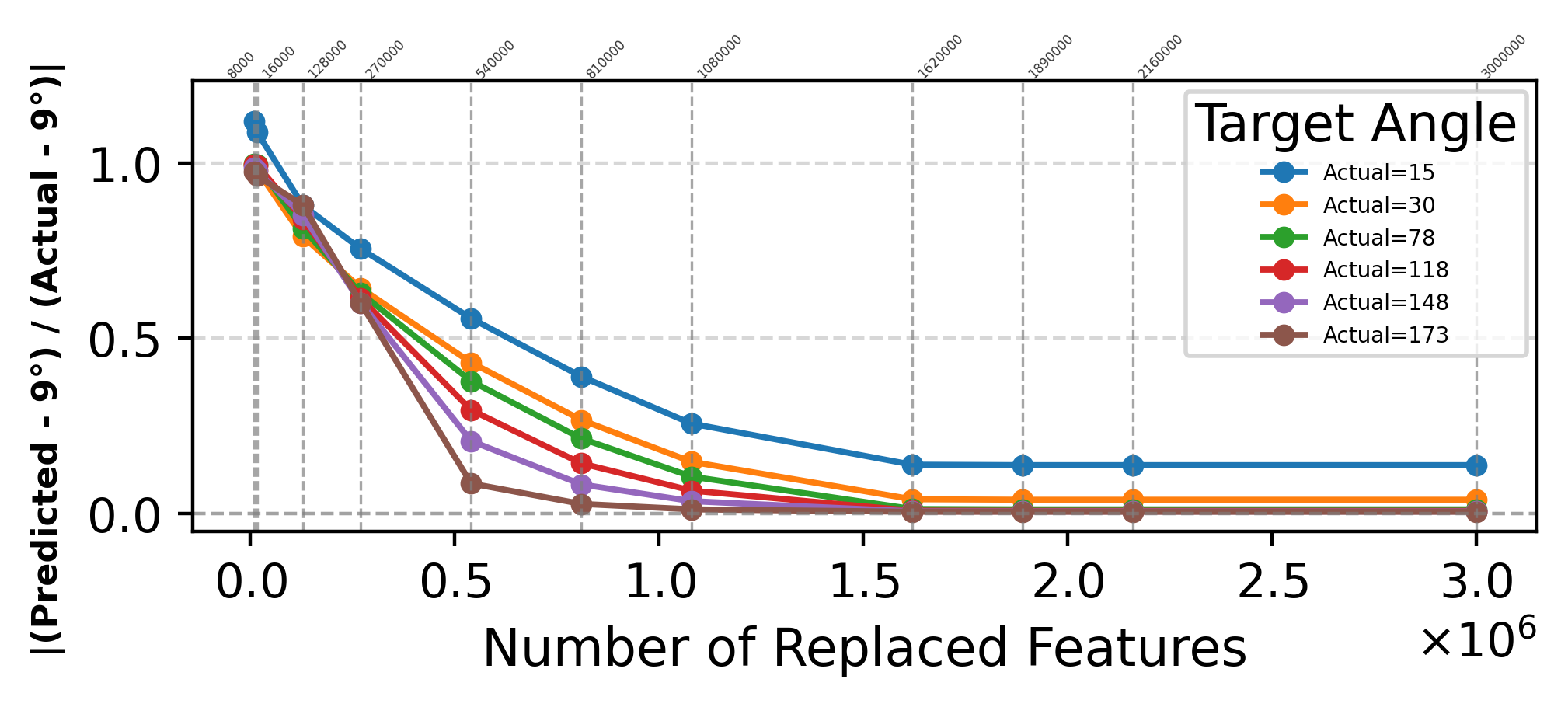}
        \caption{Ordered By Model Weight}
    \end{subfigure}%
    \hspace{-0.5em} 
    \begin{subfigure}{0.33\textwidth}
        \includegraphics[width=\linewidth]{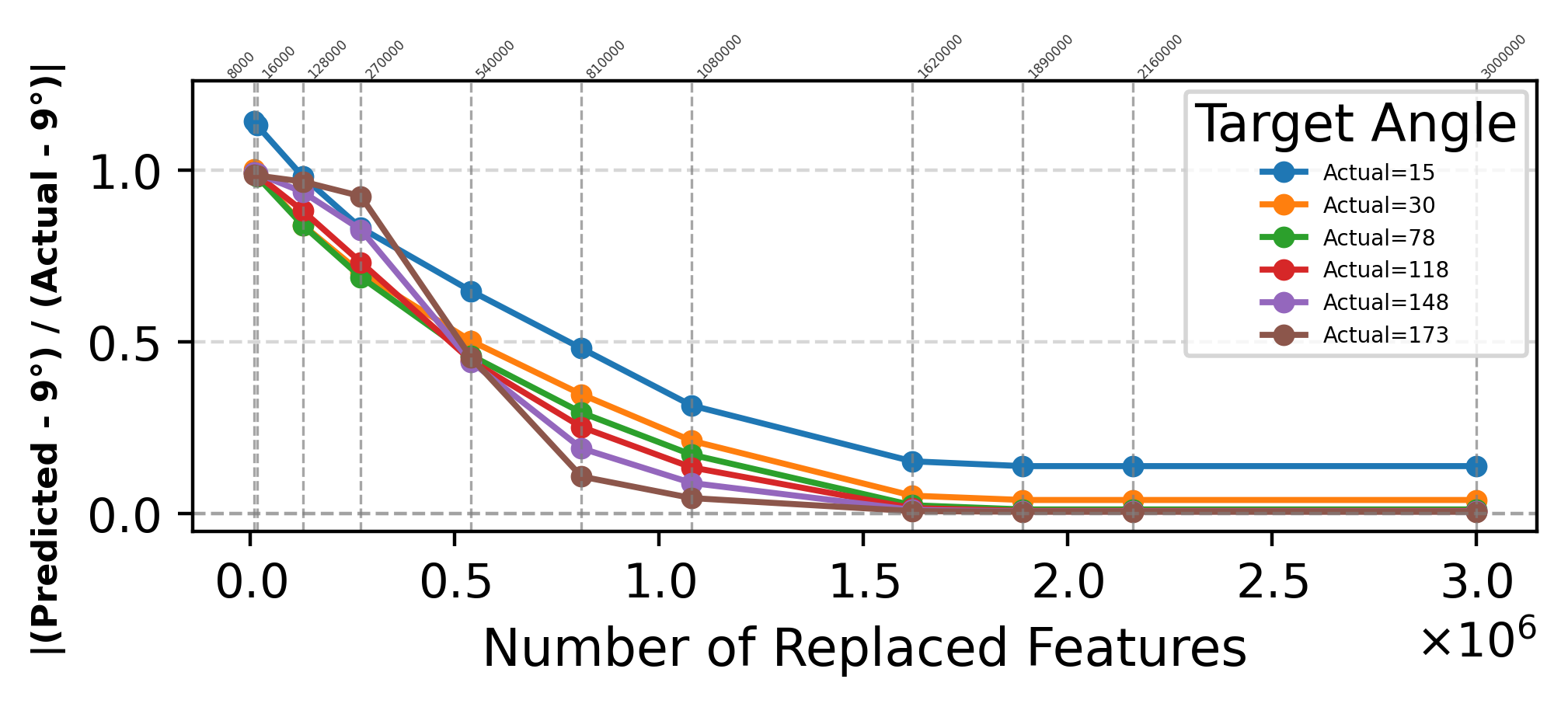}
        \caption{Ordered By Value Difference}
    \end{subfigure}%
    \hspace{-0.5em} 
    \begin{subfigure}{0.33\textwidth}
        \includegraphics[width=\linewidth]{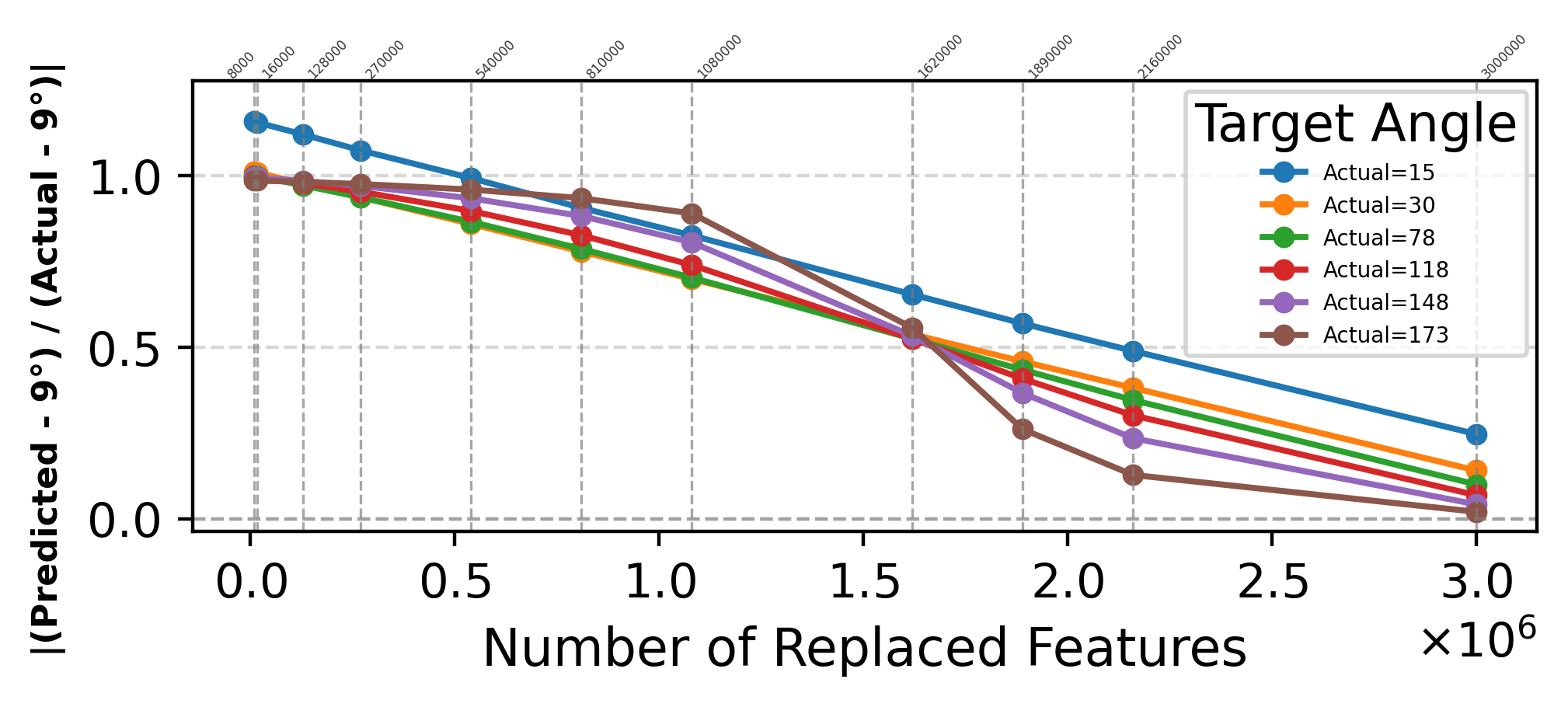}
        \caption{Picked Randomly}
    \end{subfigure}
   \vspace{-2mm}
    \caption{Incremental feature substitution for LLaVA-OneVision on images with the train scene. No matter how the features are selected (according to the magnitude of the weights in the regressor or the absolute difference between anchor and target feature values, or randomly). 540,000 features or more must be replaced to fool the predictor. (Note that the x-axis is the number of feature substitutions times $10^6$.) This implies the orientation information is highly diffuse.}
   \vspace{-2mm}
    \label{fig:patch_analysis_llava-ov_train}
\end{figure*}

\begin{figure*}[h!]
    \begin{subfigure}{0.33\textwidth}
        \includegraphics[width=\linewidth]{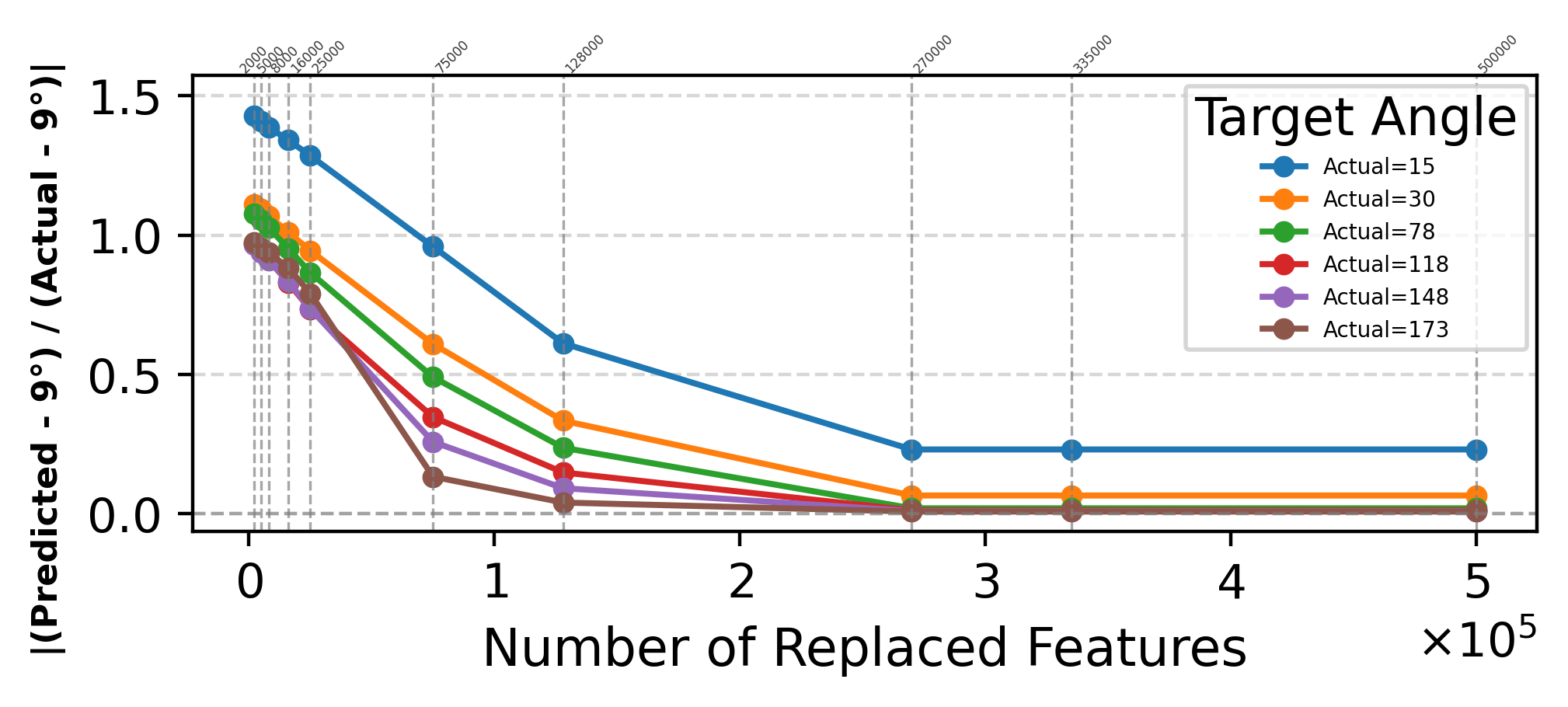}
        \caption{Ordered By Model Weight}
    \end{subfigure}%
    \hspace{-0.5em} 
    \begin{subfigure}{0.33\textwidth}
        \includegraphics[width=\linewidth]{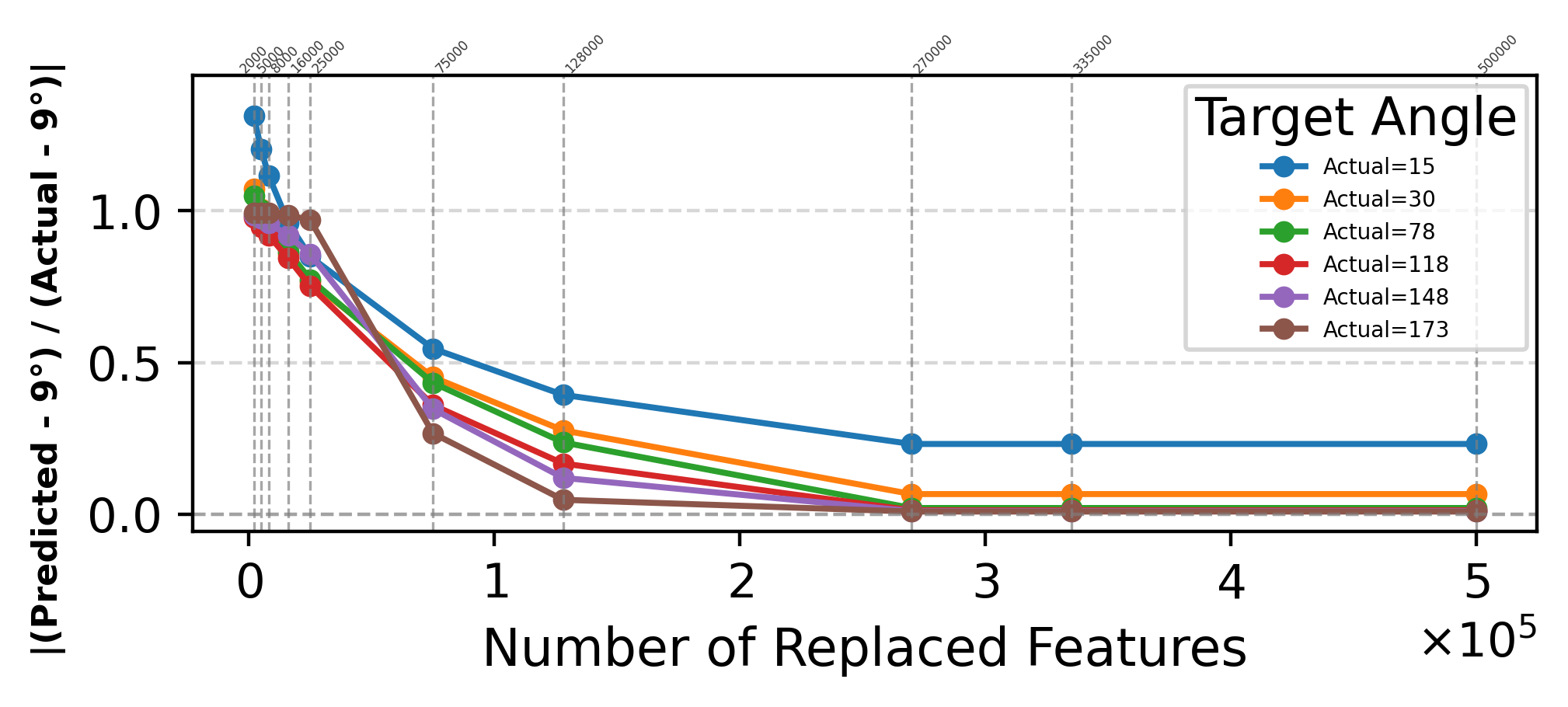}
        \caption{Ordered By Value Difference}
    \end{subfigure}%
    \hspace{-0.5em} 
    \begin{subfigure}{0.33\textwidth}
        \includegraphics[width=\linewidth]{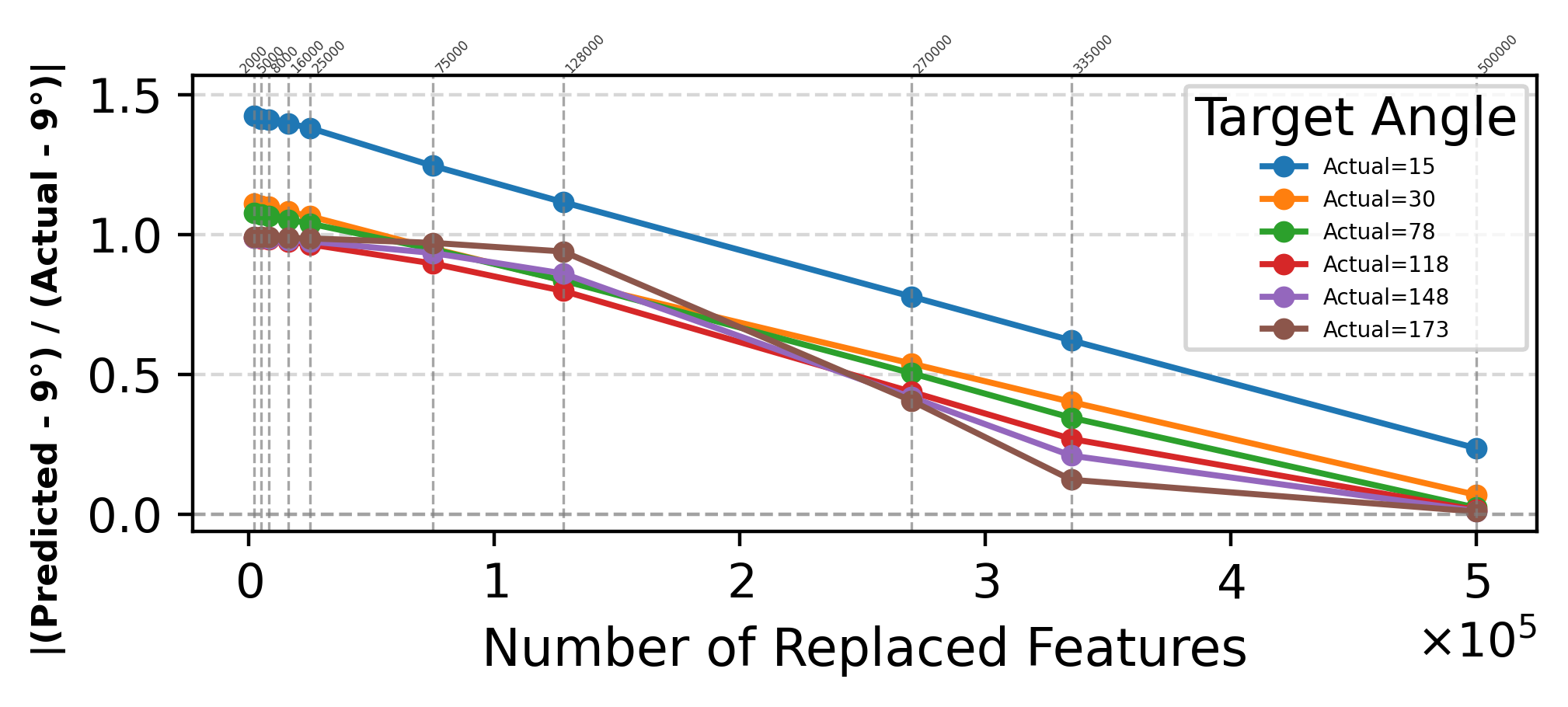}
        \caption{Picked Randomly}
    \end{subfigure}
   \vspace{-2mm}
    \caption{Incremental feature substitution for Qwen2.5-VL-7B-Instruct on images with the train scene. No matter how the features are selected (according to the magnitude of the weights in the regressor or the absolute difference between anchor and target feature values, or randomly). 128,000 features or more must be replaced to fool the predictor. (Note that the x-axis is the number of feature substitutions times $10^5$.) This implies the orientation information is highly diffuse.}
   \vspace{-2mm}
    \label{fig:patch_analysis_qwen_train}
\end{figure*}
\newpage
\begin{figure*}[h!]
    \begin{subfigure}{0.33\textwidth}
        \includegraphics[width=\linewidth]{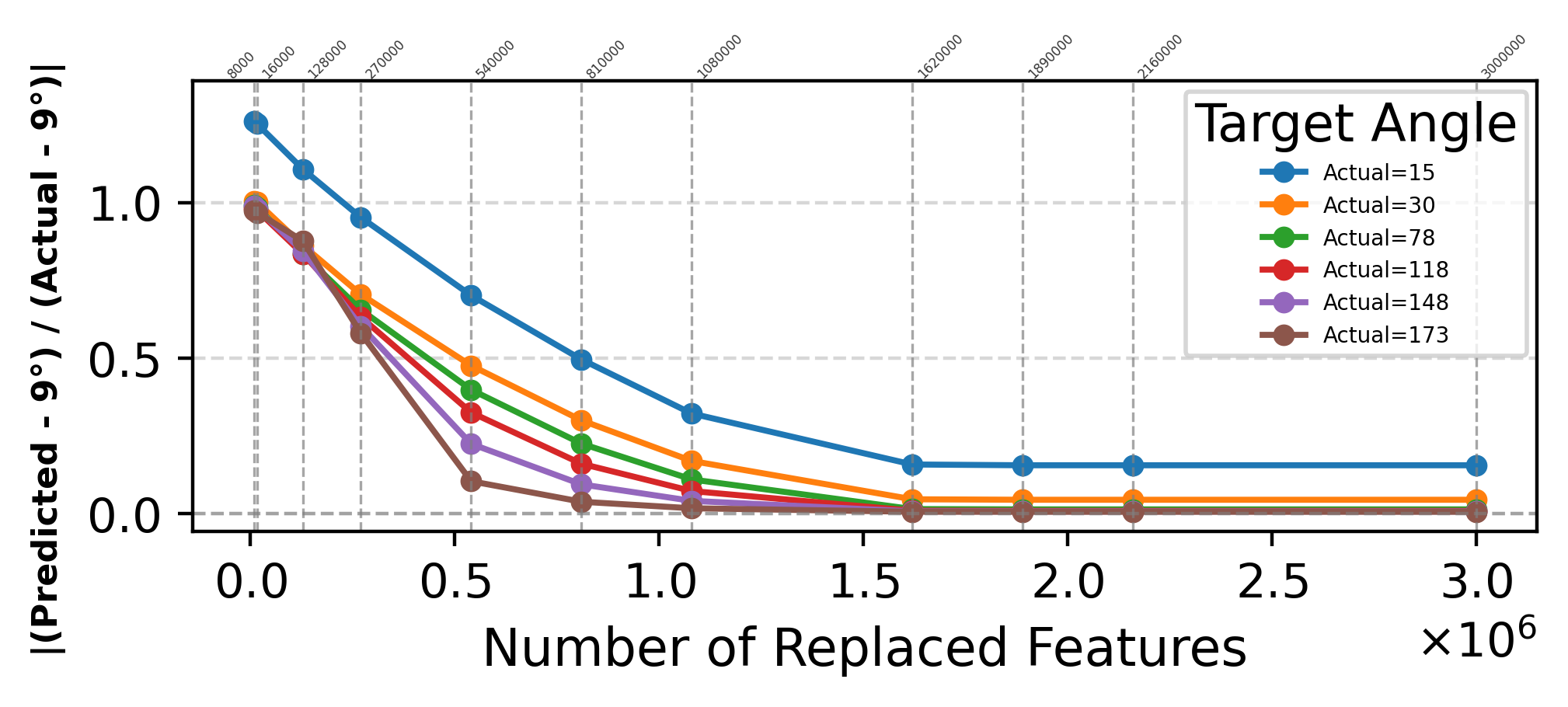}
        \caption{Ordered By Model Weight}
    \end{subfigure}%
    \hspace{-0.5em} 
    \begin{subfigure}{0.33\textwidth}
        \includegraphics[width=\linewidth]{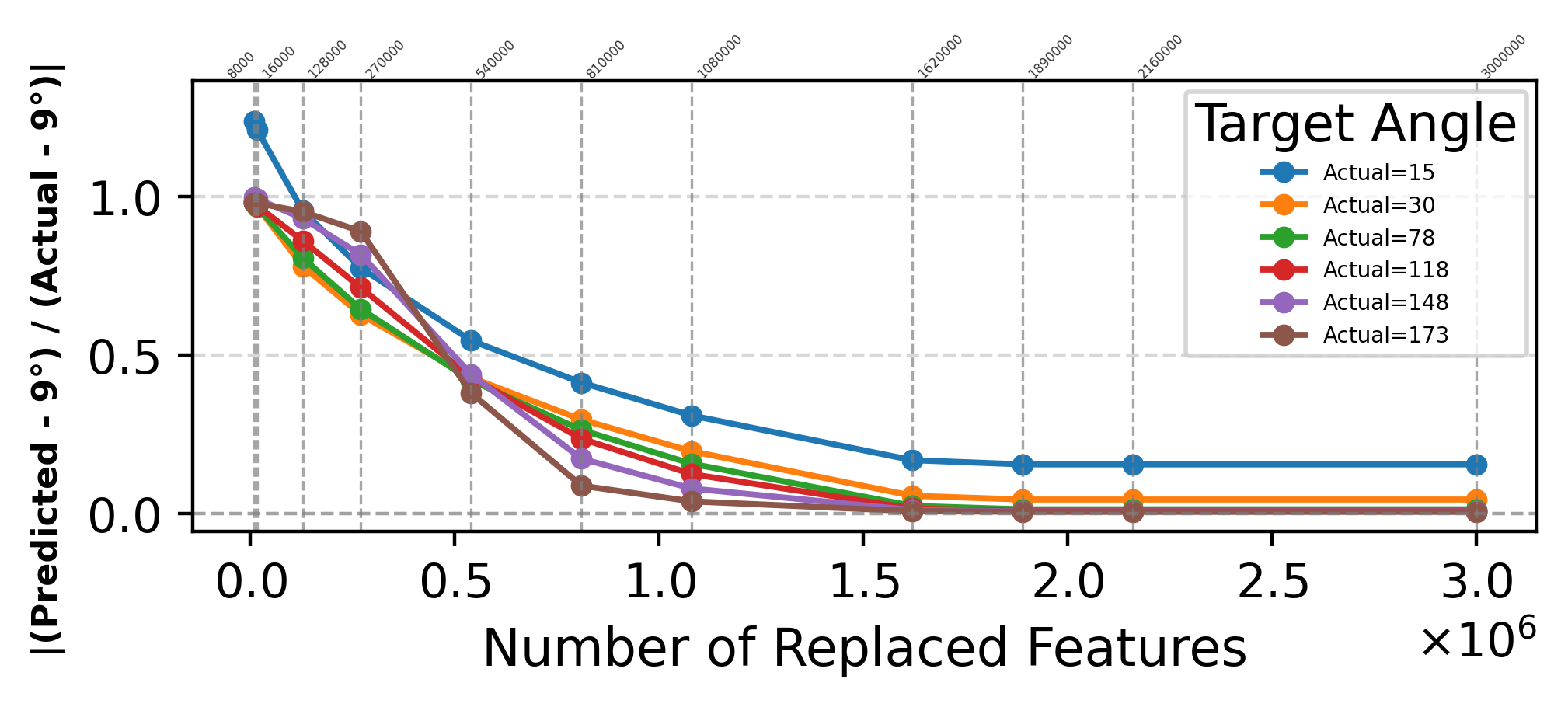}
        \caption{Ordered By Value Difference}
    \end{subfigure}%
    \hspace{-0.5em} 
    \begin{subfigure}{0.33\textwidth}
        \includegraphics[width=\linewidth]{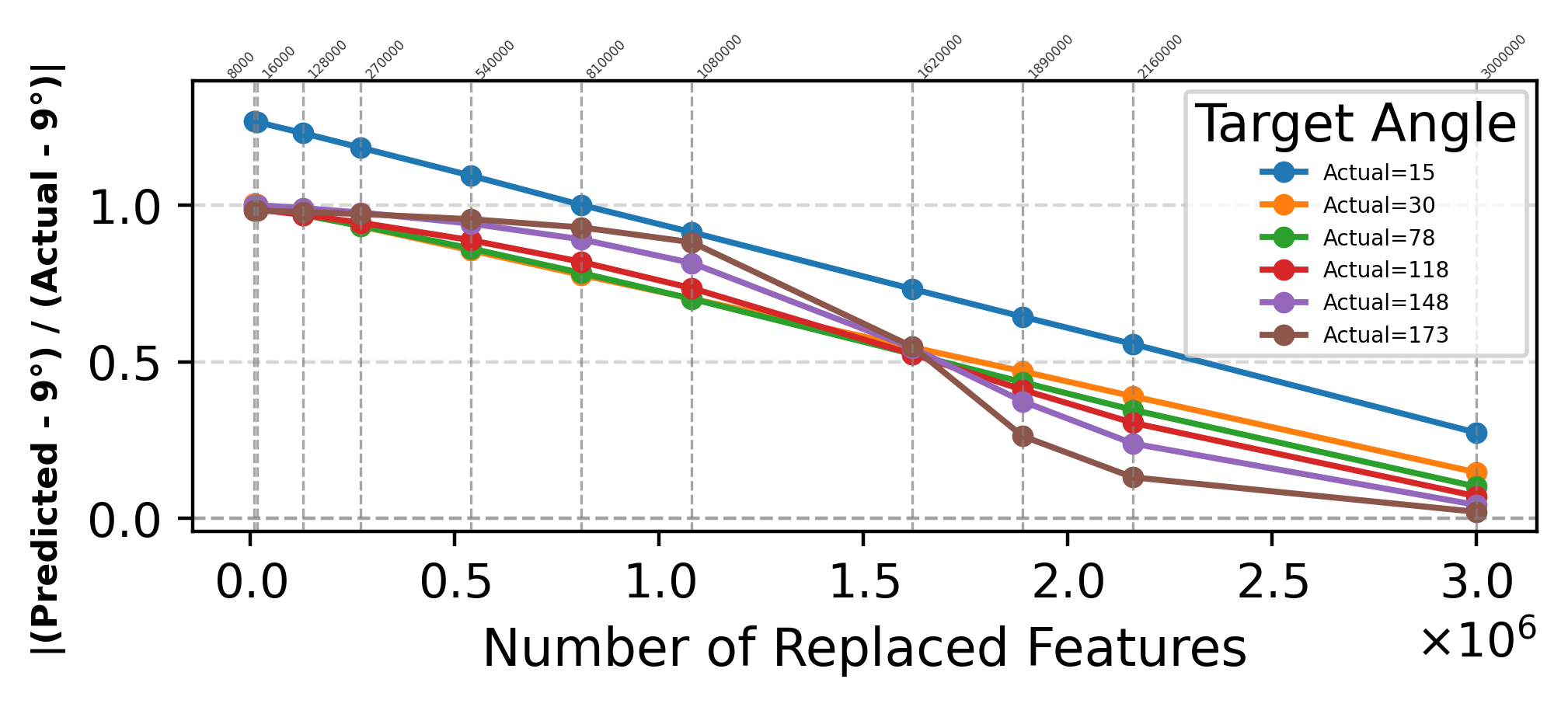}
        \caption{Picked Randomly}
    \end{subfigure}
   \vspace{-2mm}
    \caption{Incremental feature substitution for LLaVA-OneVision on images with the beach scene. No matter how the features are selected (according to the magnitude of the weights in the regressor or the absolute difference between anchor and target feature values, or randomly). 540,000 features or more must be replaced to fool the predictor. (Note that the x-axis is the number of feature substitutions times $10^6$.) This implies the orientation information is highly diffuse.}
   \vspace{-2mm}
    \label{fig:patch_analysis_llava-ov_beach}
\end{figure*}

\begin{figure*}[h!]
    \begin{subfigure}{0.33\textwidth}
        \includegraphics[width=\linewidth]{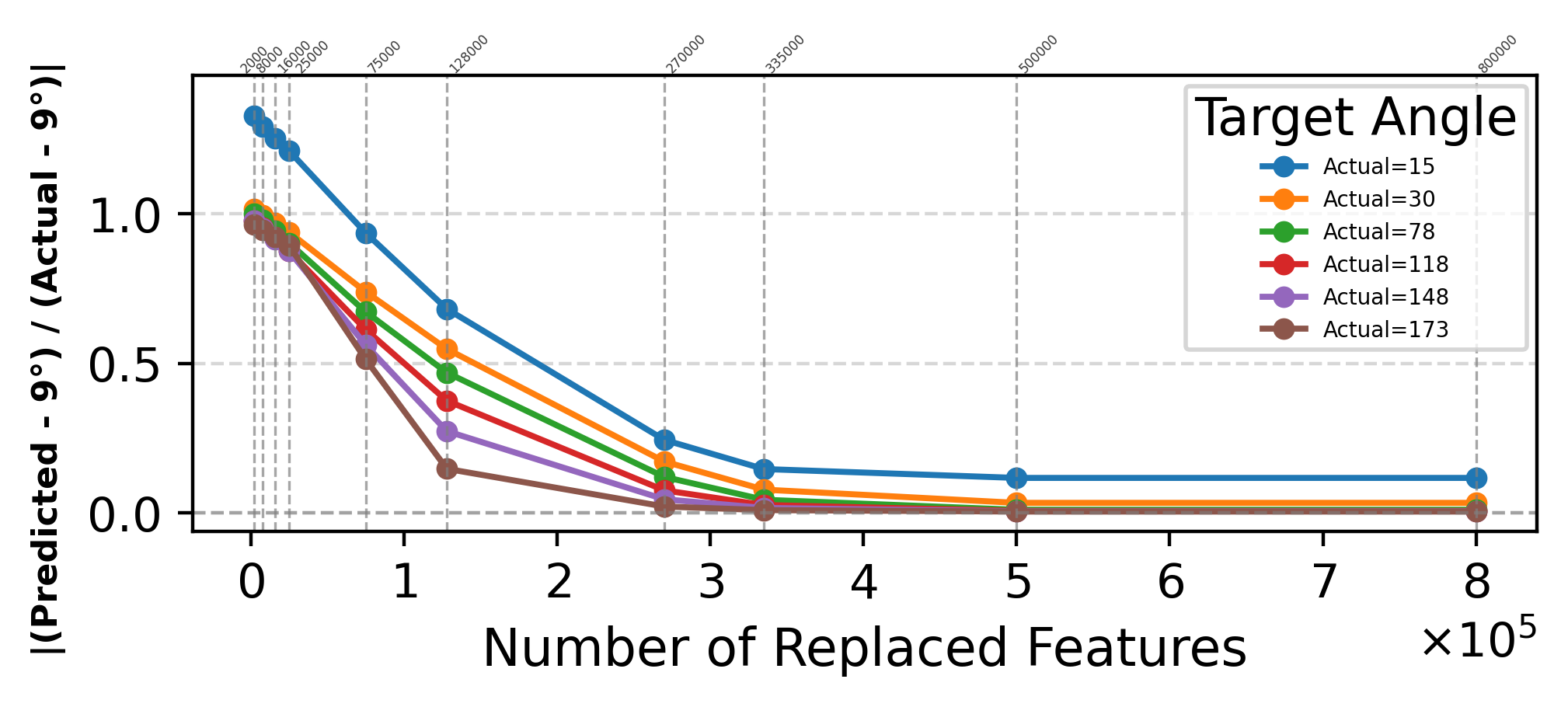}
        \caption{Ordered By Model Weight}
    \end{subfigure}%
    \hspace{-0.5em} 
    \begin{subfigure}{0.33\textwidth}
        \includegraphics[width=\linewidth]{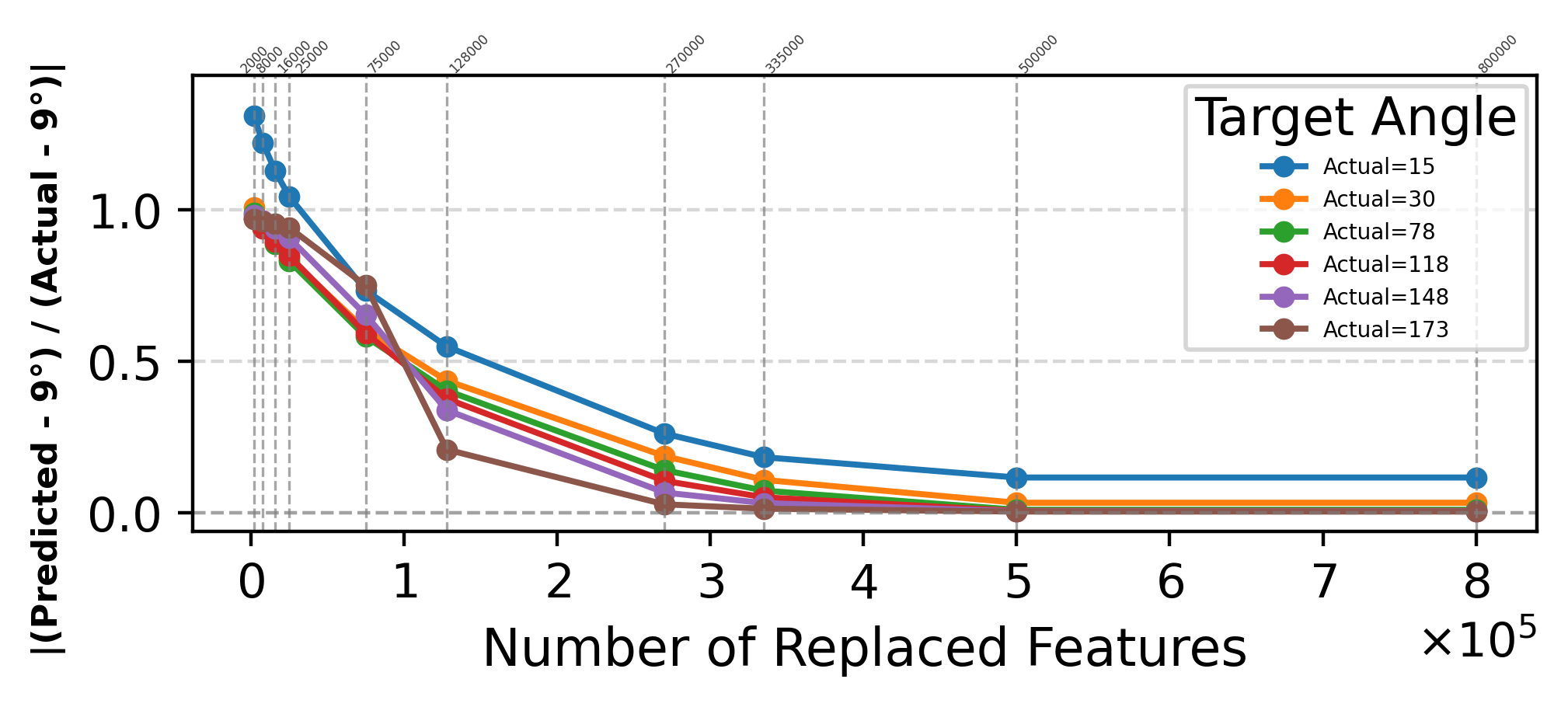}
        \caption{Ordered By Value Difference}
    \end{subfigure}%
    \hspace{-0.5em} 
    \begin{subfigure}{0.33\textwidth}
        \includegraphics[width=\linewidth]{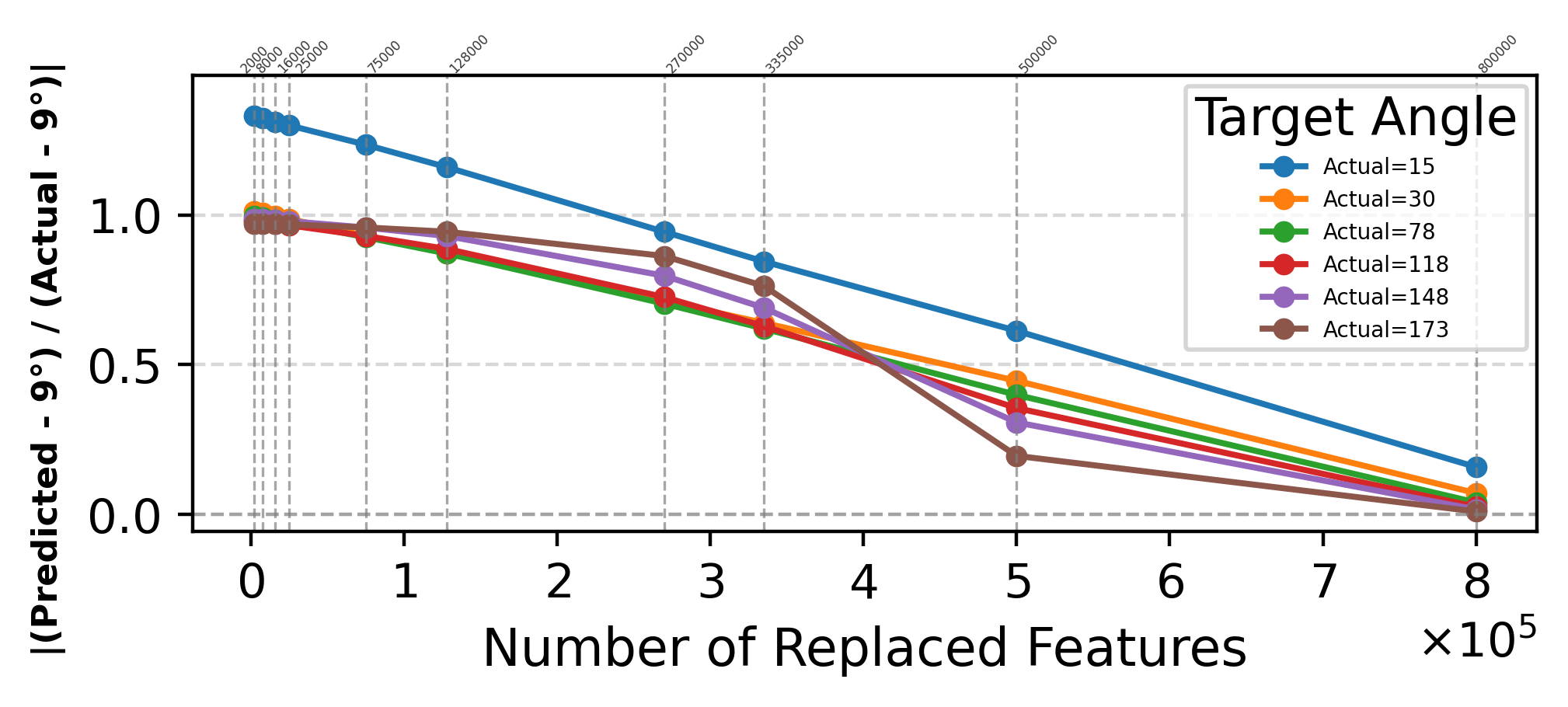}
        \caption{Picked Randomly}
    \end{subfigure}
   \vspace{-2mm}
    \caption{Incremental feature substitution for Qwen2.5-VL-7B-Instruct on images with the beach scene. No matter how the features are selected (according to the magnitude of the weights in the regressor or the absolute difference between anchor and target feature values, or randomly). 128,000 features or more must be replaced to fool the predictor. (Note that the x-axis is the number of feature substitutions times $10^5$.) This implies the orientation information is highly diffuse.}
   \vspace{-2mm}
    \label{fig:patch_analysis_qwen_beach}
\end{figure*}
\newpage
\begin{figure*}[h!]
    \begin{subfigure}{0.33\textwidth}
        \includegraphics[width=\linewidth]{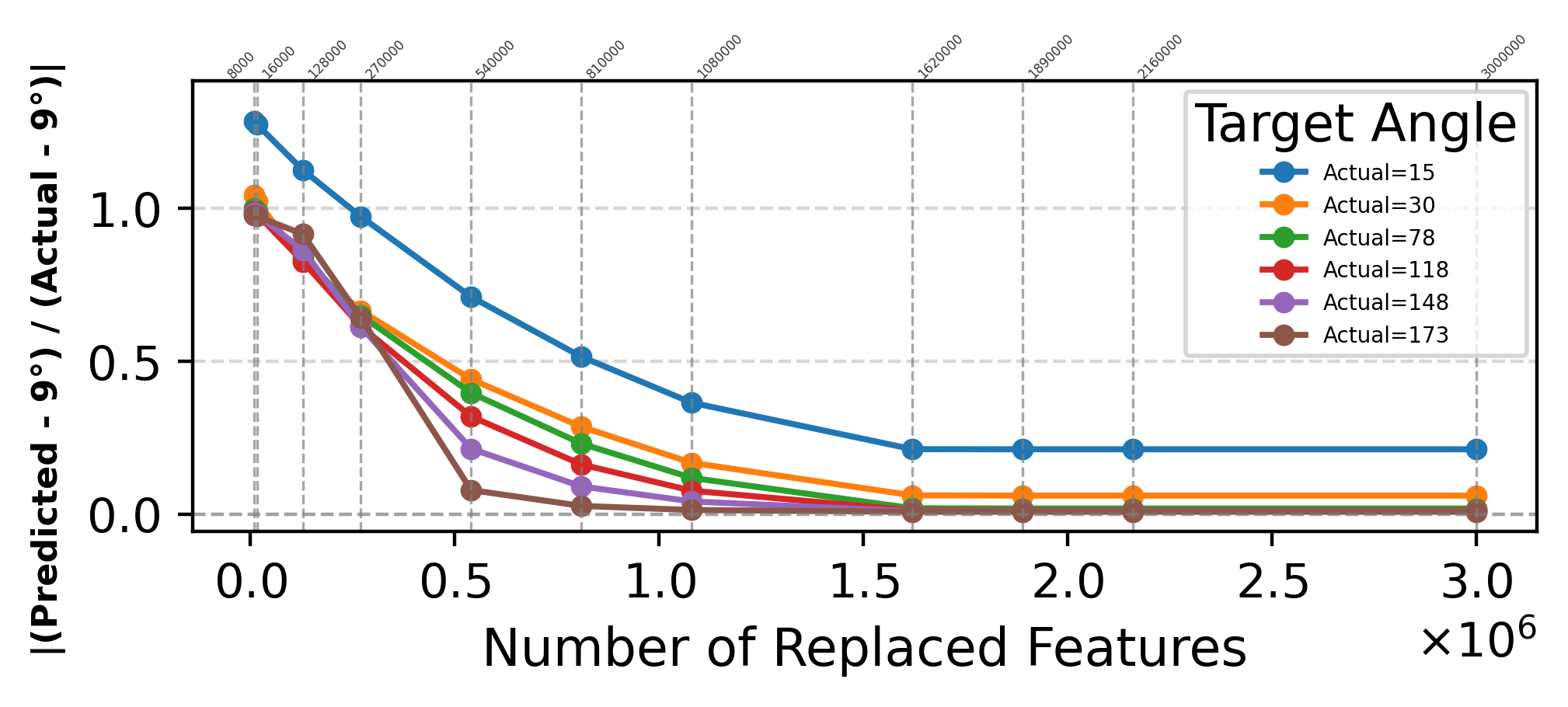}
        \caption{Ordered By Model Weight}
    \end{subfigure}%
    \hspace{-0.5em} 
    \begin{subfigure}{0.33\textwidth}
        \includegraphics[width=\linewidth]{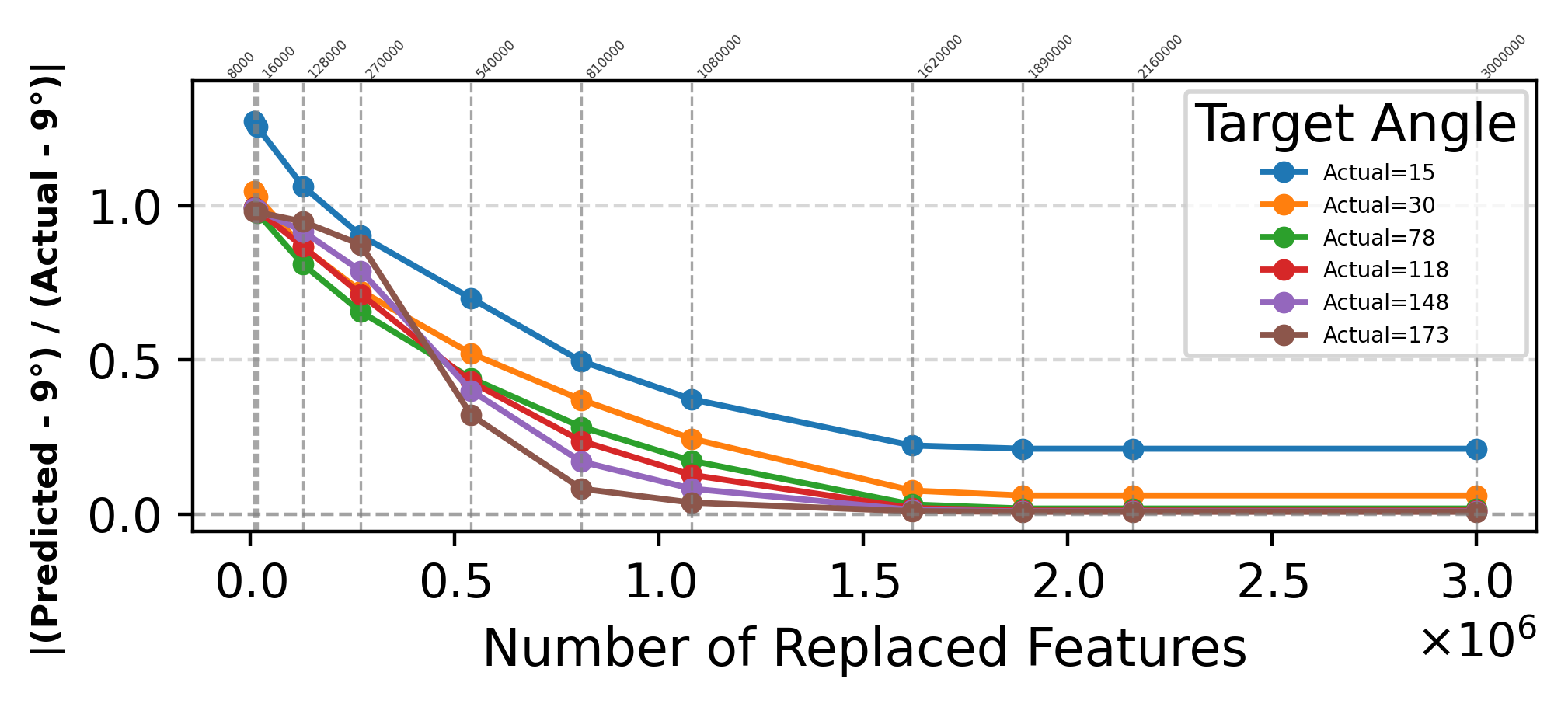}
        \caption{Ordered By Value Difference}
    \end{subfigure}%
    \hspace{-0.5em} 
    \begin{subfigure}{0.33\textwidth}
        \includegraphics[width=\linewidth]{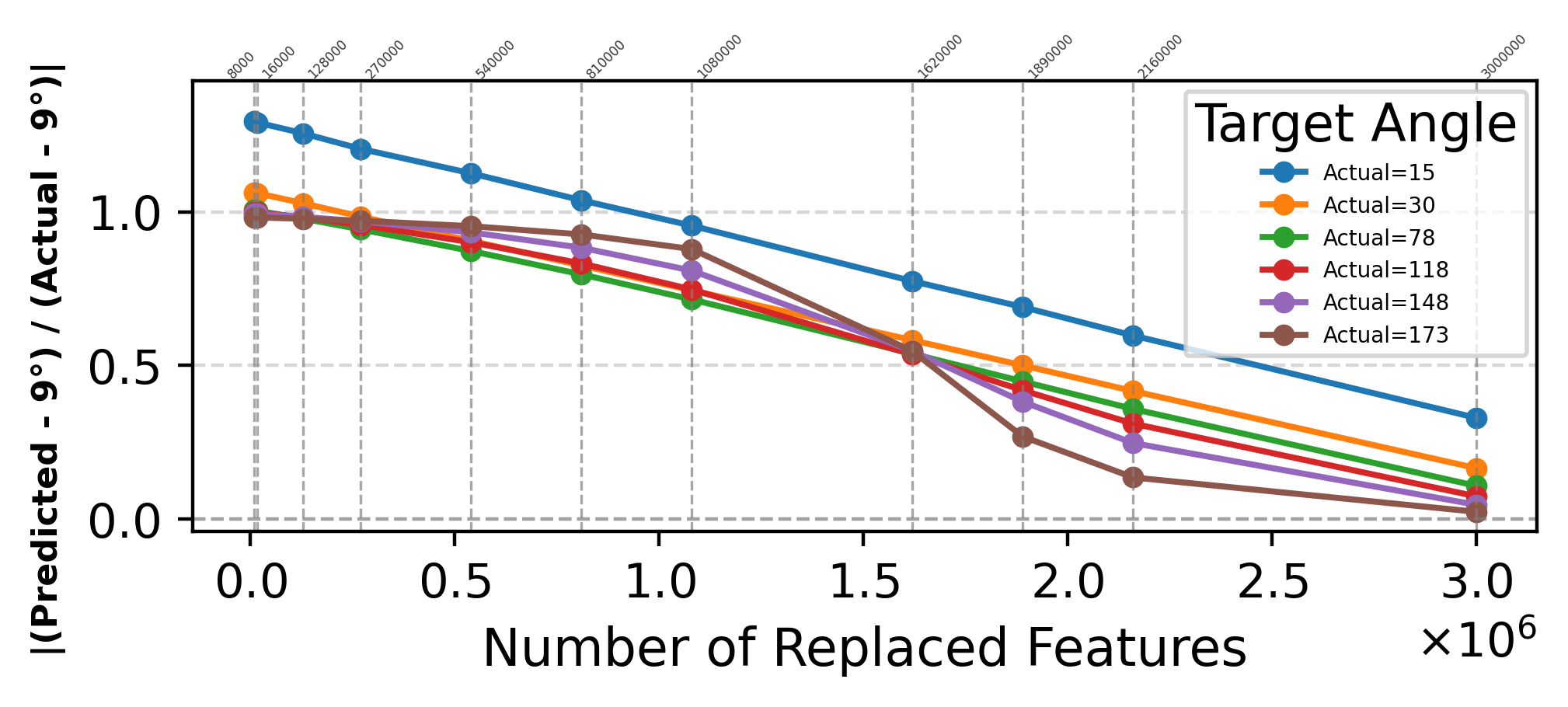}
        \caption{Picked Randomly}
    \end{subfigure}
   \vspace{-2mm}
    \caption{Incremental feature substitution for LLaVA-OneVision on images with the indoor scene. No matter how the features are selected (according to the magnitude of the weights in the regressor or the absolute difference between anchor and target feature values, or randomly). 540,000 features or more must be replaced to fool the predictor. (Note that the x-axis is the number of feature substitutions times $10^6$.) This implies the orientation information is highly diffuse.}
   \vspace{-2mm}
    \label{fig:patch_analysis_llava-ov_indoor}
\end{figure*}

\begin{figure*}[h!]
    \begin{subfigure}{0.33\textwidth}
        \includegraphics[width=\linewidth]{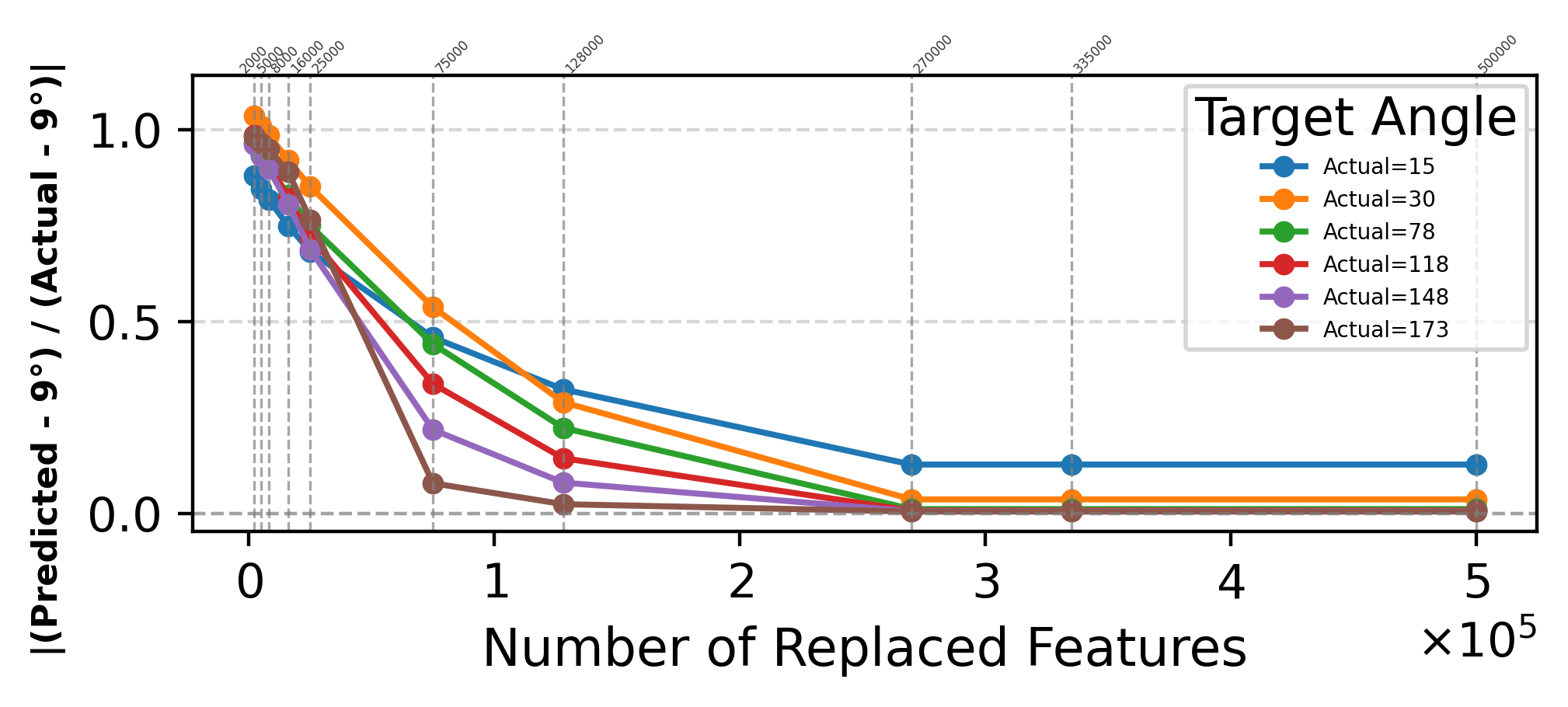}
        \caption{Ordered By Model Weight}
    \end{subfigure}%
    \hspace{-0.5em} 
    \begin{subfigure}{0.33\textwidth}
        \includegraphics[width=\linewidth]{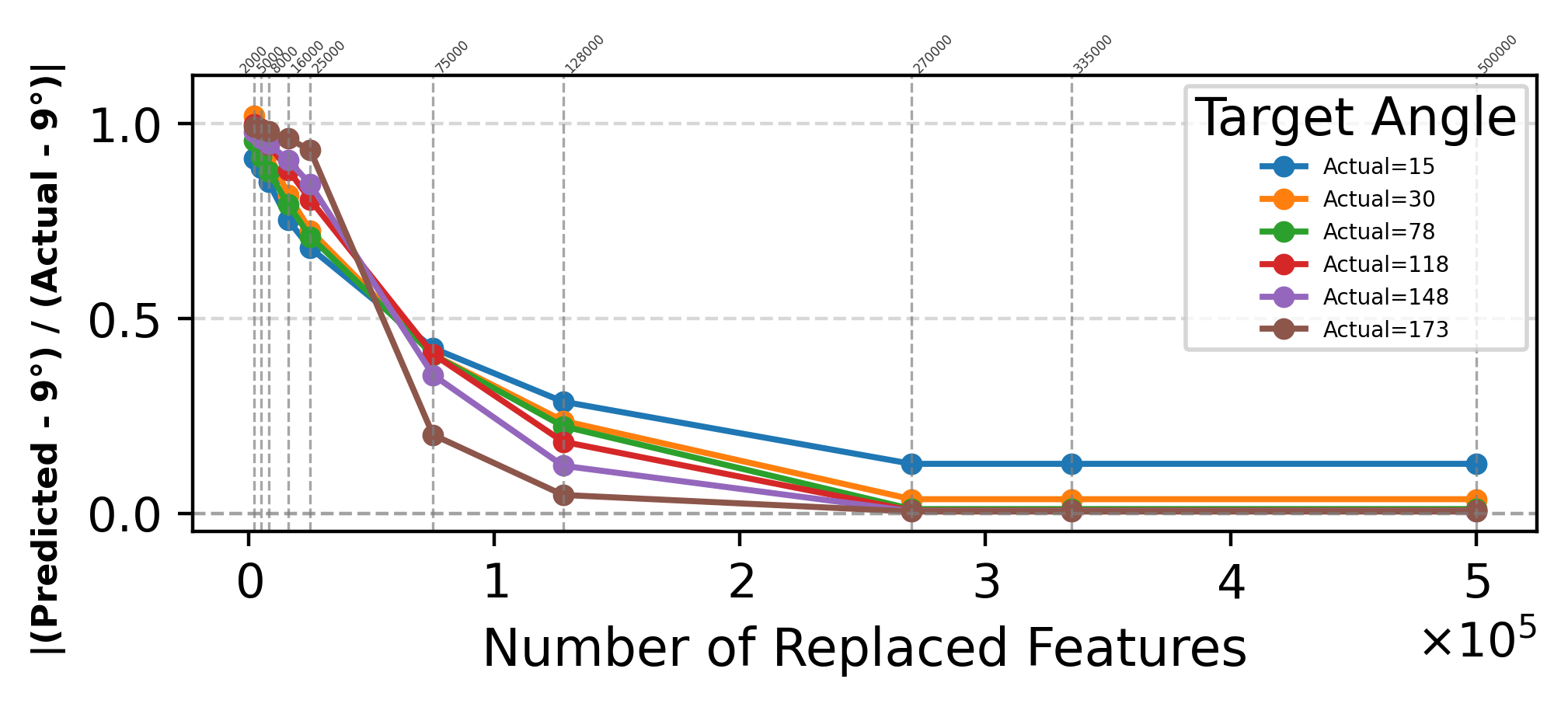}
        \caption{Ordered By Value Difference}
    \end{subfigure}%
    \hspace{-0.5em} 
    \begin{subfigure}{0.33\textwidth}
        \includegraphics[width=\linewidth]{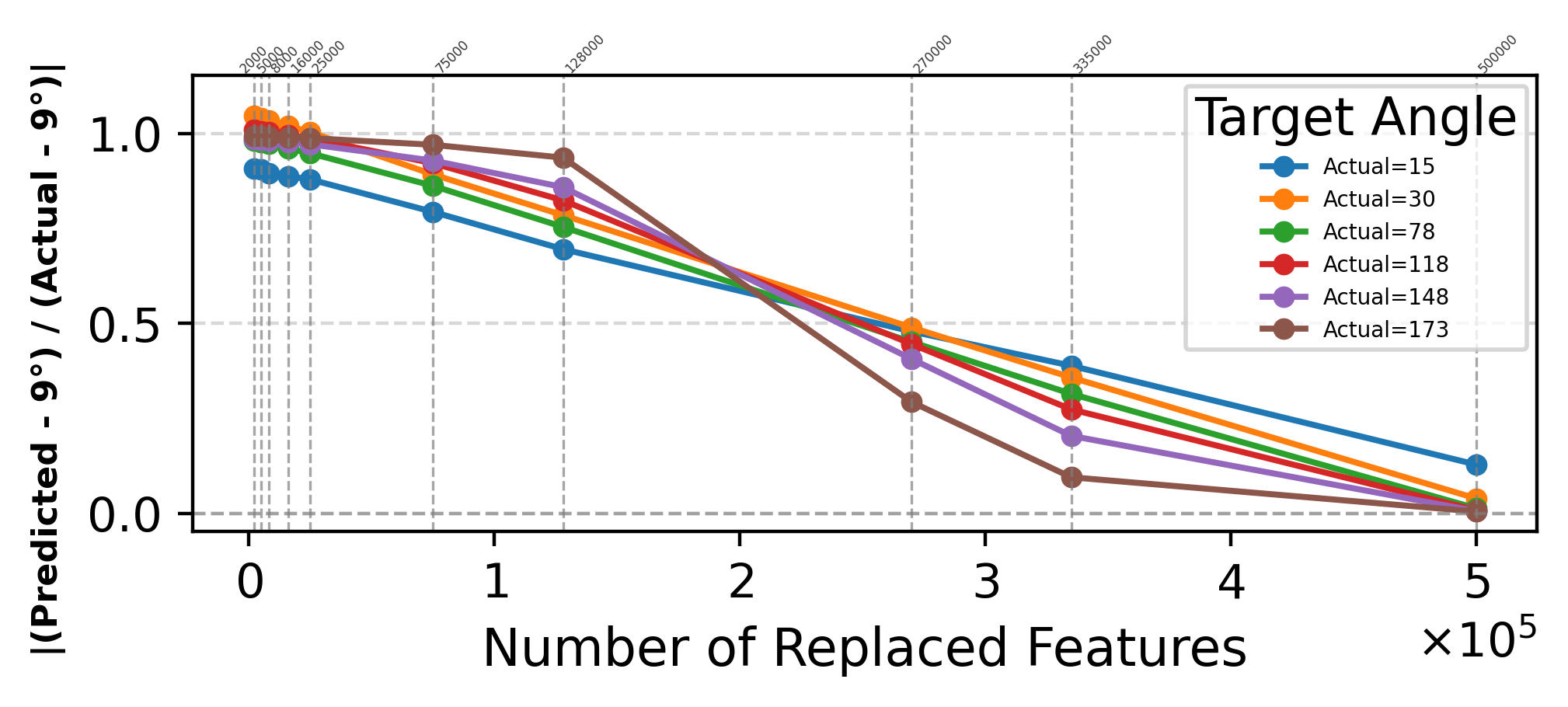}
        \caption{Picked Randomly}
    \end{subfigure}
   \vspace{-2mm}
    \caption{Incremental feature substitution for Qwen2.5-VL-7B-Instruct on images with the indoor scene. No matter how the features are selected (according to the magnitude of the weights in the regressor or the absolute difference between anchor and target feature values, or randomly). 128,000 features or more must be replaced to fool the predictor. (Note that the x-axis is the number of feature substitutions times $10^5$.) This implies the orientation information is highly diffuse.}
   \vspace{-2mm}
    \label{fig:patch_analysis_qwen_indoor}
\end{figure*}
\newpage
\begin{figure*}[h!]
    \begin{subfigure}{0.33\textwidth}
        \includegraphics[width=\linewidth]{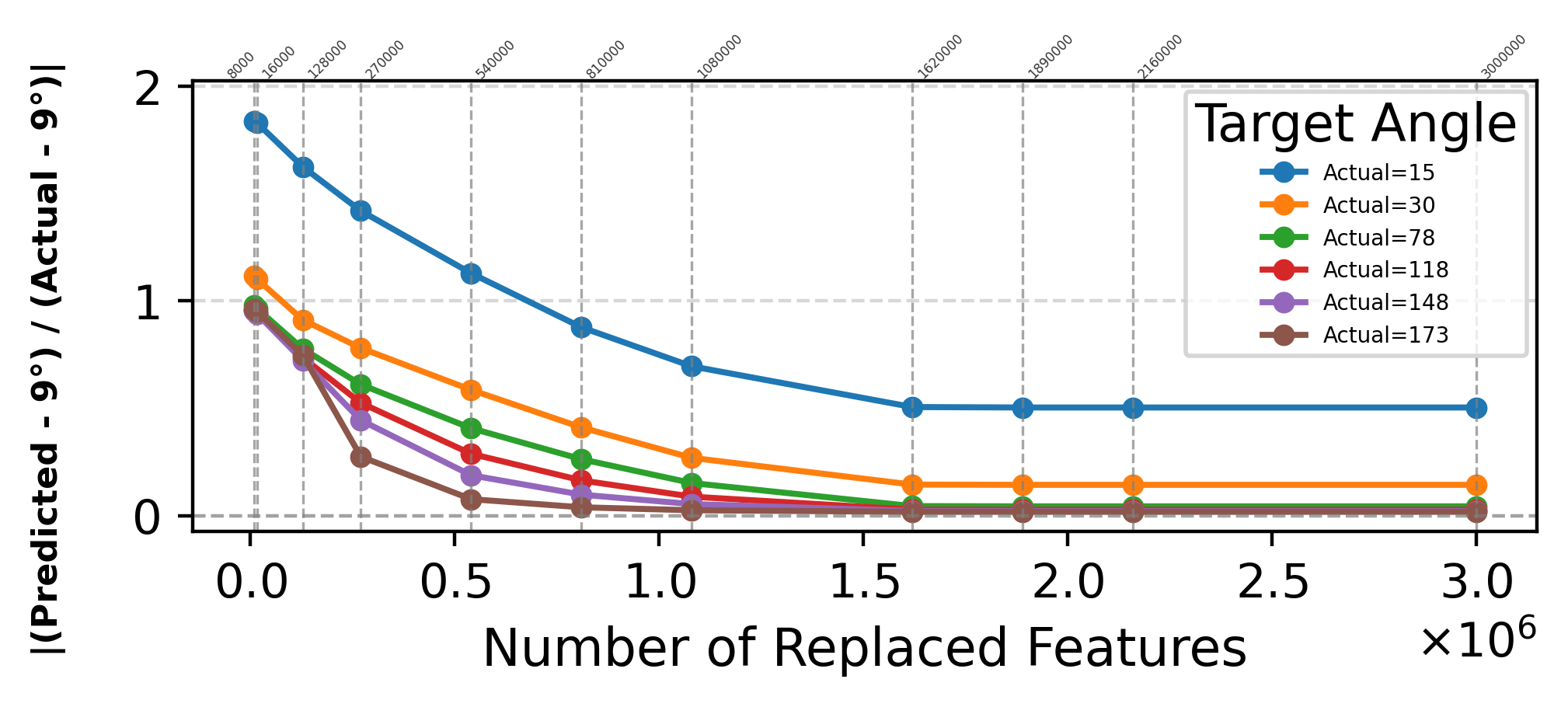}
        \caption{Ordered By Model Weight}
    \end{subfigure}%
    \hspace{-0.5em} 
    \begin{subfigure}{0.33\textwidth}
        \includegraphics[width=\linewidth]{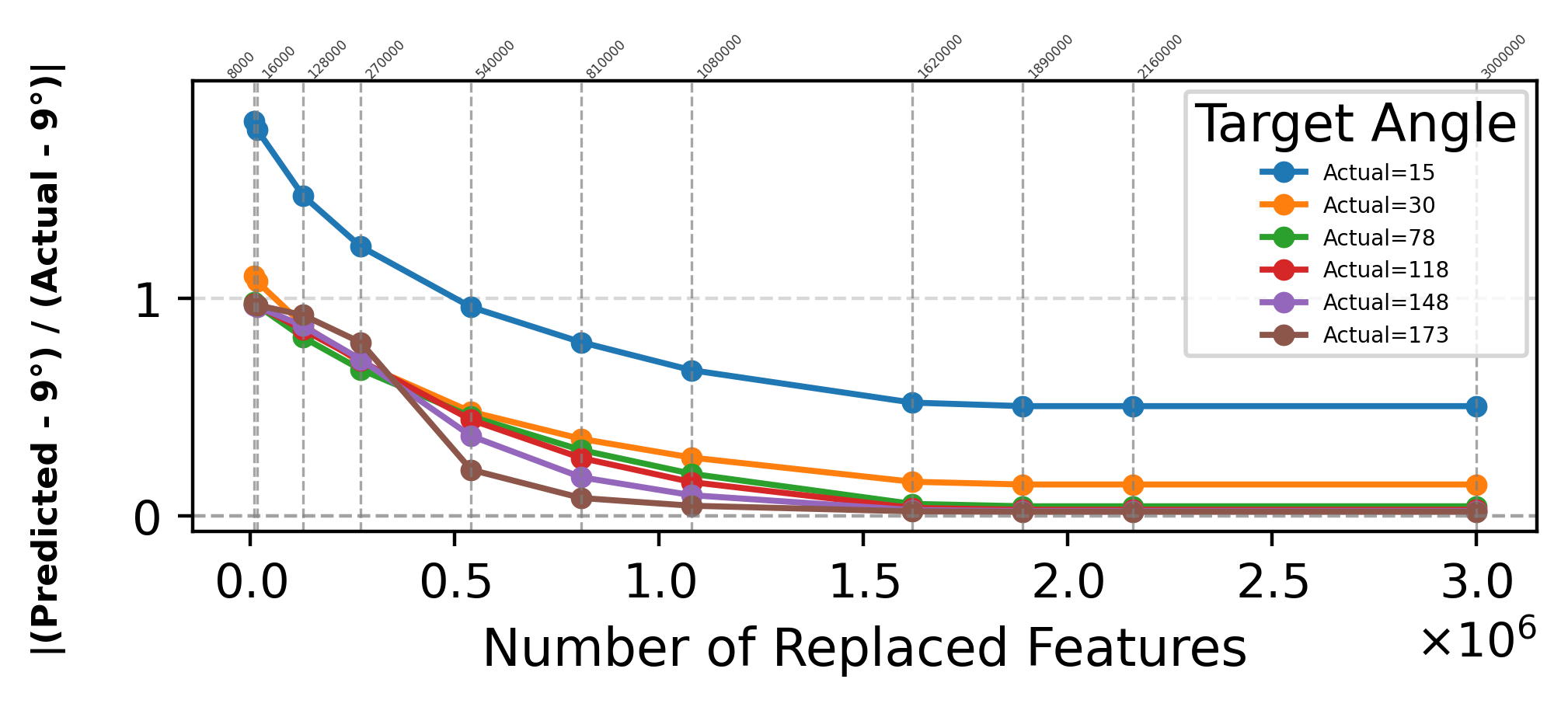}
        \caption{Ordered By Value Difference}
    \end{subfigure}%
    \hspace{-0.5em} 
    \begin{subfigure}{0.33\textwidth}
        \includegraphics[width=\linewidth]{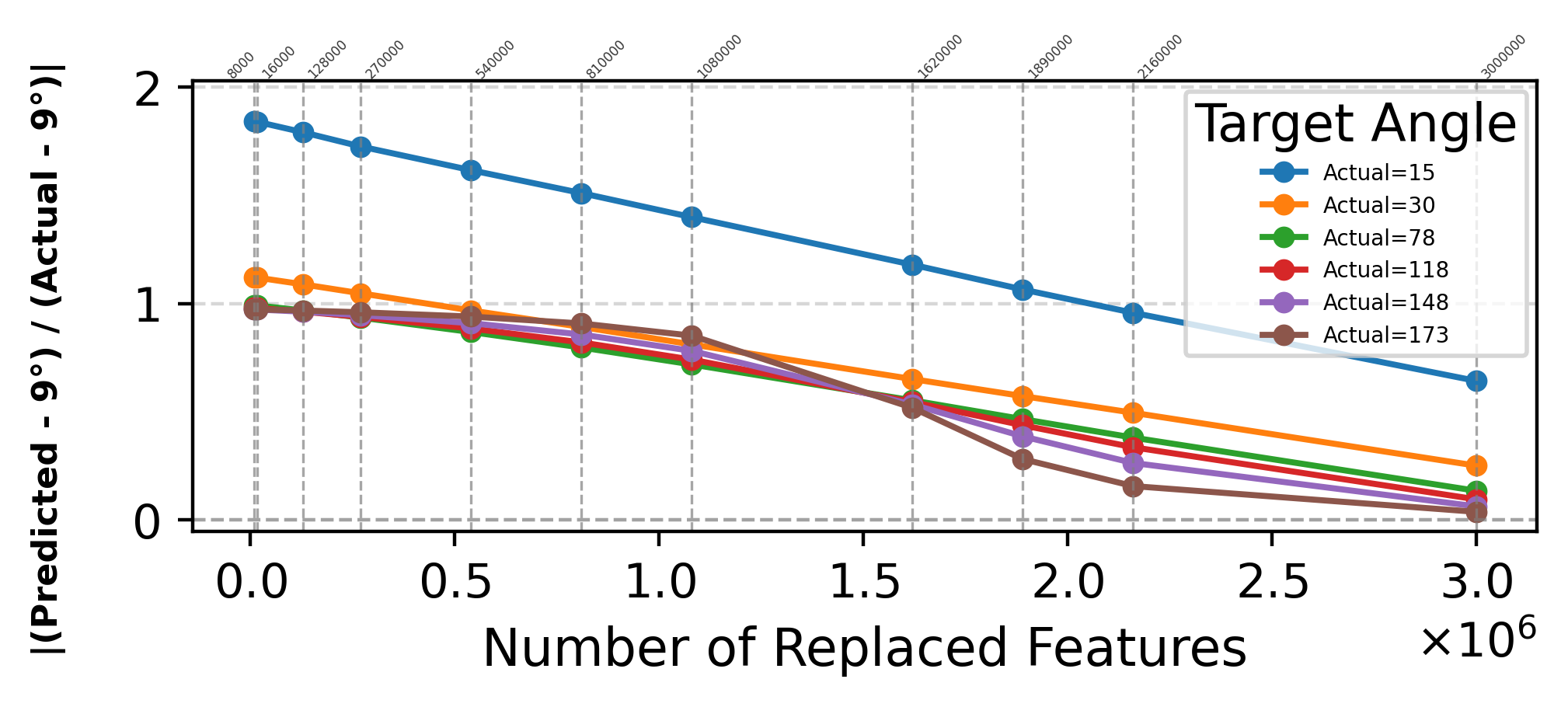}
        \caption{Picked Randomly}
    \end{subfigure}
   \vspace{-2mm}
    \caption{Incremental feature substitution for LLaVA-OneVision on images with the fish scene. No matter how the features are selected (according to the magnitude of the weights in the regressor or the absolute difference between anchor and target feature values, or randomly). 540,000 features or more must be replaced to fool the predictor. (Note that the x-axis is the number of feature substitutions times $10^6$.) This implies the orientation information is highly diffuse.}
   \vspace{-2mm}
    \label{fig:patch_analysis_llava-ov_fish}
\end{figure*}

\begin{figure*}[h!]
    \begin{subfigure}{0.33\textwidth}
        \includegraphics[width=\linewidth]{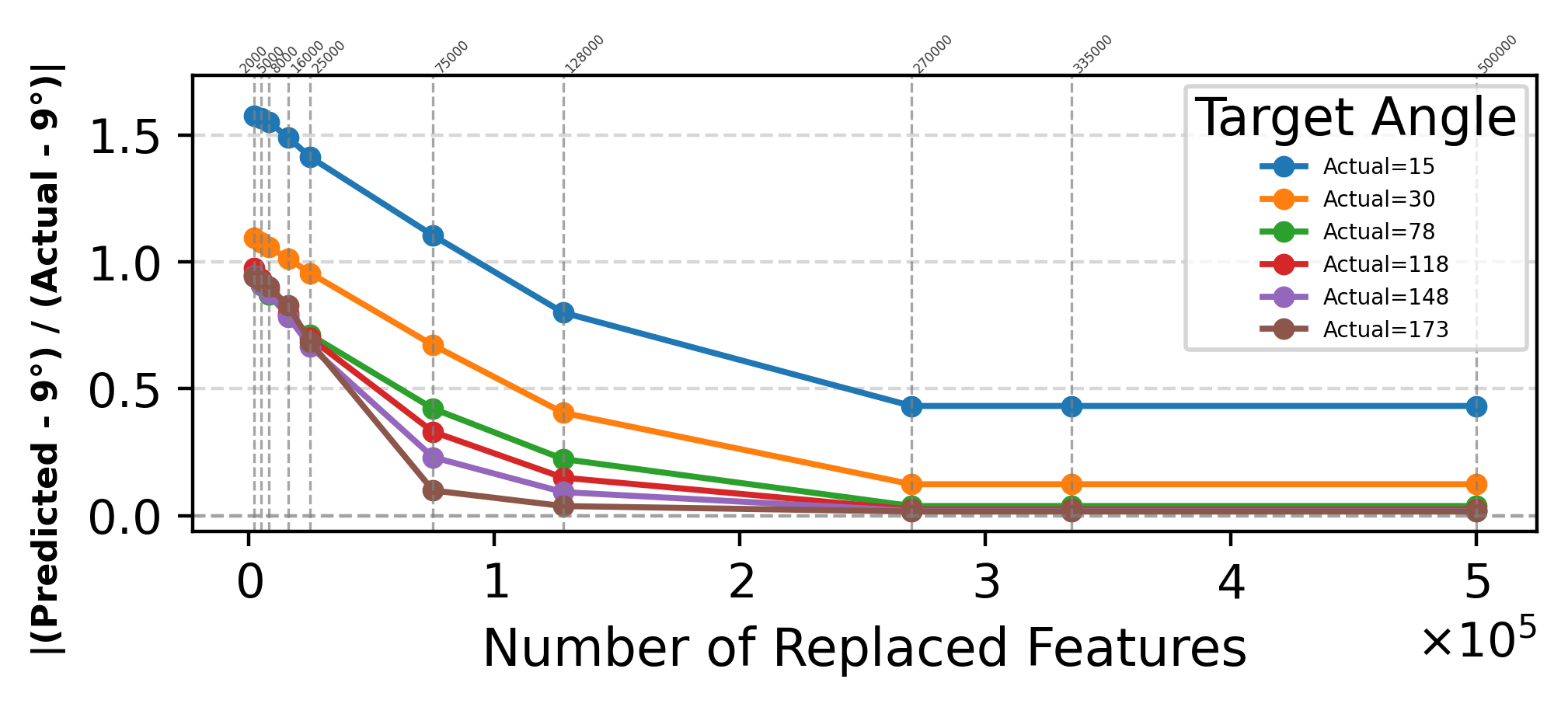}
        \caption{Ordered By Model Weight}
    \end{subfigure}%
    \hspace{-0.5em} 
    \begin{subfigure}{0.33\textwidth}
        \includegraphics[width=\linewidth]{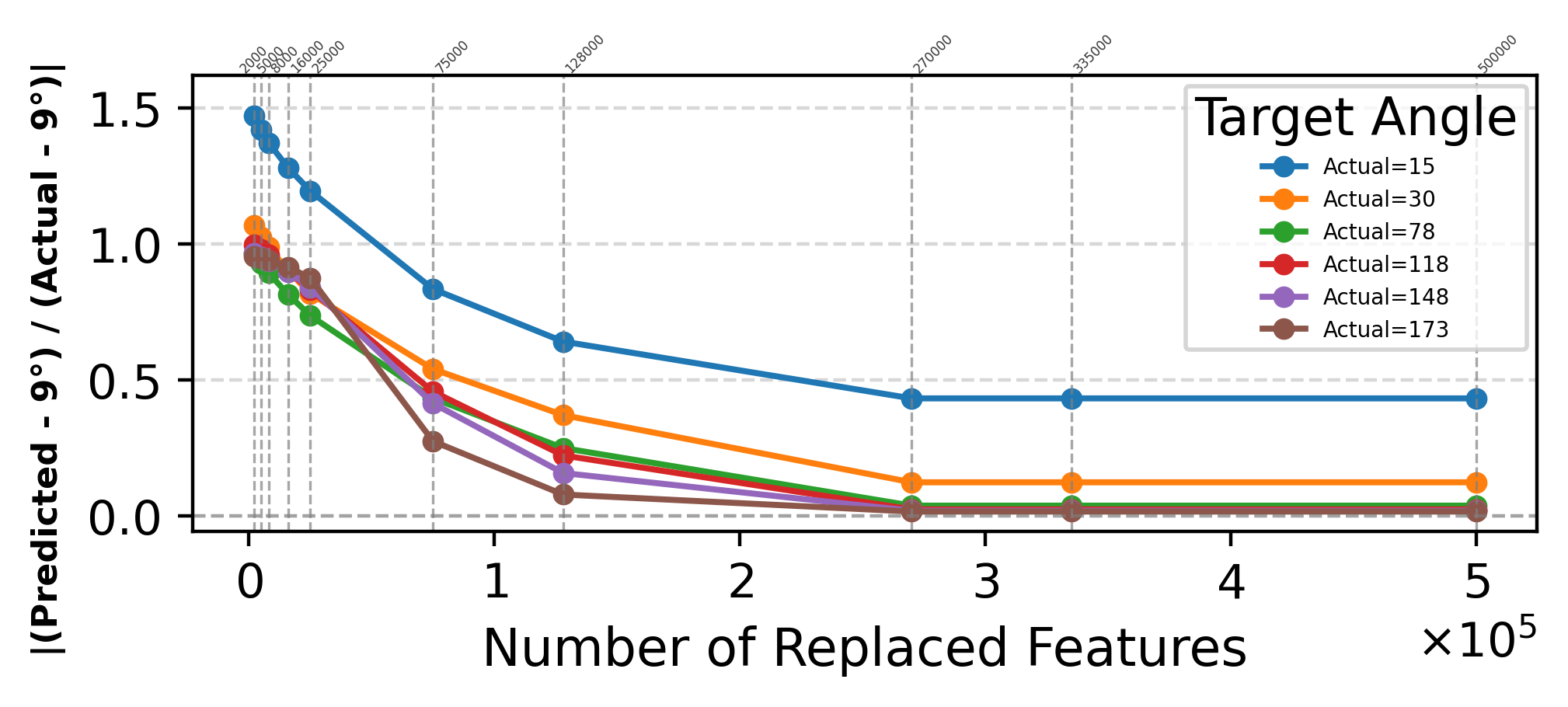}
        \caption{Ordered By Value Difference}
    \end{subfigure}%
    \hspace{-0.5em} 
    \begin{subfigure}{0.33\textwidth}
        \includegraphics[width=\linewidth]{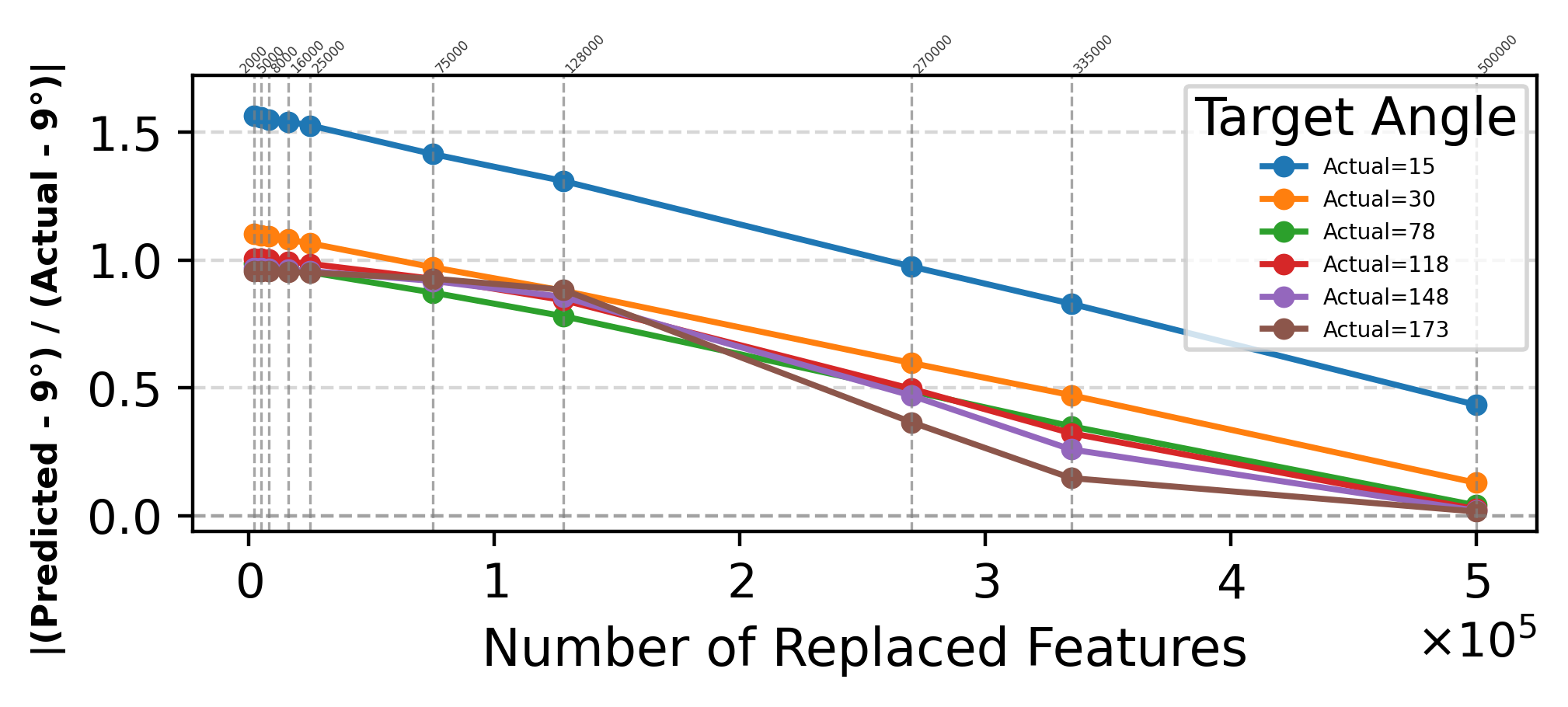}
        \caption{Picked Randomly}
    \end{subfigure}
   \vspace{-2mm}
    \caption{Incremental feature substitution for Qwen2.5-VL-7B-Instruct on images with the fish scene. No matter how the features are selected (according to the magnitude of the weights in the regressor or the absolute difference between anchor and target feature values, or randomly). 128,000 features or more must be replaced to fool the predictor. (Note that the x-axis is the number of feature substitutions times $10^5$.) This implies the orientation information is highly diffuse.}
   \vspace{-2mm}
    \label{fig:patch_analysis_qwen_fish}
\end{figure*}
\newpage
\begin{figure*}[h!]
    \begin{subfigure}{0.33\textwidth}
        \includegraphics[width=\linewidth]{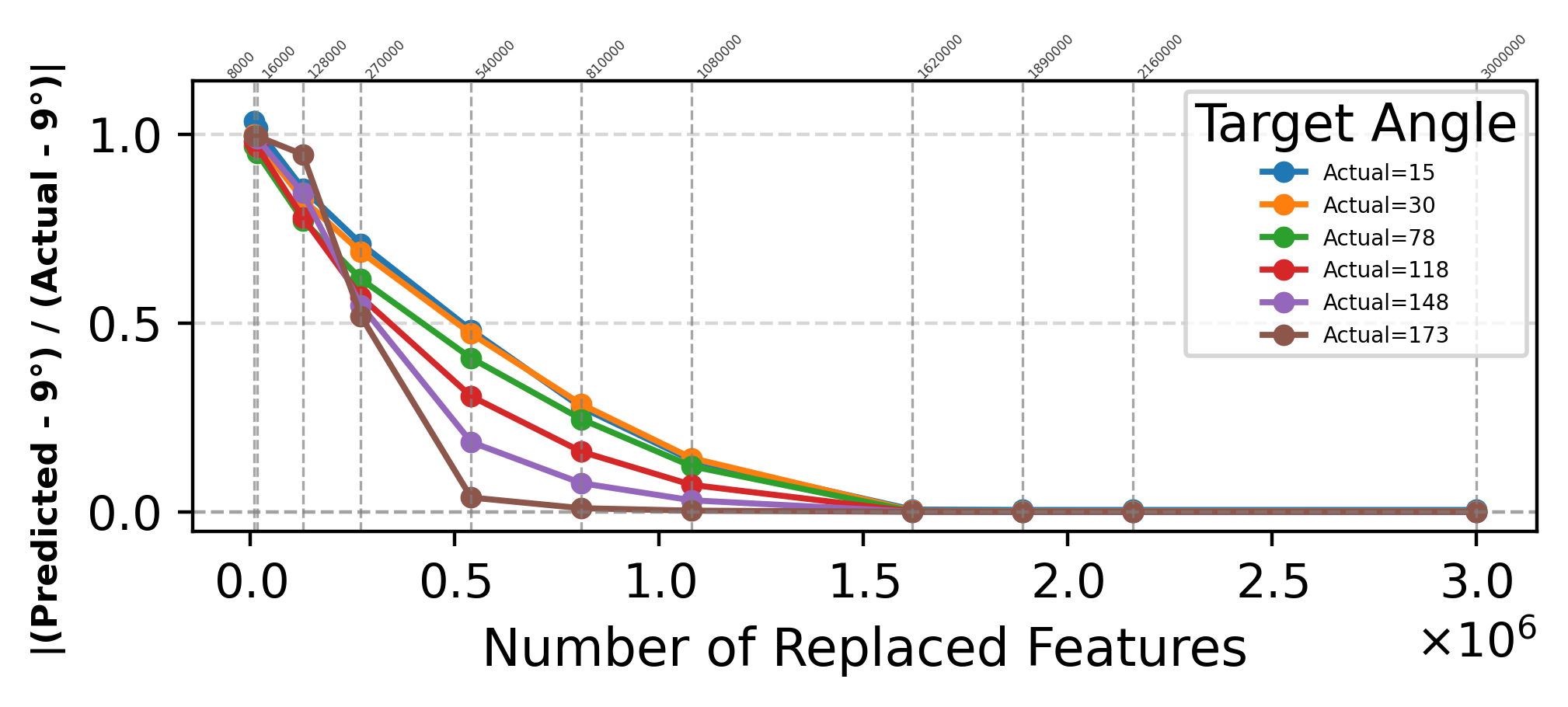}
        \caption{Ordered By Model Weight}
    \end{subfigure}%
    \hspace{-0.5em} 
    \begin{subfigure}{0.33\textwidth}
        \includegraphics[width=\linewidth]{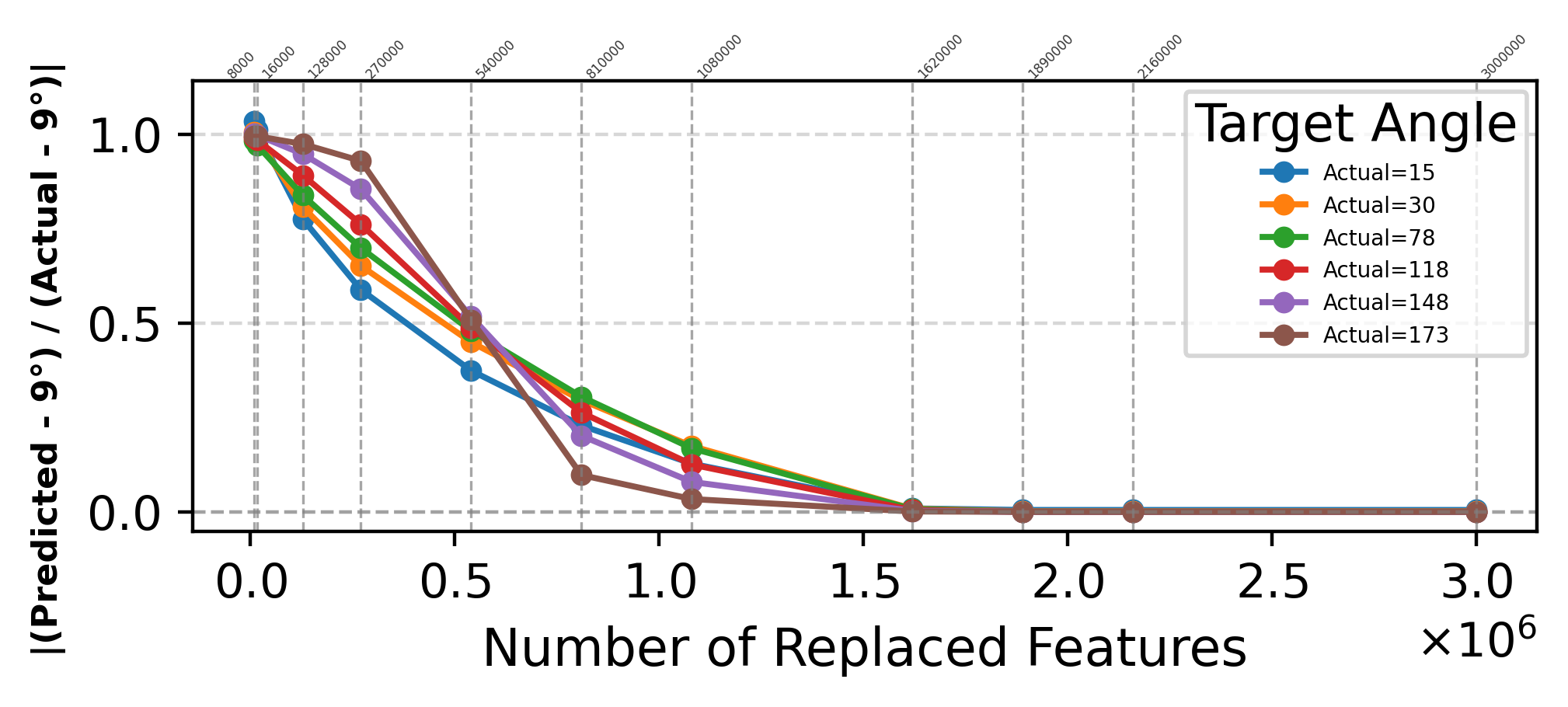}
        \caption{Ordered By Value Difference}
    \end{subfigure}%
    \hspace{-0.5em} 
    \begin{subfigure}{0.33\textwidth}
        \includegraphics[width=\linewidth]{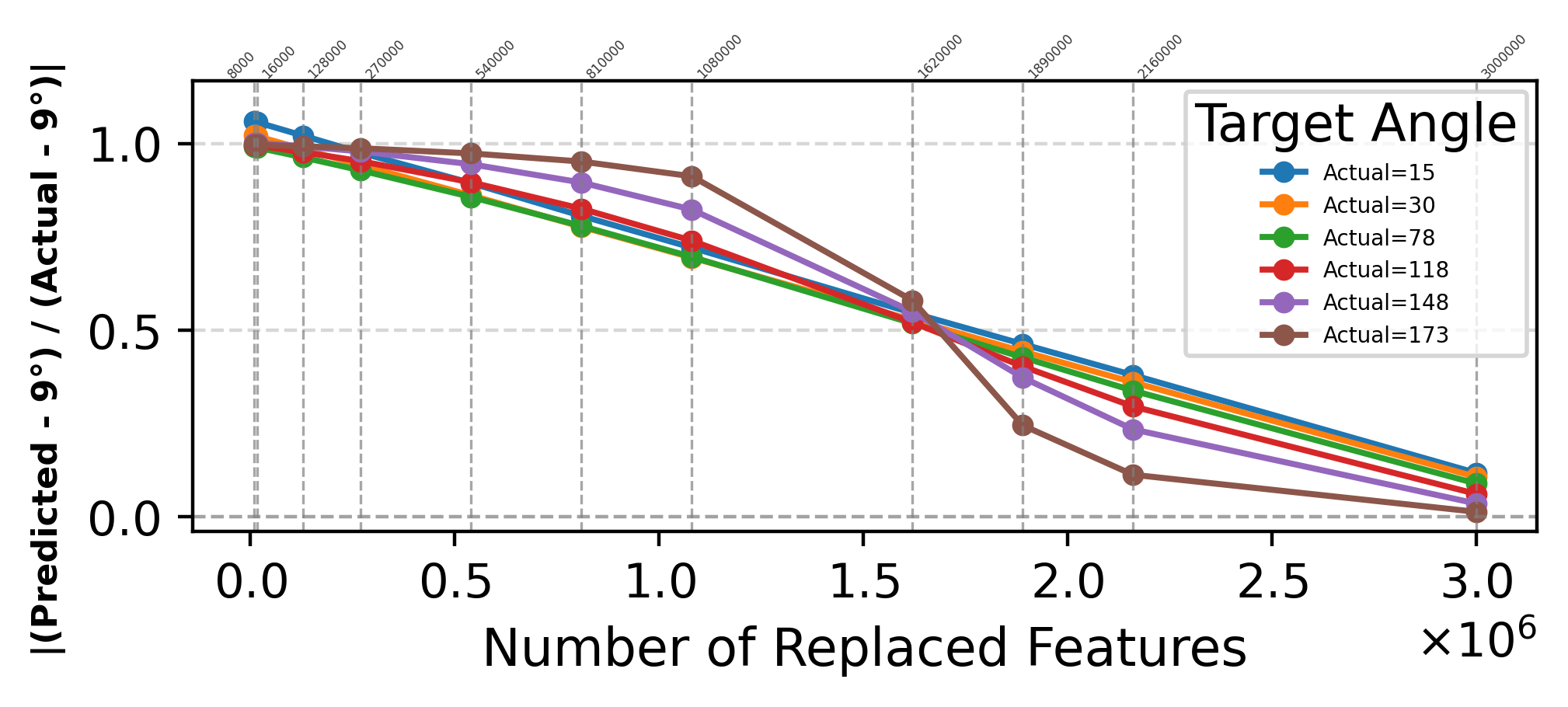}
        \caption{Picked Randomly}
    \end{subfigure}
   \vspace{-2mm}
    \caption{Incremental feature substitution for LLaVA-OneVision on images with the koala-beach scene. No matter how the features are selected (according to the magnitude of the weights in the regressor or the absolute difference between anchor and target feature values, or randomly). 540,000 features or more must be replaced to fool the predictor. (Note that the x-axis is the number of feature substitutions times $10^6$.) This implies the orientation information is highly diffuse.}
   \vspace{-2mm}
    \label{fig:patch_analysis_llava-ov_koala-beach}
\end{figure*}

\begin{figure*}[h!]
    \begin{subfigure}{0.33\textwidth}
        \includegraphics[width=\linewidth]{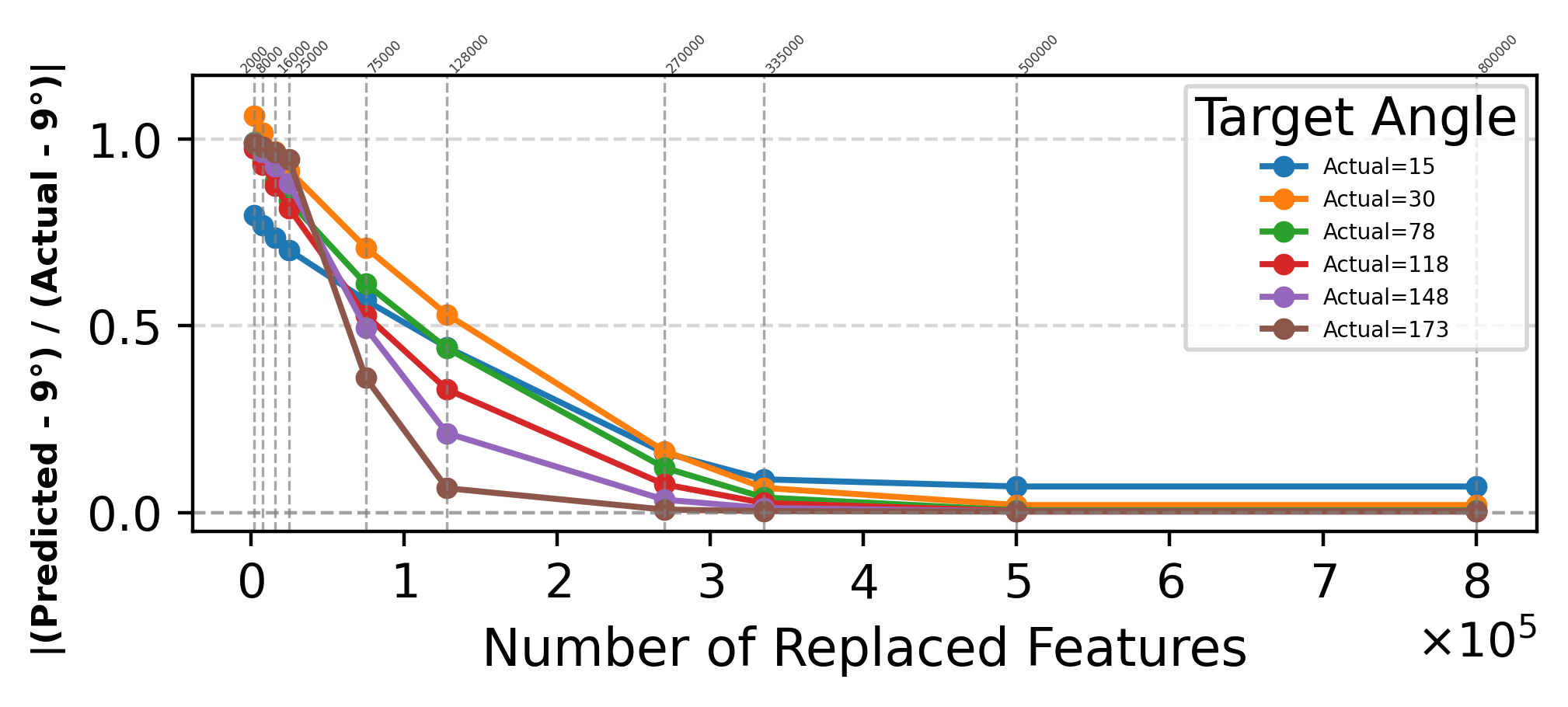}
        \caption{Ordered By Model Weight}
    \end{subfigure}%
    \hspace{-0.5em} 
    \begin{subfigure}{0.33\textwidth}
        \includegraphics[width=\linewidth]{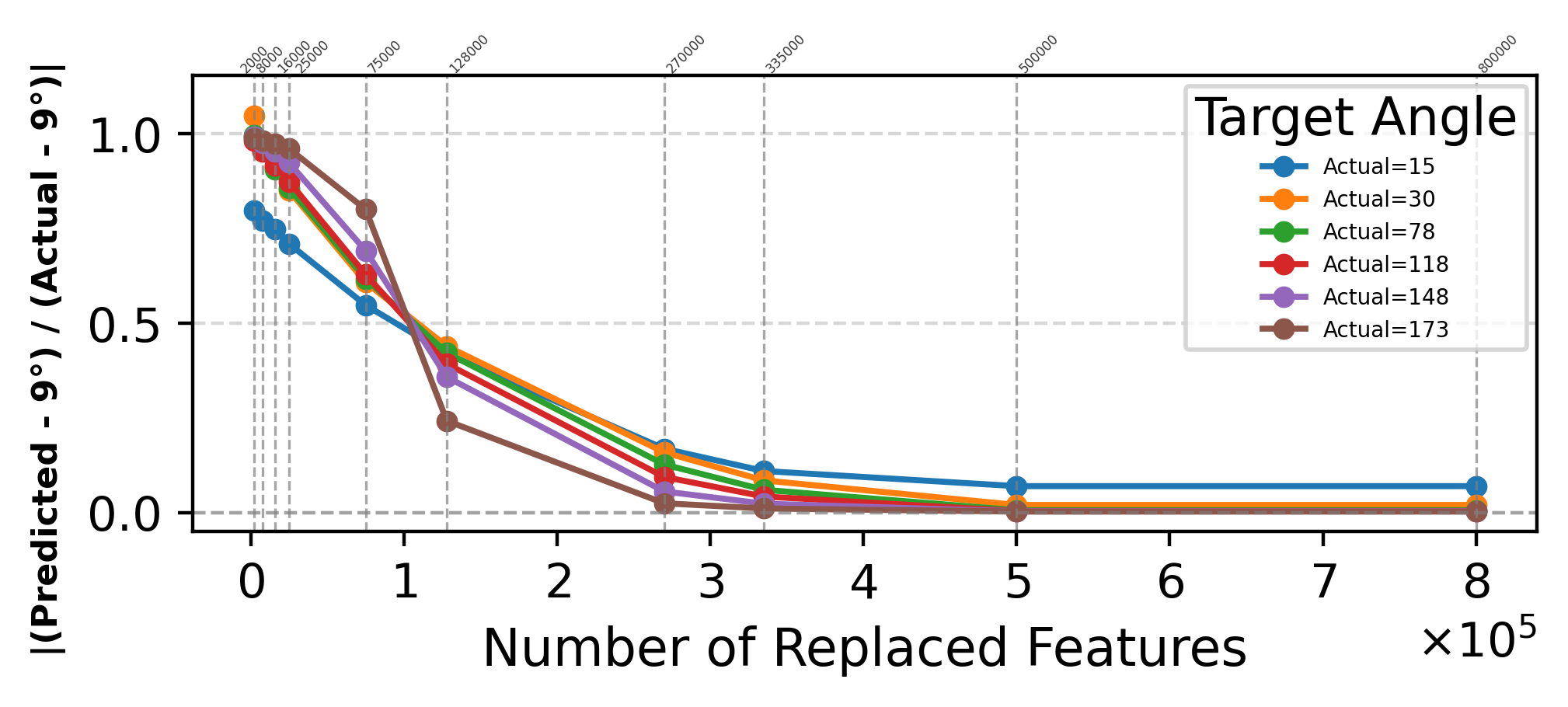}
        \caption{Ordered By Value Difference}
    \end{subfigure}%
    \hspace{-0.5em} 
    \begin{subfigure}{0.33\textwidth}
        \includegraphics[width=\linewidth]{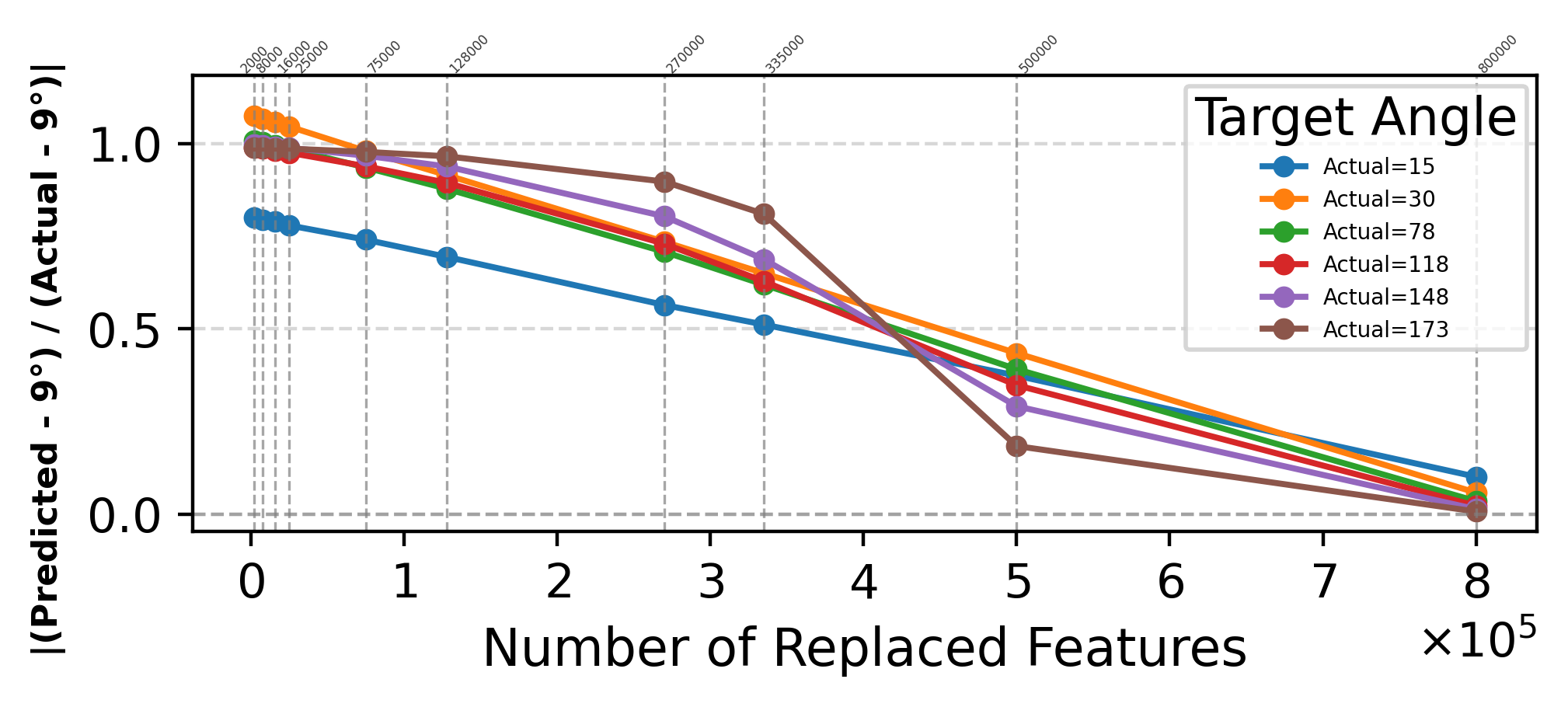}
        \caption{Picked Randomly}
    \end{subfigure}
   \vspace{-2mm}
    \caption{Incremental feature substitution for Qwen2.5-VL-7B-Instruct on images with the koala-beach scene. No matter how the features are selected (according to the magnitude of the weights in the regressor or the absolute difference between anchor and target feature values, or randomly). 128,000 features or more must be replaced to fool the predictor. (Note that the x-axis is the number of feature substitutions times $10^5$.) This implies the orientation information is highly diffuse.}
   \vspace{-2mm}
    \label{fig:patch_analysis_qwen_koala-beach}
\end{figure*}
\newpage
\begin{figure*}[h!]
    \begin{subfigure}{0.33\textwidth}
        \includegraphics[width=\linewidth]{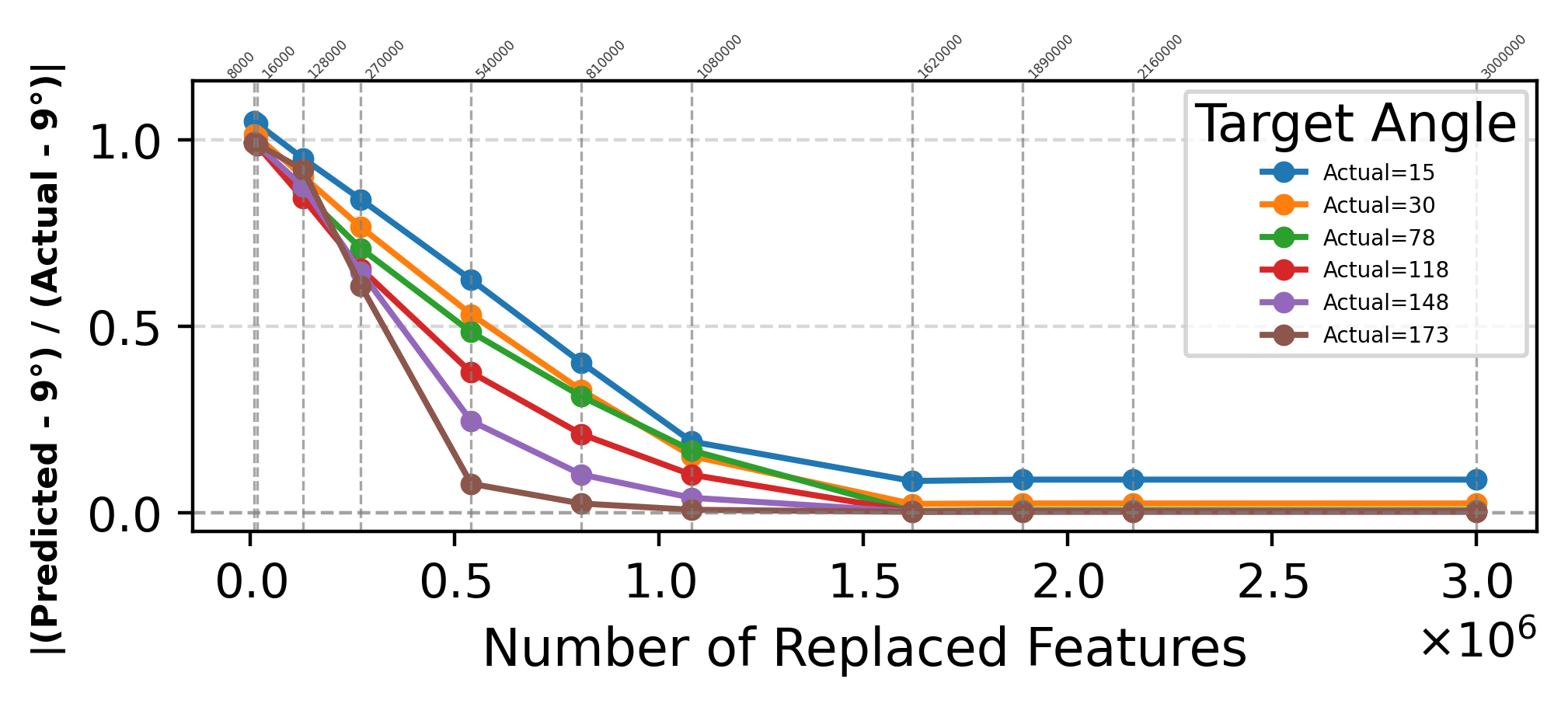}
        \caption{Ordered By Model Weight}
    \end{subfigure}%
    \hspace{-0.5em} 
    \begin{subfigure}{0.33\textwidth}
        \includegraphics[width=\linewidth]{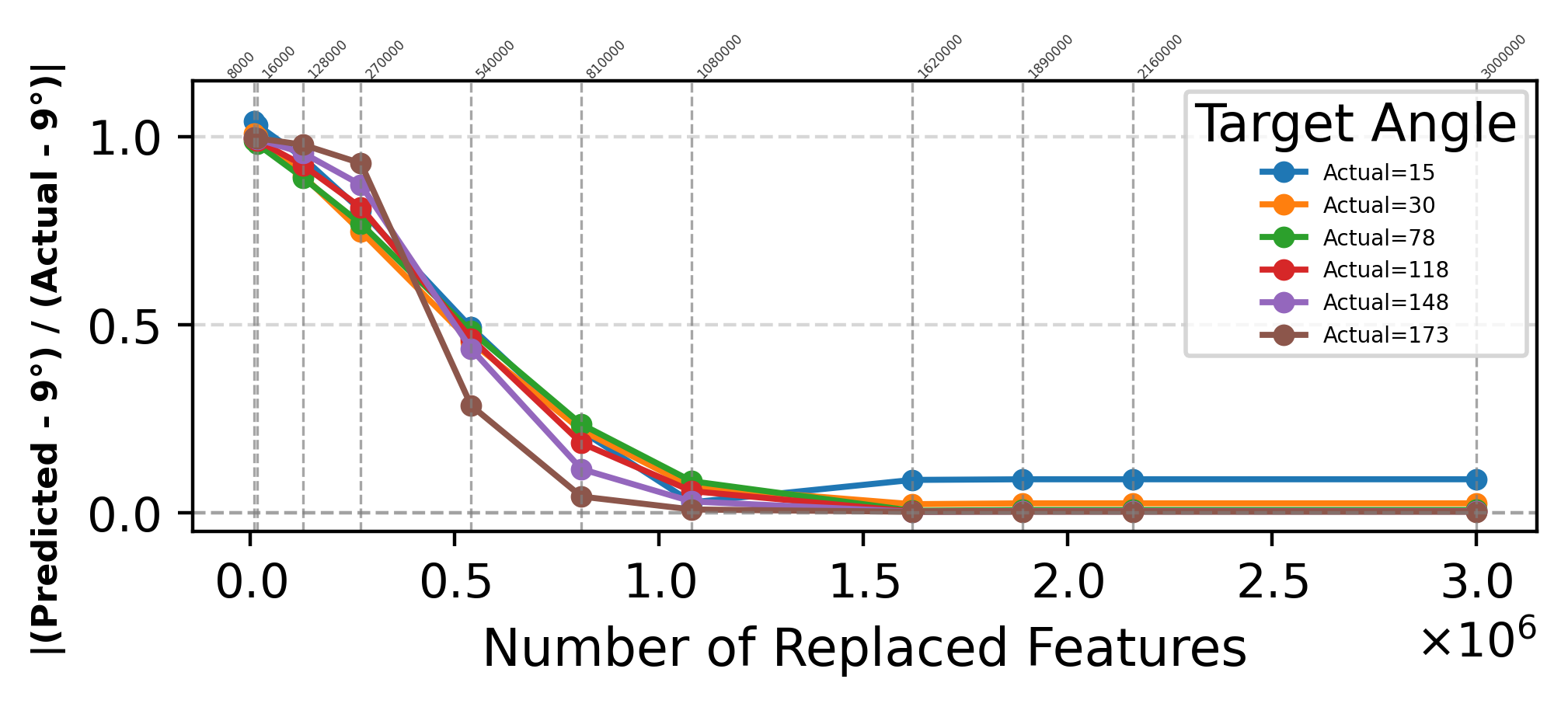}
        \caption{Ordered By Value Difference}
    \end{subfigure}%
    \hspace{-0.5em} 
    \begin{subfigure}{0.33\textwidth}
        \includegraphics[width=\linewidth]{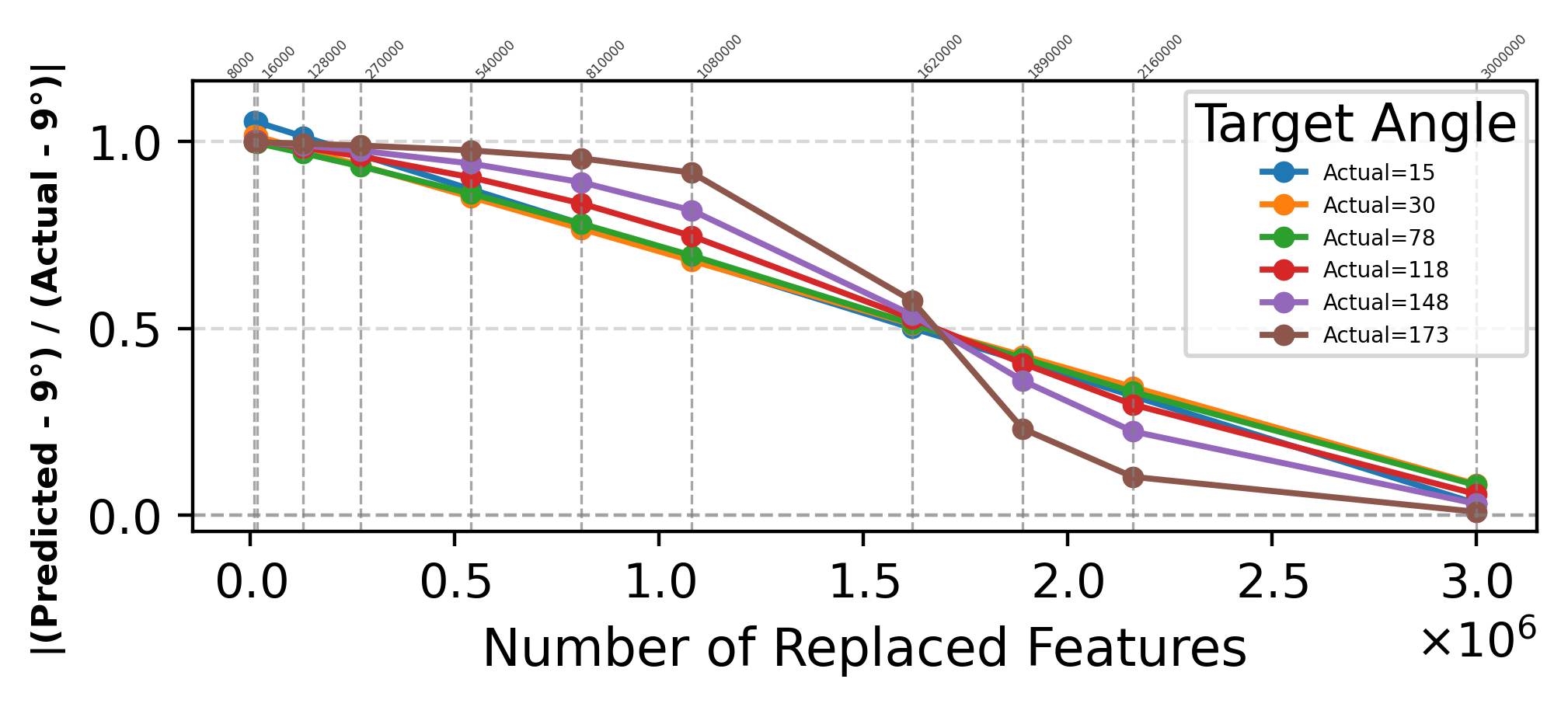}
        \caption{Picked Randomly}
    \end{subfigure}
   \vspace{-2mm}
    \caption{Incremental feature substitution for LLaVA-OneVision on images with the vase-indoor scene. No matter how the features are selected (according to the magnitude of the weights in the regressor or the absolute difference between anchor and target feature values, or randomly). 540,000 features or more must be replaced to fool the predictor. (Note that the x-axis is the number of feature substitutions times $10^6$.) This implies the orientation information is highly diffuse.}
   \vspace{-2mm}
    \label{fig:patch_analysis_llava-ov_vase-indoor}
\end{figure*}

\begin{figure*}[h!]
    \begin{subfigure}{0.33\textwidth}
        \includegraphics[width=\linewidth]{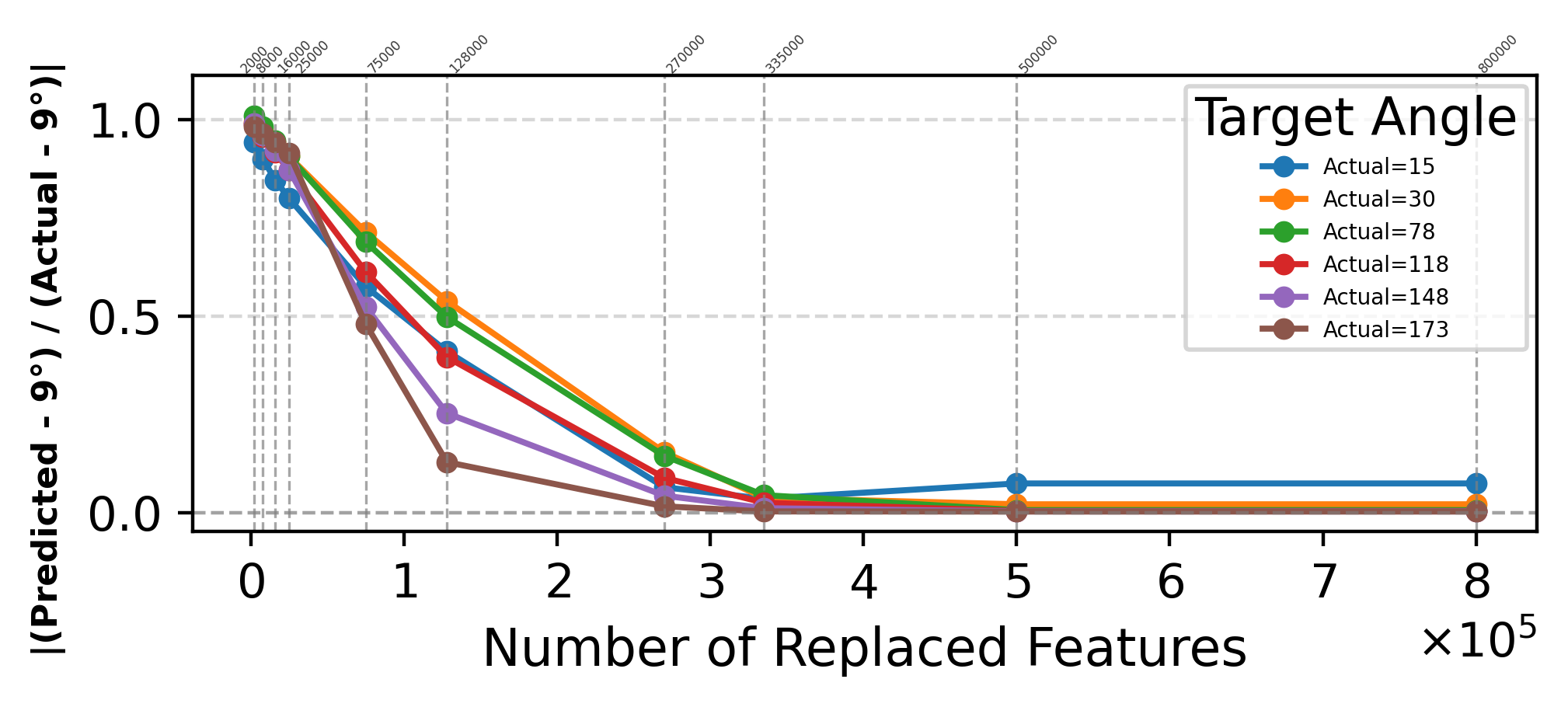}
        \caption{Ordered By Model Weight}
    \end{subfigure}%
    \hspace{-0.5em} 
    \begin{subfigure}{0.33\textwidth}
        \includegraphics[width=\linewidth]{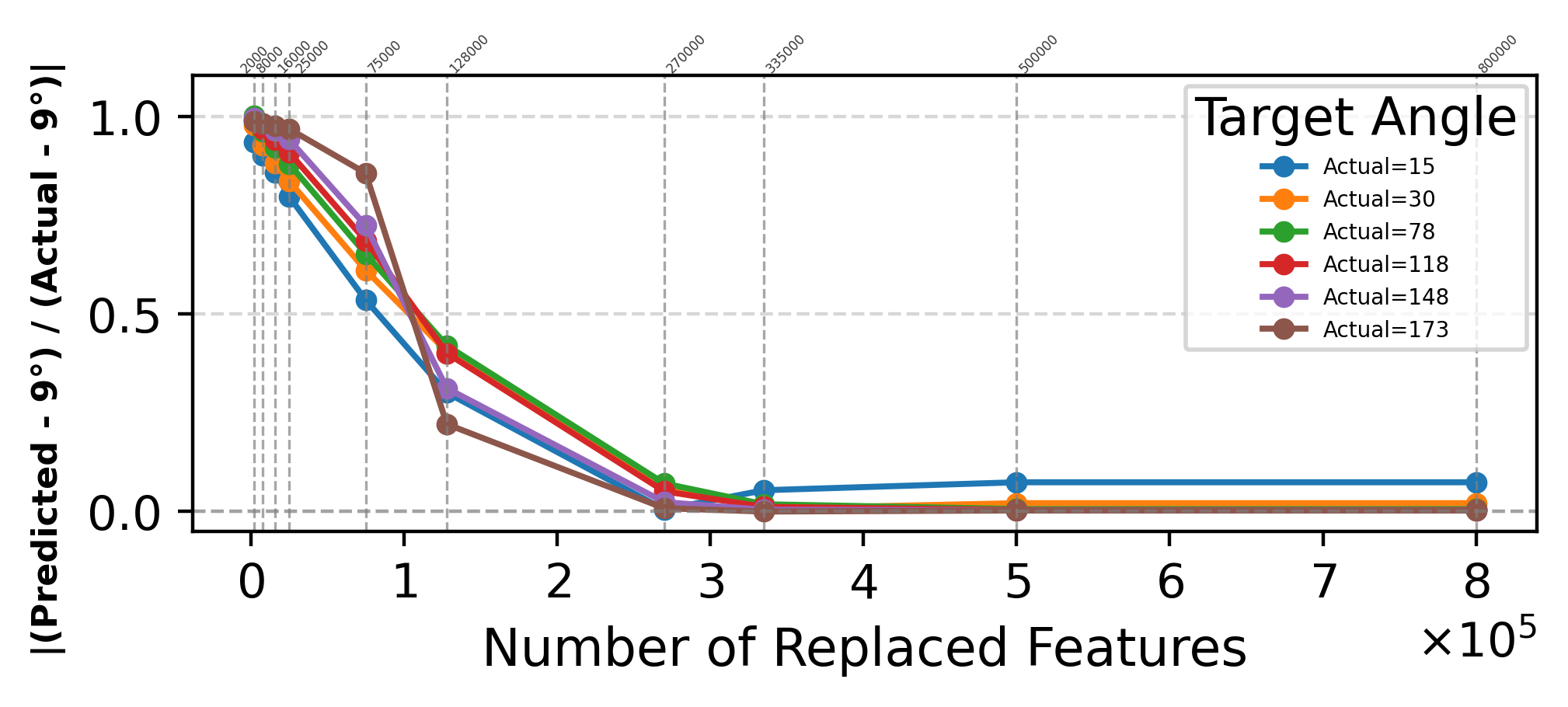}
        \caption{Ordered By Value Difference}
    \end{subfigure}%
    \hspace{-0.5em} 
    \begin{subfigure}{0.33\textwidth}
        \includegraphics[width=\linewidth]{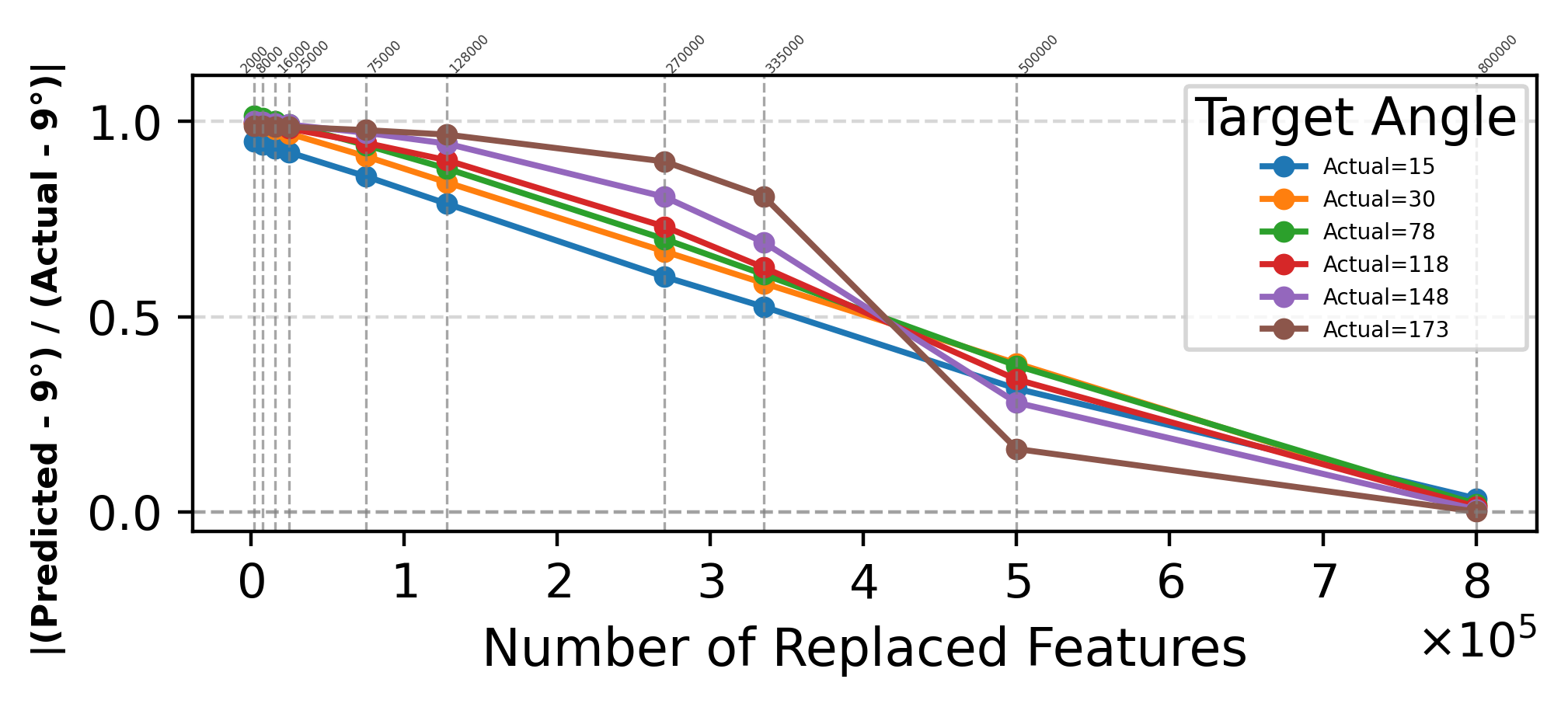}
        \caption{Picked Randomly}
    \end{subfigure}
   \vspace{-2mm}
    \caption{Incremental feature substitution for Qwen2.5-VL-7B-Instruct on images with the vase-indoor scene. No matter how the features are selected (according to the magnitude of the weights in the regressor or the absolute difference between anchor and target feature values, or randomly). 128,000 features or more must be replaced to fool the predictor. (Note that the x-axis is the number of feature substitutions times $10^5$.) This implies the orientation information is highly diffuse.}
   \vspace{-2mm}
    \label{fig:patch_analysis_qwen_vase-indoor}
\end{figure*}
\newpage
\begin{figure*}[h!]
    \begin{subfigure}{0.33\textwidth}
        \includegraphics[width=\linewidth]{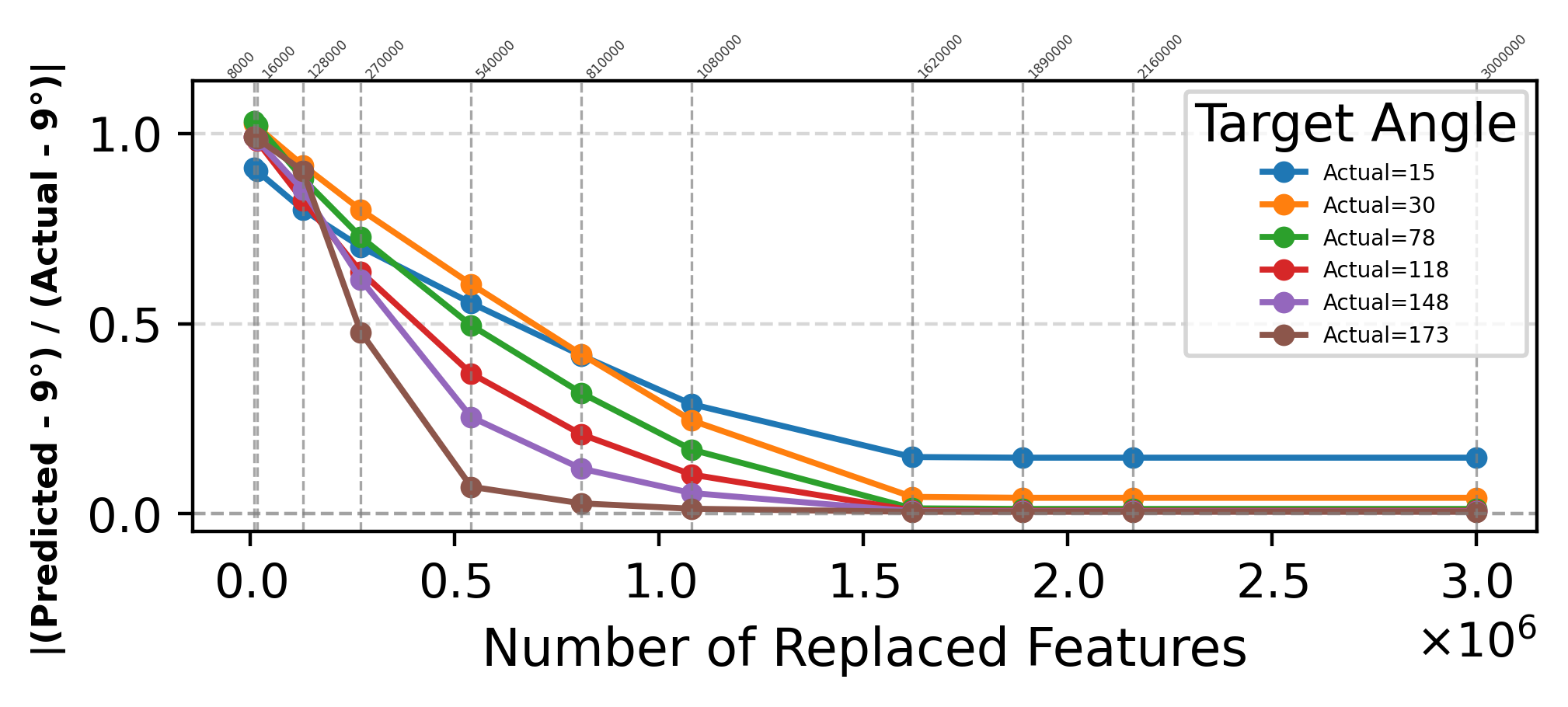}
        \caption{Ordered By Model Weight}
    \end{subfigure}%
    \hspace{-0.5em} 
    \begin{subfigure}{0.33\textwidth}
        \includegraphics[width=\linewidth]{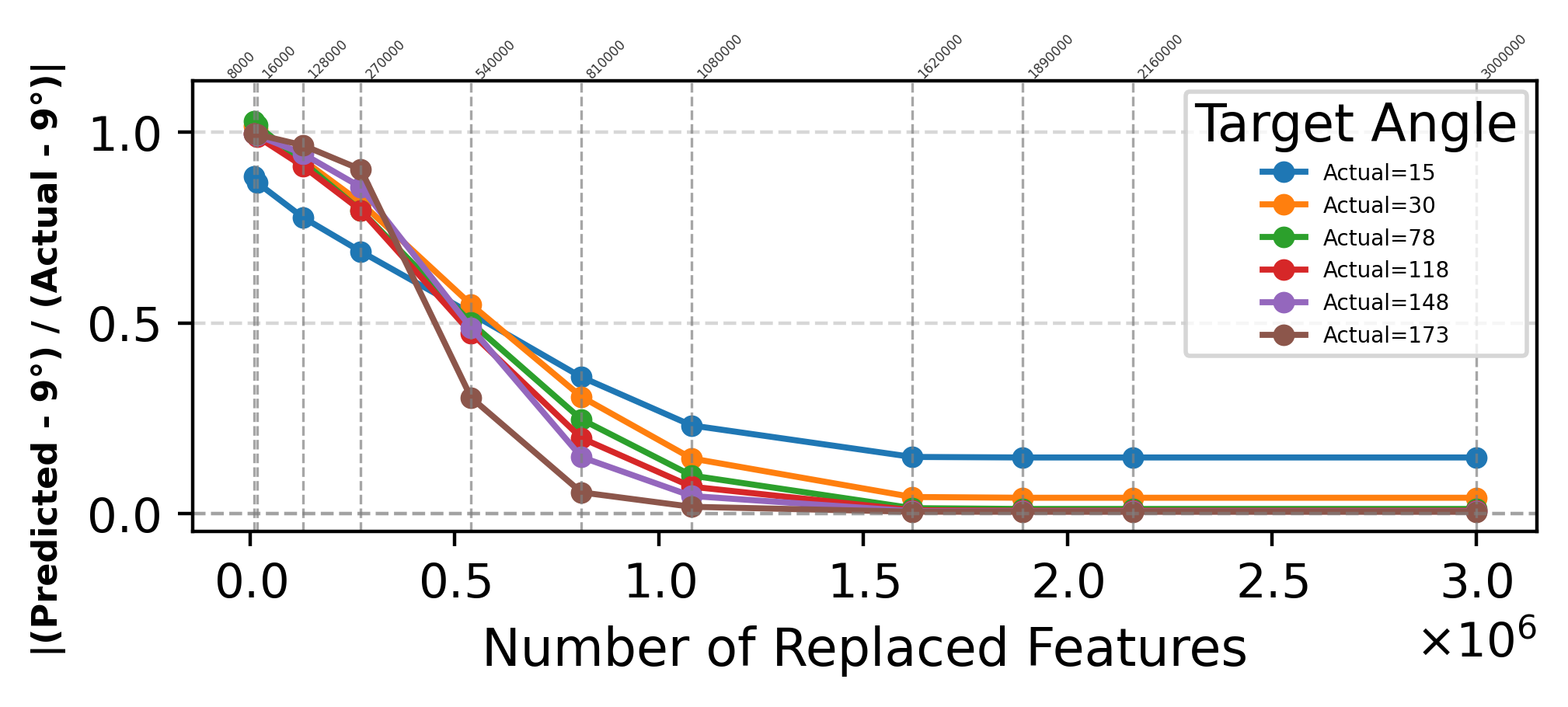}
        \caption{Ordered By Value Difference}
    \end{subfigure}%
    \hspace{-0.5em} 
    \begin{subfigure}{0.33\textwidth}
        \includegraphics[width=\linewidth]{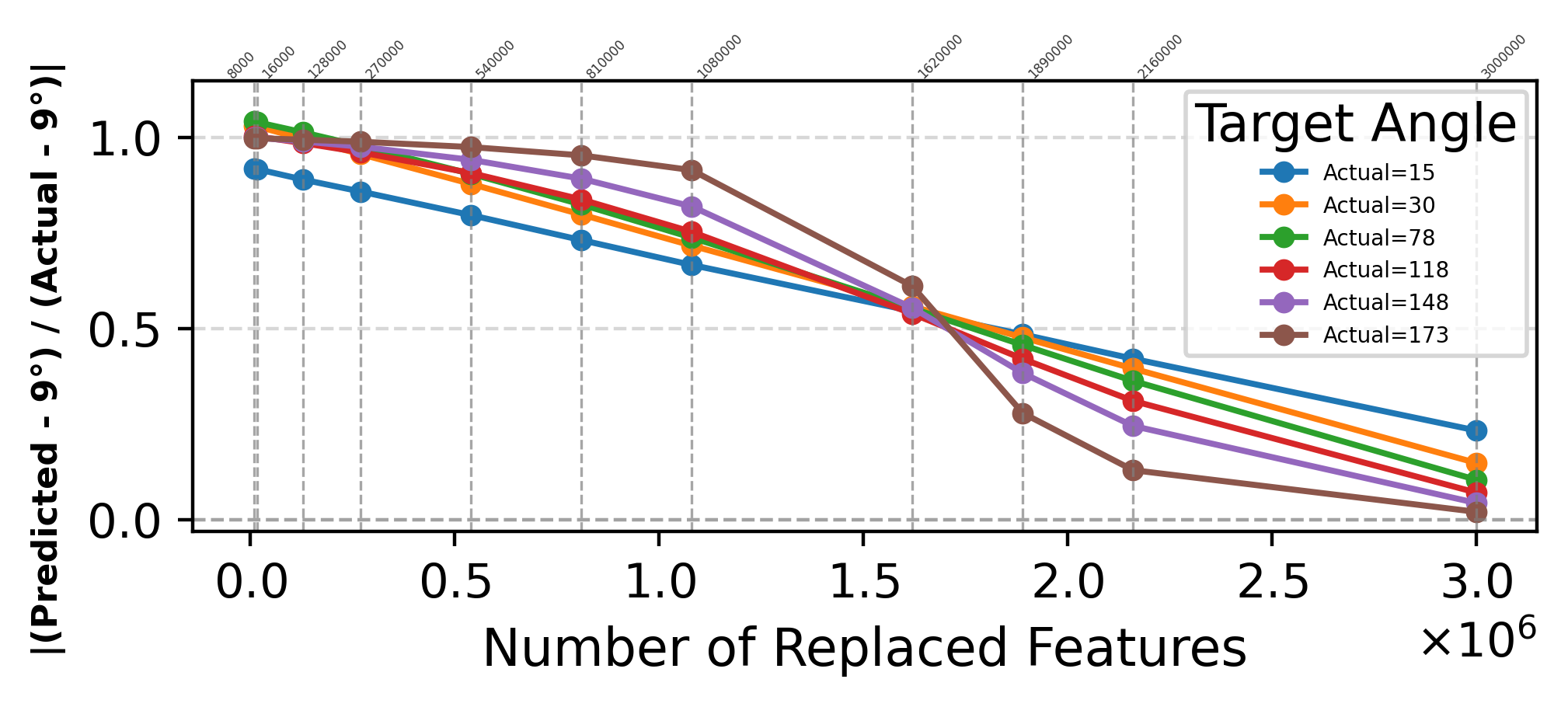}
        \caption{Picked Randomly}
    \end{subfigure}
   \vspace{-2mm}
    \caption{Incremental feature substitution for LLaVA-OneVision on images with the vase-toaster-indoor scene. No matter how the features are selected (according to the magnitude of the weights in the regressor or the absolute difference between anchor and target feature values, or randomly). 540,000 features or more must be replaced to fool the predictor. (Note that the x-axis is the number of feature substitutions times $10^6$.) This implies the orientation information is highly diffuse.}
   \vspace{-2mm}
    \label{fig:patch_analysis_llava-ov_vase-toaster-indoor}
\end{figure*}

\begin{figure*}[h!]

    \begin{subfigure}{0.33\textwidth}
        \includegraphics[width=\linewidth]{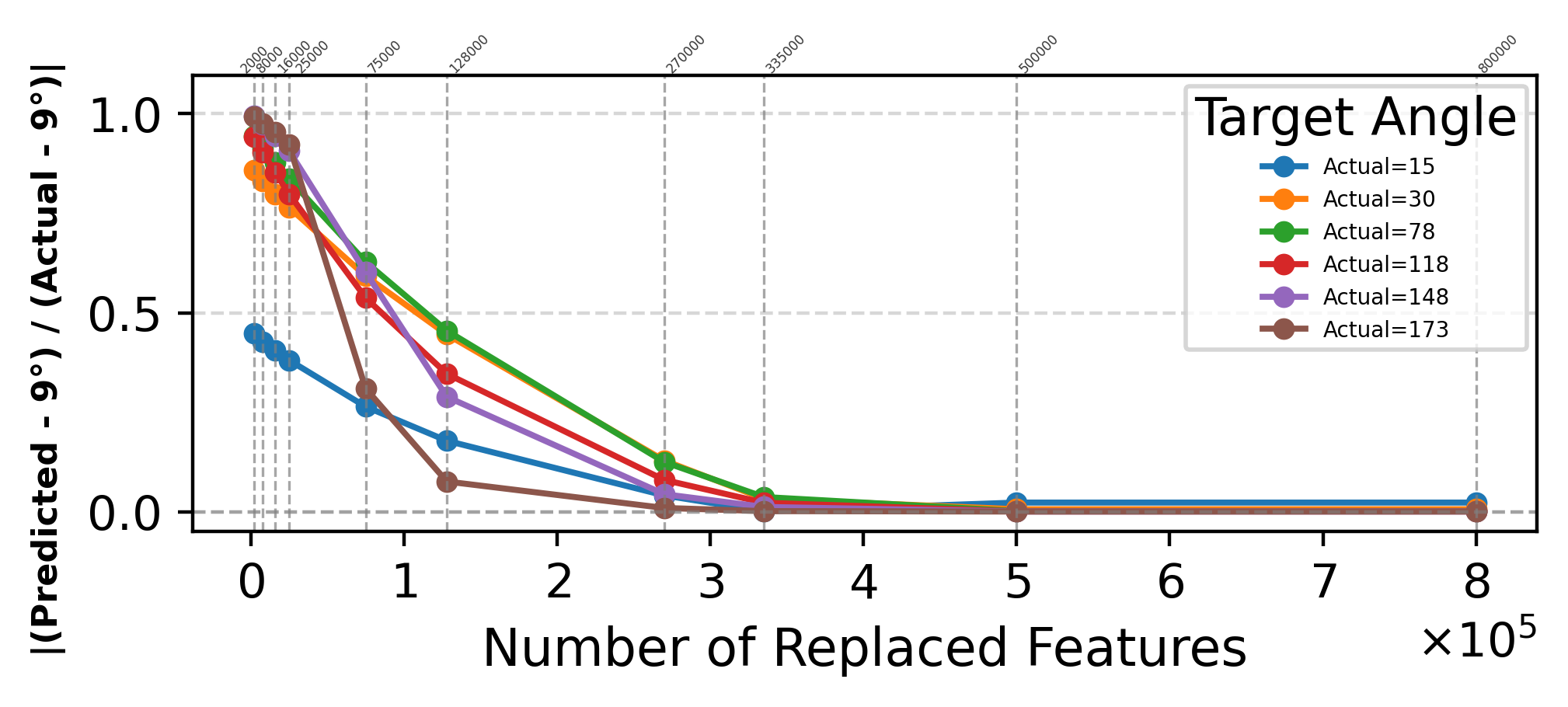}
        \caption{Ordered By Model Weight}
    \end{subfigure}%
    \hspace{-0.5em} 
    \begin{subfigure}{0.33\textwidth}
        \includegraphics[width=\linewidth]{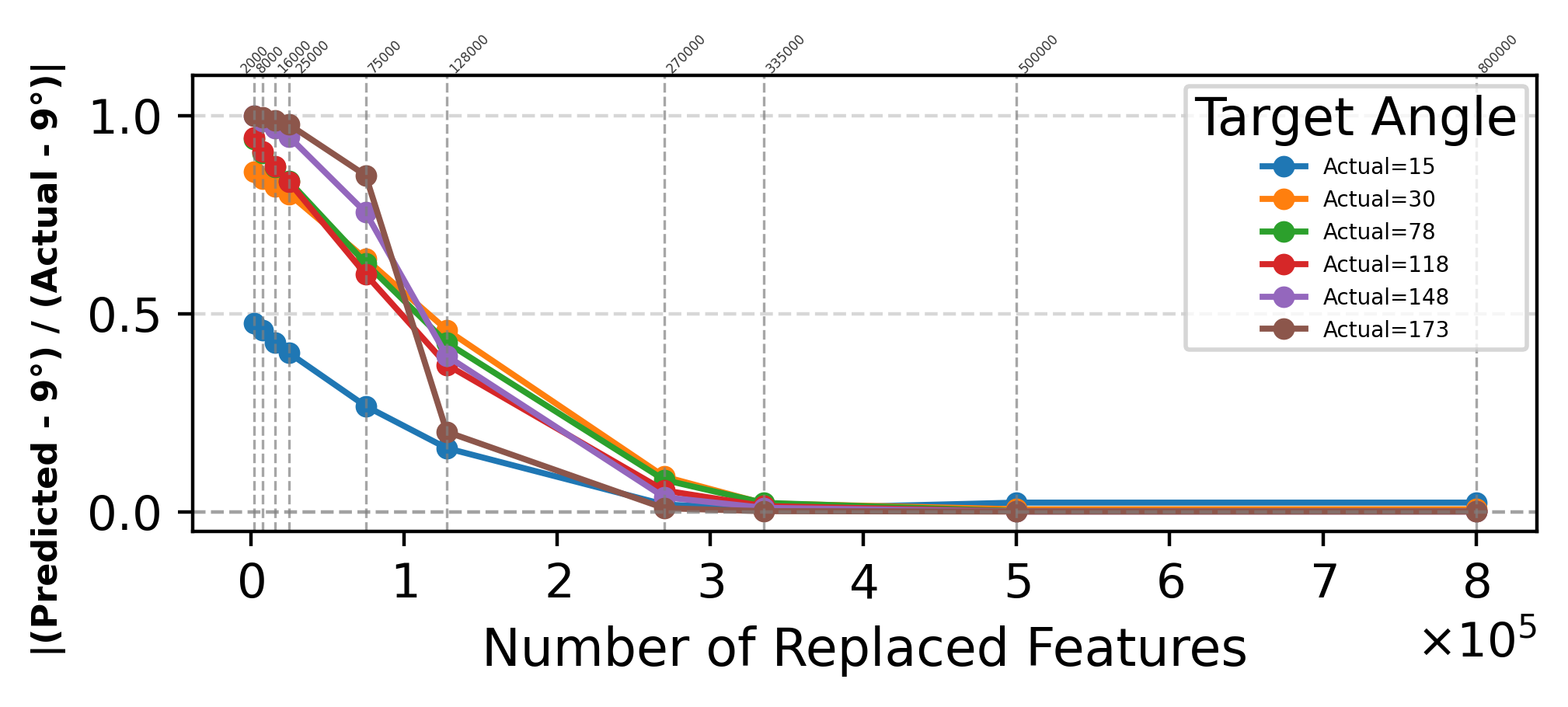}
        \caption{Ordered By Value Difference}
    \end{subfigure}%
    \hspace{-0.5em} 
    \begin{subfigure}{0.33\textwidth}
        \includegraphics[width=\linewidth]{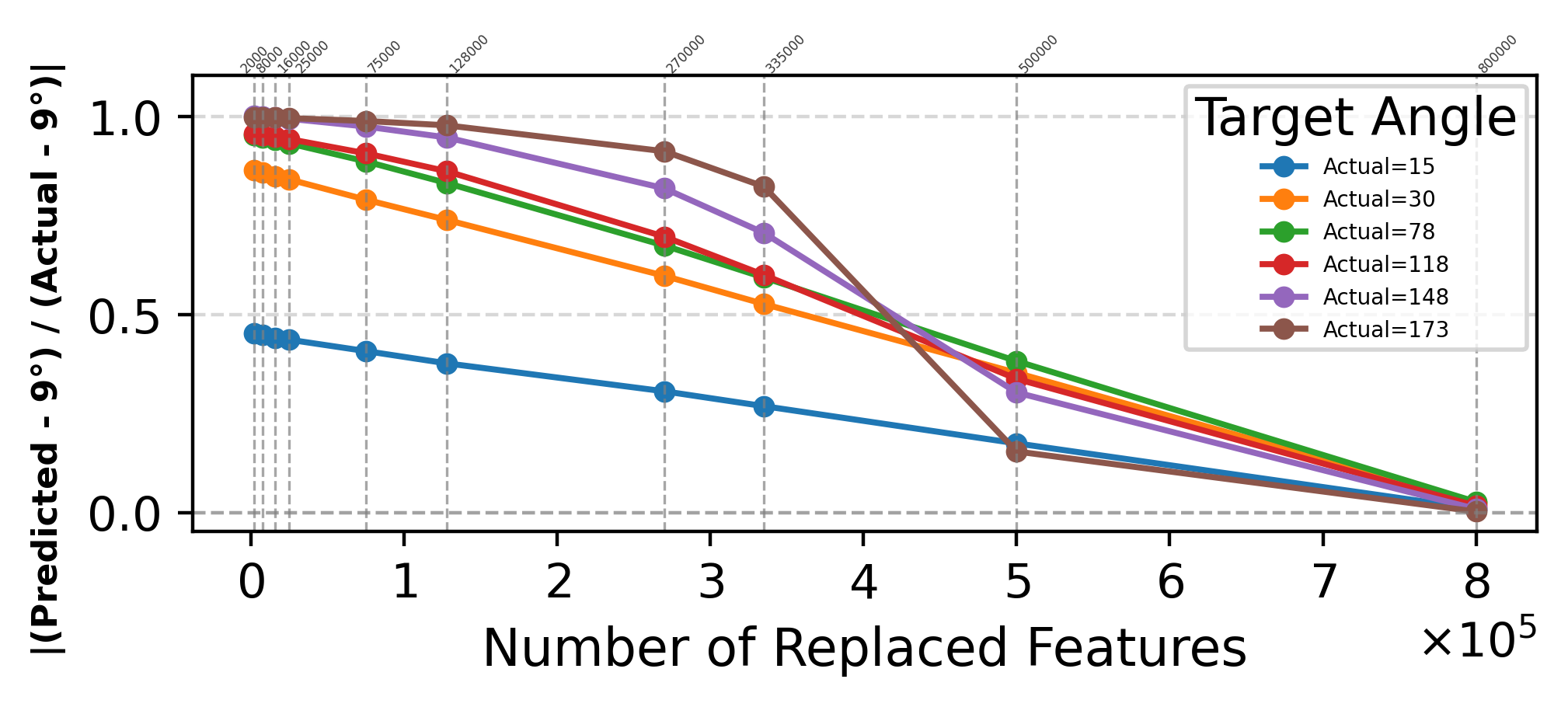}
        \caption{Picked Randomly}
    \end{subfigure}
   \vspace{-2mm}
    \caption{Incremental feature substitution for Qwen2.5-VL-7B-Instruct on images with the vase-toaster-indoor scene. No matter how the features are selected (according to the magnitude of the weights in the regressor or the absolute difference between anchor and target feature values, or randomly). 128,000 features or more must be replaced to fool the predictor. (Note that the x-axis is the number of feature substitutions times $10^5$.) This implies the orientation information is highly diffuse.}
   \vspace{-2mm}
    \label{fig:patch_analysis_qwen_vase-toaster-indoor}
\end{figure*}
\clearpage
\newpage
\subsection{Feature Substitution Plots for LLaVA 1.5 and 1.6}
\label{app:ftr-subs-llava1.5-1.6}

\begin{figure*}[h!]

    \begin{subfigure}{0.30\textwidth}
        \includegraphics[width=\linewidth]{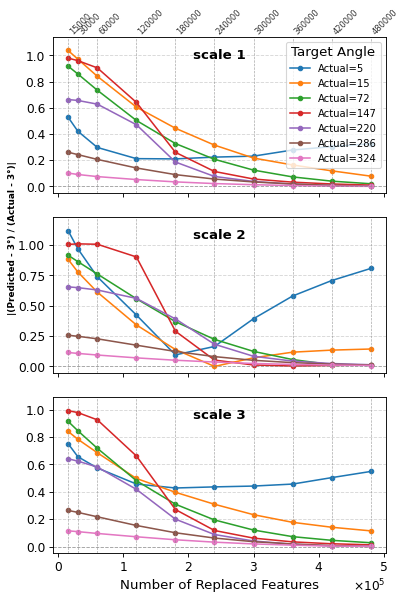}
        \caption{Ordered By Model Weight}
    \end{subfigure}%
    \hspace{-0.5em} 
    \begin{subfigure}{0.30\textwidth}
        \includegraphics[width=\linewidth]{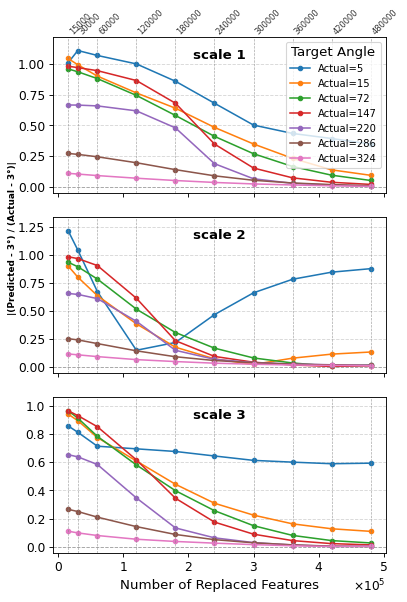}
        \caption{Ordered By Value Difference}
    \end{subfigure}%
    \hspace{-0.5em} 
    \begin{subfigure}{0.30\textwidth}
        \includegraphics[width=\linewidth]{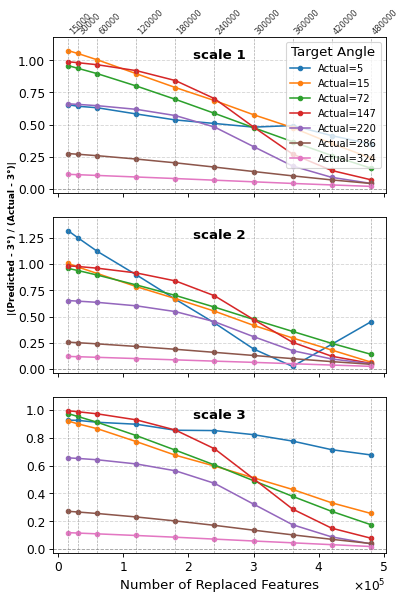}
        \caption{Picked Randomly}
    \end{subfigure}
   \vspace{-2mm}
    \caption{Incremental feature substitution for LLaVA 1.5 on images with the beach background scene. No matter how the features are selected (according to the magnitude of the weights in the regressor or the absolute difference between anchor and target feature values, or randomly). 20,000 features or more must be replaced to fool the predictor. (Note that the x-axis is the number of feature substitutions times $10^5$.) This implies the orientation information is highly diffuse.}
   \vspace{-2mm}
    \label{fig:patch_analysis_llava1.5_dog-on-beach}
\end{figure*}

\begin{figure*}[h!]
    \begin{subfigure}{0.32\textwidth}
        \includegraphics[width=\linewidth]{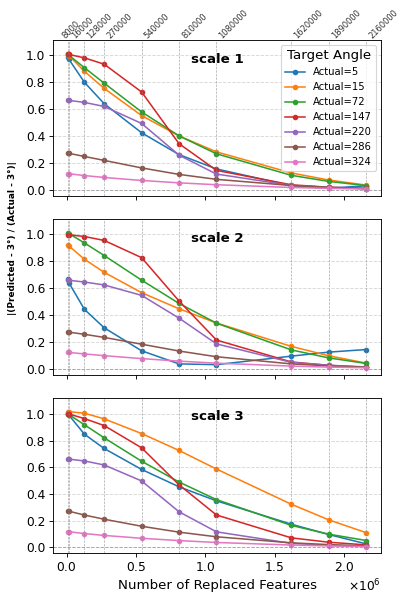}
        \caption{Ordered By Model Weight}
    \end{subfigure}%
    \hfill
    \begin{subfigure}{0.32\textwidth}
        \includegraphics[width=\linewidth]{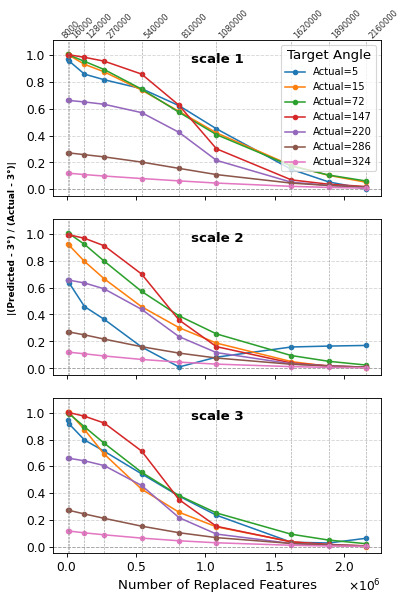}
        \caption{Ordered By Value Difference}
    \end{subfigure}%
    \hfill
    \begin{subfigure}{0.32\textwidth}
        \includegraphics[width=\linewidth]{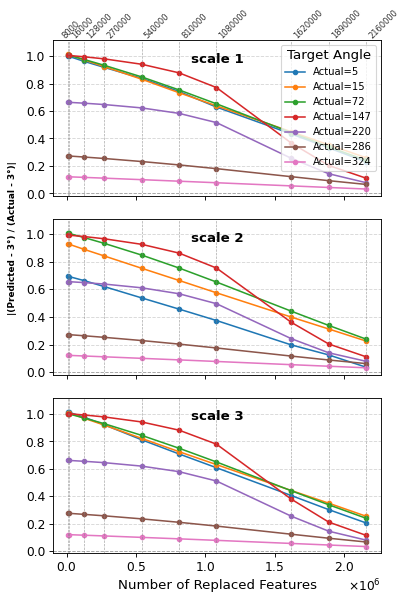}
        \caption{Picked Randomly}
    \end{subfigure}
   \vspace{-2mm}
    \caption{Incremental feature substitution for LLaVA 1.6 on images with the beach background scene. No matter how the features are selected (according to the magnitude of the weights in the regressor or the absolute difference between anchor and target feature values, or randomly). 16,000 features or more must be replaced to fool the predictor. (Note that the x-axis is the number of feature substitutions times $10^6$.) This implies the orientation information is highly diffuse.}
    \label{fig:patch_analysis_llava1.6_dog-on-beach}
   \vspace{-2mm}
\end{figure*}
\newpage

\begin{figure*}[t]
    \begin{subfigure}{0.33\textwidth}
        \includegraphics[width=\linewidth]{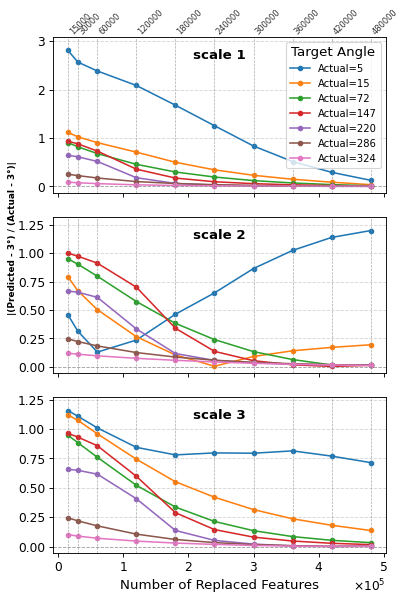}
        \caption{Ordered By Model Weight}
    \end{subfigure}%
    \hspace{-0.5em} 
    \begin{subfigure}{0.33\textwidth}
        \includegraphics[width=\linewidth]{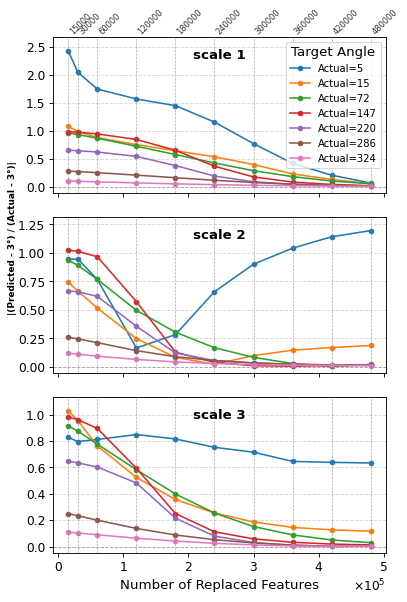}
        \caption{Ordered By Value Difference}
    \end{subfigure}%
    \hspace{-0.5em} 
    \begin{subfigure}{0.33\textwidth}
        \includegraphics[width=\linewidth]{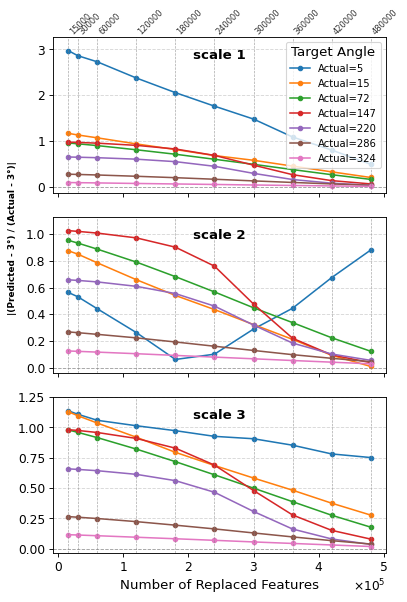}
        \caption{Picked Randomly}
    \end{subfigure}
   \vspace{-2mm}
    \caption{Incremental feature substitution for LLaVA 1.5 on images with the fish background scene. No matter how the features are selected (according to the magnitude of the weights in the regressor or the absolute difference between anchor and target feature values, or randomly). 20,000 features or more must be replaced to fool the predictor. (Note that the x-axis is the number of feature substitutions times $10^5$.) This implies the orientation information is highly diffuse.}
   \vspace{-2mm}
    \label{fig:feature_substitution_llava1.5_lizard-on-fish}
\end{figure*}

\begin{figure*}[h]
    \begin{subfigure}{0.33\textwidth}
        \includegraphics[width=\linewidth]{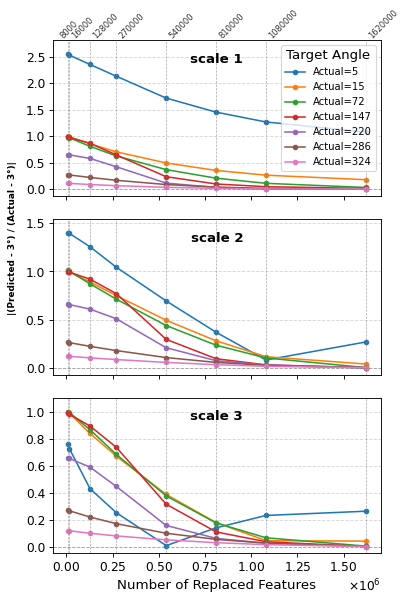}
        \caption{Ordered By Model Weight}
    \end{subfigure}%
    \hspace{-0.5em} 
    \begin{subfigure}{0.33\textwidth}
        \includegraphics[width=\linewidth]{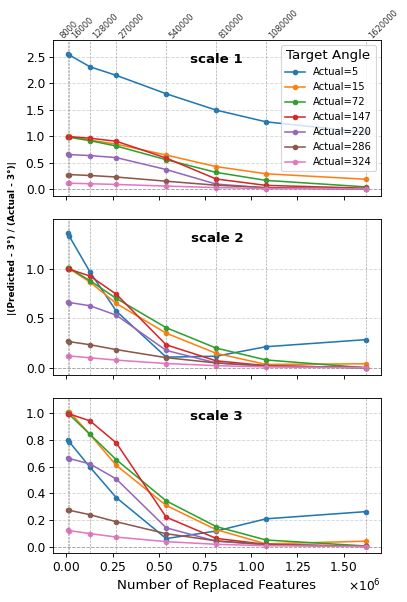}
        \caption{Ordered By Value Difference}
    \end{subfigure}%
    \hspace{-0.5em} 
    \begin{subfigure}{0.33\textwidth}
        \includegraphics[width=\linewidth]{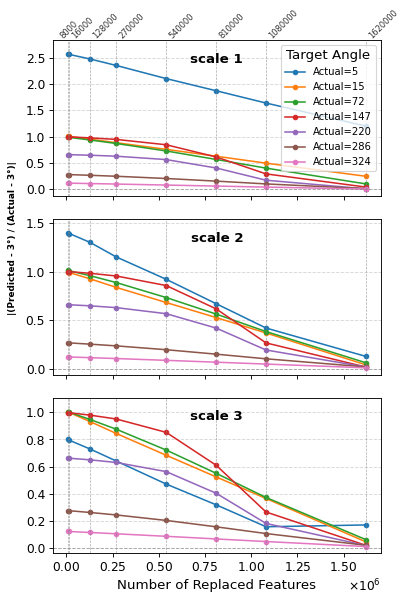}
        \caption{Picked Randomly}
    \end{subfigure}
   \vspace{-2mm}
    \caption{Incremental feature substitution for LLaVA 1.6 on images with the fish background scene. No matter how the features are selected (according to the magnitude of the weights in the regressor or the absolute difference between anchor and target feature values, or randomly). 16,000 features or more must be replaced to fool the predictor. (Note that the x-axis is the number of feature substitutions times $10^6$.) This implies the orientation information is highly diffuse.}
   \vspace{-2mm}
    \label{fig:feature_substitution_llava1.6_lizard-on-fish}
\end{figure*}
\newpage
\begin{figure*}[h]
    \begin{subfigure}{0.33\textwidth}
        \includegraphics[width=\linewidth]{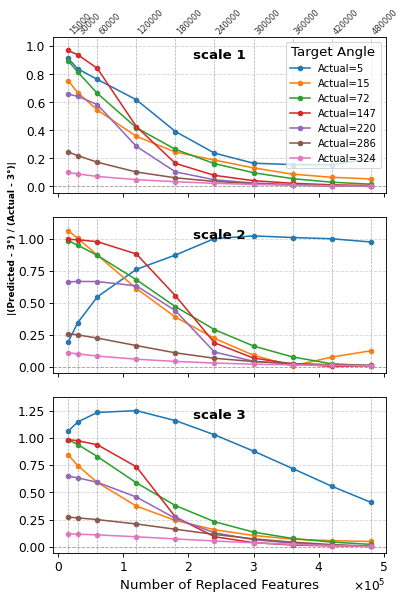}
        \caption{Ordered By Model Weight}
    \end{subfigure}%
    \hspace{-0.5em} 
    \begin{subfigure}{0.33\textwidth}
        \includegraphics[width=\linewidth]{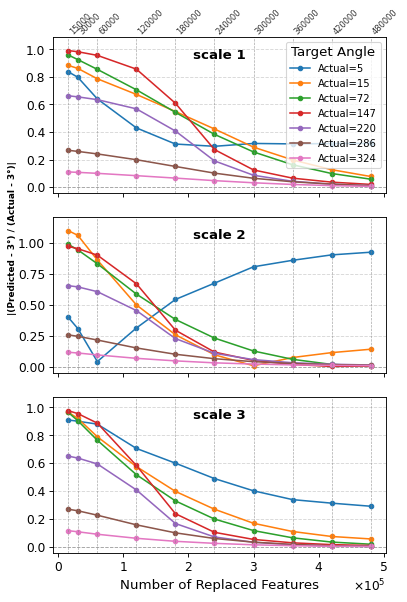}
        \caption{Ordered By Value Difference}
    \end{subfigure}%
    \hspace{-0.5em} 
    \begin{subfigure}{0.33\textwidth}
        \includegraphics[width=\linewidth]{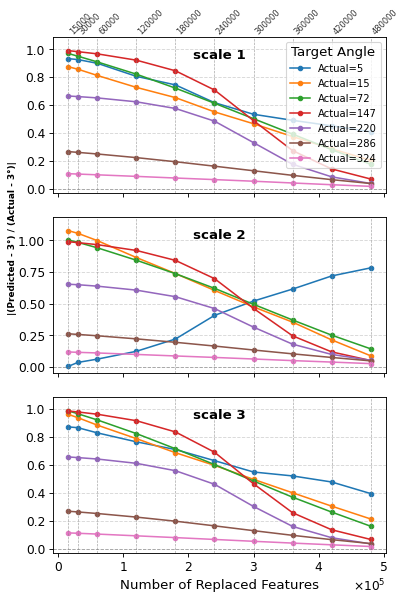}
        \caption{Picked Randomly}
    \end{subfigure}
   \vspace{-2mm}
    \caption{Incremental feature substitution for LLaVA 1.5 on images with the indoor background scene. No matter how the features are selected (according to the magnitude of the weights in the regressor or the absolute difference between anchor and target feature values, or randomly). 20,000 features or more must be replaced to fool the predictor. (Note that the x-axis is the number of feature substitutions times $10^5$.) This implies the orientation information is highly diffuse.}
   \vspace{-2mm}
    \label{fig:feature_substitution_llava1.5_train-on-indoor}
\end{figure*}

\begin{figure*}[h]
    \begin{subfigure}{0.33\textwidth}
        \includegraphics[width=\linewidth]{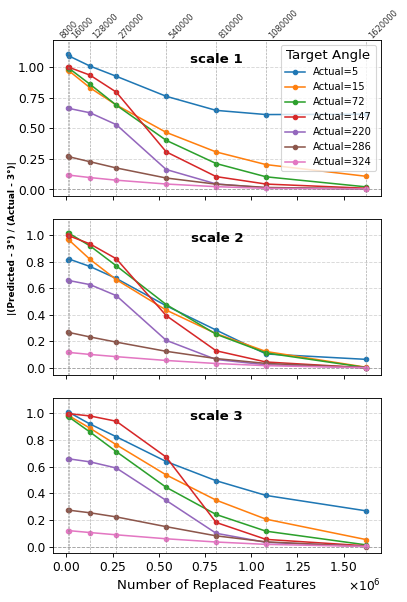}
        \caption{Ordered By Model Weight}
    \end{subfigure}%
    \hspace{-0.5em} 
    \begin{subfigure}{0.33\textwidth}
        \includegraphics[width=\linewidth]{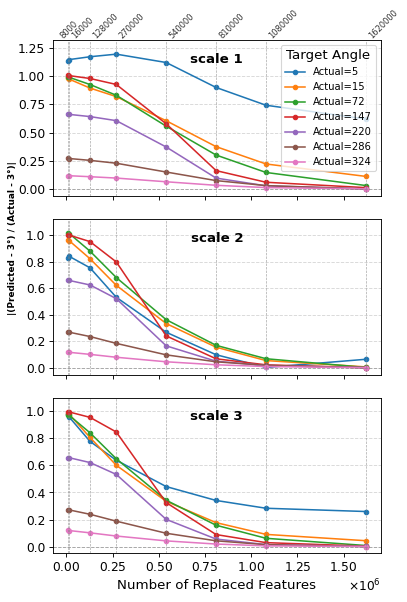}
        \caption{Ordered By Value Difference}
    \end{subfigure}%
    \hspace{-0.5em} 
    \begin{subfigure}{0.33\textwidth}
        \includegraphics[width=\linewidth]{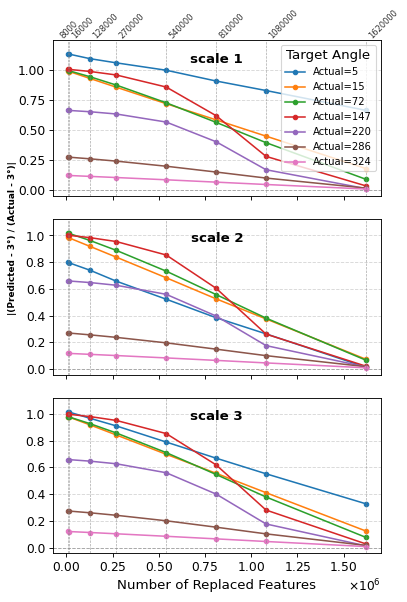}
        \caption{Picked Randomly}
    \end{subfigure}
   \vspace{-2mm}
    \caption{Incremental feature substitution for LLaVA 1.6 on images with the indoor background scene. No matter how the features are selected (according to the magnitude of the weights in the regressor or the absolute difference between anchor and target feature values, or randomly). 16,000 features or more must be replaced to fool the predictor. (Note that the x-axis is the number of feature substitutions times $10^6$.) This implies the orientation information is highly diffuse.}
   \vspace{-2mm}
    \label{fig:feature_substitution_llava1.6_train-on-indoor}
\end{figure*}

\clearpage
\newpage
\subsection{Background vs. Foreground Rotation}
\label{app:bg-fg}
In trying to better understand how orientation information is embedded by visual encoders, we looked by at Table~\ref{tab:llava_qwen_comparison} and noticed something small but odd: when estimating foreground orientations, the MAE does not get larger as the foreground patch gets smaller. In fact, although the effect is small, predictions are better for smaller image patches. This led us to investigate the relationship between backgrounds and the estimated orientation of the foreground.

\begin{figure}[h!]
  \centering
   \includegraphics[width=8cm, height=5cm]{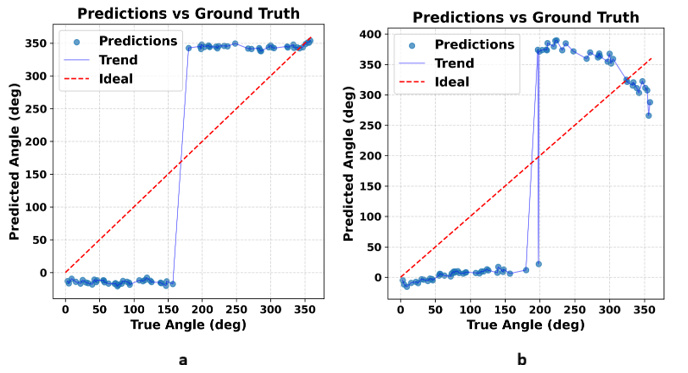}
   \vspace{-2mm}
   \caption{Results of foreground orientation estimation by LLaVA 1.5 for dog images (scale 1) when (a) only BG (b) BG and FG are rotated in the test images and training set consists of images with only FG rotated.}
   \vspace{-2mm}
\label{fig:dogOnBeach_bgROT_test_1.5-scale1}
   \hfill
\end{figure}

\begin{figure}[h!]
  \centering
   \includegraphics[width=6.5cm, height=4cm]{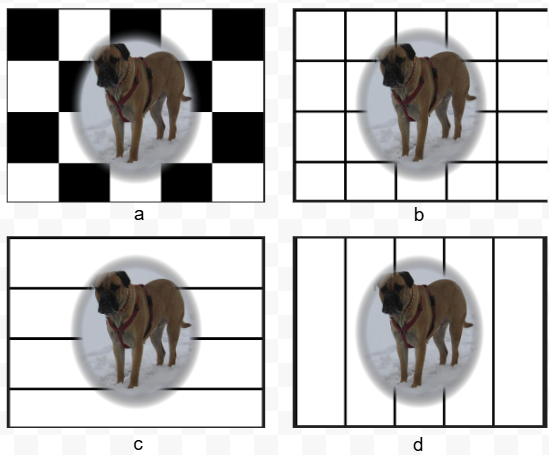}
   \vspace{-2mm}
   \caption{Images with synthetic backgrounds used to test the impact of background rotations on foreground orientation estimation.}
   \vspace{-2mm}
\label{fig:dog_syn_backgrounds}
   \hfill
\end{figure}

\begin{figure}[h!]
  \centering
   \includegraphics[width=8cm, height=5cm]{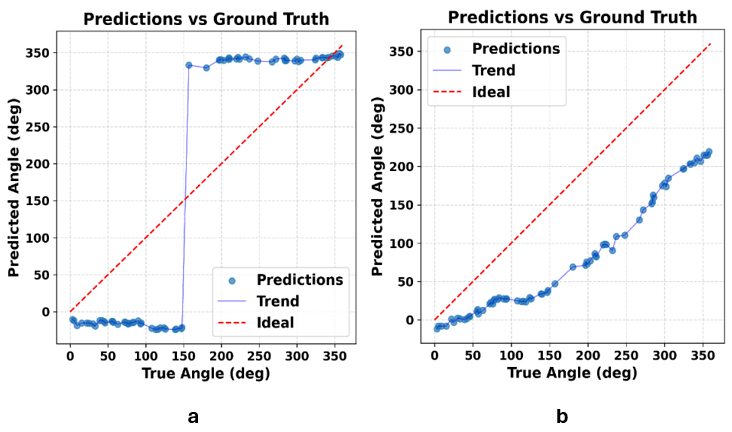}
   \vspace{-2mm}
   \caption{Results of foreground orientation estimation by LLaVA 1.6 for dog images (scale 1) when (a) only BG (b) BG and FG are rotated in the test images and training set consists of images with only FG rotated.}
   \vspace{-2mm}
\label{fig:dogOnBeach_bgROT_test_1.6-scale1}
   \hfill
\end{figure}

\begin{figure}[h!]
  \centering
   \includegraphics[width=8cm, height=5cm]{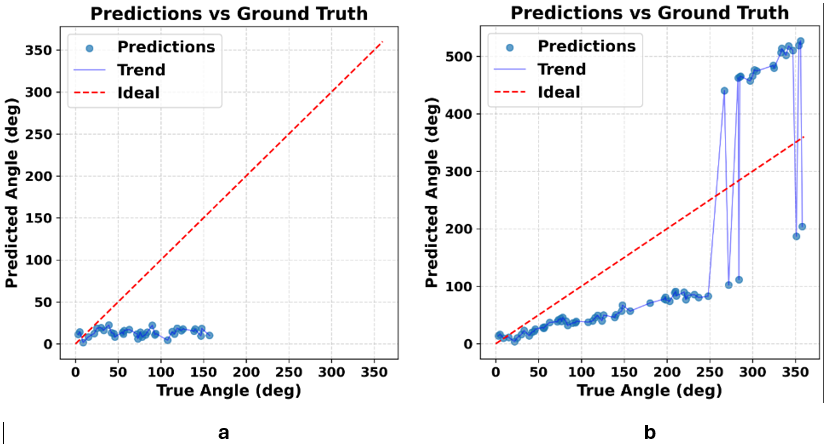}
   \vspace{-2mm}
   \caption{Results of foreground orientation estimation by LLaVA 1.5 for lizard images (scale 1) when (a) only BG (b) BG and FG are rotated in the test images and training set consists of images with only FG rotated.}
   \vspace{-2mm}
\label{fig:lizardOnFish_bgROT_test_1.5-scale1}
   \hfill
\end{figure}

\begin{figure}[h!]
  \centering
   \includegraphics[width=8cm, height=5cm]{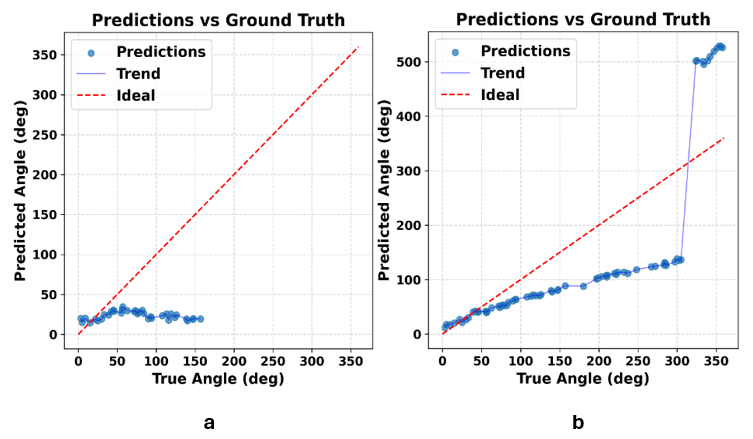}
   \vspace{-2mm}
   \caption{Results of foreground orientation estimation by LLaVA 1.6 for lizard images (scale 1) when (a) only BG (b) BG and FG are rotated in the test images and training set consists of images with only FG rotated.}
   \vspace{-2mm}
\label{fig:lizardOnFish_bgROT_test_1.6-scale1}
   \hfill
\end{figure}

\begin{figure}[h!]
  \centering
   \includegraphics[width=8cm, height=5cm]{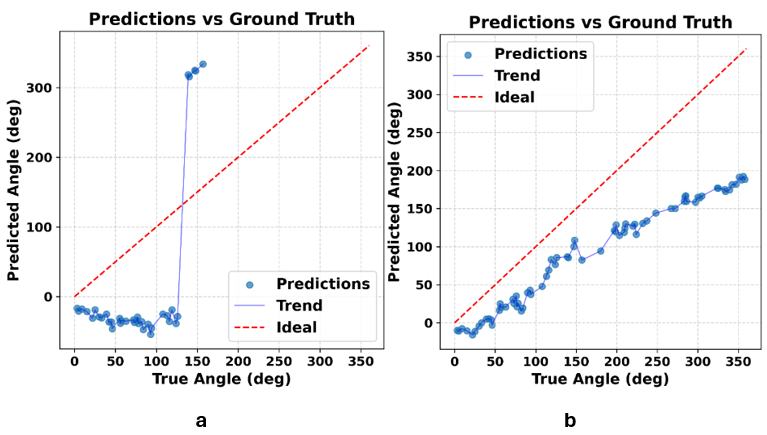}
   \vspace{-2mm}
   \caption{Results of foreground orientation estimation by LLaVA 1.5 for train images (scale 1) when (a) only BG (b) BG and FG are rotated in the test images and training set consists of images with only FG rotated.}
   \vspace{-2mm}
\label{fig:trainOnIndoor_bgROT_test_1.5-scale1}
   \hfill
\end{figure}

\begin{figure}[h!]
  \centering
   \includegraphics[width=8cm, height=5cm]{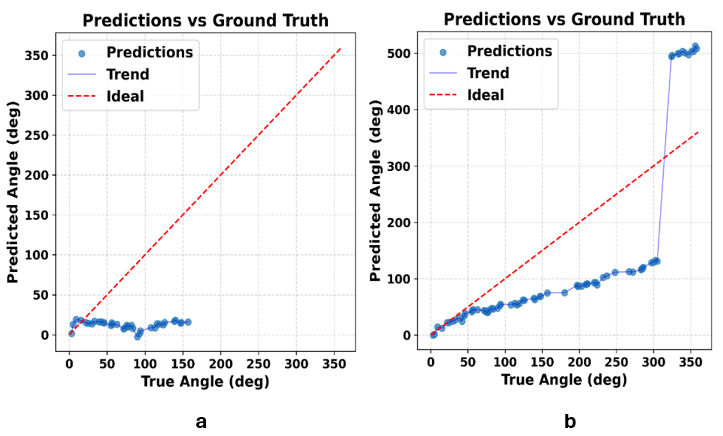}
   \vspace{-2mm}
   \caption{Results of foreground orientation estimation by LLaVA 1.6 for train images (scale 1) when (a) only BG (b) BG and FG are rotated in the test images and training set consists of images with only FG rotated.}
   \vspace{-2mm}
\label{fig:trainOnIndoor_bgROT_test_1.6-scale1}
   \hfill
\end{figure}

We evaluated the model trained on only foreground rotated images using two variant image sets - (1) background rotated and foreground static, and (2) both background and foreground rotated. The results for LLaVA 1.5 on images with dog foregrounds (scale 1) are shown in Figures \ref{fig:dogOnBeach_bgROT_test_1.5-scale1} - \ref{fig:trainOnIndoor_bgROT_test_1.6-scale1}. Both experiments fared poorly with an MAE upwards of $80^{\circ}$. 

To understand why the model is unable to predict the foreground orientation when the background is rotated, we repeat the experiments on an image set with synthetic backgrounds (see Figure \ref{fig:dog_syn_backgrounds}) with horizontal and vertical lines. Our hypothesis is that accurate foreground orientation is dependent on the background being in its canonical orientation, so this experiment gauges the sensitivity of the foreground orientation estimation to edges in the background. Results are in Tables \ref{tab:dog_synthetic_mae}-\ref{tab:train_synthetic_mae}. Our experiments show that accurately estimating the orientation of the foreground is dependent on the orientation of the background. If the training set contains background and foreground rotations, then the test with both background and foreground rotations perform very well. But this is not the case when the training and test sets contain only background rotations. This requires further investigation.


\begin{table*}[t]

 \centering
 \scriptsize

\renewcommand{\arraystretch}{1.1}
\setlength{\tabcolsep}{6pt}
\begin{tabular}{|l|l|c|c|c|}
\hline
\textbf{S/N} & \textbf{Train  Set} & \textbf{Test set} & \multicolumn{2}{|c|}{\textbf{\makecell{MAE \\ (degrees)}}} \\
\hline
\multicolumn{1}{|c|}{} & \multicolumn{1}{c|}{} & \multicolumn{1}{c|}{} & \textbf{LLaVA 1.5} & \textbf{LLaVA 1.6} \\
\hline

\multirow{12}{*}{\textbf{1}}
& \multirow{3}{*}{\makecell{dog on \\ chessboard \\(FG rotated)}}
 & baseline (FG rotated) & 1.1 & 0.71\\
&  & BG + FG rotated & 96.5 &92.34 \\
&  & BG rotated & 80.09 & 80.48\\
\cline{2-5} 

& \multirow{3}{*}{\makecell{dog on \\ grid\\ (FG rotated)}}
 & baseline (FG rotated) & 1.28 & 0.68\\
&  & BG + FG rotated & 92.13 & 88.14\\
&  & BG rotated & 81.77 & 91.29\\
\cline{2-5}

& \multirow{3}{*}{\makecell{dog on \\ horizontal lines\\(FG rotated)}}
 & baseline (FG rotated) &  1.35 & 0.7\\
&  & BG + FG rotated & 86.2 & 89.45\\
&  & BG rotated & 75.03 & 80.59\\
\cline{2-5}

& \multirow{3}{*}{\makecell{dog on \\ vertical lines\\(FG rotated)}}
 & baseline (FG rotated) & 1.4 & 0.64\\
&  & BG + FG rotated & 84.2 & 86.67\\
&  & BG rotated &92.06 & 81.28\\
\hline
\hline

\multirow{2}{*}{\textbf{2}}
& {\makecell{dog on \\ grid lines\\ - BG + FG rot.}}
& {\makecell{dog on \\ vertical lines\\ - BG + FG rot.}} & 1.87 &1.37 \\
\cline{2-5} 
& {\makecell{dog on \\ grid lines\\ - BG + FG rot.}}
& {\makecell{dog on \\ horizontal lines\\ - BG + FG rot.}} & 1.75 &2.14 \\
\hline
\hline

\multirow{2}{*}{\textbf{3}}
& {\makecell{dog on \\ chessboard\\ - BG + FG rot.}}
& {\makecell{dog on \\ vertical lines\\ - BG + FG rot.}} &2.73 &3.33 \\
\cline{2-5} 
& {\makecell{dog on \\ chessboard\\ - BG + FG rot.}}
& {\makecell{dog on \\ horizontal lines\\ - BG + FG rot.}} &2.73 &3.65 \\
\hline
\hline

\multirow{2}{*}{\textbf{4}}
& {\makecell{dog on \\ grid lines\\ - BG rot.}}
& {\makecell{dog on \\ horizontal lines\\ - BG rot.}} & 29.95 & 21.67\\
\cline{2-5} 
& {\makecell{dog on \\ grid lines\\ - BG rot.}}
& {\makecell{dog on \\ vertical lines\\ - BG rot.}} & 58.98 & 115.78\\
\hline
\end{tabular}

\caption{Impact of background (BG) image rotation on foreground (FG) rotation (rot.) estimation for dog foreground images (Scale 1) using LLaVA 1.5 and LLaVA 1.6 - Mean Absolute Error (MAE) for synthetic background image sets under different rotation conditions. When background is rotated, performance: (1) degrades sharply when trained on only FG rot. images, (2) and (3) improves significantly when trained on BG+FG rot., (4) improves moderately when trained on only BG rot. }
\label{tab:dog_synthetic_mae}
   \vspace{-2mm}
\end{table*}

\begin{table}[t]
\centering
 \scriptsize
\renewcommand{\arraystretch}{1.1}
\setlength{\tabcolsep}{6pt}
\begin{tabular}{|l|l|c|c|c|}
\hline
\textbf{S/N} & \textbf{Train  Set} & \textbf{Test set} & \multicolumn{2}{|c|}{\textbf{\makecell{MAE \\ (degrees)}}} \\
\hline
\multicolumn{1}{|c|}{} & \multicolumn{1}{c|}{} & \multicolumn{1}{c|}{} & \textbf{LLaVA 1.5} & \textbf{LLaVA 1.6} \\
\hline

\multirow{12}{*}{\textbf{1}}
& \multirow{3}{*}{\makecell{lizard on \\ chessboard \\(FG rotated)}}
 & baseline (FG rotated) &1.81  & 1.36\\
&  & BG + FG rotated & 85.2 &85.55 \\
&  & BG rotated &80.2  & 79.93\\
\cline{2-5} 

& \multirow{3}{*}{\makecell{lizard on \\ grid\\ (FG rotated)}}
 & baseline (FG rotated) &1.87  &1.16 \\
&  & BG + FG rotated &84.1  &87.22 \\
&  & BG rotated & 60.18 & 67.97\\
\cline{2-5}

& \multirow{3}{*}{\makecell{lizard on \\ horizontal lines\\(FG rotated)}}
 & baseline (FG rotated) &  2.23 &1.24 \\
&  & BG + FG rotated & 87.47 & 84.32\\
&  & BG rotated & 70.57 & 69.18\\
\cline{2-5}

& \multirow{3}{*}{\makecell{lizard on \\ vertical lines\\(FG rotated)}}
 & baseline (FG rotated) &1.68  &1.38 \\
&  & BG + FG rotated & 94.57 & 85.71\\
&  & BG rotated & 72.29&70.37 \\
\hline
\hline

\multirow{2}{*}{\textbf{2}}
& {\makecell{lizard on \\ grid lines\\ - BG + FG rot.}}
& {\makecell{lizard on \\ vertical lines\\ - BG + FG rot.}} & 1.36 &1.37 \\
\cline{2-5} 
& {\makecell{lizard on \\ grid lines\\ - BG + FG rot.}}
& {\makecell{lizard on \\ horizontal lines\\ - BG + FG rot.}} & 1.75 & 1.61\\
\hline
\hline

\multirow{2}{*}{\textbf{3}}
& {\makecell{lizard on \\ chessboard\\ - BG + FG rot.}}
& {\makecell{lizard on \\ vertical lines\\ - BG + FG rot.}} & 2.49&3.88 \\
\cline{2-5} 
& {\makecell{lizard on \\ chessboard\\ - BG + FG rot.}}
& {\makecell{lizard on \\ horizontal lines\\ - BG + FG rot.}} & 2.8&3.75 \\
\hline
\hline

\multirow{2}{*}{\textbf{4}}
& {\makecell{lizard on \\ grid lines\\ - BG rot.}}
& {\makecell{lizard on \\ horizontal lines\\ - BG rot.}} & 37.96 & 29.28\\
\cline{2-5} 
& {\makecell{lizard on \\ grid lines\\ - BG rot.}}
& {\makecell{lizard on \\ vertical lines\\ - BG rot.}} & 94.99 &81.97 \\
\hline
\end{tabular}
   \vspace{-2mm}

\caption{Impact of background (BG) image rotation on foreground (FG) rotation (rot.) estimation for lizard foreground images (Scale 1) using LLaVA 1.5 and LLaVA 1.6 - Mean Absolute Error (MAE) for synthetic background image sets under different rotation conditions. When background is rotated, performance: (1) degrades sharply when trained on only FG rot. images, (2) and (3) improves significantly when trained on BG+FG rot., (4) improves moderately when trained on only BG rot. }
\label{tab:lizard_synthetic_mae}
   \vspace{-2mm}
\end{table}

\begin{table}[t]
\centering
 \scriptsize
\renewcommand{\arraystretch}{1.1}
\setlength{\tabcolsep}{6pt}
\begin{tabular}{|l|l|c|c|c|}
\hline
\textbf{S/N} & \textbf{Train  Set} & \textbf{Test set} & \multicolumn{2}{|c|}{\textbf{\makecell{MAE \\ (degrees)}}} \\
\hline
\multicolumn{1}{|c|}{} & \multicolumn{1}{c|}{} & \multicolumn{1}{c|}{} & \textbf{LLaVA 1.5} & \textbf{LLaVA 1.6} \\
\hline

\multirow{12}{*}{\textbf{1}}
& \multirow{3}{*}{\makecell{train on \\ chessboard \\(FG rotated)}}
 & baseline (FG rotated) &1.2  & 0.77\\
&  & BG + FG rotated & 86.59 & 82.96\\
&  & BG rotated & 80.46 & 81\\
\cline{2-5} 

& \multirow{3}{*}{\makecell{train on \\ grid\\ (FG rotated)}}
 & baseline (FG rotated) & 1.3 & 0.94\\
&  & BG + FG rotated & 87.4 &87.96 \\
&  & BG rotated & 68.76 & 79.58\\
\cline{2-5}

& \multirow{3}{*}{\makecell{train on \\ horizontal lines\\(FG rotated)}}
 & baseline (FG rotated) &  1.43 & 0.85\\
&  & BG + FG rotated &82.36  & 88.84\\
&  & BG rotated &72.68  & 78.66\\
\cline{2-5}

& \multirow{3}{*}{\makecell{train on \\ vertical lines\\(FG rotated)}}
 & baseline (FG rotated) & 1.53 & 0.92\\
&  & BG + FG rotated & 77.62 &84.03 \\
&  & BG rotated &62.08 & 80.15\\
\hline
\hline

\multirow{2}{*}{\textbf{2}}
& {\makecell{train on \\ grid lines\\ - BG + FG rot.}}
& {\makecell{train on \\ vertical lines\\ - BG + FG rot.}} & 1.52 & 1.51\\
\cline{2-5} 
& {\makecell{train on \\ grid lines\\ - BG + FG rot.}}
& {\makecell{train on \\ horizontal lines\\ - BG + FG rot.}} & 1.31 & 1.25\\
\hline
\hline

\multirow{2}{*}{\textbf{3}}
& {\makecell{train on \\ chessboard\\ - BG + FG rot.}}
& {\makecell{train on \\ vertical lines\\ - BG + FG rot.}} &2.61 & 3.51\\
\cline{2-5} 
& {\makecell{train on \\ chessboard\\ - BG + FG rot.}}
& {\makecell{train on \\ horizontal lines\\ - BG + FG rot.}} &2.44 & 3.1\\
\hline
\hline

\multirow{2}{*}{\textbf{4}}
& {\makecell{train on \\ grid lines\\ - BG rot.}}
& {\makecell{train on \\ horizontal lines\\ - BG rot.}} & 55.09 & 25.01\\
\cline{2-5} 
& {\makecell{train on \\ grid lines\\ - BG rot.}}
& {\makecell{train on \\ vertical lines\\ - BG rot.}} & 102.92 &88.76 \\
\hline
\end{tabular}
   \vspace{-2mm}

\caption{Impact of background (BG) image rotation on foreground (FG) rotation (rot.) estimation for train foreground images (Scale 1) using LLaVA 1.5 and LLaVA 1.6 - Mean Absolute Error (MAE) for synthetic background image sets under different rotation conditions. When background is rotated, performance: (1) degrades sharply when trained on only FG rot. images, (2) and (3) improves significantly when trained on BG+FG rot., (4) improves moderately when trained on only BG rot. }
\label{tab:train_synthetic_mae}
\vspace{-2mm}
\end{table}

\end{document}